%% file: LIE_04_arxiv.tex
\documentclass[10pt,a4paper,oneside,titlepage,openright,fleqn,reqno,english,table]{report} 
\newcommand{\docname}{The SO(3) and SE(3) Lie Algebras of Rigid Body Rotations and Motions and their Application to Discrete Integration, Gradient Descent Optimization, and State Estimation}
\newcommand{\headname}{The SO(3) and SE(3) Lie Algebras of Rigid Body Rotations and Motions and their Application to \ldots}
\newcommand{\docdate}{August 2023}

\usepackage[hidelinks]{hyperref} % do not show links and references in red and green boxes
\usepackage{cmap} 				 % enable searches in PDF

\usepackage{multicol} % allows multiple columns

\usepackage{longtable} % Produces tables extending over several pages.
\usepackage{multirow} %Permit the use of the command \multirow to join several rows in a table
\usepackage{rotating} %Rotate text in tables
\usepackage[cp1252]{inputenc} % To able to use the spanish keyboard to introduce tildes and enyes
\usepackage{lscape} %Permits Landscape commands
\usepackage{url}    %Nicer view of the url
\usepackage[section]{placeins}  % force figures to remain in section
\usepackage{afterpage} % Big Figures are sent to a new page instead of to the end of the document
\usepackage{capt-of}        %Permit the placement of the figures right where you define them outside the floating environment
\usepackage{dirtytalk} % contains the \say instruction for quotes

\usepackage[T1]{fontenc}
\usepackage{tabularx, booktabs}   % permits linebreaks in table headings
\usepackage{makecell} % permits linebreaks in table cells
\usepackage{array} % permits centered table cells
\usepackage{xcolor} % add color options
\usepackage{xfrac} % add slanted fractions
\usepackage{tocloft} % avoid page numbering in table of contents
\usepackage{setspace} % change line spacing
\usepackage{chngcntr} % remove figure and table counter in conclusions

\graphicspath{{figs/}} 					% folder where figures are located
\usepackage{fancyhdr} 					% flexibility for header typesetting
\fancypagestyle{plain}{ % same header and footer for the first page of each chapter (otherwise they only have footer with page at center)
	\fancyhead{}																
	\fancyhead[LO,RE]{\raisebox{2mm}{\footnotesize{Eduardo Gallo}}}				
	\fancyhead[C]{}																
	\fancyhead[RO,LE]{\raisebox{2mm}{\footnotesize{\headname}}}	
											
	\fancyfoot{}																
	\fancyfoot[LO,RE]{}															
	\fancyfoot[C]{\rule{0pt}{12pt}}				 												
	\fancyfoot[RO,LE]{\thepage}													
}
  
\fancypagestyle{nopagenumber}{
	\fancyhead{}																
	\fancyhead[LO,RE]{\raisebox{2mm}{\footnotesize{Eduardo Gallo}}}				
	\fancyhead[C]{}																
	\fancyhead[RO,LE]{\raisebox{2mm}{\footnotesize{\headname}}}	
											
	\fancyfoot{}																
	\fancyfoot[LO,RE]{}															
	\fancyfoot[C]{\rule{0pt}{12pt}}				 												
} 
 
\usepackage[parfill]{parskip} 					% paragraph separation not messing up lists

\usepackage{amsmath} 			% supports more mathematical constructions
\usepackage{amsfonts}			% supports specific fonts in math environment
\usepackage{mathtools}			% supports even more math (and includes amsmath)
\everymath{\displaystyle}		% proper display (bigger size) of some math operators
\usepackage{amssymb}  			% supports even more mathematical constructs
\usepackage{nicefrac}  			% supports slanted fractions
\usepackage{accents}			% support various and stacked accents such as circles (seems not to work)

\usepackage{selinput}
\SelectInputMappings{
  aacute={á},
  eacute={é},
  iacute={í},
  oacute={ó},
  uacute={ú},
  ntilde={ñ}
}

\usepackage{tikz}  % flow diagrams
\usetikzlibrary{shapes,arrows,arrows.meta,calc,shadows,fit,intersections,datavisualization.formats.functions,external}
\tikzstyle{block}        = [draw, fill=blue!30, rectangle, text centered, minimum height=3em, minimum width=6em] % rectangle
\tikzstyle{blockred}     = [draw, fill=red!30, rectangle, text centered, minimum height=3em, minimum width=6em] % rectangle red
\tikzstyle{blockyellow}  = [draw, fill=yellow!30, rectangle, text centered, minimum height=3em, minimum width=6em] % rectangle yellow
\tikzstyle{blockyellow1} = [draw, fill=yellow!15, rectangle, text centered, minimum height=3em, minimum width=6em] % rectangle very light yellow 
\tikzstyle{blockyellow2} = [draw, fill=yellow!30, rectangle, text centered, minimum height=3em, minimum width=6em] % rectangle light yellow
\tikzstyle{blockyellow3} = [draw, fill=yellow!45, rectangle, text centered, minimum height=3em, minimum width=6em] % rectangle dark yellow
\tikzstyle{blockyellow4} = [draw, fill=yellow!60, rectangle, text centered, minimum height=3em, minimum width=6em] % rectangle very dark yellow
\tikzstyle{blockgreen}   = [draw, fill=green!30, rectangle, text centered, minimum height=3em, minimum width=6em] % rectangle green
\tikzstyle{blockgreen1}  = [draw, fill=green!15, rectangle, text centered, minimum height=3em, minimum width=6em] % rectangle very light green
\tikzstyle{blockgreen2}  = [draw, fill=green!30, rectangle, text centered, minimum height=3em, minimum width=6em] % rectangle light green
\tikzstyle{blockgreen3}  = [draw, fill=green!45, rectangle, text centered, minimum height=3em, minimum width=6em] % rectangle dark green
\tikzstyle{blockgreen4}  = [draw, fill=green!60, rectangle, text centered, minimum height=3em, minimum width=6em] % rectangle very dark green
\tikzstyle{blockgrey1}   = [draw, fill=black!8,  rectangle, text centered, minimum height=3em, minimum width=6em] % rectangle very light grey
\tikzstyle{blockgrey2}   = [draw, fill=black!16, rectangle, text centered, minimum height=3em, minimum width=6em] % rectangle light grey
\tikzstyle{blockgrey3}   = [draw, fill=black!24, rectangle, text centered, minimum height=3em, minimum width=6em] % rectangle medium grey
\tikzstyle{blockgrey4}   = [draw, fill=black!32, rectangle, text centered, minimum height=3em, minimum width=6em] % rectangle drak grey
\tikzstyle{blockgrey5}   = [draw, fill=black!40, rectangle, text centered, minimum height=3em, minimum width=6em] % rectangle very dark grey
\tikzstyle{blocknofill}  = [draw=black!50, line width=1.5pt, rectangle, rounded corners, text centered, minimum height=3em, minimum width=6em]
\tikzstyle{blockhigh}    = [draw, fill=blue!20, rectangle, text centered, minimum height=6em, minimum width=6em]
\tikzstyle{noblock}      = [draw=black!50, line width=1.5pt, rectangle, rounded corners, text centered, minimum height=3em, minimum width=6em]
\tikzstyle{sum}          = [draw, fill=blue!20, circle, node distance=1cm]
\tikzstyle{sumrest}      = [draw, fill=black, circle, radius=0.5cm]
\tikzstyle{pinstyle}     = [pin edge={to-,thin,black}]
\tikzstyle{title}        = [text centered]
\pgfdeclarelayer{background}
\pgfdeclarelayer{foreground}
\pgfsetlayers{background,main,foreground}

\usepackage{pgfplots}
\pgfplotsset{compat=1.13}
\pgfplotsset{colormap/bluered}
\usepackage{pgfplotstable}
\usepgfplotslibrary{polar}
\usepgfplotslibrary{colormaps}
\usepgfplotslibrary{external} 
 
\usepackage{graphicx, color} 	% This package makes it easy to include graphics in a LaTeX document.
\usepackage{epstopdf}  			% Convert automatically eps graphics when compiling with pdflatex

\usepackage{emptypage} % empty pages created by cleardoublepage have no page number

\usepackage{changepage} % change margins for single paraphaphs

\allowdisplaybreaks % Allows LaTeX to break pages automatically within multiline formulas as necessary.

\input{symbols}				% definition of new commands

\begin{document}
%\layout		% keep commented and activate together with package above to show page layout on document

\input{files_arxiv/ch00_title_page} 		\thispagestyle{empty} \cleardoublepage	
\input{files_arxiv/ch00_abstract}   		\thispagestyle{empty} \cleardoublepage
\input{files_arxiv/ch00_notation}   		\thispagestyle{empty} \cleardoublepage
\input{files_arxiv/ch00_acronyms} 		\thispagestyle{empty} \cleardoublepage

\pagestyle{plain}
\pagenumbering{roman}
\tableofcontents \cleardoublepage
\pagenumbering{arabic}
 
\include{files_arxiv/ch01_intro} \cleardoublepage
\include{files_arxiv/ch02_estim} \cleardoublepage 
\include{files_arxiv/ch03_algebra} \cleardoublepage
\include{files_arxiv/ch04_rotate} \cleardoublepage 
\include{files_arxiv/ch05_motion} \cleardoublepage 
\include{files_arxiv/ch06_multiple} \cleardoublepage 

\addcontentsline{toc}{chapter}{Bibliography} % so Bibliography (or anything else) appears on table of contents without numbering
\bibliographystyle{ieeetr}   % bibligraphy style (ordered according to \cite commands, others are unsrt, plain, ieeetr)
\bibliography{lie_algebra}
\cleardoublepage

\end{document}

%% file: symbols.tex
\newcommand{\ds}    {\displaystyle} 		% shortning for \displaystyle, which is used for all math anyway
\newcommand{\ttt}	{\textstyle}            % shortning for \textstyle (no math, normal text)
\newcommand{\sss}   {\scriptscriptstyle}    % shortning for \scriptscriptstyle (smaller than scriptstyle)

\DeclareMathOperator*{\argmin}{arg\,min}    % new operator
\DeclareMathOperator*{\sign}{sign}

\newcommand{\nm}   		[1] {\ensuremath{\mathrm{#1}}} 									% upright characters in math mode
\newcommand{\neweq}     [2] {\begin{equation} \mathrm{#1}\label{#2} \end{equation}} 	% numbered equation with upright characters
\renewcommand{\vec}		[1] {\mbox{\boldmath{\ensuremath{\mathrm{#1}}}}}	% redefines \vec command to turn vectors boldface
\newcommand{\pderpar}   [2] {\dfrac{\partial{#1}}{\partial{#2}}}    		% partial differential with respect to parameter #2
\newcommand{\lrp}       [1] {\left(#1\right)}								% left and right parethesis
\newcommand{\lrsb}      [1] {\left[#1\right]}								% left and right square brackets
\newcommand{\lrb}       [1] {\left\{#1\right\}}								% left and right brackets

\newcommand{\hypertt}	[1] {\hyperlink{#1}{\texttt{#1}}}

\newcommand{\Deltat}     	{\Delta t}

\newcommand{\first}         {\nm{1^{\sss st}}}	 % 1st
\newcommand{\second}        {\nm{2^{\sss nd}}}   % 2nd
\newcommand{\third}       	{\nm{3^{\sss rd}}}   % 3rd
\newcommand{\fourth}        {\nm{4^{\sss th}}}   % 4th
\newcommand{\fifth}         {\nm{5^{\sss th}}}   % 5th
\newcommand{\RNBtrans}		{{\vec R}_{\sss NB}^{T}}
\newcommand{\RNBdottrans}	{\vec{\dot R}_{\sss NB}^{T}}

\newcommand{\Tzeroone}			{\vec T_{01}}
\newcommand{\Tzerotwo}			{\vec T_{02}}
\newcommand{\Tonetwo}			{\vec T_{12}}
\newcommand{\Tzeroonezero}		{\vec T_{01}^{0}}
\newcommand{\Tzerotwozero}		{\vec T_{02}^{0}}
\newcommand{\Tonetwozero}		{\vec T_{12}^{0}}
\newcommand{\Tonetwoone}		{\vec T_{12}^{1}}
\newcommand{\Tzerotwozerodot}	{\vec {\dot T}_{02}^{0}}
\newcommand{\Tonetwoonedot}		{\vec {\dot T}_{12}^{1}}
\newcommand{\Tzeroonezerodot}	{\vec {\dot T}_{01}^{0}}

\newcommand{\Rzeroone}		{\vec R_{01}}
\newcommand{\Rzerotwo}		{\vec R_{02}}
\newcommand{\Ronetwo}		{\vec R_{12}}
\newcommand{\Rzeroonedot}	{\vec{\dot R}_{01}}
\newcommand{\Rzerotwodot}	{\vec{\dot R}_{02}}
\newcommand{\Ronetwodot}	{\vec{\dot R}_{12}}

\newcommand{\wzeroone}				{\vec \omega_{01}}
\newcommand{\wzerotwo}				{\vec \omega_{02}}
\newcommand{\wonetwo}				{\vec \omega_{12}}
\newcommand{\wonetwozero}			{\vec \omega_{12}^{0}}
\newcommand{\wzerotwozero}			{\vec \omega_{02}^{0}}
\newcommand{\wzeroonezero}			{\vec \omega_{01}^{0}}
\newcommand{\wonetwoone}			{\vec \omega_{12}^{1}}
\newcommand{\wzerotwozerodot}		{\vec{\dot \omega}_{02}^{0}}
\newcommand{\wonetwoonedot}			{\vec{\dot \omega}_{12}^{1}} 
\newcommand{\wzeroonezerodot}		{\vec{\dot \omega}_{01}^{0}}

\newcommand{\wzerooneskew}			{\widehat{\vec \omega}_{01}}
\newcommand{\wzerooneoneskew}		{\widehat{\vec \omega}_{01}^{1}}
\newcommand{\wzeroonezeroskew}		{\widehat{\vec \omega}_{01}^{0}} 
\newcommand{\wzerotwotwoskew}		{\widehat{\vec \omega}_{02}^{2}}
\newcommand{\wonetwotwoskew}		{\widehat{\vec \omega}_{12}^{2}}
\newcommand{\wzeroonetwoskew}		{\widehat{\vec \omega}_{01}^{2}}
\newcommand{\wzerooneoneskewdot}	{\dot{\widehat{\vec \omega}}_{01}^{1}}

\newcommand{\vzeroonezero}		{\vec v_{01}^{0}}
\newcommand{\vzerotwozero}		{\vec v_{02}^{0}}
\newcommand{\vonetwozero}		{\vec v_{12}^{0}}
\newcommand{\vonetwoone}		{\vec v_{12}^{1}}

\newcommand{\xvec}   			{\vec x}

\newcommand{\xvecdot}			{\dot {\vec x}}
\newcommand{\xvecest}	  		{\hat{\vec x}}

\newcommand{\FE}		 {F_{\sss E}}

\newcommand{\OECEF}    	 {O_{\sss E}}                           % Earth frame

\newcommand{\iEi}        {\vec i_{1}^{\sss E}}
\newcommand{\iEii}       {\vec i_{2}^{\sss E}}
\newcommand{\iEiii}      {\vec i_{3}^{\sss E}}

\newcommand{\FN} 		  {F_{\sss N}}
\newcommand{\ON}          {O_{\sss N}}                            % NED frame

\newcommand{\pN}          {\vec p^{\sss N}}

\newcommand{\FB}		  {F_{\sss B}}
\newcommand{\OB}          {O_{\sss B}}                            % Body frame

\newcommand{\iBi}         {\vec i_{1}^{\sss B}}
\newcommand{\iBii}        {\vec i_{2}^{\sss B}}
\newcommand{\iBiii}       {\vec i_{3}^{\sss B}}

\newcommand{\pB}          {\vec p^{\sss B}}
\newcommand{\pBbar}       {\vec {\bar p}^{\sss B}}

\newcommand{\vvec} 				{\vec v}

\newcommand{\vE}  				{\vec v^{\sss E}}
\newcommand{\vB} 	   			{\vec v^{\sss B}}

\newcommand{\vN}	    		{\vec v^{\sss N}}

\newcommand{\vzeroone}			{\vec v_{01}}
\newcommand{\vzerotwo}			{\vec v_{02}}
\newcommand{\vonetwo}			{\vec v_{12}}

\newcommand{\vzerotwozerodot}	{\dot {\vec v}_{02}^{0}}
\newcommand{\vonetwoonedot}		{\dot {\vec v}_{12}^{1}}
\newcommand{\vzeroonezerodot}	{\dot {\vec v}_{01}^{0}}
\newcommand{\vtilde}			{\widetilde{\vec v}}

\newcommand{\TEBEskew}		{\widehat{\vec T}_{\sss EB}^{\sss E}}
\newcommand{\Tskew}			{\widehat{\vec T}}
\newcommand{\pskew}			{\widehat{\vec p}}

\newcommand{\zetaEB}		{\vec \zeta_{\sss EB}}
\newcommand{\qEB}			{\vec q_{\sss EB}}

\newcommand{\TEBE}			{\vec T_{\sss EB}^{\sss E}}
\newcommand{\qEBast}		{\vec q_{\sss EB}^{\ast}}

\newcommand{\pE}          {\vec p^{\sss E}}
\newcommand{\pEbar}       {\vec {\bar p}^{\sss E}}
\newcommand{\pEbardot}    {\vec {\dot {\bar p}}^{\sss E}}

\newcommand{\omegaskew}		{\widehat{\vec \omega}}
\newcommand{\nskew}			{\widehat{\vec n}}
\newcommand{\rskew}			{\widehat{\vec r}}

\newcommand{\wEB}			{\vec \omega_{\sss EB}}
\newcommand{\wEBB}			{\vec \omega_{\sss EB}^{\sss B}}
\newcommand{\wEBBskew}		{\widehat{\vec \omega}_{\sss EB}^{\sss B}}

\newcommand{\wNBN}			{\vec \omega_{\sss NB}^{\sss N}}
\newcommand{\wNBB}			{\vec \omega_{\sss NB}^{\sss B}}

\newcommand{\wNBNskew}		{\widehat{\vec \omega}_{\sss {NB}}^{\sss N}}
\newcommand{\wNBBskew}		{\widehat{\vec \omega}_{\sss {NB}}^{\sss B}}

\newcommand{\pbar}				{\vec {\bar p}}
\newcommand{\qbar}				{\vec {\bar q}}
\newcommand{\vbar}				{\vec {\bar v}}

\newcommand{\RNB}			{\vec R_{\sss NB}}

\newcommand{\REB}			{\vec R_{\sss EB}}

\newcommand{\RNBdot}		{\dot{\vec R}_{\sss NB}}

\newcommand{\REBdot}		{\dot{\vec R}_{\sss EB}}

\newcommand{\qNB}			{\vec q_{\sss NB}}

\newcommand{\qNBdot}		{\dot{\vec q}_{\sss NB}}

\newcommand{\qast}		    {\vec q^{\ast}}

\newcommand{\qr}            {\vec q_r}
\newcommand{\qd}            {\vec q_d}
\newcommand{\qrast}         {\vec q_r^{\ast}}
\newcommand{\qdast}         {\vec q_d^{\ast}}

\newcommand{\zetaast}		{\vec \zeta^{\ast}}

\newcommand{\zetabullet}	{\vec \zeta^{\bullet}}
\newcommand{\zetacirc}		{\vec \zeta^{\circ}}

\newcommand{\phiNB}			  	{\vec \phi^{\sss NB}}

\newcommand{\xiEBBskew}		{\vec \xi_{\sss {EB}}^{\sss B\wedge}} 

\newcommand{\zetaEBast}		{\vec \zeta_{\sss EB}^{\ast}}
\newcommand{\zetaEBbullet}	{\vec \zeta_{\sss EB}^{\bullet}}
\newcommand{\zetaEBcirc}	{\vec \zeta_{\sss EB}^{\circ}}
\newcommand{\zetaBE}		{\vec \zeta_{\sss BE}}
\newcommand{\qBE}			{\vec q_{\sss BE}}

\newcommand{\Ma}			{\vec M_a}
\newcommand{\Mb}			{\vec M_b}
\newcommand{\MEB}			{\vec M_{\sss EB}}
\newcommand{\MEN}			{\vec M_{\sss EN}}
\newcommand{\MNB}			{\vec M_{\sss NB}}

\newcommand{\xiEBEskew}		{\vec \xi_{\sss {EB}}^{\sss E\wedge}}
\newcommand{\MEBdot}     	{\vec {\dot M}_{\sss EB}}
\newcommand{\MEBinv}     	{\vec M_{\sss EB}^{-1}}
\newcommand{\MEBdotinv}		{\vec {\dot M}_{\sss EB}^{-1}}

\newcommand{\xiEBE}			{\vec \xi_{\sss EB}^{\sss E}}
\newcommand{\xiEBB}			{\vec \xi_{\sss EB}^{\sss B}}
\newcommand{\pBbardot}    {\vec {\dot {\bar p}}^{\sss B}}
\newcommand{\vpEbar}        {\vec {\bar v}_{\sss p}^{\sss E}}
\newcommand{\vpBbar}        {\vec {\bar v}_{\sss p}^{\sss B}}

\newcommand{\TBEB}			{\vec T_{\sss BE}^{\sss B}}
\newcommand{\zetaEBdot}		{\vec{\dot \zeta}_{\sss EB}}
\newcommand{\nuEBB}			{\vec \nu_{\sss EB}^{\sss B}}

\newcommand{\REBtrans}		{{\vec R}_{\sss EB}^{T}}
\newcommand{\wEBEskew}		{\widehat{\vec \omega}_{\sss {EB}}^{\sss E}}
\newcommand{\nuEBE}			{\vec \nu_{\sss EB}^{\sss E}}
\newcommand{\wEBE}			{\vec \omega_{\sss EB}^{\sss E}}

\newcommand{\azeroone}		{\vec a_{01}}
\newcommand{\azerotwo}		{\vec a_{02}}
\newcommand{\aonetwo}		{\vec a_{12}}
\newcommand{\azeroonezero}	{\vec a_{01}^{0}}
\newcommand{\azerotwozero}	{\vec a_{02}^{0}}
\newcommand{\aonetwoone}	{\vec a_{12}^{1}}
\newcommand{\aonetwozero}	{\vec a_{12}^{0}}

\newcommand{\alphazeroone}				{\vec \alpha_{01}}
\newcommand{\alphazerotwo}				{\vec \alpha_{02}}
\newcommand{\alphaonetwo}				{\vec \alpha_{12}}
\newcommand{\alphazerotwozero}			{\vec \alpha_{02}^{0}}
\newcommand{\alphaonetwoone}			{\vec \alpha_{12}^{1}}
\newcommand{\alphazeroonezero}			{\vec \alpha_{01}^{0}}
\newcommand{\alphaonetwozero}			{\vec \alpha_{12}^{0}}
\newcommand{\alphazerooneskew}			{\widehat{\vec \alpha}_{01}}
\newcommand{\alphazerooneoneskew}		{\widehat{\vec \alpha}_{01}^{1}}
\newcommand{\alphazeroonezeroskew}		{\widehat{\vec \alpha}_{01}^{0}}

\newcommand{\uvec}						{\vec u}
\newcommand{\utilde}					{\widetilde{\vec u}}
\newcommand{\wvec}						{\vec w}
\newcommand{\wtilde}					{\widetilde{\vec w}}
\newcommand{\Avec}						{\vec A}
\newcommand{\Bvec}						{\vec B}
\newcommand{\Ivec}						{\vec I}
\newcommand{\Lvec}						{\vec L}
\newcommand{\Qvec}						{\vec Q}
\newcommand{\Qtilde}					{\widetilde{\vec Q}}
\newcommand{\zvec}						{\vec z}
\newcommand{\Fvec}						{\vec F}
\newcommand{\Gvec}						{\vec G}
\newcommand{\Rvec}						{\vec R}
\newcommand{\Pvec}						{\vec P}

\newcommand{\yvec}						{\vec y}
\newcommand{\Hvec}						{\vec H}
\newcommand{\Mvec}						{\vec M}
\newcommand{\Rtilde}					{\widetilde{\vec R}}
\newcommand{\Kvec}						{\vec K}
\newcommand{\rvec}						{\vec r}
\newcommand{\Svec}						{\vec S}

%% file: files_arxiv/ch00_title_page.tex
\begin{titlepage}
	\begin{center}
		\begin{Huge} \hspace*{0cm} \bigskip \bigskip \bigskip \bigskip \end{Huge}
		\\[1.5cm]
		
		\begin{Huge}
			\begin{spacing}{1.5}
				\textcolor{black!80!gray}{\textbf{\docname}}
			\end{spacing}
		\end{Huge}
		
		\vspace{2.5cm}
		
		\begin{Large}
			\begin{spacing}{1.4}
				\textcolor{black!80!gray}{\textbf{Universidad Politécnica de Madrid \\ Centro de Automática y Robótica}}
			\end{spacing}
		\end{Large}
		
		\vspace{1.5cm}
		
		\begin{Large}
			\begin{spacing}{1.4}
				\textcolor{black!80!gray}{\textbf{\docdate}}
			\end{spacing}
		\end{Large}
		
		\vspace{2.5cm}
		
		\begin{large}
			\begin{tabular}{p{1.7cm}p{4.0cm}}
			%\begin{tabular}{p{2.5cm}p{4.0cm}}
			Author: & \rule[-12pt]{0pt}{12pt}\textbf{Eduardo Gallo} \\
			%Reviewed by: & \textbf{Antonio Barrientos} \\
			\end{tabular}
		\end{large}
		
	\end{center}
\end{titlepage}

%% file: files_arxiv/ch00_abstract.tex
\addtocontents{toc}{\cftpagenumbersoff{chapter}}
\chapter*{Abstract} \label{cha:Abstract} % The * avoids the automatic inclusion of the section name in the table of contents and no number is assigned
\addcontentsline{toc}{chapter}{Abstract} % only to include it in table of contents

Classical mathematical techniques such as discrete integration, gradient descent optimization, and state estimation (exemplified by the Runge-Kutta method, Gauss-Newton minimization, and extended Kalman filter or \hypertt{EKF}, respectively), rely on linear algebra and hence are only applicable to state vectors belonging to Euclidean spaces when implemented as described in the literature. This document discusses how to modify these methods so they can be applied to non-Euclidean state vectors, such as those containing rotations and full motions of rigid bodies.

To do so, this document provides an in-depth review of the concept of manifolds or Lie groups, together with their tangent spaces or Lie algebras, their exponential and logarithmic maps, the analysis of perturbations, the treatment of uncertainty and covariance, and in particular the definitions of the Jacobians required to employ the previously mentioned calculus methods. These concepts are particularized to the specific cases of the \nm{\mathbb{SO}\lrp{3}} and \nm{\mathbb{SE}\lrp{3}} Lie groups, known as the special orthogonal and special Euclidean groups of \nm{\mathbb{R}^3}, which represent the rigid body rotations and motions, describing their various possible parameterizations as well as their advantages and disadvantages.

\textbf{\emph{Keywords}}: Lie algebra, SO(3), SE(3), manifold, tangent space, state estimation, EKF, discrete integration, Runge-Kutta, gradient descent optimization, minimization, Gauss-Newton

%% file: files_arxiv/ch00_notation.tex
\addtocontents{toc}{\cftpagenumbersoff{chapter}}
\chapter*{Notation} \label{cha:Notation} % The * avoids the automatic inclusion of the section name in the table of contents and no number is assigned
\addcontentsline{toc}{chapter}{Notation} % only to include it in table of contents
\addtocontents{toc}{\cftpagenumberson{chapter}}

This document is organized into chapters (e.g., chapter \ref{cha:Rotate}), sections (e.g., section \ref{sec:RigidBody_rotation_rodrigues}), subsections (e.g., section \ref{subsec:RigidBody_rotation_rodrigues_quat}),   and so on. Note that the prefix \say{sub} is not employed when referring to subsections. Referencing tables (e.g., table \ref{tab:Rotate_lie_unit_quat}) and figures (e.g., figure \ref{fig:Error_robust_estimators_bis}) is also straightforward. When a reference appears in between parenthesis, it refers to an equation or mathematical expression, as in (\ref{eq:SO3_quat_concatenation}), and when in between square brackets, to a bibliographic reference, as in \cite{Sola2018}. Acronyms are displayed in uppercase teletype font, and reference to the proper location within the acronym list, as in \hypertt{EKF}.

With respect to the mathematical notation, vectors and matrices appear in bold (e.g., \nm{\vec x} or \nm{\vec R}), and in general the former are lowercase and the latter uppercase, although not always. A variable or vector with a hat over it \nm{< \hat{\cdot} >} refers to its estimated value, and with a dot \nm{< \dot{\cdot} >} to its time derivative. Other commonly employed symbols for vectors are the wide hat \nm{< \widehat{\cdot} >}, which refers to its skew-symmetric form, the double vertical bars \nm{< \| \cdot \| >}, which refer to its norm, and the wedge \nm{<\cdot^\wedge>}, which implies a tangent space representation. The superindex \texttt{T} denotes the transpose of a vector or matrix. In the case of a scalar, the vertical bars \nm{< | \cdot | >} refer to its absolute value. The left arrow \nm{\longleftarrow} represents an update operation, in which the value on the right of the arrow is assigned to the variable on the left.

Superindexes are employed over vectors to specify the reference frame in which they are viewed or evaluated (e.g., \nm{\xvec^{\sss A}} refers to the vector \nm{\xvec} viewed in frame \texttt{A}, while \nm{\xvec^{\sss B}} is the same vector but viewed in frame \texttt{B}). Where two reference frames appear as subindexes to a vector, it means that the vector goes from the first frame to the second. For example, \nm{\xvec_{\sss AB}^{\sss C}} refers to the vector \nm{\xvec} from frame \texttt{A} to frame \texttt{B} viewed in frame \texttt{C}\footnote{This is analogous to the vector \nm{\xvec} of frame \texttt{B} with respect to frame \texttt{A} viewed in frame \texttt{C}.}.

Many different types of Jacobians or function partial derivative matrices can be found in this document (formal definitions can be found in sections \ref{subsec:algebra_lie_derivatives} and \ref{sec:algebra_lie_jacobians}), and a special notation is hence required. Jacobians are represented by a \nm{\vec J} combined with a subindex and a superindex:
\begin{itemize}
\item The subindex provides information about the function domain, and is composed by a symbol representing how increments are added to the domain, followed by the domain itself. A \nm{+} is employed when the domain belongs to an Euclidean space\footnote{The subindex symbol may sometimes be omitted when it is clear from the context that the domain is Euclidean.} but not when it is a manifold or Lie group; in these cases, \nm{\oplus} means that increments are added in the local tangent space and \nm{\boxplus} that they are added in the global tangent space. 
\item The superindex is composed by a second symbol representing how differences are computed in the function image or codomain, followed by the codomain itself. A \nm{-} is employed when the codomain is Euclidean\footnote{The superindex symbol may sometimes be omitted when it is clear from the context that the codomain is Euclidean.}, \nm{\ominus} when the differences are evaluated in the local tangent space to the codomain manifold, and \nm{\boxminus} when evaluated in the global tangent space.
\end{itemize}

%% file: files_arxiv/ch00_acronyms.tex
\addtocontents{toc}{\cftpagenumbersoff{chapter}}
\chapter*{Acronyms} \label{cha:Acronyms} % The * avoids the automatic inclusion of the section name in the table of contents and no number is assigned
\addcontentsline{toc}{chapter}{Acronyms} % only to include it in table of contents
\addtocontents{toc}{\cftpagenumberson{chapter}}

\renewcommand{\arraystretch}{1.25}
\begin{table}[ht]
\begin{tabular}{lp{6.0cm}p{0.1cm}lp{6.0cm}}
	\hypertarget{CDF}{\texttt{CDF}}			& Cumulative Distribution Functions	& & \hypertarget{PDF}{\texttt{PDF}}	  		& Probability Density Function		\\ 
	\hypertarget{ECEF}{\texttt{ECEF}}		& Earth Centered Earth Fixed		& & \hypertarget{PMF}{\texttt{PMF}}		  	& Probability Mass Function		  	\\
	\hypertarget{EKF}{\texttt{EKF}}			& Extended Kalman Filter			& & \hypertarget{PSD}{\texttt{PSD}}			& Power Spectral Density			\\
	\hypertarget{NED}{\texttt{NED}}			& North East Down					& & \hypertarget{ScLERP}{\texttt{ScLERP}}	& Screw linear interpolation		\\	
	\hypertarget{ODE}{\texttt{ODE}}			& Ordinary Differential Equation	& & \hypertarget{SLERP}{\texttt{SLERP}}		& Spherical linear interpolation	\\
\end{tabular}
\end{table}
\renewcommand{\arraystretch}{1.0}

%% file: files_arxiv/ch01_intro.tex
\chapter{Introduction and Outline} \label{cha:Outline}

The classical implementations of common calculus techniques such as discrete integration (exemplified by the Runge-Kutta method), gradient descent optimization (Gauss-Newton or Levenberg-Marquardt), and state estimation (extended Kalman filter or \hypertt{EKF}), are designed to work on state vectors belonging to Euclidean spaces\footnote{Although Euclidean spaces are formally defined in section \ref{sec:algebra_structures}, they can be informally understood as those that comply with the five axioms of Euclidean geometry (\nm{\first} things that are equal to the same thing are also equal to one another; \nm{\second} if equals be added to equals, the wholes are equal; \nm{\third} if equals be subtracted from equals, the remainders are equal; \nm{\fourth} things that coincide with one another are equal to one another; \nm{\fifth} the whole is greater than the part).} and hence rely of linear algebra \cite{Blanco2020}. There exist two possible approaches for those cases in which the state vector contains non-Euclidean components, such as rigid body motions or rotations:
\begin{itemize}

\item The solution most commonly observed in the literature is to incorporate each component of the pose\footnote{The pose of an object includes both its position and attitude.} (attitude) as an unconstrained real number into the state vector. The above techniques can then be employed without difficulties, but the resulting state vector does not comply with the constraints imposed by having some of its members being components of a rigid body pose (attitude), as these constraints have not been taken into account when integrating, optimizing, or estimating. The solutions hence need to be projected back into the space of valid rigid body poses (attitudes), but this does not hide the fact that the whole process has been performed without respecting the motion (rotation) constraints, which often has negative consequences for the accuracy, stability, and consistency of the solution \cite{Sola2018}.

\item The approach taken in some recent robotics literature, in particular in the field of motion estimation for navigation, consists on reformulating the above calculus techniques (integration, optimization, filtering) taking into account that some members of the state vector represent rigid body poses (attitudes) and hence can not be treated as Euclidean. By modeling these states properly, the quality of the solution can be improved. The use of Lie theory, with its manifolds and tangent spaces, enables the construction of rigorous calculus techniques to handle uncertainties, derivatives, and integrals of non-Euclidean elements with precision and ease \cite{Sola2018}.
\end{itemize}

Chapter \ref{cha:StateEstimation} discusses key statistical concepts such as random variables, stochastic processes, white noise, and robust estimators, and describes the classical approaches (this is, when applied to Euclidean spaces) to discrete integration, gradient descent optimization, and state estimation. Chapter \ref{cha:Algebra} introduces the concepts of Lie groups and their tangent spaces or Lie algebras, and adapts the three calculus techniques so they can be applied when the state vector contains non-Euclidean Lie group components. Chapters \ref{cha:Rotate} and \ref{cha:Motion} particularize the generic concepts of chapter \ref{cha:Algebra} to the specific cases of both rigid body rotations and complete motions, placing special emphasis on the Lie Jacobians that enable the application of the modified calculus methods to non-Euclidean spaces. Last, chapter \ref{cha:Composition} discusses the relationships among the velocities and accelerations (both linear and angular) of various rigid bodies or references frames.

%% file: files_arxiv/ch02_estim.tex
\chapter{Stochastic Processes and Euclidean Space Methods} \label{cha:StateEstimation}

The material in this chapter can be found in many textbooks and technical documents. It is summarized here to act as an introduction to the next chapters, which make frequent use of the concepts and equations discussed here. Section \ref{sec:Error_Random} provides an introduction to statistics and the concepts of random variables, stochastic processes, and white noise, followed by an overview of robust statistics and its associated estimators in section \ref{sec:Error_Robust}. The following sections describe the classical approaches to three frequent calculus problems, such as discrete integration (section \ref{sec:euclidean_integration}), gradient descent optimization (section \ref{sec:euclidean_gradient_descent}), and state estimation (section \ref{sec:SS}). The solutions, known as the Runge-Kutta integration method, the Gauss-Newton minimization, and the extended Kalman filter or \hypertt{EKF}, are intended for state vectors in which all components can be considered Euclidean. Note that the main objective of this document is how to modify these three techniques so they can cope with non-Euclidean state vectors.

%%%%%%%%%%%%%%%%%%%%%%%%%%%%%%%%%%%%%%%%%%%%%%%%%%%%%%%%%%%%%%%%%%%%%%%%
%%%%%%%%%%%%%%%%%%%%%%%%%%%%%%%%%%%%%%%%%%%%%%%%%%%%%%%%%%%%%%%%%%%%%%%%
%%%%%%%%%%%%%%%%%%%%%%%%%%%%%%%%%%%%%%%%%%%%%%%%%%%%%%%%%%%%%%%%%%%%%%%%
% SECTION		RANDOM VARIABLES, STOCHASTIC PROCESSES AND WHITE NOISE
%%%%%%%%%%%%%%%%%%%%%%%%%%%%%%%%%%%%%%%%%%%%%%%%%%%%%%%%%%%%%%%%%%%%%%%%
%%%%%%%%%%%%%%%%%%%%%%%%%%%%%%%%%%%%%%%%%%%%%%%%%%%%%%%%%%%%%%%%%%%%%%%%
%%%%%%%%%%%%%%%%%%%%%%%%%%%%%%%%%%%%%%%%%%%%%%%%%%%%%%%%%%%%%%%%%%%%%%%%

\section{Random Variables, Stochastic Processes, and White Noise}\label{sec:Error_Random}

This section provides an introduction to the random variables and processes required to model those physical systems that can not be represented by deterministic models due to their inherent randomness, which results in the same set of parameter values and initial conditions leading to different outputs.

%%%%%%%%%%%%%%%%%%%%%%%%%%%%%%%%%%%%%%%%%%%%%%%%%%%%%%%%%%%%%%%%%%%%%%%%
%%%%%%%%%%%%%%%%%%%%%%%%%%%%%%%%%%%%%%%%%%%%%%%%%%%%%%%%%%%%%%%%%%%%%%%%
% SUBSECTION		RANDOM VARIABLES
%%%%%%%%%%%%%%%%%%%%%%%%%%%%%%%%%%%%%%%%%%%%%%%%%%%%%%%%%%%%%%%%%%%%%%%%
%%%%%%%%%%%%%%%%%%%%%%%%%%%%%%%%%%%%%%%%%%%%%%%%%%%%%%%%%%%%%%%%%%%%%%%%

\subsection{Random Variables}\label{subsec:Error_RandomVariables}

Consider a random experiment (one in which the outcome is uncertain) with a sample space \nm{\Omega} (collection of possible elementary outcomes of the experiment), and let \nm{\omega} be a sample point belonging to \nm{\Omega}. A \emph{random variable} \nm{X\lrp{\omega}} (generally just \nm{X}) is a single valued real function that assigns a real number, called the value of \nm{X\lrp{\omega}}, to each sample point \nm{\omega \in \Omega} \cite{Ibe2005,Papoulis2002}. A random variable hence represents a map between the sample and real spaces \nm{\{X : \Omega \rightarrow \mathbb{R} \ | \ \omega \in \Omega \rightarrow X\lrp{\omega} \in \mathbb{R}\}}. The \emph{realization} of a random variable is the real variable obtained after a given experiment.

A random variable \nm{X} is completely described by its \emph{cumulative distribution function} (\hypertt{CDF}) \nm{F_X}, which represents the probability that the value of \nm{X} is less or equal than the function input \cite{Farrell2008}:
\neweq{F_X\lrp{x} = P\lrsb{X \leq x} \ \ \ \ \ \ \ -\infty < x < \infty}{eq:Error_rvar_CDF}

Random variables can also be described by the \hypertt{CDF} derivative, known as the \emph{probability mass function} (\hypertt{PMF}) \nm{p_X} in case of discrete random variables (those that can take at most a countable number of possible values) or as \emph{probability density function} (\hypertt{PDF}) \nm{f_X} for continuous ones (those that can take an uncountable number of possible values):
\begin{eqnarray}
\nm{F_X\lrp{x}} & = & \nm{\sum_{x_k \leq x} p_X\lrp{x_k}}\label{eq:Error_rvar_PMF} \\
\nm{F_X\lrp{x}} & = & \nm{\int_{- \infty}^x f_X\lrp{y} \, dy} \label{eq:Error_rvar_PDF}
\end{eqnarray}

The \emph{expected value} \nm{E\lrsb{X}} or \emph{mean} \nm{\mu_X} of a random variable \nm{X} is a function defined as its average value over a large number of experiments, and represents its central or typical value:
\neweq{E\lrsb{X} = \mu_X = \begin{dcases*}
\nm{\sum_k x_k \, p_X\lrp{x_k}} & when \nm{X} is discrete \\
\nm{\int_{- \infty}^{\infty} y \, f_X\lrp{y} \, dy} & when \nm{X} is continuous 
\end{dcases*}} {eq:Error_rvar_Mean}

If a function acts on a random variable, then its output is also a random variable \cite{Simon2006}, and it is hence possible to compute the expected value of the output random variable\footnote{Note that the mean can be considered as the expected value of the \nm{f\lrp{X} = X} function.}. The \emph{variance} \nm{\sigma_X^2}, \nm{Var\lrp{X}}, or second central moment of a random variable \nm{X}, is the expected value of the squared deviation of  \nm{X} from its mean, and measures the spread of its \hypertt{PMF} or \hypertt{PDF} about its expected value \cite{Ibe2005}, this is, how much the random variable is expected to deviate from its mean. The square root of the variance is called the \emph{standard deviation} \nm{\sigma_X}.
\neweq{Var\lrp{X} = \sigma_X^2 = E\lrsb{\lrp{X - \mu_X}^2} = \begin{dcases*}
\nm{\sum_k \lrp{x_k - \mu_X}^2 \, p_X\lrp{x_k}} & when \nm{X} is discrete \\
\nm{\int_{- \infty}^{\infty} \lrp{y - \mu_X}^2 \, f_X\lrp{y} \, dy} & when \nm{X} is continuous 
\end{dcases*}} {eq:Error_rvar_Variance}

The variance and mean of a random variable are related by the following expression, where \nm{E\lrsb{X^2}} is the second moment of \nm{X}.
\neweq{\sigma_X^2 = E\lrsb{\lrp{X - \mu_X}^2} = E\lrsb{X^2} - 2 \, \mu_X \, E\lrsb{X} + \mu_X^2 = E\lrsb{X^2} - \mu_X^2 = \mu_{\small{X}\ds{^2}} - \mu_X^2}{eq:Error_rvar_VarianceMean}

The notation \nm{X \sim \lrp{\mu_X, \ \sigma_X^2}} means that the random variable X has \nm{\mu_X} mean and \nm{\sigma_X^2} variance. A \emph{normal} or \emph{Gaussian} random variable X of parameters \nm{\mu_X} and \nm{\sigma_X^2} \cite{Farrell2008}, represented as \nm{X \sim N\lrp{\mu_X, \ \sigma_X^2}}, is one whose \hypertt{PDF} responds to:
\neweq{f_X\lrp{x} = \dfrac{1}{\sqrt{2 \, \pi \, \sigma_X^2}} \, exp\lrp{- \, \dfrac{\lrp{x - \mu_X}^2}{2 \; \sigma_X^2}} \ \ \ \ \ \ \ -\infty < x < \infty}{eq:Error_Normal}

The expected value and variance of a normal random variable \nm{N\lrp{\mu_X, \ \sigma_X^2}} are \nm{\mu_X} and \nm{\sigma_X^2}, respectively. A normal random variable \nm{N\lrp{0, \, 1}} of zero mean and unit variance is called a \emph{standard normal random variable}. It is worth noting that any affine\footnote{In this contest affine means a function of the form \nm{y = a \, x + b}, while linear means \nm{y = a \, x}.} function of a Gaussian random variable results in a Gaussian random variable \cite{Simon2006}.

The \emph{discrete uniform distribution} assigns the same probability to each of its N possible values \cite{Ibe2005}. Represented by \nm{X \sim U\lrp{a, \ a + N - 1}}, its expected value is \nm{a + (N - 1) / 2} and its variance is \nm{(N^2 - 1) / 12}. Its \hypertt{PMF} responds to:
\neweq{p_X\lrp{x_k} = \begin{dcases*}
\nm{\frac{1}{N}} & \nm{x_k = a, \ a+1, \ldots, \ a+N-1}\\
\nm{0}           & otherwise 
\end{dcases*}} {eq:Error_uniform}

Consider now two random variables \nm{X} and \nm{Y} defined in the same sample space \nm{\Omega} with expected values \nm{\mu_X} and \nm{\mu_Y}, respectively, and variances \nm{\sigma_X^2} and \nm{\sigma_Y^2}. They are called \emph{independent} if their results do not depend on each other:
\neweq{F_{XY}\lrp{x, \, y} = P\lrsb{X \leq x, \, Y \leq y} = P\lrsb{X \leq x} \, P\lrsb{Y \leq y} = F_X\lrp{x} \, F_Y\lrp{y}}{eq:Error_rvar_indep}

The \emph{central limit theorem} states that the sum of independent random variables tends towards a Gaussian random variable, regardless of the \hypertt{CDF} of the individual random variables that contribute to the sum \cite{Ibe2005,Simon2006}. If a given random variable \nm{X} is realized many times, the \emph{law of large numbers} states that the average of the realizations is close to the random variable expected value \nm{\mu_X}, and tends to it as the numbers of realizations grows \cite{Ibe2005}.

\emph{Stochastic simulations}, also known as \emph{Monte Carlo simulations}, use randomness to solve complex problems that may have a deterministic nature \cite{Ripley1987, Sawilowsky2003} and that are based on multiple unknown parameters, many of which are difficult to obtain experimentally \cite{Shojaeefard2003}. They rely on defining the domain of possible inputs, randomly generating inputs from a probability distribution over the domain, performing a deterministic computation based on those inputs, and finally aggregating the results by means of a set of metrics.

The \emph{correlation} \nm{R_{XY}} and the \emph{covariance} \nm{C_{XY}} of the random variables X and Y are two measures of the linear correlation between both random variables \cite{Ibe2005}. They are defined as:
\begin{eqnarray}
\nm{R\lrp{X, \, Y} = R_{XY}} & = & \nm{E\lrsb{X \cdot Y} = \mu_{X \cdot Y}}\label{eq:Error_rvar_Correlation} \\
\nm{C\lrp{X, \, Y} = C_{XY}} & = & \nm{E\lrsb{\lrp{X - \mu_X} \cdot \lrp{Y - \mu_Y}} = E\lrsb{X \cdot Y} - \mu_X \, \mu_Y = R_{XY} - \mu_X \, \mu_Y}\label{eq:Error_rvar_Covariance}
\end{eqnarray}

Two random variables are \emph{uncorrelated} if their covariance is zero \nm{\lrp{C_{XY} = 0 \rightarrow R_{XY} = \mu_X \, \mu_Y}}. Independent random variables are always uncorrelated, but not the other way around, as two uncorrelated random variables may not necessarily be independent if there exists a nonlinear dependence between them. \emph{Orthogonal} random variables are those whose correlation is zero \nm{\lrp{R_{XY} = 0}}, so they may or may not also be uncorrelated. If they are, at least one of them is zero mean.

The expected value and variance of the sum and product of two random variables are of particular interest. Given two random variables \nm{X} and \nm{Y} with expected values \nm{\lrb{\mu_X, \mu_Y}} and variances \nm{\lrb{\sigma_X^2, \sigma_Y^2}}, its sum \nm{X + Y} and product \nm{X \cdot Y} are also random variables, as indicated above. The following expressions can be easily obtained by applying the equations above\footnote{These expressions can be further simplified when \nm{X} and \nm{Y} are uncorrelated.} \cite{Frishman1971}:
\begin{eqnarray}
\nm{\mu_{X+Y}} & = & \nm{E\lrsb{X + Y} = \mu_X + \mu_Y}\label{eq:Error_sum_mean} \\
\nm{\sigma_{X+Y}^2} & = & \nm{E\lrsb{\lrp{X + Y}^2} - \mu_{X+Y}^2 = \sigma_X^2 + 2 \, C_{XY} + \sigma_Y^2}\label{eq:Error_sum_variance} \\
\nm{\mu_{X \cdot Y}} & = & \nm{E\lrsb{X \cdot Y} = R_{XY} = C_{XY} + \mu_X \, \mu_Y}\label{eq:Error_product_mean} \\
\nm{\sigma_{X \cdot Y}^2} & = & \nm{E\lrsb{\lrp{X \cdot Y}^2} - \mu_{X \cdot Y}^2 = C_{{\small{X}}{\ds{^2}}{\small{Y}}{\ds{^2}}} - C_{XY}^2 - 2 \, C_{XY} \, \mu_X \, \mu_Y + \sigma_X^2 \, \sigma_Y^2 + \sigma_X^2 \, \mu_Y^2 + \mu_X^2 \, \sigma_Y^2} \label{eq:Error_product_variance}
\end{eqnarray}

%%%%%%%%%%%%%%%%%%%%%%%%%%%%%%%%%%%%%%%%%%%%%%%%%%%%%%%%%%%%%%%%%%%%%%%%
%%%%%%%%%%%%%%%%%%%%%%%%%%%%%%%%%%%%%%%%%%%%%%%%%%%%%%%%%%%%%%%%%%%%%%%%
% SUBSECTION		RANDOM VECTORS
%%%%%%%%%%%%%%%%%%%%%%%%%%%%%%%%%%%%%%%%%%%%%%%%%%%%%%%%%%%%%%%%%%%%%%%%
%%%%%%%%%%%%%%%%%%%%%%%%%%%%%%%%%%%%%%%%%%%%%%%%%%%%%%%%%%%%%%%%%%%%%%%%

\subsection{Random Vectors}\label{subsec:Error_RandomVectors}

A \emph{random vector} \nm{\vec X = \lrsb{X_1 \ \dotsc \ X_n}^T} is a collection of random variables obtained from the same sample space \nm{\Omega} \cite{Ibe2005,Papoulis2002}. The random vector joint \hypertt{CDF}, \hypertt{PMF}, and \hypertt{PDF} are defined as follows:
\begin{eqnarray}
\nm{F_{X_1,\dotsc,X_n}\lrp{x_1,\dotsc,x_n}} & = & \nm{P\lrsb{\{X_1 \leq x_1\} \cap \ldots \cap \{X_n \leq x_n\}} = P\lrsb{X_1 \leq x_1,\dotsc,X_n \leq x_n}} \label{eq:Error_rvec_CDF} \\
\nm{F_{X_1,\dotsc,X_n}\lrp{x_1,\dotsc,x_n}} & = & \nm{\sum_{k_1 \leq x_1} \dots \sum_{k_n \leq x_n} p_{X_1,\dotsc,X_n}\lrp{k_1,\dotsc,k_n}} \label{eq:Error_rvec_PMF2_Joint} \\
\nm{F_{X_1,\dotsc,X_n}\lrp{x_1,\dotsc,x_n}} & = & \nm{\int_{- \infty}^{x_1} \dots \int_{- \infty}^{x_n} f_{X_1,\dotsc,X_n}\lrp{y_1,\dotsc,y_n}\, dy_n \ldots dy_1} \label{eq:Error_rvec_PDF_Joint}
\end{eqnarray}

When the components \nm{X_1,\dotsc,X_n} of the random vector \nm{\vec X} are independent from each other, its joint \hypertt{CDF}, \hypertt{PMF}, and \hypertt{PDF} are just the product of the respective functions of each of the random vector components \cite{Farrell2008}. The expected value \nm{E\lrsb{\vec X}} or mean \nm{\vec \mu_X} and the variance \nm{\vec \sigma_X^2} of a random vector \nm{\vec X} are defined as the vectors of those of its components:
\begin{eqnarray}
\nm{E\lrsb{\vec X}} & = & \nm{\vec \mu_X = \lrsb{\mu_{X1} \ \dotsc \ \mu_{Xn}}^T} \label{eq:Error_rvec_mean} \\
\nm{Var\lrp{\vec X}} & = & \nm{\vec \sigma_X^2 = \lrsb{\sigma_{X1}^2 \ \dotsc \ \sigma_{Xn}^2}^T} \label{eq:Error_rvec_var} 
\end{eqnarray}

Given two random vectors \nm{\vec X \in \mathbb{R}^m} and \nm{\vec Y \in \mathbb{R}^n}, their \emph{correlation matrix} \nm{\vec R_{XY}} is defined so \nm{R_{ij} = R\lrp{X_i, \, Y_j}}, while their \emph{covariance matrix} \nm{\vec C_{XY}} verifies that \nm{C_{ij} = C\lrp{X_i, \, Y_j}} \cite{Farrell2008}:
\begin{eqnarray}
\nm{\vec R\lrp{\vec X, \, \vec Y} = \vec R_{XY}} & = & \nm{E\lrsb{\vec X \, \vec Y^T}}\label{eq:Error_rvec_CorrelationMatrix} \\
\nm{\vec C\lrp{\vec X, \, \vec Y} = \vec C_{XY}} & = & \nm{ E\lrsb{\lrp{\vec X - \vec \mu_X}\lrp{\vec Y - \vec \mu_Y}^T} = E\lrsb{\vec X \, \vec Y^T} - \vec \mu_X \, \vec \mu_Y^T = \vec R_{XY} - \vec \mu_X \, \vec \mu_Y^T}\label{eq:Error_rvec_CovarianceMatrix}
\end{eqnarray}

The \emph{autocorrelation} and \emph{autocovariance} matrices \nm{\vec R_{XX}} and \nm{\vec C_{XX}} of a random vector \nm{\vec X \in \mathbb{R}^m} are defined as the correlation and covariance matrices of that vector with itself. Both are square, symmetric \nm{\lrp{\vec R_{XX} = \vec R_{XX}^T, \, \vec C_{XX} = \vec C_{XX}^T}}, positive semidefinite \nm{\lrp{\vec z^T \, \vec R_{XX} \, \vec z \geq 0, \, \vec z^T \, \vec C_{XX} \, \vec z \geq 0, \forall \, \vec z \in \mathbb{R}^m}}, and their diagonals contain the second moments and variances of each of the random variables \nm{X_i} within \nm{\vec X}:
\begin{eqnarray}
\nm{\vec R\lrp{\vec X} = \vec R_{XX}} & = & \nm{E\lrsb{\vec X \, \vec X^T}}\label{eq:Error_rvec_autoCorrelationMatrix} \\
\nm{\vec C\lrp{\vec X} = \vec C_{XX}} & = & \nm{E\lrsb{\lrp{\vec X - \vec \mu_X}\lrp{\vec X - \vec \mu_X}^T} = E\lrsb{\vec X \, \vec X^T} - \vec \mu_X \, \vec \mu_X^T = \vec R_{XX} - \vec \mu_X \, \vec \mu_X^T}\label{eq:Error_rvec_autoCovarianceMatrix}
\end{eqnarray}

A normal or Gaussian random vector is that whose components are all normal random variables. As in the case of scalar random variables, an affine transformation of a Gaussian random vector results in a new Gaussian random vector.

%%%%%%%%%%%%%%%%%%%%%%%%%%%%%%%%%%%%%%%%%%%%%%%%%%%%%%%%%%%%%%%%%%%%%%%%
%%%%%%%%%%%%%%%%%%%%%%%%%%%%%%%%%%%%%%%%%%%%%%%%%%%%%%%%%%%%%%%%%%%%%%%%
% SUBSECTION		STOCHASTIC PROCESSES
%%%%%%%%%%%%%%%%%%%%%%%%%%%%%%%%%%%%%%%%%%%%%%%%%%%%%%%%%%%%%%%%%%%%%%%%
%%%%%%%%%%%%%%%%%%%%%%%%%%%%%%%%%%%%%%%%%%%%%%%%%%%%%%%%%%%%%%%%%%%%%%%%

\subsection{Stochastic Processes}\label{subsec:Error_StochasticProcesses}

A \emph{random process} or \emph{stochastic process} enlarges the concept of random vector (or random variable when the vector size is one) to include time. Given a sample vector \nm{\vec \omega} of the sample space \nm{\Omega} and a parameter \emph{t} belonging to a parameter set \nm{\mathbb{T}} (generally time), a stochastic process assigns a real vector \nm{\{\vec X : \mathbb{T}, \Omega \rightarrow \mathbb{R}^m \ | \ t \in \mathbb{T}, \, \vec \omega \in \Omega \rightarrow \vec X\lrp{t, \, \vec \omega} \in \mathbb{R}^m\}} \cite{Ibe2005,Papoulis2002,Hoel1972}. If the sample vector \nm{\vec \omega} is fixed, the random process \nm{\vec X\lrp{t}} behaves as a function of time; on the other hand, if the time is fixed, the stochastic process \nm{\vec X\lrp{\vec \omega}} defaults to a random vector. A stochastic process is thus a family of random vectors (either discrete or continuous) indexed by a continuous parameter \nm{t \in \mathbb{T}}. If the parameter is discrete \nm{t \in \mathbb{Z}^+} \footnote{\nm{\mathbb{Z}^+} represents the set of positive integers.}, then the appropriate name is \emph{stochastic sequence} \cite{Farrell2008, Simon2006}.

Size one stochastic processes \nm{X\lrp{t, \, \omega}}, generally represented just by \nm{X\lrp{t}}, are completely described by their \hypertt{CDF}, while if \nm{\vec X\lrp{t}} is instead a random vector, it is represented by its joint \hypertt{CDF}: 
\begin{eqnarray}
\nm{F_X\lrp{x; \, t}} & = & \nm{P\lrsb{X\lrp{t} \leq x}}\label{eq:Error_rpro_CDF_single} \\
\nm{F_X\lrp{x_1,\dotsc,x_n; \, t}} & = & \nm{P\lrsb{X_1\lrp{t} \leq x_1,\dotsc,X_n\lrp{t} \leq x_n}}\label {eq:Error_rpro_CDF} 
\end{eqnarray}

The joint \hypertt{PMF} and \hypertt{PDF} are also defined in similar fashion. The \emph{ensemble average} or mean of a random process becomes a function of time: 
\neweq{\vec \mu_X\lrp{t} = E\lrsb{\vec X\lrp{t}}} {eq:Error_rpro_Mean}

Note that the random process \nm{\vec X\lrp{t}} evaluated at different times comprises different random vectors of the same size. It is then possible to apply the concepts of autocorrelation and autocovariance introduced in section \ref{subsec:Error_RandomVectors} to any two of these random vectors, providing quantitative measures of the similarity of the random process at two different times, this is, measuring by how much a signal is similar to its time shifted version \cite{Ibe2005}. This results in the \emph{autocorrelation} \nm{\vec R_{XX}\lrp{t,\, t + \tau}} and the \emph{autocovariance} \nm{\vec C_{XX}\lrp{t,\, t + \tau}}:
\begin{eqnarray}
\nm{\vec R_{XX}\lrp{t, \, t + \tau}} & = & \nm{E\lrsb{\vec X\lrp{t} \, \vec X^T\lrp{t + \tau}}}\label{eq:Error_rpro_Autocorrelation} \\
\nm{\vec C_{XX}\lrp{t, \, t + \tau}} & = & \nm{E\lrsb{\big(\vec X\lrp{t} - \vec \mu_X\lrp{t}\big)\big(\vec X\lrp{t + \tau} - \vec \mu_X\lrp{t + \tau}\big)^T}} \nonumber \\
& = & \nm{E\lrsb{\vec X\lrp{t} \, \vec X^T\lrp{t + \tau}} - \vec \mu_X\lrp{t} \, \vec \mu_X^T\lrp{t + \tau} = \vec R_{XX}\lrp{t, \, t + \tau} - \vec \mu_X\lrp{t} \, \vec \mu_X^T\lrp{t + \tau}}\label{eq:Error_rpro_Autocovariance}
\end{eqnarray}

The autocovariance is zero when the two observations of \nm{\vec X} are independent, meaning that there is no coupling between \nm{\vec X\lrp{t}} and \nm{\vec X\lrp{t + \tau}}, and they are called uncorrelated. As with the random vectors, the reverse is not true, as two uncorrelated observations does not necessarily mean that they are independent.

A \emph{wide sense stationary process} is that in which the mean does not vary with time and the autocorrelation depends exclusively on the time difference\footnote{Strict sense stationary processes are those in which the complete \hypertt{CDF} is time invariant, not only its mean and autocorrelation. The definition is usually too restrictive for any practical use.}:
\begin{eqnarray}
\nm{\vec \mu_{X\sss{WSS}}\lrp{t}} & = & \nm{E\lrsb{\vec X_{\sss {WSS}}\lrp{t}} = \vec \mu_{X\sss{WSS}}}\label{eq:Error_rpro_Mean_Stationary} \\
\nm{\vec R_{XX\sss{WSS}}\lrp{t, \, t + \tau}} & = & \nm{E\lrsb{\vec X_{\sss{WSS}}\lrp{t} \, \vec X_{\sss{WSS}}^T\lrp{t + \tau}} = \vec R_{XX\sss{WSS}}\lrp{\tau}}\label{eq:Error_rpro_Autocorrelation_Stationary} 
\end{eqnarray}
 
Consider now two stochastic processes \nm{\vec X\lrp{t}} and \nm{\vec Y\lrp{t}} defined in the same sample space \nm{\Omega}. The \emph{crosscorrelation} \nm{\vec R_{XY}\lrp{t,\, t + \tau}} and \emph{crosscovariance} \nm{\vec C_{XY}\lrp{t,\, t + \tau}} measure how similar two different processes (or signals) are when one is time shifted with respect to the other \cite{Ibe2005}:
\begin{eqnarray}
\nm{\vec R_{XY}\lrp{t, \, t + \tau}} & = & \nm{E\lrsb{\vec X\lrp{t} \, \vec Y^T\lrp{t + \tau}}} \label{eq:Error_rpro_Crosscorrelation} \\
\nm{\vec C_{XY}\lrp{t, \, t + \tau}} & = & \nm{E\lrsb{\big(\vec X\lrp{t} - \vec \mu_X\lrp{t}\big)\big(\vec Y\lrp{t + \tau} - \vec \mu_Y\lrp{t + \tau}\big)^T}} \nonumber \\
& = & \nm{E\lrsb{\vec X\lrp{t} \, \vec Y^T\lrp{t + \tau}} - \vec \mu_X\lrp{t} \, \vec \mu_Y^T\lrp{t + \tau} = \vec R_{XY}\lrp{t, \, t + \tau} - \vec \mu_X\lrp{t} \, \vec \mu_Y^T\lrp{t + \tau}}\label{eq:Error_rpro_Crosscovariance}
\end{eqnarray}

Two processes \nm{\vec X\lrp{t}} and \nm{\vec Y\lrp{t}} are orthogonal if their crosscorrelation is zero for all t and \nm{t + \tau}, while they are uncorrelated if their crosscovariance is zero. They are jointly wide sense stationary if their crosscorrelation is independent of the absolute time:
\neweq{\vec R_{XY\sss{WSS}}\lrp{t, \, t + \tau} = \vec R_{XY_{\sss{WSS}}}\lrp{\tau}} {eq:Error_rpro_Crosscorrelation_Stationary}

Consider also a stochastic process \nm{\vec X\lrp{t}} that has one realization \nm{\vec x\lrp{t}}. It is then possible to define the \emph{time average} \nm{A\big[\vec X\lrp{t}\big]} and the \emph{time autocorrelation} \nm{R\big[\vec X\lrp{t}, \, \tau\big]} for continuous processes as:
\begin{eqnarray}
\nm{A\big[\vec X\lrp{t}\big]} & = & \nm{\lim\limits_{T \to \infty} \frac{1}{2\, T} \int_{- T}^{T} \vec x\lrp{t} \, \mathrm{d}t} \label{eq:Error_rpro_TimeAverage} \\
\nm{R\big[\vec X\lrp{t}, \, \tau\big]} & = & \nm{A\lrsb{\vec X\lrp{t} \, \vec X^T\lrp{t + \tau}}}\label{eq:Error_rpro_TimaAutoCorrelation}
\end{eqnarray}

The discrete time definitions can be derived accordingly. Finally, an \emph{ergodic} process \cite{Simon2006} is a stationary random process for which
\begin{eqnarray}
\nm{A_{\sss{ERG}}\big[\vec X\lrp{t}\big]} & = & \nm{E_{\sss{ERG}}\lrsb{\vec X}} \label{eq:Error_rpro_Ergodic1} \\
\nm{R_{\sss{ERG}}\big[\vec X\lrp{t}, \, \tau\big]} & = & \nm{\vec R_{XX\sss{ERG}}\lrp{\tau}}\label{eq:Error_rpro_Ergodic2}
\end{eqnarray}

%%%%%%%%%%%%%%%%%%%%%%%%%%%%%%%%%%%%%%%%%%%%%%%%%%%%%%%%%%%%%%%%%%%%%%%%
%%%%%%%%%%%%%%%%%%%%%%%%%%%%%%%%%%%%%%%%%%%%%%%%%%%%%%%%%%%%%%%%%%%%%%%%
% SUBSECTION		WHITE NOISE
%%%%%%%%%%%%%%%%%%%%%%%%%%%%%%%%%%%%%%%%%%%%%%%%%%%%%%%%%%%%%%%%%%%%%%%%
%%%%%%%%%%%%%%%%%%%%%%%%%%%%%%%%%%%%%%%%%%%%%%%%%%%%%%%%%%%%%%%%%%%%%%%%

\subsection{White Noise}\label{subsec:Error_WhiteNoise}

If any two random vectors \nm{\vec X\lrp{t_1}} and \nm{\vec X\lrp{t_2}} taken from a stochastic process \nm{\vec X\lrp{t}} are independent for all \nm{t_1 \neq t_2}, then the random process \nm{\vec X\lrp{t}} is called \emph{white noise}. Otherwise, it is known as \emph{colored noise} \cite{Simon2006}. 

The whiteness or color content of a stochastic process can be characterized by its \emph{power spectrum} or \emph{power spectral density} (\hypertt{PSD}) \nm{S_{XX}\lrp{\omega}}. For wide sense stationary processes, it is defined as the Fourier transform of its autocorrelation function \nm{R_{XX}\lrp{\tau}} \cite{Simon2006}:
\neweq{\vec S_{XX}\lrp{\omega} = \begin{dcases*}
\nm{\sum_{k = - \infty}^{\infty} \vec R_{XX}\lrp{k} \, exp\lrp{- i \, \omega \, k} \ \ \ \omega \in \lrsb{- \pi, \, \pi}} & when \nm{X} is discrete \\
\nm{\int_{- \infty}^{\infty} \vec R_{XX}\lrp{\tau} \, exp\lrp{- i \, \omega \, \tau} \, \mathrm{d}\tau} & when \nm{X} is continuous 
\end{dcases*}} {eq:Error_whitenoise_Fourier}

The autocorrelation can be recovered by means of the inverse Fourier transform:
\neweq{\begin{dcases*}
\nm{\vec R_{XX}\lrp{k} = \dfrac{1}{2\pi}\int_{- \infty}^{\infty} \vec S_{XX}\lrp{\omega} \, exp\lrp{i \, \omega \, k} \, \mathrm{d}\omega} & when \nm{\vec X\lrp{t}} is discrete \\
\nm{\vec R_{XX}\lrp{\tau} = \dfrac{1}{2\pi}\int_{- \infty}^{\infty} \vec S_{XX}\lrp{\omega} \, exp\lrp{i \, \omega \, \tau} \, \mathrm{d}\omega} & when \nm{\vec X\lrp{t}} is continuous 
\end{dcases*}} {eq:Error_whitenoise_Fourier_inverse}

In case of continuous time wide sense stationary stochastic processes\footnote{Similar expressions can be easily obtained for discrete time processes.}, the \emph{power} is defined as:
\neweq{\vec P_{XX} = \dfrac{1}{2\pi}\int_{- \infty}^{\infty} \vec S_{XX}\lrp{\omega} \, \mathrm{d}\omega}{eq:Error_whitenoise_Fourier_power}

In the case of two continuous time jointly wide sense stationary stochastic processes \nm{\vec X\lrp{t}} and \nm{\vec Y\lrp{t}}, the \emph{cross power spectrum} \nm{S_{XY}\lrp{\omega}} is defined as the Fourier transform of their crosscorrelation \nm{R_{XY}\lrp{\tau}} \cite{Simon2006}:
\begin{eqnarray}
\nm{\vec S_{XY}\lrp{\omega}} & = & \nm{\int_{- \infty}^{\infty} \vec R_{XY}\lrp{\tau} \, exp\lrp{- i \, \omega \, \tau} \, \mathrm{d}\tau}\label{eq:Error_whitenoise_Fourier_cross} \\
\nm{\vec R_{XY}\lrp{\tau}} & = & \nm{\dfrac{1}{2\pi}\int_{- \infty}^{\infty} \vec S_{XY}\lrp{\omega} \, exp\lrp{i \, \omega \, \tau} \, \mathrm{d}\omega}\label{eq:Error_whitenoise_Fourier_inverse_cross}
\end{eqnarray}

A white noise process \nm{\vec N\lrp{t}} (continuous time) or \nm{\vec N\lrp{k}} (discrete time) is one whose \hypertt{PSD} is constant for all frequencies, this is, a random process having equal power at all frequencies \cite{Farrell2008}. These processes do not have any correlation with themselves except at the present time \cite{Simon2006}. The definition for discrete time processes relies on the \emph{Kronecker delta function} \nm{\delta_k}\footnote{The Kronecker delta function \nm{\delta\lrp{k}} is valued 0 for all \emph{k} except at \nm{k = 0}, where it is \nm{1}.}:
\begin{eqnarray}
\nm{\vec R_{NN}\lrp{k}}      & = & \nm{\vec \sigma^2 \, \delta_k} \label{eq:Error_whitenoise_DiscreteWhiteNoise2} \\
\nm{\vec S_{NN}\lrp{\omega}} & = & \nm{\vec \sigma^2 = \vec R_{NN}\lrp{0} \ \ \ \forall \, \omega \in \lrsb{- \pi, \, \pi}  } \label{eq:Error_whitenoise_DiscreteWhiteNoise} 
\end{eqnarray}

while that for continuous time processes makes use of the \emph{impulse Dirac delta function} \nm{\delta\lrp{\tau}}\footnote{The Dirac delta function \nm{\delta\lrp{\tau}} is valued 0 everywhere except at \nm{\tau = 0}, where it is \nm{\infty}. Its integral over any space containing \nm{\tau = 0} is 1.}:
\begin{eqnarray}
\nm{\vec R_{NN}\lrp{\tau}}   & = & \nm{\vec \sigma^2 \, \delta\lrp{\tau}} \label{eq:Error_whitenoise_ContinuousWhiteNoise2} \\
\nm{\vec S_{NN}\lrp{\omega}} & = & \nm{\vec \sigma^2 = \vec R_{NN}\lrp{0} \ \forall \omega \in \mathbb{R}} \label{eq:Error_whitenoise_ContinuousWhiteNoise}
\end{eqnarray}

%%%%%%%%%%%%%%%%%%%%%%%%%%%%%%%%%%%%%%%%%%%%%%%%%%%%%%%%%%%%%%%%%%%%%%%%
%%%%%%%%%%%%%%%%%%%%%%%%%%%%%%%%%%%%%%%%%%%%%%%%%%%%%%%%%%%%%%%%%%%%%%%%
% SUBSECTION		WHITE NOISE GAUSSIAN PROCESSES AND THEIR INTEGRATION
%%%%%%%%%%%%%%%%%%%%%%%%%%%%%%%%%%%%%%%%%%%%%%%%%%%%%%%%%%%%%%%%%%%%%%%%
%%%%%%%%%%%%%%%%%%%%%%%%%%%%%%%%%%%%%%%%%%%%%%%%%%%%%%%%%%%%%%%%%%%%%%%%

\subsection{White Noise Gaussian Processes and their Integration}\label{subsec:Error_WNGauss}

A \emph{Gaussian process} is a stochastic process \nm{\{X\lrp{t}, t \in \mathbb{T}\}} in which for any choice of \emph{n} real coefficients \nm{a_1,\dotsc,a_n} and choice of \emph{k} time instants \nm{t_1,\dotsc,t_k} in the time set \nm{\mathbb{T}}, the random variable \nm{a_1 \, X\lrp{t_1} + \ldots + a_k \, X\lrp{t_k}} is a normal random variable. It can be proven that the random variables \nm{X\lrp{t_1},\dotsc,X\lrp{t_k}} are mutually uncorrelated and independent. A Gaussian process hence associates the random variable occurring at any time instant with a normal distribution, and is fully defined by its expected value and covariance function \cite{Ibe2005}, so strict and wide sense stationary processes coincide in the case of Gaussian processes.

Considering now\footnote{It is necessary to point out that many of the expressions obtained in this section are the result of long but not necessarily complex mathematical operations relying on the repeated application of the expressions derived in section \ref{sec:Error_Random}. The interested reader should not have many problems in obtaining the same results as shown here.} that \nm{\mu\lrp{t}} is a zero mean white noise Gaussian process with \hypertt{PSD} \nm{\sigma^2}:
\neweq{E\lrsb{\mu\lrp{t}} = 0 \ \ \ \ \ \ \ E\lrsb{\mu\lrp{t} \, \mu\lrp{\tau}} = \sigma^2 \, \delta\lrp{t - \tau}}{eq:Error_WNGauss_white_noise}

and also consider a standard normal random variable \nm{N\lrp{0, \, 1}} so that \nm{\sigma \, N} is identically distributed to \nm{\mu\lrp{t}}:
\neweq{E\lrsb{\sigma \, N} = 0 \ \ \ \ \ \ \ Var\lrp{\sigma \, N} = E\lrsb{\sigma^2 \, N^2} = \sigma^2}{eq:Error_WNGauss_normal}

The integration of a white noise Gaussian process is known as a \first \emph{order random walk} \cite{Rogers2007}, characterized by variances that grow linearly with time as well as \hypertt{PSD}s that fall off as the inverse of the square of the sampling frequency \cite{Grewal2010}. \nm{\mu\lrp{t}} can then be integrated over a timespan \nm{k \, \Deltat} by means of the rectangular rule, and its expected value and variance evaluated:
\begin{eqnarray}
\nm{a\lrp{t}} & = & \nm{a\lrp{k \, \Deltat} = \int_0^{k \, \Deltat} \mu\lrp{\tau} \, d\tau = \Deltat \, \sigma \sum_{i=1}^k N_i} \label{eq:Error_WNGauss_1st} \\
\nm{E\lrsb{a\lrp{t}}} & = & \nm{\sigma \, k \, \Deltat \, E\lrsb{N} = 0} \label{eq:Error_WNGauss_1st_mean} \\
\nm{Var\big(a\lrp{t}\big)} & = & \nm{E\Big[\big(a\lrp{t} - E\lrsb{a\lrp{t}}\big)^2\Big] = \sigma^2 \, \Deltat^2 \sum_{i=1}^k E\lrsb{N_i^2} = \sigma^2 \, \Deltat^2 \, k = \sigma^2 \, t \, \Deltat} \label{eq:Error_WNGauss_1st_variance} 
\end{eqnarray}

The second integration of a white noise Gaussian process is known as a \second \emph{order random walk}:
\begin{eqnarray}
\nm{b\lrp{t}} & = & \nm{b\lrp{k \, \Deltat} = \int_0^{k \, \Deltat} a\lrp{\tau} \, d\tau = \int_0^t \int_0^\upsilon \mu\lrp{\tau} \, d\tau \, d\upsilon} \nonumber \\
 & = & \nm{\Deltat \sum_{i=1}^k \Deltat \, \sigma \sum_{j=1}^i N_j = \Deltat^2 \, \sigma \sum_{i=1}^k \lrp{k - i + 1} \, N_i} \label{eq:Error_WNGauss_2nd} \\
\nm{E\lrsb{b\lrp{t}}} & = & \nm{\Deltat^2 \, \sigma \sum_{i=1}^k \lrp{k - i + 1} \, E\lrsb{N_i} = 0} \label{eq:Error_WNGauss_2nd_mean} \\
\nm{Var\big(b\lrp{t}\big)} & = & \nm{E\Big[\big(b\lrp{t} - E\lrsb{b\lrp{t}}\big)^2\Big] = \Deltat^4 \, \sigma^2 \sum_{i=1}^k \lrp{k-i+1}^2 \, E\lrsb{N_i^2}} \nonumber \\
& = & \nm{\dfrac{\sigma^2}{6} \, \Deltat^4 \, k \, \lrp{k+1} \, \lrp{2 \, k + 1} \approx \dfrac{\sigma^2}{3} \, t^3 \, \Deltat}\label{eq:Error_WNGauss_2nd_variance} 
\end{eqnarray}

The third integration of a white noise Gaussian process is known as a \third \emph{order random walk}:
\begin{eqnarray}
\nm{c\lrp{t}} & = & \nm{c\lrp{k \, \Deltat} = \int_0^{k \, \Deltat} b\lrp{\tau} \, d\tau = \int_0^t \int_0^\upsilon \int_0^\kappa \mu\lrp{\tau} \, d\tau \, d\kappa \, d\upsilon} \nonumber \\
 & = & \nm{\Deltat \sum_{i=1}^k \Deltat \sum_{j=1}^i \Deltat \, \sigma \sum_{k=1}^j N_k = \Deltat^3 \, \sigma \sum_{i=1}^k \lrp{\sum_{j=1}^{k-i+1} j \, N_i}} \label{eq:Error_WNGauss_3rd} \\
\nm{E\lrsb{c\lrp{t}}} & = & \nm{\Deltat^3 \, \sigma \sum_{i=1}^k \sum_{j=1}^{k-i+1} j \, E\lrsb{N_i} = 0} \label{eq:Error_WNGauss_3rd_mean} \\
\nm{Var\big(c\lrp{t}\big)} & = & \nm{E\Big[\big(c\lrp{t} - E\lrsb{c\lrp{t}}\big)^2\Big] = \Deltat^6 \, \sigma^2 \sum_{i=1}^k \lrp{\sum_{j=1}^{k-i+1} j}^2 \, E\lrsb{N_i^2} \approx \dfrac{\sigma^2}{20} \, \Deltat^6 \, k^5 = \dfrac{\sigma^2}{20} \, t^5 \, \Deltat}\label{eq:Error_WNGauss_3rd_variance} 
\end{eqnarray}

Note that while the expected values or means of all random walks are zero, their variances grow with the \first, \third, and \fifth\ powers of time, respectively, so the expected deviations with respect to the zero means at any given time are proportional to \nm{t^{1/2}}, \nm{t^{3/2}}, and \nm{t^{5/2}} respectively.

%%%%%%%%%%%%%%%%%%%%%%%%%%%%%%%%%%%%%%%%%%%%%%%%%%%%%%%%%%%%%%%%%%%%%%%%
%%%%%%%%%%%%%%%%%%%%%%%%%%%%%%%%%%%%%%%%%%%%%%%%%%%%%%%%%%%%%%%%%%%%%%%%
%%%%%%%%%%%%%%%%%%%%%%%%%%%%%%%%%%%%%%%%%%%%%%%%%%%%%%%%%%%%%%%%%%%%%%%%
% SECTION		ROBUST STATISTICS and M-ESTIMATORS
%%%%%%%%%%%%%%%%%%%%%%%%%%%%%%%%%%%%%%%%%%%%%%%%%%%%%%%%%%%%%%%%%%%%%%%%
%%%%%%%%%%%%%%%%%%%%%%%%%%%%%%%%%%%%%%%%%%%%%%%%%%%%%%%%%%%%%%%%%%%%%%%%
%%%%%%%%%%%%%%%%%%%%%%%%%%%%%%%%%%%%%%%%%%%%%%%%%%%%%%%%%%%%%%%%%%%%%%%%

\section{Robust Statistics and M-Estimators} \label{sec:Error_Robust}

Classic statistical methods rely heavily on conditions often not met in practice, such as data errors being normally distributed, or that it is possible to apply the central limit theorem to obtain normally distributed estimates. As a result, they work poorly when there exist outliers in the data. \emph{Robust} statistical methods are those that emulate the classic ones but are not unduly affected by outliers or other small departures from the model assumptions, in particular the lack of normally distributed inputs. A robust statistic is resistant to errors in the results produced by deviations from the normality assumption.

A measure of \emph{central tendency} provides the central or typical value of a random variable or probability distribution, and can also be called the center, location, or average of the distribution. The classic measure of central tendency is the mean \nm{\mu_X} introduced in section \ref{subsec:Error_RandomVariables}, but the \emph{median} \nm{\eta_X}, defined as the middle value that separates the higher valued half from the lower half in the data set, provides better results if outliers are present\footnote{An additional measure of central tendency is the \emph{mode}, or most frequent value in the data set.}.

The dispersion, variability, scatter, or spread of a random variable or distribution measures the extent to which a distribution is stretched or squeezed, quantifying the statistical dispersion of the data, and is often used to estimate the scale, and hence these are known as \emph{measures of scale}. The classic measures such as the variance \nm{\sigma_X^2} and standard deviation \nm{\sigma_X} introduced in section \ref{subsec:Error_RandomVariables} are very sensitive to contaminated data. On the other hand, the \emph{median absolute deviation} \nm{MAD_X}, defined as the median of the absolute values of the differences between the data values and the overall median of the data set, is more robust and should replace the standard deviation if the normality of the input data can not be guaranteed\footnote{An additional measure of scale is the \emph{inter quartile range} (\hypertt{IQR}) or difference between the 75\% and 25\% percentile of a sample.}. To do so, it must be corrected with the following factor to ensure results coincide if applied to a normal distribution:
\neweq{\sigma_X \approx 1.4826 \ MAD_X = \dfrac{MAD_X}{0.6745}}{eq:Error_robust_mad}
\input{scripts/a4_estim/robust_estimators}

An statistical \emph{estimator} is a method to estimate the coefficients of a parametric model based on observations of a data set that comply with certain assumptions. \emph{Extremum estimators} of parametric models are those that can be calculated by minimization of a certain objective function \nm{Q_n} that depends on the \emph{n} samples:
\neweq{\hat{\vec \theta} = \argmin_{\ds{\vec \theta \in \vec \Theta}} Q_n\lrp{\vec \theta}}{eq:Error_robust_extreme}

where \nm{\vec \Theta} is the parameter space. If the objective function is a sampled measure of central tendency, the estimator is known as an \emph{M-estimator}, and can be expressed as follows \cite{Huber1981}:
\neweq{\hat{\vec \theta} = \argmin_{\ds{\vec \theta \in \vec \Theta}} \lrp{\sum_{i=1}^n \rho\lrp{x_i, \vec \theta}}}{eq:Error_robust_m_estimator}

where \nm{\rho} is a function with certain properties \cite{Huber1981}. \emph{Least squares} (minimum of the sum of squares of the differences between the samples and a given value) and \emph{maximum likelihood} (\hypertt{ML}) estimators (derivative of likelihood function with respect to parameters is zero) are both special cases of M-estimators. The function \nm{\rho} and its derivative \nm{\Upsilon} can be chosen to provide the estimator with desirable properties when the data are truly from the assumed distribution, and to be robust when it comes from a model that is close but does not comply with the input assumptions. The minimization of \nm{\sum \rho\lrp{x_i, \vec \theta}} can always be done directly, but it is often simpler to derivate with respect to \emph{x} and solve for the root of the derivative. When this is possible (most practical cases), the M-estimator if of \nm{\Upsilon}-type; if not, \nm{\rho}-type.

Given a function \nm{\rho\lrp{x}}, known as the \emph{objective} or \emph{error function}, and its derivative with \emph{x} \nm{\Upsilon\lrp{x} = d\rho\lrp{x} / dx}, the \emph{weight function} \nm{\omega\lrp{x} = \Upsilon\lrp{x} / x} represents the relative weight given to each observation. When the observed samples can only be positive and do not need to be squared, the objective function is usually replaced by \nm{\varrho\lrp{x} = \rho\lrp{\sqrt{x}}}; its derivative \nm{\varrho^\prime\lrp{x} = d\varrho\lrp{x}/dx} coincides with the weight function once the scale (1/2) is removed and \emph{x} replaced by \nm{x^2}, but it represents the same concept.
\input{scripts/a4_estim/robust_estimators_bis}

The most common M-estimators \cite{Fox2013} are listed below, where the expressions consider with no loss of generality that the samples are errors and hence \nm{\rho\lrp{x}} should have a minimum at zero\footnote{If this is not the case, just replace \emph{x} by \nm{x - x_{MIN}}.}:
\begin{itemize} 

\item The \emph{absolute error} or \emph{median estimator} employs \nm{\rho_{\ds{\eta}}\lrp{x} = | x |}. Although not differentiable, methods that rely of its \nm{\Upsilon}-type can use \nm{\Upsilon_{\ds{\eta}}\lrp{x} = \sign{\lrp{x}}}. The weight \nm{w_{\ds{\eta}}\lrp{x} = 1 / | x |} of each sample is inversely proportional to its absolute value.

\item The \emph{squared error} or \emph{mean estimator} relies on \nm{\rho_{\ds{\mu}}\lrp{x} = x^2 / 2} or \nm{\Upsilon_{\ds{\mu}}\lrp{x} = x}, and it represents the least squares method employed to minimize the sum of squared differences. All data samples are assigned the same weight as \nm{w_{\ds{\mu}}\lrp{x} = 1}.

\item The \emph{Huber estimator} behaves as the mean (squared error) when the sample absolute value is inferior to a certain threshold \nm{\delta_{HUB}}, where all samples are weighted equally, and as a median estimator (absolute error) when superior, reducing the sample weight as its absolute value increases. As outliers are generally characterized by a big value \emph{x}, their relative weight with respect to the remaining samples is reduced.
\begin{eqnarray}
\nm{\rho_{HUB}\lrp{x}} & = & \begin{dcases*}
\nm{x^2 / \, 2}                                         & \nm{| x | \leq \delta_{HUB}} \\
\nm{\delta_{HUB} \, \big[| x | - \delta_{HUB} / 2\big]} & \nm{| x | > \delta_{HUB}} 
\end{dcases*} \label{eq:Error_robust_huber_rho} \\
\nm{w_{HUB}\lrp{x}} & = & \begin{dcases*}
\nm{1}                          & \nm{| x | \leq \delta_{HUB}} \\
\nm{\delta_{HUB} \, / \, | x |} & \nm{| x | > \delta_{HUB}} 
\end{dcases*} \label{eq:Error_robust_huber_w} 
\end{eqnarray}

\item The \emph{bisquare} or \emph{Tukey estimator} provides a much quicker weight reduction with absolute value than the Huber estimator, as the diminution starts at 0 and is complete at \nm{\delta_{TUK}}, above which the samples are neglected.
\begin{eqnarray}
\nm{\rho_{TUK}\lrp{x}} & = & \begin{dcases*}
\nm{\dfrac{x^2}{2} \lrsb{1 - \dfrac{x^2}{{\delta_{TUK}}^2} + \dfrac{x^4}{3 \, {\delta_{TUK}}^4}}} & \nm{| x | \leq \delta_{TUK}} \\
\nm{{\delta_{TUK}}^2 / \, 6}                                                                      & \nm{| x | > \delta_{TUK}} 
\end{dcases*} \label{eq:Error_robust_tukey_rho} \\
\nm{w_{TUK}\lrp{x}} & = & \begin{dcases*}
\nm{\lrsb{1 - x^2 / \, {\delta_{TUK}}^2}^2} & \nm{| x | \leq \delta_{TUK}} \\
\nm{0}                                      & \nm{| x | > \delta_{TUK}} 
\end{dcases*} \label{eq:Error_robust_tukey_w} 
\end{eqnarray}
\end{itemize}

Smaller values of the tuning constants \nm{\delta_{HUB}} and \nm{\delta_{TUK}} provide more resistance to outliers at the expense of lower efficiency when the errors are normally distributed; their values are usually set to \nm{\delta_{HUB} = 1.345 \, \sigma_X} and \nm{\delta_{TUK} = 4.685 \, \sigma_X} \cite{Fox2013}, where \nm{\sigma_X} represents the inputs standard deviation provided by (\ref{eq:Error_robust_mad}).

%%%%%%%%%%%%%%%%%%%%%%%%%%%%%%%%%%%%%%%%%%%%%%%%%%%%%%%%%%%%%%%%%%%%%%%%
%%%%%%%%%%%%%%%%%%%%%%%%%%%%%%%%%%%%%%%%%%%%%%%%%%%%%%%%%%%%%%%%%%%%%%%%
%%%%%%%%%%%%%%%%%%%%%%%%%%%%%%%%%%%%%%%%%%%%%%%%%%%%%%%%%%%%%%%%%%%%%%%%
% SECTION      DISCRETE INTEGRATION IN EUCLIDEAN SPACES
%%%%%%%%%%%%%%%%%%%%%%%%%%%%%%%%%%%%%%%%%%%%%%%%%%%%%%%%%%%%%%%%%%%%%%%%
%%%%%%%%%%%%%%%%%%%%%%%%%%%%%%%%%%%%%%%%%%%%%%%%%%%%%%%%%%%%%%%%%%%%%%%%
%%%%%%%%%%%%%%%%%%%%%%%%%%%%%%%%%%%%%%%%%%%%%%%%%%%%%%%%%%%%%%%%%%%%%%%%

\section{Discrete Integration in Euclidean Spaces}\label{sec:euclidean_integration}

Let \nm{\vec x\lrp{t} \in \mathbb{R}^m} be an Euclidean space time varying state vector for which its value at a given discrete time \nm{\vec x_k = \vec x\lrp{t_k}} is known. The objective is to determine the state vector value at a later time \nm{\vec x_{k+1} = \vec x\lrp{t_{k+1}} = \vec x\lrp{t_k + \Delta t}} by relying on evaluations of the state vector derivative with time:
\neweq{\xvecdot\lrp{t} = f\big(\xvec\lrp{t}, \, t\big)} {eq:algebra_integration_derivative}

The initial value first order \emph{ordinary differential equation} (\hypertt{ODE}) problem can be solved with varying degrees of complexity and accuracy \cite{Press2002}, three of which are described below:
\begin{itemize}

\item \emph{Euler's method} is a first order approach that relies on evaluating the time derivative at \nm{t_k} and considering that its value does not change for the duration of the integration interval \nm{\Delta t}. Its error is proportional to the square of the integration interval:
\neweq{\vec x_{k+1} \approx \vec x_k + \Delta t \ \xvecdot(\vec x_k, \, t_k)}{eq:algebra_integration_euler}

\item \emph{Heun's method} is a second order approach that requires two evaluations of the time derivative. The constant gradient is estimated as the average between the time derivative evaluation at the initial state and that at the result of Euler's method, and results in an error proportional to the cube of the integration interval:
\begin{eqnarray}
\nm{\vec v_1} & = & \nm{\xvecdot(\vec x_k, \, t_k)} \label{eq:algebra_integration_heun_one} \\
\nm{\vec v_2} & = & \nm{\xvecdot(\vec x_k + \Delta t \ \vec v_1, t_k + \Delta t)} \label{eq:algebra_integration_heun_two} \\
\nm{\vec x_{k+1}} & \nm{\approx} & \nm{\vec x_k + \dfrac{\Delta t}{2} \ \lrsb{\vec v_1 + \vec v_2}} \label{eq:algebra_integration_heun}
\end{eqnarray}

\item The \nm{\vec 4^{th}} \emph{order Runge-Kutta method} is the de facto standard and relies on four evaluations of the state vector time derivative to obtain an error proportional to the fifth power of the integration interval:
\begin{eqnarray}
\nm{\vec v_1} & = & \nm{\xvecdot(\vec x_k, \, t_k)} \label{eq:algebra_integration_rk4th_one} \\
\nm{\vec v_2} & = & \nm{\xvecdot(\vec x_k + \dfrac{\Delta t \ \vec v_1}{2}, t_k + \dfrac{\Delta t}{2})} \label{eq:algebra_integration_rk4th_two} \\
\nm{\vec v_3} & = & \nm{\xvecdot(\vec x_k + \dfrac{\Delta t \ \vec v_2}{2}, t_k + \dfrac{\Delta t}{2})} \label{eq:algebra_integration_rk4th_three} \\
\nm{\vec v_4} & = & \nm{\xvecdot(\vec x_k + \Delta t \ \vec v_3, t_k + \Delta t)} \label{eq:algebra_integration_rk4th_four} \\
\nm{\vec x_{k+1}} & \nm{\approx} & \nm{\vec x_k + \Delta t \ \lrsb{\dfrac{\vec v_1}{6} + \dfrac{\vec v_2}{3} + \dfrac{\vec v_3}{3} + \dfrac{\vec v_4}{6}}} \label{eq:algebra_integration_rk4th}
\end{eqnarray}
\end{itemize}

%%%%%%%%%%%%%%%%%%%%%%%%%%%%%%%%%%%%%%%%%%%%%%%%%%%%%%%%%%%%%%%%%%%%%%%%
%%%%%%%%%%%%%%%%%%%%%%%%%%%%%%%%%%%%%%%%%%%%%%%%%%%%%%%%%%%%%%%%%%%%%%%%
%%%%%%%%%%%%%%%%%%%%%%%%%%%%%%%%%%%%%%%%%%%%%%%%%%%%%%%%%%%%%%%%%%%%%%%%
% SECTION      GRADIENT DESCENT OPTIMIZATION IN EUCLIDEAN SPACES
%%%%%%%%%%%%%%%%%%%%%%%%%%%%%%%%%%%%%%%%%%%%%%%%%%%%%%%%%%%%%%%%%%%%%%%%
%%%%%%%%%%%%%%%%%%%%%%%%%%%%%%%%%%%%%%%%%%%%%%%%%%%%%%%%%%%%%%%%%%%%%%%%
%%%%%%%%%%%%%%%%%%%%%%%%%%%%%%%%%%%%%%%%%%%%%%%%%%%%%%%%%%%%%%%%%%%%%%%%

\section{Gradient Descent Optimization in Euclidean Spaces}\label{sec:euclidean_gradient_descent}

Let \nm{\vec x \in \mathbb{R}^m} be an Euclidean vector, \nm{\lrb{\vec f: \mathbb{R}^m \rightarrow \mathbb{R}^n \ | \ \vec f\lrp{\vec x} \in \mathbb{R}^n, \forall \ \vec x \in \mathbb{R}^m}} a nonlinear map for which it is also possible to evaluate its Jacobian \nm{\lrb{\vec J: \mathbb{R}^m \rightarrow \mathbb{R}^{nxm} \ | \ \vec J\lrp{\vec x} = \partial{\vec f\lrp{\vec x}} / \partial{\vec x} \in \mathbb{R}^{nxm}, \forall \ \vec x \in \mathbb{R}^m}}, and \nm{\vec{\mathcal E}\lrp{\vec x} = \vec f\lrp{\vec x}} \nm{- \, \vec f_T \ \in \mathbb{R}^n} an error or cost function containing the difference between the map \nm{\vec f} and a target result \nm{\vec f_T}.

The objective is to determine an input vector \nm{\vec x = \vec x_0 + \Delta \vec x} in the vicinity of a known initial value \nm{\vec x_0}, for which the cost function norm \nm{\| \vec{\mathcal E}\lrp{\vec x} \| \in \mathbb{R}} holds a local minimum, this is, \nm{\| \vec{\mathcal E}\lrp{\vec x_0 + \Delta \vec x} \| < \| \vec{\mathcal E}\lrp{\vec x_0} \|, \, \forall \ \vec x_0 \in \mathbb{R}^m}. The \emph{Gauss-Newton} optimization method provides a solution to this problem that relies on iteratively advancing the solution per (\ref{eq:algebra_gradient_descent_iterative}) starting with \nm{\vec x_0}:
\neweq{\vec x_{k+1} \longleftarrow \vec x_k + \Delta \vec x_k}{eq:algebra_gradient_descent_iterative}

Adopting a lighter notation in which \nm{\vec f_k = \vec f\lrp{\vec x_k}}, \nm{\vec J_k = \vec J\lrp{\vec x_k}}, and \nm{\vec{\mathcal E}_k = \vec{\mathcal E} \lrp{\vec x_k}}, the process concludes when the step diminution of the cost function norm falls below a given threshold (\nm{\| \vec{\mathcal E}_k \| - \| \vec{\mathcal E}_{k+1} \| < \delta}). The Gauss-Newton method consists on linearizing each step by performing a first order Taylor expansion of the cost function before minimizing its norm by equaling its derivative with respect to \nm{\Delta \vec x_k} to zero \cite{Hartley2003}:
\begin{eqnarray}
\nm{\vec{\mathcal E}_{k+1}} & = & \nm{\vec f_{k+1} - \vec f_T \approx \vec f_k + \vec J_k \, \Delta \vec x_k - \vec f_T = \vec{\mathcal E}_k + \vec J_k \, \Delta \vec x_k} \label{eq:algebra_gradient_descent_taylor} \\
\nm{\| \vec{\mathcal E}_{k+1} \|} & = & \nm{\vec{\mathcal E}_{k+1}^T \, \vec{\mathcal E}_{k+1} = \vec{\mathcal E}_k^T \, \vec{\mathcal E}_k + \Delta \vec x_k^T \, \vec J_k^T \, \vec J_k \, \Delta \vec x_k + 2 \, \Delta \vec x_k^T \, \vec J_k^T \, \vec{\mathcal E}_k} \label{eq:algebra_gradient_descent_norm} \\
\nm{\pderpar{\| \vec{\mathcal E}_{k+1} \|}{\Delta \vec x_k}} & = & \nm{0 \ \longrightarrow \ 2 \, \vec J_k^T \, \vec J_k \, \Delta \vec x_k + 2 \, \vec J_k^T \, \vec{\mathcal E}_k = 0 \ \longrightarrow \ \Delta \vec x_k = - \big(\vec J_k^T \, \vec J_k\big)^{-1} \, \vec J_k^T \, \vec{\mathcal E}_k} \label{eq:algebra_gradient_descent_solution}
\end{eqnarray}           

The Gauss-Newton algorithm is just one type of a more generic class of iterative minimization methods grouped under the name of \emph{gradient descent methods}. The \emph{Newton} method relies on minimizing (equaling its \nm{\Delta \vec x_k} derivative to zero) a second order Taylor expansion of the cost function norm \nm{N\lrp{\vec x} = \| \vec{\mathcal E}\lrp{\vec x} \| \ \in \mathbb{R}}, which requires the computation of both its gradient \nm{\lrb{\vec \nabla: \mathbb{R}^m \rightarrow \mathbb{R}^{1xm} \ | \ \vec \nabla\lrp{\vec x} = \partial{N\lrp{\vec x}} / \partial{\vec x} \in \mathbb{R}^{1xm}, \forall \ \vec x \in \mathbb{R}^m}} and its Hessian \nm{\lrb{\vec H: \mathbb{R}^m \rightarrow \mathbb{R}^{mxm} \ | \ \vec H\lrp{\vec x} = \partial^2{N\lrp{\vec x}} / \partial{\vec x^2} \in \mathbb{R}^{mxm}, \forall \ \vec x \in \mathbb{R}^m}} at each step, resulting in:
\neweq{\Delta \vec x_k = - \vec H_k^{-1} \ \vec \nabla_k^T}{eq:algebra_newton_solution}

As the Hessian \nm{\vec H} can be difficult or expensive to compute, there exist several approximations that reduce the computational cost of each step, such as the \emph{steepest descent} method, which replaces the Hessian with the product of a constant and the identity matrix \nm{\vec I_m \in \mathbb{R}^{mxm}}, and the \emph{diagonal approximation}, which sets to zero all \nm{\vec H} components outside its main diagonal. In this sense, the Gauss-Newton method is just another approximation that employs a first order simplification of the Hessian, as proven in \cite{Baker2004}.

The convergence of none of these methods is guaranteed. In general, both Gauss-Newton and Newton work better near the local minimum, where the quadratic approximation is good, but may diverge when the initial value is further away, where the steepest descent and diagonal approximation methods may be more robust. To ensure that the error gets smaller in each iteration, it may be convenient to advance with a smaller \nm{\Delta \vec x_k} step. The \emph{Levenberg-Marquardt} algorithm employs a varying ratio between the Gauss-Newton (or Newton) and diagonal approximations to the Hessian, moving towards the former when the error \nm{\| \vec{\mathcal E}_k \|} decreases, and towards the latter while repeating the step if it increases.

%%%%%%%%%%%%%%%%%%%%%%%%%%%%%%%%%%%%%%%%%%%%%%%%%%%%%%%%%%%%%%%%%%%%%%%%
%%%%%%%%%%%%%%%%%%%%%%%%%%%%%%%%%%%%%%%%%%%%%%%%%%%%%%%%%%%%%%%%%%%%%%%%
%%%%%%%%%%%%%%%%%%%%%%%%%%%%%%%%%%%%%%%%%%%%%%%%%%%%%%%%%%%%%%%%%%%%%%%%
% SECTION		STATE ESTIMATION IN EUCLIDEAN SPACES
%%%%%%%%%%%%%%%%%%%%%%%%%%%%%%%%%%%%%%%%%%%%%%%%%%%%%%%%%%%%%%%%%%%%%%%%
%%%%%%%%%%%%%%%%%%%%%%%%%%%%%%%%%%%%%%%%%%%%%%%%%%%%%%%%%%%%%%%%%%%%%%%%
%%%%%%%%%%%%%%%%%%%%%%%%%%%%%%%%%%%%%%%%%%%%%%%%%%%%%%%%%%%%%%%%%%%%%%%%

\section{State Estimation in Euclidean Spaces}\label{sec:SS}

\emph{State estimation} is the problem of determining the value of the state of a dynamic system based on a series of noisy equations that describe the evolution of the state with time, together with a series of noisy measurements or observations of variables that also depend on the state. \emph{State} or \emph{state vector} refers to those variables that provide a representation of the condition or status of the system at a given instant in time. Section \ref{subsec:SS_SampledDataSystems} discusses the equations that describe the system dynamics, this is, the state evolution with time, and what is the best possible estimation of the state that can be obtained from them. Section \ref{subsec:SS_SampledObservations} describes the measurement or observation equations, and also reaches the best possible state estimate from the information they contain. Both approaches are combined in section \ref{subsec:SS_EKF}, which describes the extended Kalman filter or \hypertt{EKF}, the most widely used nonlinear state estimation algorithm.

%%%%%%%%%%%%%%%%%%%%%%%%%%%%%%%%%%%%%%%%%%%%%%%%%%%%%%%%%%%%%%%%%%%%%%%%
%%%%%%%%%%%%%%%%%%%%%%%%%%%%%%%%%%%%%%%%%%%%%%%%%%%%%%%%%%%%%%%%%%%%%%%%
% SUBSECTION		SAMPLED DATA SYSTEMS
%%%%%%%%%%%%%%%%%%%%%%%%%%%%%%%%%%%%%%%%%%%%%%%%%%%%%%%%%%%%%%%%%%%%%%%%
%%%%%%%%%%%%%%%%%%%%%%%%%%%%%%%%%%%%%%%%%%%%%%%%%%%%%%%%%%%%%%%%%%%%%%%%

\subsection{Sampled Data Systems}\label{subsec:SS_SampledDataSystems}

A \emph{state space system} is a mathematical representation of a physical process in which the variables (both state and input) are related by first order differential equations (for continuous systems) or difference equations (for discrete ones). If the state of the system (the value of the state variables) is known at a given time, and so are all the present and future inputs (the evolution with time of the input variables), it is then possible to obtain the evolution with time of all the state variables.

A \emph{sampled data system} is one whose dynamics are described by continuous time differential equations, but whose inputs only change at discrete time instants. Additionally, it is only necessary to estimate the state variables, or to be precise its mean and covariance\footnote{Although the term covariance is traditionally employed in state estimation, it is in fact referring to the state random vector autocovariance provided by (\ref{eq:Error_rvec_autoCovarianceMatrix}), or to the state random process autocovariance given by (\ref{eq:Error_rpro_Autocovariance}), depending on context.}, at those same discrete time instants \cite{Simon2006}. A continuous time nonlinear state system can be written as
\neweq{\xvecdot\lrp{t} = \vec f\big(\xvec\lrp{t}, \, \uvec\lrp{t}, \ \wvec\lrp{t}, \, t\big)} {eq:SS_cont_time_system}

where \nm{\xvec \in \mathbb{R}^m} is the state vector, \nm{\uvec \in \mathbb{R}^n} is the known \emph{control} or \emph{input vector}, and \nm{\wvec \in \mathbb{R}^p} is the \emph{process noise}. These three vectors may have different sizes. Consider also that the process noise \nm{\wvec\lrp{t}} can be modeled by a zero mean continuous time white noise random process\footnote{Note that the process noise does not need to be Gaussian.} of covariance \nm{\Qvec_c} (sections \ref{subsec:Error_RandomVariables} and \ref{subsec:Error_WhiteNoise}):
\begin{eqnarray}
\nm{\wvec\lrp{t}} & \nm{\sim} & \nm{\lrp{\vec 0, \, \Qvec_c}}\label{eq:SS_cont_time_system_noise1} \\
\nm{\vec R_{ww}\lrp{t, \, \tau}} & = & \nm{E\lrsb{\wvec\lrp{t} \, \wvec^T\lrp{\tau}} = \Qvec_c \, \delta\lrp{t - \tau}}\label{eq:SS_cont_time_system_noise2}
\end{eqnarray}

%%%%%%%%%%%%%%%%%%%%%%%%%%%%%%%%%%%%%%%%%%%%%%%%%%%%%%%%%%%%%%%%%%%%%%%%
% subSUBSECTION		LINEARIZATION OF CONTINUOUS TIME SYSTEM
%%%%%%%%%%%%%%%%%%%%%%%%%%%%%%%%%%%%%%%%%%%%%%%%%%%%%%%%%%%%%%%%%%%%%%%%

\subsubsection{Linearization of Continuous Time Systems}\label{subsubsec:SS_SampledDataSystems_linearization}

The dynamics represented by (\ref{eq:SS_cont_time_system}) can be linearized by performing a Taylor expansion around an unknown nominal state \nm{\xvec_N\lrp{t}} and process noise \nm{\wvec_N\lrp{t}}\footnote{As the input vector \nm{\uvec\lrp{t}} is known, there is no need to expand around it.}, assuming without loss of generality that \nm{\wvec_N\lrp{t} = \vec 0}. If it is not, it can be written as the sum of a zero mean part and a known deterministic part, which can then be added to the control vector. The expansion is truncated so only the first order terms remain, introducing linearization errors; these are higher the more nonlinear that \nm{\vec f\lrp{\xvec, \, \uvec, \ \wvec, \, t}} is with respect to \nm{\xvec} and \nm{\wvec}, and the farther away that \nm{\xvec\lrp{t}} lies from \nm{\xvec_N\lrp{t}} and \nm{\wvec\lrp{t}} from \nm{\wvec_N = \vec 0} \cite{Simon2006}.
\neweq{\xvecdot\lrp{t} \approx \vec f\rvert_N + \pderpar{\vec f}{\xvec}\Bigr\rvert_N \, \lrp{\xvec - \xvec_N} + \pderpar{\vec f}{\wvec}\Bigr\rvert_N \, \wvec = \pderpar{\vec f}{\xvec}\Bigr\rvert_N \, \xvec + \lrp{\vec f\rvert_N - \pderpar{\vec f}{\xvec}\Bigr\rvert_N \, \xvec_N} + \pderpar{\vec f}{\wvec}\Bigr\rvert_N \, \wvec} {eq:SS_cont_time_system_taylor}

where \nm{\mid_N} stands for evaluation at \nm{\big(\xvec_N\lrp{t}, \, \uvec\lrp{t}, \, \vec 0, \, t\big)}. The state system is now continuous time but linear:
\begin{eqnarray}
\nm{\xvecdot\lrp{t}} & \nm{\approx} & \nm{\Avec\lrp{t} \, \xvec\lrp{t} + \Bvec\lrp{t} \, \utilde\lrp{t} + \wtilde\lrp{t}}\label{eq:SS_cont_time_system_linear} \\
\nm{\Avec\lrp{t}} & = & \nm{\pderpar{\vec f}{\xvec}\big(\xvec_N\lrp{t}, \, \uvec\lrp{t}, \, \vec 0, \, t\big)}\label{eq:SS_cont_time_system_linear_system_matrix} \\
\nm{\Bvec\lrp{t}} & = & \nm{\Ivec}\label{eq:SS_cont_time_system_linear_other_matrix} \\
\nm{\Lvec\lrp{t}} & = & \nm{\pderpar{\vec f}{\wvec}\big(\xvec_N\lrp{t}, \, \uvec\lrp{t}, \, \vec 0, \, t\big)}\label{eq:SS_cont_time_system_linear_input_matrix} \\
\nm{\utilde\lrp{t}} & = & \nm{\vec f\big(\xvec_N\lrp{t}, \, \uvec\lrp{t}, \, \vec 0, \, t\big) - \Avec\lrp{t} \, \xvec_N\lrp{t}}\label{eq:SS_cont_time_system_linear_input_vector} \\
\nm{\wtilde\lrp{t}} & = & \nm{\Lvec\lrp{t} \, \wvec\lrp{t} \sim \lrp{\vec 0, \, \Lvec \, \Qvec_c \, \Lvec^T} = \lrp{\vec 0, \, \Qtilde_c\lrp{t}}}\label{eq:SS_cont_time_system_linear_noise1} \\
\nm{\vec R_{\widetilde{w}\widetilde{w}}\lrp{t, \, \tau}} & = & \nm{E\lrsb{\wtilde\lrp{t} \, \wtilde^T\lrp{\tau}} = \Qtilde_c\lrp{t} \, \delta\lrp{t - \tau}}\label{eq:SS_cont_time_system_linear_noise2} 
\end{eqnarray}

The above linear state system is based on a unitary \emph{input matrix} \nm{\Bvec \in \mathbb{R}^{mxm}} and a \emph{system matrix} \nm{\Avec\lrp{t} \in \mathbb{R}^{mxm}} that is the Jacobian of the nonlinear system with respect to the state vector evaluated at the unknown nominal state. It also employs modified input \nm{\utilde\lrp{t} \in \mathbb{R}^m} and process noise \nm{\wtilde\lrp{t} \in \mathbb{R}^m} vectors. 

%%%%%%%%%%%%%%%%%%%%%%%%%%%%%%%%%%%%%%%%%%%%%%%%%%%%%%%%%%%%%%%%%%%%%%%%
% subSUBSECTION		COMPARISON OF INTEGRATED CONTINUOUS AND DISCRETE WHITE NOISE PROCESSES
%%%%%%%%%%%%%%%%%%%%%%%%%%%%%%%%%%%%%%%%%%%%%%%%%%%%%%%%%%%%%%%%%%%%%%%%

\subsubsection{Comparison of Integrated Continuous and Discrete White Noise Processes}\label{subsubsec:SS_SampledDataSystems_comparison}

Before continuing, this section compares the behavior of an integrated continuous white noise process with that of a discrete one, as the result is essential to the discretization of the continuous time state system (\ref{eq:SS_cont_time_system_linear}). According to section \ref{subsec:Error_WhiteNoise}, a continuous zero mean white noise is defined by \nm{\wvec\lrp{t} \sim \lrp{\vec 0, \, \Qvec_c}} and \nm{E\lrsb{\wvec\lrp{t} \, \wvec^T\lrp{\tau}} = \Qvec_c \, \delta\lrp{t - \tau}}, while a zero mean discrete time white noise process responds to \nm{\wvec_k \sim \lrp{\vec 0, \, \Qvec_d}} and \nm{E\lrsb{\wvec_k \, \wvec_l^T} = \Qvec_d \, \delta_{k-l}}. The variation with time of the mean \nm{\vec \mu_z\lrp{t}} and covariance \nm{\vec C_{zz}\lrp{t}} of the noise \nm{\zvec\lrp{t}} resulting from the integration of the continuous white noise with \nm{\dot{\zvec}\lrp{t} = \wvec\lrp{t}, \ \zvec\lrp{0} = \vec 0} are the following:
\begin{eqnarray}
\nm{\vec \mu_z\lrp{t}} & = & \nm{E\lrsb{\zvec\lrp{t}} = E\lrsb{\int_0^t \wvec\lrp{\alpha} \, \mathrm{d}\alpha} = \int_0^t E\lrsb{\wvec\lrp{\alpha}} \, \mathrm{d}\alpha = \vec 0}\label{eq:SS_cont_integr_noise_mean} \\
\nm{\vec C_{zz}\lrp{t}} & = & \nm{E\lrsb{\zvec\lrp{t} \, \zvec^T\lrp{t}} - \vec \mu_z\lrp{t} \, \vec \mu_z^T\lrp{t} = E\lrsb{\int_0^t \wvec\lrp{\alpha} \, \mathrm{d}\alpha \, \int_0^t \wvec^T\lrp{\beta} \, \mathrm{d}\beta}} \nonumber \\
& = & \nm{\int_0^t \int_0^t E\lrsb{\wvec\lrp{\alpha} \, \wvec^T\lrp{\beta}} \, \mathrm{d}\alpha \, \mathrm{d}\beta = \Qvec_c \, \int_0^t \int_0^t \delta\lrp{\alpha - \beta} \, \mathrm{d}\alpha \, \mathrm{d}\beta = \Qvec_c \, \int_0^t \mathrm{d}\beta = \Qvec_c \, t}\label{eq:SS_cont_integr_noise_covariance}
\end{eqnarray}

This expression shows that the mean of an integrated continuous white noise is always zero, but its covariance grows linearly with time. Integrating now the difference equation \nm{\zvec_k = \zvec_{k-1} + \wvec_{k-1}, \ \zvec_0 = \vec 0}, the variation with time of the mean \nm{\vec \mu_k} and covariance \nm{\vec C_{zz,k}} of the integrated noise \nm{\zvec_k} are the following:
\begin{eqnarray}
\nm{\vec \mu_k} & = & \nm{E\lrsb{\zvec_k} = E\lrsb{\sum_{l=0}^{k-1} \, \wvec_l} = \sum_{l=0}^{k-1} \, E\lrsb{\wvec_l} = \vec 0}\label{eq:SS_discr_integr_noise_mean} \\
\nm{\vec C_{zz,k}} & = & \nm{E\lrsb{\zvec_k \, \zvec_k^T} - \vec \mu_k \, \vec \mu_k^T = E\lrsb{\sum_{l=0}^{k-1} \, \wvec_l \, \sum_{m=0}^{k-1} \, \wvec_m^T} = \sum_{l=0}^{k-1} \, \sum_{m=0}^{k-1} \, E\lrsb{\wvec_l \, \wvec_m^T}} \nonumber \\
 & = & \nm{\Qvec_d \, \sum_{l=0}^{k-1} \, \sum_{m=0}^{k-1} \, \delta_{l-m} = \Qvec_d \, \sum_{l=0}^{k-1} \, 1 = \Qvec_d \, k}\label{eq:SS_discr_integr_noise_covariance}
\end{eqnarray}

The covariance of the integrated discrete white noise process also grows linearly with time. Considering a sampling period of \nm{\Deltat, \, t = k \cdot \Deltat}, a discrete zero mean white noise process can be considered equivalent \cite{Simon2006} to a continuous one if their covariances are related by:
\neweq{\Qvec_d = \Qvec_c \cdot \Deltat}{eq:SS_integr_white_noise_cov_equiv}

%%%%%%%%%%%%%%%%%%%%%%%%%%%%%%%%%%%%%%%%%%%%%%%%%%%%%%%%%%%%%%%%%%%%%%%%
% subSUBSECTION		DISCRETIZATION OF LINEAR CONTINUOUS TIME SYSTEM
%%%%%%%%%%%%%%%%%%%%%%%%%%%%%%%%%%%%%%%%%%%%%%%%%%%%%%%%%%%%%%%%%%%%%%%%

\subsubsection{Discretization of Linear Continuous Time Systems}\label{subsubsec:SS_SampledDataSystems_discretization}

Returning to the main argument, and considering that the state vector needs to be known only at a series of discrete time points, it is possible to discretize the (\ref{eq:SS_cont_time_system_linear}) linear continuous time system if \nm{\Avec\lrp{t}}, \nm{\Bvec\lrp{t}}, and \nm{\utilde\lrp{t}} are considered constant during the integration interval, which starts at \nm{t_{k-1} = \lrp{k-1} \cdot \Deltat} and concludes at \nm{t_k = k \cdot \Deltat}. The introduced discretization errors are higher the farther away this assumption is from reality. Introducing (\ref{eq:SS_integr_white_noise_cov_equiv}), the state system is now discrete and linear \cite{Simon2006}: 
\begin{eqnarray}
\nm{\xvec_k} & \nm{\approx} & \nm{\Fvec_{k-1} \, \xvec_{k-1} + \Gvec_{k-1} \, \utilde_{k-1} + \wtilde_{k-1}}\label{eq:SS_discr_time_system_linear} \\
\nm{\xvec_k} & = & \nm{\xvec\lrp{t_k} = \xvec \lrp{k \, \Deltat}}\label{eq:SS_discr_time_system_linear_x} \\
\nm{\Fvec_k} & = & \nm{exp\lrp{\Avec_k \, \Deltat} = exp\big(\Avec\lrp{k \, \Deltat} \, \Deltat\big)}\label{eq:SS_discr_time_system_linear_system_matrix} \\
\nm{\Gvec_k} & = & \nm{\Fvec_k \, \int_0^{\Deltat} \, exp\big(- \Avec\lrp{\tau} \ \tau\big) \, \mathrm{d}\tau \, \Bvec\lrp{k \, \Deltat}} \nonumber \\
& = & \nm{\Fvec_k \, \lrsb{\Ivec - exp\big(- \Avec\lrp{k \, \Deltat} \Deltat\big)} \Avec^{-1}\lrp{k \, \Deltat} \, \Bvec\lrp{k \, \Deltat}}\label{eq:SS_discr_time_system_linear_input_matrix} \\
\nm{\utilde_k} & = & \nm{\utilde\lrp{t_k} = \utilde \lrp{k \, \Deltat}}\label{eq:SS_discr_time_system_linear_u} \\
\nm{\wtilde_k} & = & \nm{\wtilde\lrp{k \, \Deltat} \sim \lrp{\vec 0, \, \Qtilde_c\lrp{k \cdot \Deltat} \cdot \Deltat} = \lrp{\vec 0, \, \Lvec_k \, \Qvec_c \, \Lvec_k^T \, \Deltat} = \lrp{\vec 0, \, \Qtilde_{d,k}}}\label{eq:SS_discr_time_system_linear_noise1} \\
\nm{\Rvec_{\widetilde{w}\widetilde{w},kj}} & = & \nm{E\lrsb{\wtilde_k \, \wtilde_j^T} = \Qtilde_{d,k} \, \delta_{k-j}}\label{eq:SS_discr_time_system_linear_noise2} 
\end{eqnarray}

Note that both the \emph{system state transition matrix} \nm{\Fvec_k \in \mathbb{R}^{mxm}} and the \emph{input transition matrix} \nm{\Gvec_k \in \mathbb{R}^{mxm}} make use of the matrix exponential function, although computing the later is not required, as shown in section \ref{subsubsec:SS_EKF_nominal}.

%%%%%%%%%%%%%%%%%%%%%%%%%%%%%%%%%%%%%%%%%%%%%%%%%%%%%%%%%%%%%%%%%%%%%%%%
% subSUBSECTION		MEAN AND COVARIANCE OF STATE VECTOR
%%%%%%%%%%%%%%%%%%%%%%%%%%%%%%%%%%%%%%%%%%%%%%%%%%%%%%%%%%%%%%%%%%%%%%%%

\subsubsection{Mean and Covariance of State Vector}\label{subsubsec:SS_SampledDataSystems_mean_covariance}

It is possible to evaluate the mean \nm{\vec \mu_{x,k}} and covariance \nm{\vec C_{xx,k} = \Pvec_k} of the state vector given by (\ref{eq:SS_discr_time_system_linear}), which provide their variation with time\footnote{To compute the covariance, note that there is no correlation between \nm{\lrp{\xvec_{k-1} - \mu_{\xvec_{k-1}}}} and \nm{\wtilde_{k-1}}.}:
\begin{eqnarray}
\nm{\vec \mu_{x,k}} & = & \nm{E\lrsb{\xvec_k} = \Fvec_{k-1} \, \vec \mu_{x,k-1} + \Gvec_{k-1} \, \utilde_{k-1}}\label{eq:SS_discr_time_system_state_mean} \\
\nm{\vec C_{xx,k}} & = & \nm{\Pvec_k = E\lrsb{\lrp{\xvec_k - \vec \mu_{x,k}} \, \lrp{\xvec_k - \vec \mu_{x,k}}^T}} \nonumber \\
& = & \nm{E\lrsb{\Big(\Fvec_{k-1} \, \lrp{\xvec_{k-1} - \vec \mu_{x,k-1}} + \wtilde_{k-1}\Big) \Big(\Fvec_{k-1} \, \lrp{\xvec_{k-1} - \vec \mu_{x,k-1}} + \wtilde_{k-1}\Big)^T}} \nonumber \\
& = & \nm{\Fvec_{k-1} \, \vec C_{xx,k-1} \, \Fvec_{k-1}^T + \Qtilde_{d,k-1} = \Fvec_{k-1} \, \Pvec_{k-1} \, \Fvec_{k-1}^T + \Qtilde_{d,k-1}}\label{eq:SS_discr_time_system_state_cov}
\end{eqnarray}

Based on (\ref{eq:SS_discr_time_system_linear}), \nm{\xvec_k} is a linear combination of a series of known real vectors \nm{\utilde_0, \dots, \utilde_{k-1}} plus a series of independent random vectors \nm{\xvec_0, \, \wtilde_0, \dots, \wtilde_{k-1}}. According to the central limit theorem stated in section \ref{subsec:Error_RandomVariables}, \nm{\xvec_k \sim N\lrp{\vec \mu_{x,k}, \, \vec C_{xx,k}} = N\lrp{\vec \mu_{x,k}, \, \Pvec_k}} is a normal or Gaussian random vector completely characterized by its mean and covariance.

The summary of this section is that given a continuous time nonlinear state space system such as (\ref{eq:SS_cont_time_system}), it is possible, with some linearization and discretization errors, to transform it into an equivalent discrete time linear system (\ref{eq:SS_discr_time_system_linear}) that can be integrated to obtain the estimated value of the state vector \nm{\xvec\lrp{t}} at a series of discrete times \nm{t_k = k \, \Deltat} characterized by its mean \nm{\vec \mu_{x,k}} (\ref{eq:SS_discr_time_system_state_mean}) and covariance \nm{\vec C_{xx,k} = \Pvec_k} (\ref{eq:SS_discr_time_system_state_cov}). Without further assistance, (\ref{eq:SS_discr_time_system_state_cov}) shows that the uncertainty of the results grows with time because of the accumulation of the white noise present in the system \cite{Simon2006}. The next section shows how the addition of measurements can solve this problem.

%%%%%%%%%%%%%%%%%%%%%%%%%%%%%%%%%%%%%%%%%%%%%%%%%%%%%%%%%%%%%%%%%%%%%%%%
%%%%%%%%%%%%%%%%%%%%%%%%%%%%%%%%%%%%%%%%%%%%%%%%%%%%%%%%%%%%%%%%%%%%%%%%
% SUBSECTION		SAMPLED OBSERVATIONS
%%%%%%%%%%%%%%%%%%%%%%%%%%%%%%%%%%%%%%%%%%%%%%%%%%%%%%%%%%%%%%%%%%%%%%%%
%%%%%%%%%%%%%%%%%%%%%%%%%%%%%%%%%%%%%%%%%%%%%%%%%%%%%%%%%%%%%%%%%%%%%%%%

\subsection{Sampled Observations}\label{subsec:SS_SampledObservations}

Given the sampled data system of section \ref{subsec:SS_SampledDataSystems}, it is possible to consider that there exist a series of sensors capable of measuring certain variables related to the state vector at the same time points at which the state system is discretized in section \ref{subsubsec:SS_SampledDataSystems_discretization}:
\neweq{\yvec_k = \vec h\lrp{\xvec_k, \, \vvec_k, \, t_k}}{eq:SS_measur_nonlinear}

where \nm{\yvec_k = \yvec\lrp{t_k} \in \mathbb{R}^q} is the \emph{measurement} or \emph{observation vector} provided by the sensors, \nm{\xvec_k = \xvec\lrp{t_k} \in \mathbb{R}^m} is the state vector, \nm{t_k = t\lrp{k \, \Deltat}} is the discrete time at which the measurements are taken, and \nm{\vvec_k = \vvec\lrp{t_k} \in \mathbb{R}^q} is the \emph{measurement} or \emph{observation noise}, which can be modeled by a zero mean white noise random process\footnote{Note that the measurement noise does not need to be Gaussian.} of covariance \nm{\Rvec} (sections \ref{subsec:Error_RandomVariables} and \ref{subsec:Error_WhiteNoise}):
\begin{eqnarray}
\nm{\vvec_k} & \nm{\sim} & \nm{\lrp{\vec 0, \, \Rvec}}\label{eq:SS_measur_nonlinear_noise1} \\
\nm{\Rvec_{vv,kj}} & = & \nm{E\lrsb{\vvec_k \, \vvec_j^T} = \Rvec \, \delta_{k-j}}\label{eq:SS_measur_nonlinear_noise2}
\end{eqnarray}

Consider also that the measurement noise and the process noise of section \ref{subsec:SS_SampledDataSystems} are orthogonal:
\neweq{\Rvec_{vw,kj} = E\lrsb{\vvec_k \, \wvec_j^T} = \vec 0}{eq:SS_measur_nonlinear_noise3}

%%%%%%%%%%%%%%%%%%%%%%%%%%%%%%%%%%%%%%%%%%%%%%%%%%%%%%%%%%%%%%%%%%%%%%%%
% subSUBSECTION		LINEARIZATION OF OBSERVATIONS
%%%%%%%%%%%%%%%%%%%%%%%%%%%%%%%%%%%%%%%%%%%%%%%%%%%%%%%%%%%%%%%%%%%%%%%%

\subsubsection{Linearization of Observations}\label{subsubsec:SS_SampledObservations_linearization}

The discrete observations represented by (\ref{eq:SS_measur_nonlinear}) can be linearized by performing a Taylor expansion around an unknown nominal state \nm{\xvec_{Nk} = \xvec_N\lrp{t_k}} and observation noise \nm{\vvec_{Nk} = \vvec_N\lrp{t_k}}, assuming without loss of generality that \nm{\vvec_{Nk} = \vec 0}. If it is not, it can be written as the sum of a zero mean part and a known deterministic part, which can then be included in the nonlinear function \emph{h}. The expansion is truncated so only the first order terms remain, introducing linearization errors; these are higher the more nonlinear that \nm{\vec h\lrp{\xvec_k, \ \vvec_k, \, t_k}} is with respect to \nm{\xvec_k} and \nm{\vvec_k}, and the farther away that \nm{\xvec_k} is from \nm{\xvec_{Nk}} and \nm{\vvec_k} from \nm{\vvec_{Nk} = \vec 0} \cite{Simon2006}.
\neweq{\yvec_k \approx \vec h\rvert_N + \pderpar{\vec h}{\xvec_k}\Bigr\rvert_N \, \lrp{\xvec_k - \xvec_{Nk}} + \pderpar{\vec h}{\vvec_k}\Bigr\rvert_N \, \vvec_k = \pderpar{\vec h}{\xvec_k}\Bigr\rvert_N \, \xvec_k + \lrp{\vec h\rvert_N - \pderpar{\vec h}{\xvec_k}\Bigr\rvert_N \, \xvec_{Nk}} + \pderpar{\vec h}{\vvec_k}\Bigr\rvert_N \, \vvec_k} {eq:SS_measur_taylor}

where \nm{\mid_N} stands for evaluation at \nm{\lrp{\xvec_{Nk}, \, \vec 0, \, t_k}}. The observations system is now discrete time and linear:
\begin{eqnarray}
\nm{\yvec_k} & \nm{\approx} & \nm{\Hvec_k \, \xvec_k + \zvec_k + \vtilde_k}\label{eq:SS_measur_linear} \\
\nm{\Hvec_k} & = & \nm{\Hvec\lrp{t_k} = \pderpar{\vec h}{\xvec_k}\lrp{\xvec_{Nk}, \, \vec 0, \, t_k}}\label{eq:SS_measur_linear_output_matrix} \\
\nm{\zvec_k} & = & \nm{\zvec\lrp{t_k} = \vec h\lrp{\xvec_{Nk}, \, \vec 0, \, t_k} - \Hvec_k \, \xvec_{Nk}}\label{eq:SS_measur_linear_extra_vector} \\
\nm{\Mvec_k} & = & \nm{\Mvec\lrp{t_k} = \pderpar{\vec h}{\vvec_k}\lrp{\xvec_{Nk}, \, \vec 0, \, t_k}}\label{eq:SS_measur_linear_input_matrix} \\
\nm{\vtilde_k} & = & \nm{\vtilde\lrp{t_k} = \Mvec_k \, \vvec_k \sim \lrp{\vec 0, \, \Mvec_k \, \Rvec \, \Mvec_k^T} = \lrp{\vec 0, \, \Rtilde_k}}\label{eq:SS_measure_linear_noise1} \\
\nm{\Rvec_{\widetilde{v}\widetilde{v},kj}} & = & \nm{E\lrsb{\vtilde_k \, \vtilde_j^T} = \Rtilde_k \, \delta_{k-j}}\label{eq:SS_measure_linear_noise2} 
\end{eqnarray}

The above observation system is based on an \emph{output matrix} \nm{\Hvec_k \in \mathbb{R}^{qxm}} that is the Jacobian of the nonlinear system with respect to the state vector evaluated at the unknown nominal state, and a vector \nm{\zvec_k \in \mathbb{R}^q} that depends exclusively of the nominal state. It also employs a modified observation noise vector \nm{\vtilde_k \in \mathbb{R}^q}. It is worth noting that computation of \nm{\zvec_k} is not necessary to obtain the solution, as shown in section \ref{subsubsec:SS_EKF_nominal}.

%%%%%%%%%%%%%%%%%%%%%%%%%%%%%%%%%%%%%%%%%%%%%%%%%%%%%%%%%%%%%%%%%%%%%%%%
% subSUBSECTION		CONSTANT STATE VECTOR ESTIMATION BASED ON OBSERVATIONS
%%%%%%%%%%%%%%%%%%%%%%%%%%%%%%%%%%%%%%%%%%%%%%%%%%%%%%%%%%%%%%%%%%%%%%%%

\subsubsection{Constant State Vector Estimation based on Observations}\label{subsubsec:SS_SampledObservations_estimation}

The objective of this section is to obtain the best possible estimate \nm{\xvecest_k} of a constant\footnote{Note that this is the only section where the state vector \nm{\xvec_k} is required to be constant, this is, \nm{\xvec = \xvec_0 = \xvec_k \ \forall \, k}.} state vector \nm{\xvec} based on the observations \nm{\yvec_k} provided by (\ref{eq:SS_measur_linear}) and the previous estimate \nm{\xvecest_{k-1}}. It is possible to employ an expression like (\ref{eq:SS_measur_linear_estimate}), where \nm{\Kvec_k \in \mathbb{R}^{mxq}} is called the \emph{gain matrix} and \nm{\rvec_k \in \mathbb{R}^q} the \emph{innovations vector}:
\neweq{\xvecest_k = \xvecest_{k-1} + \Kvec_k \, \rvec_k = \xvecest_{k-1} + \Kvec_k \, \lrp{\yvec_k - \Hvec_k \, \xvecest_{k-1} - \zvec_k}}{eq:SS_measur_linear_estimate}

The \emph{estimation error} \nm{\vec \varepsilon_{x,k}} and its mean can then be computed based on (\ref{eq:SS_measur_linear_estimate}) and (\ref{eq:SS_measur_linear}):
\begin{eqnarray}
\nm{\vec \varepsilon_{x,k}} & = & \nm{\xvec - \xvecest_k = \lrp{\Ivec - \Kvec_k \, \Hvec_k} \, \vec \varepsilon_{x,k-1}  - \Kvec_k \, \vtilde_k}\label{eq:SS_measur_linear_estimation_error} \\
\nm{\vec \mu_{\varepsilon x,k}} & = & \nm{E\lrsb{\vec \varepsilon_{x,k}} = \lrp{\Ivec - \Kvec_k \, \Hvec_k} \, \vec \mu_{\varepsilon x,k-1}}\label{eq:SS_measur_linear_estimation_error_mean}
\end{eqnarray}

As the linearized discrete noise \nm{\vtilde_k} is zero mean per (\ref{eq:SS_measure_linear_noise1}), (\ref{eq:SS_measur_linear_estimate}) is called an \emph{unbiased estimator} \cite{Simon2006}, because if the initial estimate \nm{\xvecest_0} is set equal to the expected value of the state vector \nm{\lrp{\xvecest_0 = \vec \mu_x \rightarrow \vec \mu_{\varepsilon x,0} = \vec 0}}, then \nm{\vec \mu_{\varepsilon x,k} = E\lrsb{\varepsilon_{x,k}} = \vec 0 \ \forall \, k}, this is, the expected value of \nm{\xvecest_k} is equal to \nm{\vec \mu_x = E\lrsb{\xvec}} for all \nm{t_k}. This is regardless of the value of the gain matrix \nm{\Kvec_k}.

A similar process is followed to compute the covariance of the estimation error \nm{\vec C_{xx,k} = \Pvec_k}. To do so, note that the observation noise is independent from the estimation error \nm{\lrp{E\lrsb{\vtilde_k \, \vec \varepsilon_{x,k-1}^T} = \vec 0}}:
\neweq{\vec C_{xx,k} = \Pvec_k = E\lrsb{\vec \varepsilon_{x,k} \, \vec \varepsilon_{x,k}^T} - \vec \mu_{\varepsilon x,k} \, \vec \mu_{\varepsilon x,k}^T = \lrp{\Ivec - \Kvec_k \, \Hvec_k}\, \Pvec_{k-1} \, \lrp{\Ivec - \Kvec_k \, \Hvec_k}^T + \Kvec_k \, \Rtilde_k \, \Kvec_k^T}{eq:SS_measur_linear_covariance}

This expression guarantees that \nm{\vec C_{xx,k} = \Pvec_k} is positive definite (as all covariance matrices) given that so are \nm{\Pvec_{k-1}} and \nm{\Rtilde_k}. The criterion to obtain the gain matrix \nm{\Kvec_k} and hence fill up (\ref{eq:SS_measur_linear_estimate}) and (\ref{eq:SS_measur_linear_covariance}) to obtain the estimation of the state vector as well as the covariance of its error, is the minimization of the sum of the variances of the estimation errors. That way, the estimation error is not only zero mean but it is also consistently as close as possible to zero \cite{Simon2006}.
\begin{eqnarray}
\nm{\vec J_k} & = & \nm{E\lrsb{\lrp{x_1 - \hat{x}_{1,k}}^2 + \ldots + \lrp{x_m - \hat{x}_{m,k}}^2} = E\lrsb{\varepsilon_{x1,k}^2 + \ldots + \varepsilon_{xm,k}^2}}\nonumber \\
& = & \nm{E\lrsb{\vec \varepsilon_{x,k}^T \, \vec \varepsilon_{x,k}} = E\lrsb{Tr\lrp{\vec \varepsilon_{x,k} \, \vec \varepsilon_{x,k}^T}} = Tr \, \vec C_{xx,k} = Tr \, \Pvec_k}\label{eq:SS_measur_linear_J} \\
\nm{\pderpar{\vec J_k}{\Kvec_k}} & = & \nm{2 \, \lrp{\Ivec - \Kvec_k \, \Hvec_k} \, \Pvec_{k-1} \, \lrp{- \Hvec_k^T} + 2 \, \Kvec_k \, \Rtilde_k}\label{eq:SS_measur_linear_Jderiv}
\end{eqnarray}

where \nm{Tr\lrp{}} stands for trace of a matrix, and some not so obvious matrix algebra properties have been employed. Setting the (\ref{eq:SS_measur_linear_Jderiv}) derivative to zero provides the optimum gain matrix:
\neweq{\Kvec_k = \Pvec_{k-1} \, \Hvec_k^T \lrp{\Hvec_k \, \Pvec_{k-1} \, \Hvec_k^T + \Rtilde_k}^{-1}}{eq:SS_measur_linear_optimal_gain}

%%%%%%%%%%%%%%%%%%%%%%%%%%%%%%%%%%%%%%%%%%%%%%%%%%%%%%%%%%%%%%%%%%%%%%%%
%%%%%%%%%%%%%%%%%%%%%%%%%%%%%%%%%%%%%%%%%%%%%%%%%%%%%%%%%%%%%%%%%%%%%%%%
% SUBSECTION		EXTENDED KALMAN FILTER
%%%%%%%%%%%%%%%%%%%%%%%%%%%%%%%%%%%%%%%%%%%%%%%%%%%%%%%%%%%%%%%%%%%%%%%%
%%%%%%%%%%%%%%%%%%%%%%%%%%%%%%%%%%%%%%%%%%%%%%%%%%%%%%%%%%%%%%%%%%%%%%%%

\subsection{Extended Kalman Filter}\label{subsec:SS_EKF}

Provided with discrete time and linear state dynamics (\ref{eq:SS_discr_time_system_linear}) and observations (\ref{eq:SS_measur_linear}), the goal of state estimation is to obtain the best possible estimate of the state vector \nm{\xvec_k = \xvec\lrp{t_k}} based on the knowledge of the system provided by the state dynamics and the availability of observations \cite{Simon2006}. At a given time  \nm{t_k = k \, \Deltat}, the \emph{a priori estimation} \nm{\xvecest_k^-} is defined as the estimation of \nm{\xvec_k}, this is, the estimation of the state vector at time \nm{t_k}, making use of all measurements taken before \nm{t_k} but not including those at \nm{t_k}. The \emph{a posteriori estimation} \nm{\xvecest_k^+} is defined as the estimation of \nm{\xvec_k} that makes use of all measurements up and including \nm{t_k}. In the same way, it is possible to define the \emph{a priori} and \emph{a posteriori covariances} of the estimation error \nm{\Pvec_k^-} and \nm{\Pvec_k^+}:
\begin{eqnarray}
\nm{\Pvec_k^-} & = & \nm{E\lrsb{\lrp{\xvec_k - \xvecest_k^-} \, \lrp{\xvec_k - \xvecest_k^-}^T} - E\lrsb{\xvec_k - \xvecest_k^-} \, E\lrsb{\xvec_k - \xvecest_k^-}^T} \label{eq:SS_EKF_covariance_apriori_definition} \\
\nm{\Pvec_k^+} & = & \nm{E\lrsb{\lrp{\xvec_k - \xvecest_k^+} \, \lrp{\xvec_k - \xvecest_k^+}^T} - E\lrsb{\xvec_k - \xvecest_k^+} \, E\lrsb{\xvec_k - \xvecest_k^+}^T} \label{eq:SS_EKF_covariance_aposteriori_definition} 
\end{eqnarray}

The process starts with an initial estimation of the state vector \nm{\xvecest_0^+} before any measurements are available (they start at \nm{k = 1}). Since no measurements are available, it is reasonable to form \nm{\xvecest_0^+} as the expected value of the initial state \nm{\xvec_0} \cite{Simon2006}:
\neweq{\xvecest_0^+ = \vec \mu_{x,0} = E\lrsb{\xvec_0} = E\big[\xvec\lrp{t_0}\big]}{eq:SS_EKF_x0_initial_state} 

The covariance of the initial estimation error \nm{\Pvec_0^+} is also required, representing the uncertainty in the initial estimation \nm{\xvecest_0^+} \footnote{If the initial state is known with exactitude, use \nm{\Pvec_0^+ = 0}. Otherwise, use higher values the less confidence the user has in the accuracy of \nm{\xvecest_0^+}.}: 
\neweq{\Pvec_0^+ = E\lrsb{\lrp{\xvec_0 - \xvecest_0^+} \, \lrp{\xvec_0 - \xvecest_0^+}^T} - E\lrsb{\xvec_0 - \xvecest_0^+} \, E\lrsb{\xvec_0 - \xvecest_0^+}^T = E\lrsb{\lrp{\xvec_0 - \vec \mu_{x,0}} \, \lrp{\xvec_0 - \vec \mu_{x,0}}^T}}{eq:SS_EKF_P0_initial_covariance} 

%%%%%%%%%%%%%%%%%%%%%%%%%%%%%%%%%%%%%%%%%%%%%%%%%%%%%%%%%%%%%%%%%%%%%%%%
% subSUBSECTION		TIME UPDATE AND MEASUREMENT UPDATE EQUATIONS
%%%%%%%%%%%%%%%%%%%%%%%%%%%%%%%%%%%%%%%%%%%%%%%%%%%%%%%%%%%%%%%%%%%%%%%%

\subsubsection{Time Update and Measurement Update Equations}\label{subsubsec:SS_EKF_equations}

The next step is to propagate the state estimation without the use of any observations from \nm{\xvecest_0^+} to \nm{\xvecest_1^-}, with the objective of obtaining an estimation that coincides with the state vector mean, this is, \nm{\xvecest_1^- = \vec \mu_{x,1} = E\lrsb{\xvec_1}}. Recalling the evolution of the state vector expected value provided by (\ref{eq:SS_discr_time_system_state_mean}), and extending the same reasoning to all steps, it makes sense intuitively to propagate the state estimate the same way that the mean of the state propagates \cite{Simon2006}. Hence, the time propagation for the state estimate results in:
\neweq{\xvecest_k^- = \Fvec_{k-1} \, \xvecest_{k-1}^+ + \Gvec_{k-1} \, \utilde_{k-1}}{eq:SS_EKF_xest_minus_propagate}

A similar reasoning is employed for the propagation of the covariance of the estimation error in the absence of observations. Recalling the evolution of the state vector covariance provided by (\ref{eq:SS_discr_time_system_state_cov}) and extending the same reasoning to all steps, the time propagation of the covariance results in \cite{Simon2006}:
\neweq{\Pvec_k^- = \Fvec_{k-1} \, \Pvec_{k-1}^+ \, \Fvec_{k-1}^T + \Qtilde_{d,k-1}}{eq:SS_EKF_P_minus_propagate}

The above equations are called the \emph{time update equations}. Once the a priori estimation and error covariance have been computed, it is possible to update them with the information contained in the observation. This is done with the expressions derived in section \ref{subsubsec:SS_SampledObservations_estimation}, replacing \nm{\xvecest_{k-1}} with \nm{\xvecest_k^-}, \nm{\xvecest_k} with \nm{\xvecest_k^+}, \nm{\Pvec_{k-1}} with \nm{\Pvec_k^-}, and \nm{\Pvec_k} with \nm{\Pvec_k^+} \cite{Simon2006}. These are called the \emph{measurement update equations}:
\begin{eqnarray}
\nm{\Kvec_k} & = & \nm{\Pvec_k^- \, \Hvec_k^T \lrp{\Hvec_k \, \Pvec_k^- \, \Hvec_k^T + \Rtilde_k}^{-1}}\label{SS_EKF_kalman_gain} \\
\nm{\xvecest_k^+} & = & \nm{\xvecest_k^- + \Kvec_k \, \rvec_k = \xvecest_k^- + \Kvec_k \, \lrp{\yvec_k - \Hvec_k \, \xvecest_k^- - \zvec_k}}\label{eq:SS_EKF_xest_plus_propagate} \\
\nm{\Pvec_k^+} & = & \nm{\lrp{\Ivec - \Kvec_k \, \Hvec_k}\, \Pvec_k^- \, \lrp{\Ivec - \Kvec_k \, \Hvec_k}^T + \Kvec_k \, \Rtilde_k \, \Kvec_k^T}\label{eq:SS_EKF_P_plus_propagate}
\end{eqnarray}

%%%%%%%%%%%%%%%%%%%%%%%%%%%%%%%%%%%%%%%%%%%%%%%%%%%%%%%%%%%%%%%%%%%%%%%%
% subSUBSECTION		INTRODUCTION OF THE NOMINAL TRAJECTORY
%%%%%%%%%%%%%%%%%%%%%%%%%%%%%%%%%%%%%%%%%%%%%%%%%%%%%%%%%%%%%%%%%%%%%%%%

\subsubsection{Introduction of the Nominal Trajectory}\label{subsubsec:SS_EKF_nominal}

The time and measurement update equations developed in the previous section provide the means to compute the variation with time of the estimated state vector \nm{\lrp{\xvecest_k^-, \, \xvecest_k^+}} as well as that of the covariance of the estimation errors \nm{\lrp{\Pvec_k^-, \, \Pvec_k^+}}. However, to do so, it is necessary to define what is the nominal point \nm{\xvec_{Nk} = \xvec_N\lrp{t_k}, \, \wvec_{Nk} = \wvec_N\lrp{t_k} = \vec 0, \, \vvec_{Nk} = \vvec_N\lrp{t_k} = \vec 0} around which the dynamics system is linearized in section \ref{subsubsec:SS_SampledDataSystems_linearization} and the observations in section \ref{subsubsec:SS_SampledObservations_linearization}.

The \emph{extended Kalman filter} (\hypertt{EKF}) provides a solution to this problem that is simple but not too intuitive. The \hypertt{EKF} considers its own a priori state estimate as the nominal trajectory, this is, the nonlinear state system and observations are linearized around the \hypertt{EKF} estimate, and simultaneously that same estimate depends on the linearized system \cite{Simon2006}:
\neweq{\xvec_{Nk} = \xvec_N\lrp{t_k} = \xvecest_k^-}{eq:SS_EKF_assumption}

This assumption can be introduced into the expressions for the observations output matrix \nm{\Hvec_k} and observations input vector \nm{\zvec_k} provided by (\ref{eq:SS_measur_linear_output_matrix}) and (\ref{eq:SS_measur_linear_extra_vector}), with the results introduced into the state estimation measurement update equation (\ref{eq:SS_EKF_xest_plus_propagate}):
\begin{eqnarray}
\nm{\xvecest_k^+} & = & \nm{\xvecest_k^- + \Kvec_k \, \lrp{\yvec_k - \pderpar{\vec h}{\xvec_k}\lrp{\xvecest_k^-, \, \vec 0, \, t_k} \, \xvecest_k^- - \vec h\lrp{\xvecest_k^-, \, \vec 0, \, t_k} + \pderpar{\vec h}{\xvec_k}\lrp{\xvecest_k^-, \, \vec 0, \, t_k} \, \xvecest_k^-}}\nonumber \\
& = & \nm{\xvecest_k^- + \Kvec_k \, \big[\yvec_k - \vec h\lrp{\xvecest_k^-, \, \vec 0, \, t_k}\big]}\label{eq:SS_EKF_xest_plus_propagate_bis}
\end{eqnarray}

Note that, in contrast with (\ref{eq:SS_EKF_xest_plus_propagate}), it is no longer necessary to compute the observations input vector \nm{\zvec_k}. In order to diminish the state system linearization errors described in section \ref{subsubsec:SS_SampledDataSystems_linearization}, it is also possible to replace the state estimate time update equation (\ref{eq:SS_EKF_xest_minus_propagate}) with a zeroth order forward integration of the continuous time state system (\ref{eq:SS_cont_time_system}), with the time derivative evaluated at \nm{\xvecest_{k-1}^+}:
\neweq{\xvecest_k^- = \xvecest_{k-1}^+ + \Deltat \cdot \vec f \, \big(\xvecest_{k-1}^+, \, \uvec_{k-1}, \ \vec 0, \, t_{k-1}\big)} {eq:SS_xest_minus_propagate_bis}

An extra benefit of this approach compared with (\ref{eq:SS_EKF_xest_minus_propagate}) is that it is no longer necessary to perform the expensive computations required to evaluate \nm{\Gvec_{k-1}} (\ref{eq:SS_discr_time_system_linear_input_matrix}).

%%%%%%%%%%%%%%%%%%%%%%%%%%%%%%%%%%%%%%%%%%%%%%%%%%%%%%%%%%%%%%%%%%%%%%%%
% subSUBSECTION		EKF SUMMARY
%%%%%%%%%%%%%%%%%%%%%%%%%%%%%%%%%%%%%%%%%%%%%%%%%%%%%%%%%%%%%%%%%%%%%%%%

\subsubsection{EKF Summary}\label{subsubsec:SS_EKF_summary}

Given a continuous time nonlinear state system (\ref{eq:SS_cont_time_system}) with process noise provided by (\ref{eq:SS_cont_time_system_noise1}) and (\ref{eq:SS_cont_time_system_noise2}), together with a series of discrete time nonlinear observations (\ref{eq:SS_measur_nonlinear}) with measurement noise given by (\ref{eq:SS_measur_nonlinear_noise1}) and (\ref{eq:SS_measur_nonlinear_noise2}), and considering no correlation between both noises (\ref{eq:SS_measur_nonlinear_noise3}), it is possible to compute estimations of the state at the same time points at which the observations are provided, in such a way that their errors (difference with respect to the true state) are zero mean and have a covariance that is also computed by means of the following equations:
\begin{eqnarray}
\nm{\xvecest_0^+} & = & \nm{\vec \mu_{x,0} = E\lrsb{\xvec_0}}\label{eq:SS_EKF_x0_initial_state_FINAL} \\ 
\nm{\Pvec_0^+}    & = & \nm{E\lrsb{\lrp{\xvec_0 - \vec \mu_{x,0}} \, \lrp{\xvec_0 - \vec \mu_{x,0}}^T}}\label{eq:SS_EKF_P0_initial_covariance_FINAL} \\
\nm{\xvecest_k^-} & = & \nm{\xvecest_{k-1}^+ + \Deltat \cdot \vec f \, \big(\xvecest_{k-1}^+, \, \uvec_{k-1}, \ \vec 0, \, t_{k-1}\big)}\label{eq:SS_xest_minus_propagate_bis_FINAL} \\
\nm{\Pvec_k^-}    & = & \nm{\Fvec_{k-1} \, \Pvec_{k-1}^+ \, \Fvec_{k-1}^T + \Qtilde_{d,k-1}}\label{eq:SS_EKF_P_minus_propagate_FINAL} \\
\nm{\Kvec_k}      & = & \nm{\Pvec_k^- \, \Hvec_k^T \lrp{\Hvec_k \, \Pvec_k^- \, \Hvec_k^T + \Rtilde_k}^{-1}}\label{eq:SS_EKF_kalman_gain_FINAL} \\
\nm{\xvecest_k^+} & = & \nm{\xvecest_k^- + \Kvec_k \, \lrsb{\yvec_k - \vec h\lrp{\xvecest_k^-, \, \vec 0, \, t_k}}}\label{eq:SS_EKF_xest_plus_propagate_bis_FINAL} \\
\nm{\Pvec_k^+}    & = & \nm{\lrp{\Ivec - \Kvec_k \, \Hvec_k}\, \Pvec_k^- \, \lrp{\Ivec - \Kvec_k \, \Hvec_k}^T + \Kvec_k \, \Rtilde_k \, \Kvec_k^T}\label{eq:SS_EKF_P_plus_propagate_FINAL}
\end{eqnarray}

In the discussion that follows, \nm{\xvecest_k} is employed to refer to both the a priori and a posteriori state vector estimations \nm{\lrp{\xvecest_k^-, \, \xvecest_k^+}}, and \nm{\vec \varepsilon_k = \xvec_k - \xvecest_k} for the state estimation errors. The above equations show that \nm{\xvecest_k} is a linear combination of a random vector \nm{\xvec_0^+} plus a series of random processes \nm{\utilde_k}, so it is itself a random process, and so is \nm{\vec \varepsilon_k}.

Leaving aside for the time being the linearization and discretization errors of sections \ref{subsubsec:SS_SampledDataSystems_linearization}, \ref{subsubsec:SS_SampledDataSystems_discretization}, and \ref{subsubsec:SS_SampledObservations_linearization}, results in a problem composed by a discrete time linear state system (\ref{eq:SS_discr_time_system_linear}) and discrete time linear observations (\ref{eq:SS_measur_linear}). Provided with any user defined positive definite weighting matrix \nm{\Svec_k}, it can be proven that the solution provided verifies (\ref{eq:SS_EKF_KF_minimum}), this is, results in a state estimation that always minimizes the weighted sum of squared estimation errors \cite{Simon2006}, as long as the process and observations noises are Gaussian zero mean uncorrelated white noise processes. If they are not Gaussian, then \nm{\xvecest_k} provides the best linear (in the sense of the previous paragraph) solution to the (\ref{eq:SS_EKF_KF_minimum}) minimization, although there may be a better nonlinear solution.
\neweq{\xvecest_k = \argmin E\lrsb{\vec \varepsilon_k^T \, \Svec_k \, \vec \varepsilon_k}}{eq:SS_EKF_KF_minimum}

The errors induced by the discretization of the linear continuous time dynamics system in section \ref{subsubsec:SS_SampledDataSystems_discretization} generally do not result in significant errors as modern systems are capable of running the estimation algorithms at elevated frequencies. The system matrix \nm{\Avec\lrp{t}} and input vector \nm{\utilde\lrp{t}} in the continuous time system (\ref{eq:SS_cont_time_system_linear}) generally do not vary much during the integration interval, and hence the discretization errors are small. 

The linearization errors of sections \ref{subsubsec:SS_SampledDataSystems_linearization} and \ref{subsubsec:SS_SampledObservations_linearization} are a different story, and can induce the \hypertt{EKF} to provide unreliable estimates or even to diverge in case the nonlinearities are severe \cite{Simon2006}.

%% file: scripts/a4_estim/robust_estimators.tex
\begin{figure}[h]
\centering
\begin{tikzpicture}
\begin{axis}[
cycle list={{blue,no markers},{red,no markers},{violet, no markers},{green, no markers}},
width=6.4cm,
ticks=none,
xlabel={\nm{x}},
ymax=9.0,
ylabel={\nm{\rho\lrp{x}}},
axis lines=left,
axis line style={-stealth},
legend style={at={(0.5,0.97)},anchor=north,font=\footnotesize},
legend entries={Median, Mean, Huber, Tukey},
legend cell align=left,
]
\pgfplotstableread{figs/a4_estim/weight_rho.txt}\mytable
\addplot table [header=false, x index=0,y index=1] {\mytable};
\addplot table [header=false, x index=0,y index=2] {\mytable};
\addplot table [header=false, x index=0,y index=3] {\mytable};
\addplot table [header=false, x index=0,y index=4] {\mytable};
\end{axis}	
\end{tikzpicture}%
\hskip 10pt
\begin{tikzpicture}
\begin{axis}[
cycle list={{blue,no markers},{red,no markers},{violet, no markers},{green, no markers}},
width=6.4cm,
ticks=none,
xlabel={\nm{x}},
ymax=3.0,
ymin=-3.0,
ylabel={\nm{\Upsilon\lrp{x}}},
axis lines=left,
axis line style={-stealth},
]
\pgfplotstableread{figs/a4_estim/weight_upsilon.txt}\mytable
\addplot table [header=false, x index=0,y index=1] {\mytable};
\addplot table [header=false, x index=0,y index=2] {\mytable};
\addplot table [header=false, x index=0,y index=3] {\mytable};
\addplot table [header=false, x index=0,y index=4] {\mytable};
\end{axis}		
\end{tikzpicture}
\begin{tikzpicture}
\begin{axis}[
cycle list={{blue,no markers},{red,no markers},{violet, no markers},{green, no markers}},
width=6.4cm,
ticks=none,
xlabel={\nm{x}},
ymax=2.5,
ylabel={\nm{w\lrp{x}}},
axis lines=left,
axis line style={-stealth},
]
\pgfplotstableread{figs/a4_estim/weight_weight.txt}\mytable
\addplot table [header=false, x index=0,y index=1] {\mytable};
\addplot table [header=false, x index=0,y index=2] {\mytable};
\addplot table [header=false, x index=0,y index=3] {\mytable};
\addplot table [header=false, x index=0,y index=4] {\mytable};
\end{axis}		
\end{tikzpicture}
\caption{Robust estimators (not to scale)}
\label{fig:Error_robust_estimators}
\end{figure}

%% file: scripts/a4_estim/robust_estimators_bis.tex
\begin{figure}[h]
\centering
\begin{tikzpicture}
\begin{axis}[
cycle list={{blue,no markers},{red,no markers},{violet, no markers},{green, no markers}},
width=6.4cm,
ticks=none,
ymax=9.0,
xlabel={\nm{x^2}},
ylabel={\nm{\varrho\lrp{x^2}}},
axis lines=left,
axis line style={-stealth},
]
\pgfplotstableread{figs/a4_estim/weight_varrho.txt}\mytable
\addplot table [header=false, x index=0,y index=1] {\mytable};
\addplot table [header=false, x index=0,y index=2] {\mytable};
\addplot table [header=false, x index=0,y index=3] {\mytable};
\addplot table [header=false, x index=0,y index=4] {\mytable};
\end{axis}	
\end{tikzpicture}%
\hskip 10pt
\begin{tikzpicture}
\begin{axis}[
cycle list={{blue,no markers},{red,no markers},{violet, no markers},{green, no markers}},
width=6.4cm,
ticks=none,
ymax=2.0,
ymin=0.0,
xlabel={\nm{x^2}},
ylabel={\nm{\varrho^\prime\lrp{x^2}}},
axis lines=left,
axis line style={-stealth},
legend style={at={(0.7,0.9)},anchor=north,font=\footnotesize},
legend entries={Median, Mean, Huber, Tukey},
legend cell align=left,
]
\pgfplotstableread{figs/a4_estim/weight_varrho_prima.txt}\mytable
\addplot table [header=false, x index=0,y index=1] {\mytable};
\addplot table [header=false, x index=0,y index=2] {\mytable};
\addplot table [header=false, x index=0,y index=3] {\mytable};
\addplot table [header=false, x index=0,y index=4] {\mytable};
\end{axis}		
\end{tikzpicture}
\caption{Robust estimators (not to scale)}
\label{fig:Error_robust_estimators_bis}
\end{figure}

%% file: files_arxiv/ch03_algebra.tex
\chapter{Introduction to Lie Algebra} \label{cha:Algebra}

This chapter begins with some basic abstract and linear algebra concepts in sections \ref{sec:algebra_structures} and \ref{sec:algebra_points_and_vectors}, and then introduces Lie algebra in section \ref{sec:algebra_lie}, followed by the derivation of some useful Lie Jacobians in section \ref{sec:algebra_lie_jacobians}. Its application to the discrete integration of states is discussed in section \ref{sec:algebra_integration}, to gradient descent optimization in section \ref{sec:algebra_gradient_descent}, and to state estimation in section \ref{sec:algebra_SS}. The contents of this chapter are generic to any Lie group without making further mention to rigid bodies. It is only in chapters \ref{cha:Rotate} and \ref{cha:Motion} where the Lie theory concepts are applied first to rotational motion and then to the more generic rigid body motion.

%%%%%%%%%%%%%%%%%%%%%%%%%%%%%%%%%%%%%%%%%%%%%%%%%%%%%%%%%%%%%%%%%%%%%%%%
%%%%%%%%%%%%%%%%%%%%%%%%%%%%%%%%%%%%%%%%%%%%%%%%%%%%%%%%%%%%%%%%%%%%%%%%
%%%%%%%%%%%%%%%%%%%%%%%%%%%%%%%%%%%%%%%%%%%%%%%%%%%%%%%%%%%%%%%%%%%%%%%%
% SECTION      ALGEBRAIC STRUCTURES, MAPS, AND METRIC SPACES
%%%%%%%%%%%%%%%%%%%%%%%%%%%%%%%%%%%%%%%%%%%%%%%%%%%%%%%%%%%%%%%%%%%%%%%%
%%%%%%%%%%%%%%%%%%%%%%%%%%%%%%%%%%%%%%%%%%%%%%%%%%%%%%%%%%%%%%%%%%%%%%%%
%%%%%%%%%%%%%%%%%%%%%%%%%%%%%%%%%%%%%%%%%%%%%%%%%%%%%%%%%%%%%%%%%%%%%%%%

\section{Algebraic Structures, Maps, and Metric Spaces}\label{sec:algebra_structures}

In algebra, a \emph{set} is a well defined collection of objects, named elements, while an \emph{operation} \nm{\ast} is a uniquely defined rule that assigns to each ordered pair of elements exactly a third element \nm{\lrb{\ast : \mathbb{A} \times \mathbb{B} \rightarrow \mathbb{C} \ | \ a \ast b = c \in \mathbb{C}, \forall \ a \in \mathbb{A}, \forall \ b \in \mathbb{B}}} \cite{Pinter1990}. Although an operation may involve up to three different sets \nm{\lrp{\mathbb{A}, \mathbb{B}, \mathbb{C}}}, often two or even the three of them coincide. A set \nm{\mathbb{A}} is a \emph{subset} of a set \nm{\mathbb{B}} if all elements of \nm{\mathbb{A}} are also elements of \nm{\mathbb{B}}. An \emph{algebraic structure} is a combination of a set and one or multiple operations that complies with certain axioms.

A set \nm{\mathbb{A}} has \emph{group} structure under operation \nm{\lrb{\ast: \mathbb{A} \times \mathbb{A} \rightarrow \mathbb{A}}} if it complies with the following four axioms \nm{\forall \ a, b, c \in \mathbb{A}} \cite{Pinter1990}:
\begin{enumerate}
\item Closure: \nm{a \ast b \in \mathbb{A}}
\item Associativity: \nm{\lrp{a \ast b} \ast c = a \ast \lrp{b \ast c}}
\item Identity: \nm{\exists \ e \in \mathbb{A} \ | \ e \ast a = a \ast e =  a}
\item Inverse: \nm{\exists \ f \in \mathbb{A} \ | \ f \ast a = a \ast f = e}
\end{enumerate}

An \emph{abelian group} is that which in addition also complies with commutativity \nm{\lrb{a \ast b = b \ast a}}. A set \nm{\mathbb{A}} has \emph{ring} structure under two operations, usually named addition \nm{\lrb{+: \mathbb{A} \times \mathbb{A} \rightarrow \mathbb{A}}} and multiplication \nm{\lrb{\cdot: \mathbb{A} \times \mathbb{A} \rightarrow \mathbb{A}}}, if, in addition to being an abelian group under addition, complies with the following four axioms \nm{\forall \ a, b, c \in \mathbb{A}} \cite{Pinter1990}: 
\begin{enumerate}
\item Closure of \nm{\cdot} : \nm{a \cdot b \in \mathbb{A}}
\item Associativity of \nm{\cdot} : \nm{\lrp{a \cdot b} \cdot c = a \cdot \lrp{b \cdot c}}
\item Distributivity of \nm{\cdot} with respect to \nm{+} : \nm{a \cdot \lrp{b + c} = a \cdot b + a \cdot c, \ \lrp{a + b} \cdot c = a \cdot c + b \cdot c}
\item Identity of \nm{\cdot} : \nm{\exists \ 1 \in \mathbb{A} \ | \ 1 \cdot a = a \cdot 1 =  a}
\end{enumerate}

An \emph{abelian ring} is that which in addition also complies with commutativity over multiplication \nm{\lrb{a \cdot b = b \cdot a}}. Note that by convention, the identity and inverse of addition are denoted 0 and \nm{-a}, respectively, while those of multiplication are denoted 1 and \nm{a^{-1}}. A set \nm{\mathbb{A}} has \emph{field} structure under operations \nm{+} and \nm{\cdot} if \nm{\mathbb{A}} is an abelian group under \nm{+} and \nm{\mathbb{A} - \lrb{0}} (the set \nm{\mathbb{A}} without the additive identity 0) is an abelian group under \nm{\cdot} \cite{Pinter1990}. In an \emph{ordered field}, the implementation of the addition and multiplication operations enables determining if one element is greater, equal, or lower than a second element. The set of real numbers \nm{\mathbb{R}} endowed with the operations of addition \nm{+} and multiplication \nm{\cdot} forms an ordered field, known as the field of real numbers \nm{\langle \mathbb{R}, +, \cdot \rangle}, nearly always abbreviated to simply \nm{\mathbb{R}}.

A \emph{topological space} is an ordered pair \nm{\lrp{\mathbb{A}, \mathbb{\tau}}}, where \nm{\mathbb{A}} is a set and \nm{\mathbb{\tau}} is a collection of subsets of \nm{\mathbb{A}}, satisfying the following axioms \cite{Armstrong1983}:
\begin{enumerate}
\item The empty set and \nm{\mathbb{A}} itself belong to \nm{\mathbb{\tau}}.
\item Any arbitrary (finite or infinite) union of members of \nm{\mathbb{\tau}} still belongs to \nm{\mathbb{\tau}}.
\item The intersection of any finite number of members of \nm{\mathbb{\tau}} still belongs to \nm{\mathbb{\tau}}.
\end{enumerate}

The elements of \nm{\mathbb{\tau}} are called open sets and the collection \nm{\mathbb{\tau}} is called a topology on \nm{\mathbb{A}}. Topological spaces comprise the most general notion of a mathematical space; all other spaces defined below are specializations with extra structure or constraints.

A \emph{vector space} (\emph{linear space}) over a field \nm{\mathbb{F}} is a set \nm{\mathbb{V}} together with two operations, addition \nm{\lrb{+ : \mathbb{V} \times \mathbb{V} \rightarrow \mathbb{V}}} and scalar multiplication \nm{\lrb{\cdot : \mathbb{F} \times \mathbb{V} \rightarrow \mathbb{V}}} that, in addition of \nm{\mathbb{V}} being an abelian group under +, satisfies the following axioms \nm{\forall \ u, v \in \mathbb{V}} and \nm{\forall \ a, b \in \mathbb{F}} \cite{Shuster1993, Roman2005}. Elements of \nm{\mathbb{F}} are called scalars, while those of \nm{\mathbb{V}} vectors.
\begin{enumerate}
\item Closure of \nm{\cdot} : \nm{a \cdot u \in \mathbb{V}}
\item Compatibility of \nm{\cdot} with field \nm{\cdot} : \nm{a \cdot \lrp{b \cdot v} = \lrp{a \cdot b} \cdot  v}
\item Identity of \nm{\cdot} : \nm{1 \cdot v = v}, where 1 denotes the field \nm{\cdot} identity.
\item Distributivity of \nm{\cdot} with respect to \nm{+} : \nm{a \cdot \lrp{u + v} = a \cdot u + a \cdot v}
\item Distributivity of \nm{\cdot} with respect to field \nm{+} : \nm{\lrp{a + b} \cdot v = a \cdot v + b \cdot v}
\end{enumerate}

A \emph{map} or \emph{morphism} is a rule that to every element in a set \nm{\mathbb{A}} assigns a unique element in a different set \nm{\mathbb{B}} \nm{\lrb{f : \mathbb{A} \rightarrow \mathbb{B} \ | \ f\lrp{a} = b \in \mathbb{B}, \forall \ a \in \mathbb{A}}} \cite{Pinter1990}. A map is \emph{injective} if each element in \nm{\mathbb{B}} is the image or map output of no more than one element of \nm{\mathbb{A}}, \emph{surjective} if each element in \nm{\mathbb{B}} is the image of at least one element of \nm{\mathbb{A}}, and \emph{bijective} is the map is simultaneously injective and surjective.

A \emph{homomorphism} is a structure preserving map between two algebraic structures of the same type (groups, rings, fields, vector spaces, etc.) \cite{Pinter1990}. Note that neither the sets nor the operations of the structures need to coincide, and that a homomorphism preserves every operation contained in the algebraic structures. In the case of groups, considering a homomorphism \nm{\lrb{f : \langle \mathbb{A}, \ + \rangle \rightarrow \langle \mathbb{B}, \ \cdot \rangle}}, it complies with all group axioms \nm{\forall \ a, b, c \in \mathbb{A}}:
\begin{enumerate}
\item Closure: \nm{f\lrp{a + b} = f\lrp{a} \cdot f\lrp{b}}
\item Associativity: \nm{f\big(\lrp{a + b} + c\big) = f\big(a + \lrp{b + c}\big) = \big(f\lrp{a} \cdot f\lrp{b}\big) \cdot f\lrp{c} = f\lrp{a} \cdot \big(f\lrp{b} \cdot f\lrp{c}\big)}
\item Identity: \nm{f\lrp{0} = 1 \rightarrow  f\lrp{a} = f\lrp{a + 0} = f\lrp{0 + a} = f\lrp{a} \cdot f\lrp{0} = f\lrp{0} \cdot f\lrp{a} = f\lrp{a} \cdot 1 = 1 \cdot f\lrp{a}}
\item Inverse: \nm{f\lrp{-a} = f^{-1}\lrp{a} \rightarrow  f\lrp{0} = f\lrp{a - a} = f\lrp{a} \cdot f\lrp{-a} = f\lrp{a} \cdot f^{-1}\lrp{a} = f^{-1}\lrp{a} \cdot f\lrp{a} = 1}
\end{enumerate}

Homomorphisms for other algebraic structures are defined similarly. A bijective homomorphism is known as an \emph{isomorphism} \cite{Pinter1990}. A \emph{metric} is an operation between two elements of the same set onto a field \nm{\lrb{d : \mathbb{V} \times \mathbb{V} \rightarrow \mathbb{F}}}. It defines the concept of distance between any two members of the set and complies with the following axioms \nm{\forall \ u, v, w \in \mathbb{V}}:
\begin{enumerate}
\item Identity of indiscernibles: \nm{d\lrp{u, v} = 0 \Leftrightarrow u = v}
\item Symmetry: \nm{d\lrp{u, v} = d\lrp{v, u}}
\item Subadditivity: \nm{d\lrp{u, w} \leq d\lrp{u, v} + d\lrp{v, w}}
\end{enumerate}

Based on these three axioms, it is straightforward to prove that \nm{d\lrp{u,v} \geq 0 \ \forall \ u, v \in \mathbb{V}}. The most common metric is the \emph{inner product} \nm{\lrb{\langle \cdot \, , \cdot \rangle: \mathbb{V} \times \mathbb{V} \rightarrow \mathbb{F}}}, usually associated to a vector space, which satisfies the following three axioms \nm{\forall \ u, v, w \in \mathbb{V}} and \nm{\forall \ a, b \in \mathbb{F}} \cite{Shuster1993}:
\begin{enumerate}
\item Commutativity: \nm{\langle u, v\rangle = \langle v, u\rangle}
\item Linearity with respect to \nm{+} and \nm{\cdot} : \nm{\langle u, a \cdot v + b \cdot w\rangle = a \cdot \langle u, v\rangle + b \cdot \langle u, w\rangle}
\item Positive definiteness: \nm{\langle v, v\rangle \geq 0} and \nm{\langle v, v\rangle = 0 \Leftrightarrow v = 0}
\end{enumerate}

A \emph{metric space} is a combination of a set with a metric on that same set \cite{Bryant1985}, while an \emph{inner product space} restricts the definition to the case of a vector space \nm{\mathbb{V}} over a field \nm{\mathbb{F}} endowed with an inner product metric over the same field \nm{\mathbb{F}} \cite{Jain1995}. An \emph{Euclidean space} is a finite-dimensional inner product space over the field of the real numbers \nm{\mathbb{R}} \cite{Artin1957}. 

A \emph{group action} on a space is a group homomorphism of a given group \nm{\langle \mathbb{A}, \ast \rangle} into the group of transformations of the space \nm{\lrb{g\lrp{}: \mathbb{A} \times \mathbb{V} \rightarrow \mathbb{V} \ | \ g_a\lrp{u} = v \in \mathbb{V}, \forall \ a \in \mathbb{A}, \forall \ u \in \mathbb{V}}},  and needs to verify two axioms \nm{\forall \ a, b \in \mathbb{A}, \forall \ u \in \mathbb{V}}:
\begin{enumerate}
\item Identity: \nm{g_e\lrp{u} = u}, where \emph{e} is the identity of \nm{\mathbb{G}}.
\item Compatibility: \nm{g_{a \ast b}\lrp{u} = g_a\big(g_b\lrp{u}\big)}
\end{enumerate}

Returning to the case of inner product spaces, two vectors are \emph{orthogonal} if their inner product is zero, while the length or \emph{norm} of a vector is \nm{\| v \| = \sqrt{\langle v, v\rangle}}. A unit vector is that whose norm is one. An inner product space can also be endowed with an additional operation, the \emph{cross product} \nm{\lrb{\times : \mathbb{V} \times \mathbb{V} \rightarrow \mathbb{V}}}, which complies with the following axioms \cite{Shuster1993}:
\begin{enumerate}
\item Anti commutativity: \nm{u \times v = - v \times u}
\item Compatibility with \nm{\cdot} : \nm{\lrp{a \cdot u} \times v = u \times \lrp{a \cdot v} = a \cdot \lrp{u \times v}}
\item Distributivity with + : \nm{\lrp{u + v} \times w = \lrp{u \times w} + \lrp{v \times w}}
\end{enumerate}

It can also be quickly derived that \nm{\langle u \times v , u\rangle = \langle u \times v , v\rangle = 0}. 

%%%%%%%%%%%%%%%%%%%%%%%%%%%%%%%%%%%%%%%%%%%%%%%%%%%%%%%%%%%%%%%%%%%%%%%%
%%%%%%%%%%%%%%%%%%%%%%%%%%%%%%%%%%%%%%%%%%%%%%%%%%%%%%%%%%%%%%%%%%%%%%%%
%%%%%%%%%%%%%%%%%%%%%%%%%%%%%%%%%%%%%%%%%%%%%%%%%%%%%%%%%%%%%%%%%%%%%%%%
% SECTION      POINTS, VECTORS, and AXES
%%%%%%%%%%%%%%%%%%%%%%%%%%%%%%%%%%%%%%%%%%%%%%%%%%%%%%%%%%%%%%%%%%%%%%%%
%%%%%%%%%%%%%%%%%%%%%%%%%%%%%%%%%%%%%%%%%%%%%%%%%%%%%%%%%%%%%%%%%%%%%%%%
%%%%%%%%%%%%%%%%%%%%%%%%%%%%%%%%%%%%%%%%%%%%%%%%%%%%%%%%%%%%%%%%%%%%%%%%

\section{Points, Vectors, and Axes}\label{sec:algebra_points_and_vectors}

Note that in the abstract discussion above it has not yet been defined what a vector is. This section focuses on the three-dimensional Euclidean space \nm{\mathbb{E}^3} \cite{Soatto2001}, which can be represented by a Cartesian frame, where every \emph{point} \nm{\vec p \in \mathbb{E}^3} can be identified by its three coordinates \nm{\vec p = \lrsb{p_1 \ \ p_2 \ \ p_3}^T \in \mathbb{R}^3}. A \emph{vector} in \nm{\mathbb{E}^3} is defined by a pair of points \nm{\vec p, \vec q \in \mathbb{E}^3} with a directed arrow connecting \nm{\vec p} to \nm{\vec q}, where the vector \nm{\vec v = \lrsb{v_1 \ \ v_2 \ \ v_3}^T \in \mathbb{R}^3} is a triplet of numbers, each one being the difference between the corresponding coordinates of the two points \nm{\vec q} and \nm{\vec p} (\nm{\vec v = \vec q - \vec p \in \mathbb{R}^3}). Although they share notation, points and vectors are different geometric objects. A \emph{free vector} is one that does not depend on its starting or base point.

The set of all free vectors in \nm{\mathbb{R}^3} form an inner product space with cross product over the field of real numbers \nm{\mathbb{R}}, with both products defined as follows by making use of matrix notation:
\begin{eqnarray}
\nm{\langle \vec u , \vec v \rangle} & = & \nm{\vec u \cdot \vec v = {\vec u}^T \, \vec v = u_1 \, v_1 + u_2 \, v_2 +  u_3 \, v_3} \label{eq:SO3_inner_product} \\
\nm{\vec u \times \vec v} & = & \nm{\widehat{\vec u} \; \vec v = \begin{bmatrix} \nm{0} & \nm{- u_3} & \nm{+ u_2} \\ \nm{+ u_3} & \nm{0} & \nm{- u_1} \\ \nm{- u_2} & \nm{+ u_1} & \nm{0} \end{bmatrix} \begin{bmatrix} \nm{v_1} \\ \nm{v_2} \\ \nm{v_3} \end{bmatrix} = \begin{bmatrix} \nm{u_2 \, v_3 - u_3 \, v_2} \\ \nm{u_3 \, v_1 - u_1 \, v_3} \\ \nm{u_1 \, v_2 - u_2 \, v_1} \end{bmatrix} = - \vec v \times \vec u = - \widehat{\vec v} \; \vec u} \label{eq:SO3_cross_product} 
\end{eqnarray}

where \nm{{\vec v}^T} is the transpose of \nm{\vec v} and \nm{\widehat{\vec v}} its skew-symmetric form\footnote{An skew-symmetric matrix is one whose negative equals its transpose.}. The inner product or Euclidean metric can measure distances and angles, while the cross product defines orientation.

Any vector \nm{\vec v} in \nm{\mathbb{R}^3} can be written as \nm{\vec v = v_1 \ \vec e_1 + v_2 \ \vec e_2 + v_3 \ \vec e_3}, where \nm{\vec e_1}, \nm{\vec e_2}, and \nm{\vec e_3} are the three linearly independent basis vectors and \nm{v_1, \ v_2, \ v_3} the coordinates or components of \nm{\vec v} with respect to that basis \cite{Strang2006}. The \emph{basis} is called orthogonal if \nm{{\vec e_i}^T \ \vec e_j = 0} when \nm{i \neq j}, orthonormal if additionally \nm{{\vec e_i}^T \ \vec e_j = 1} when \nm{i = j}, and right handed if additionally \nm{\epsilon_{ijk} = {\vec e_i}^T \ \widehat{\vec e}_j \ \vec e_k} is 1 for \nm{\epsilon_{123}}, \nm{\epsilon_{231}}, and \nm{\epsilon_{312}}, -1 for \nm{\epsilon_{132}}, \nm{\epsilon_{213}}, and \nm{\epsilon_{321}}, and 0 in all other cases \cite{Shuster1993}.

An \emph{axis} or \emph{line} is defined by its direction \nm{\vec n} (provided by a free vector) and a point \nm{\vec p} that it passes through. Its coordinates are \nm{\lrp{\vec n, \, \vec m} \in \mathbb{R}^6}, where \nm{ \vec m = \widehat{\vec p} \, \vec n} is called the \emph{moment} of the line. The coordinates \nm{\lrp{\vec n, \, \vec m}} are independent of \nm{\vec p}. The moment \nm{\vec m} is normal to the plane through the line and the origin with norm equal to the distance from the line to the origin. The point belonging to the line that is closest to the origin responds to \nm{\vec p_{\perp} = \widehat{\vec n} \, \vec m}. A line has four degrees of freedom and hence two redundancies, provided by \nm{\vec n} being a direction and hence a unit vector (\nm{\|\vec n\| = 1}) and \nm{\vec n} being orthogonal to \nm{\vec m} by definition (\nm{\vec n^T \, \vec m = 0}) \cite{Jia2013}.

It is important to remark that although this is the formal definition of an axis, in this document unless otherwise specified an axis is synonymous with just a direction \nm{\vec n} with two degrees of freedom (\nm{\|\vec n\| = 1}), passing through the origin \nm{\lrp{\vec p = \vec 0 \rightarrow \vec m = \vec 0}}. The reason is that in most occasions it is convenient to consider that a rigid body rotates about the origin of the reference frame representing it.

%%%%%%%%%%%%%%%%%%%%%%%%%%%%%%%%%%%%%%%%%%%%%%%%%%%%%%%%%%%%%%%%%%%%%%%%
%%%%%%%%%%%%%%%%%%%%%%%%%%%%%%%%%%%%%%%%%%%%%%%%%%%%%%%%%%%%%%%%%%%%%%%%
%%%%%%%%%%%%%%%%%%%%%%%%%%%%%%%%%%%%%%%%%%%%%%%%%%%%%%%%%%%%%%%%%%%%%%%%
% SECTION      LIE GROUPS AND LIE ALGEBRAS
%%%%%%%%%%%%%%%%%%%%%%%%%%%%%%%%%%%%%%%%%%%%%%%%%%%%%%%%%%%%%%%%%%%%%%%%
%%%%%%%%%%%%%%%%%%%%%%%%%%%%%%%%%%%%%%%%%%%%%%%%%%%%%%%%%%%%%%%%%%%%%%%%
%%%%%%%%%%%%%%%%%%%%%%%%%%%%%%%%%%%%%%%%%%%%%%%%%%%%%%%%%%%%%%%%%%%%%%%%

\section{Lie Groups and Lie Algebras}\label{sec:algebra_lie}

A \emph{manifold} is a topological space that locally resembles Euclidean space near each element, so each element of an \emph{m} dimensional manifold has a neighborhood that is homeomorphic\footnote{A homeomorphism or topological isomorphism is a continuous function between topological spaces that has a continuous inverse function.} to the \emph{m} dimensional Euclidean space \cite{Gamelin1999}. Manifolds, which are embedded in spaces of higher dimension, are curved, smooth (hyper) surfaces with no edges or spikes; they are defined by the constraints imposed on the state \cite{Sola2018}, this is, the state vector is restricted to moving within the manifold. 

A \emph{Lie group} \nm{\langle \mathcal{G}, \circ \rangle} is a smooth manifold whose elements satisfy the group axioms. They combine the local properties of smooth manifolds, enabling the use of calculus, with the global properties of groups, allowing the nonlinear composition of distant objects \cite{Sola2018}. Elements of \nm{\mathcal{G}} are denoted with \nm{\mathcal{X}}, the identity with \nm{\mathcal E}, and the inverse with \nm{\mathcal{X}^{-1}}. As in any other group, Lie groups are capable of transforming elements of other sets by means of their actions. In this sense, the group operation \nm{\lrb{\circ: \mathcal{G} \times \mathcal{G} \rightarrow \mathcal{G}}}, generally called \emph{composition}, can be considered as an action of the group on itself.

If \nm{\mathcal{X}\lrp{t}} is an element or point of the Lie group moving on the manifold, its derivative with time belongs to the space tangent to \nm{\mathcal{G}} at \nm{\mathcal{X}}, denoted by \nm{T_{\mathcal X}\mathcal G}. There exists a unique tangent space at each point, but the structure of such tangent spaces is the same everywhere \cite{Sola2018}. The \emph{tangent space} at a point is a real vector space of the same dimension as the manifold that intuitively contains all the possible directions in which one can tangentially pass through the point.

The \emph{Lie algebra} \nm{\mathfrak{m}} is defined as the tangent space at the identity \nm{\lrb{\mathfrak{m} = T_{\mathcal E}\mathcal G}}, and it is a vector space whose elements can be identified with vectors in \nm{\mathbb{R}^m}, with \emph{m} being the number of degrees of freedom of the Lie group \nm{\mathcal G} \cite{Sola2018}. Elements of the tangent space are usually denoted \nm{\vec v} when referring to velocities and \nm{\vec \tau = \vec v \cdot t} for more general elements. Lie algebras can be defined at any manifold point \nm{\mathcal{X}}, establishing local coordinates for \nm{T_{\mathcal X}\mathcal G}, and its elements are denoted by the \nm{<\cdot^{\wedge}>} symbol, such as \nm{\vec v^{\mathcal{X}\wedge} \in T_{\mathcal X}\mathcal G} or \nm{\vec \tau^{\mathcal{E}\wedge} \in T_{\mathcal E}\mathcal G}.

As all tangent spaces or Lie algebras have the same structure no matter the position of the element \nm{\mathcal X} within the Lie group \nm{\mathcal G}, it can be considered with no loss of generality\footnote{This is just a convention, and the opposite one is employed in some texts.} that all group  actions at \nm{\mathcal X \in \mathcal G}, noted as \nm{g_{\mathcal X}()}, transform elements viewed in the \emph{local} or \emph{body} frame represented by \nm{\mathcal X} into the \emph{global} or \emph{space} frame represented by \nm{\mathcal E} \cite{Sola2018}. The opposite is true for the inverse operations noted as \nm{g_{\mathcal X}^{-1}() = g_{\mathcal{X}^{-1}}()}. When the group action is the composition \nm{\circ} itself, elements to the right of \nm{\mathcal X} belong to the body frame, while those on the left are viewed on the global frame.

%%%%%%%%%%%%%%%%%%%%%%%%%%%%%%%%%%%%%%%%%%%%%%%%%%%%%%%%%%%%%%%%%%%%%%%%
%%%%%%%%%%%%%%%%%%%%%%%%%%%%%%%%%%%%%%%%%%%%%%%%%%%%%%%%%%%%%%%%%%%%%%%%
% SUBSECTION      LIE ALGEBRA VELOCITIES, HAT AND VEE OPERATORS
%%%%%%%%%%%%%%%%%%%%%%%%%%%%%%%%%%%%%%%%%%%%%%%%%%%%%%%%%%%%%%%%%%%%%%%%
%%%%%%%%%%%%%%%%%%%%%%%%%%%%%%%%%%%%%%%%%%%%%%%%%%%%%%%%%%%%%%%%%%%%%%%%

\subsection{Lie Algebra Velocities, Hat and Vee Operators}\label{subsec:algebra_lie_velocities}

The structure of the Lie algebra can be obtained by time derivating the group inverse constraint \cite{Sola2018}. Every Lie group employed in this document (refer to chapters \ref{cha:Rotate} and \ref{cha:Motion}) has a \nm{\circ} composition operator realized by some type of multiplication. If this is the case, the group inverse constraint responds to \nm{\mathcal X \circ \mathcal X^{-1} = \mathcal X^{-1} \circ \mathcal X = \mathcal E}, and its derivation with time leads to the Lie algebra \emph{velocities} viewed in either the body or local frames\footnote{Note that \nm{\mathcal X^{\dot{-}1}} represents the time derivative of the inverse, not the inverse of the time derivative.}:
\begin{eqnarray}
\nm{\dot{\mathcal X} \circ \mathcal X^{-1} + \mathcal X \circ \mathcal X^{\dot{-}1} = 0} & \nm{\rightarrow} & \nm{\vec v^{\mathcal{E}\wedge} = \dot{\mathcal X} \circ \mathcal X^{-1} = - \mathcal X \circ \mathcal X^{\dot{-}1}} \label{eq:algebra_vE} \\
\nm{\mathcal X^{-1} \circ \dot{\mathcal X} + \mathcal X^{\dot{-}1} \circ \mathcal X = 0} & \nm{\rightarrow} & \nm{\vec v^{\mathcal{X}\wedge} = \mathcal X^{-1} \circ \dot{\mathcal X} = - \mathcal X^{\dot{-}1} \circ \mathcal X} \label{eq:algebra_vX}
\end{eqnarray}

The \nm{\vec \tau^\wedge} or \nm{\vec v^\wedge} elements of the Lie algebra hence do not have trivial structures but can always be expressed as linear combinations of some base elements \nm{\vec e_i}, which are called the \emph{generators} of \nm{\mathfrak{m}} \cite{Sola2018}. It is generally more convenient to manipulate them as vectors \nm{\vec \tau \in \mathbb R^m}, as they can then be grouped together in larger state vectors and operated by means of linear algebra. The isomorphisms that linearly convert between them are called \emph{hat} \nm{\lrb{\cdot^\wedge: \mathbb{R}^m \rightarrow \mathfrak{m} \ | \ \vec \tau \rightarrow \vec \tau^\wedge}} and \emph{vee} \nm{\lrb{\cdot^\vee: \mathfrak{m} \rightarrow \mathbb{R}^m \ | \ \lrp{\vec \tau^\wedge}^\vee \rightarrow \vec \tau}}.

%%%%%%%%%%%%%%%%%%%%%%%%%%%%%%%%%%%%%%%%%%%%%%%%%%%%%%%%%%%%%%%%%%%%%%%%
%%%%%%%%%%%%%%%%%%%%%%%%%%%%%%%%%%%%%%%%%%%%%%%%%%%%%%%%%%%%%%%%%%%%%%%%
% SUBSECTION      EXPONENTIAL AND LOGARITHMIC MAPS, PLUS AND MINUS OPERATORS
%%%%%%%%%%%%%%%%%%%%%%%%%%%%%%%%%%%%%%%%%%%%%%%%%%%%%%%%%%%%%%%%%%%%%%%%
%%%%%%%%%%%%%%%%%%%%%%%%%%%%%%%%%%%%%%%%%%%%%%%%%%%%%%%%%%%%%%%%%%%%%%%%

\subsection{Exponential and Logarithmic Maps, Plus and Minus Operators}\label{subsec:algebra_lie_exp_plus}

The \emph{exponential map} \nm{\lrb{exp\lrp{} : \mathfrak{m} \rightarrow \mathcal{G} \ | \ \mathcal{X} = exp\lrp{\vec \tau^\wedge}}} wraps the tangent element around the manifold following the geodesic or minimum distance line, effectively converting elements of the Lie algebra into those of the manifold or Lie group. The unwrapping or inverse operation is the \emph{logarithmic map} \nm{\lrb{log\lrp{} : \mathcal{G} \rightarrow \mathfrak{m} \ | \ \vec \tau^\wedge = log\lrp{\mathcal X}}}. The hat and vee operators can be incorporated into these maps, resulting in the \emph{capitalized maps} \nm{\lrb{Exp\lrp{} : \mathbb{R}^m \rightarrow \mathcal{G} \ | \ \mathcal{X} = Exp\lrp{\vec \tau}}} and \nm{\lrb{Log\lrp{} : \mathcal{G} \rightarrow \mathbb{R}^m \ | \ \vec \tau = Log\lrp{\mathcal X}}}. Note that the exponential map complies with the following properties \nm{\forall \ t \in \mathbb R} \cite{Sola2018}:
\begin{eqnarray}
\nm{exp\lrp{t \, \vec \tau^{\wedge}}} & = & \nm{exp\lrp{\vec \tau^\wedge}^t} \label{eq:algebra_power} \\
\nm{exp\lrp{\mathcal X \circ \vec \tau^{\wedge} \circ \mathcal{X}^{-1}}} & = & \nm{\mathcal X \circ exp\lrp{\vec \tau^{\wedge}} \circ \mathcal{X}^{-1}} \label{eq:algebra_exp}
\end{eqnarray}

The \emph{plus} and \emph{minus operators} enable operating with increments of the nonlinear manifold expressed in the corresponding linear tangent vector space \cite{Sola2018}. As the Lie group \nm{\mathcal G} is not abelian, there exist right \nm{\oplus} and \nm{\ominus} operators as well as left \nm{\boxplus} and \nm{\boxminus} ones. Note that the addition of (usually) small perturbations \nm{\Delta \vec \tau} to a given manifold \nm{\mathcal X} result in a perturbed manifold \nm{\mathcal Y}:
\begin{eqnarray}
\nm{\mathcal Y} & = & \nm{\mathcal X \oplus \Delta \vec \tau^{\mathcal X} = \mathcal X \circ Exp\lrp{\Delta \vec \tau^{\mathcal X}} \in \mathcal G} \label{eq:algebra_plus_right} \\
\nm{\Delta \vec \tau^{\mathcal X}} & = & \nm{\mathcal Y \ominus \mathcal X = Log\lrp{\mathcal X^{-1} \circ \mathcal Y} \in T_{\mathcal X}\mathcal G}\label{eq:algebra_minus_right} \\
\nm{\mathcal Y} & = & \nm\Delta {\vec \tau^{\mathcal E} \boxplus \mathcal X = Exp\lrp{\Delta \vec \tau^{\mathcal E}} \circ \mathcal X \in \mathcal G} \label{eq:algebra_plus_left} \\ 
\nm{\Delta \vec \tau^{\mathcal E}} & = & \nm{\mathcal Y \boxminus \mathcal X = Log\lrp{\mathcal Y \circ \mathcal X^{-1}} \in T_{\mathcal E}\mathcal G}\label{eq:algebra_minus_left} 
\end{eqnarray}

%%%%%%%%%%%%%%%%%%%%%%%%%%%%%%%%%%%%%%%%%%%%%%%%%%%%%%%%%%%%%%%%%%%%%%%%
%%%%%%%%%%%%%%%%%%%%%%%%%%%%%%%%%%%%%%%%%%%%%%%%%%%%%%%%%%%%%%%%%%%%%%%%
% SUBSECTION      ADJOINT ACTION
%%%%%%%%%%%%%%%%%%%%%%%%%%%%%%%%%%%%%%%%%%%%%%%%%%%%%%%%%%%%%%%%%%%%%%%%
%%%%%%%%%%%%%%%%%%%%%%%%%%%%%%%%%%%%%%%%%%%%%%%%%%%%%%%%%%%%%%%%%%%%%%%%

\subsection{Adjoint Action}\label{subsec:algebra_lie_adjoint}

The vectors or elements of the tangent space at \nm{\mathcal{X}} can be transformed to the tangent space at the identity \nm{\mathcal E} by means of the \emph{adjoint}
\nm{\lrb{\vec{Ad}\lrp{}: \mathcal{G} \times \mathfrak{m} \rightarrow \mathfrak{m}}} \cite{Sola2018}. The adjoint is hence an action of the Lie group that operates on its own Lie algebra. The adjoint action can be obtained by the equivalence of the perturbed state \nm{\mathcal Y} in (\ref{eq:algebra_plus_right}, \ref{eq:algebra_plus_left}) by means of (\ref{eq:algebra_exp}):
\neweq{\vec \tau^{\mathcal{E}\wedge} = \vec{Ad}_{\mathcal X}\lrp{\vec \tau^{\mathcal{X}\wedge}} = \mathcal X \circ \vec \tau^{\mathcal{X}\wedge} \circ \mathcal{X}^{-1}} {eq:algebra_adjoint}

The adjoint action is a linear homomorphism, and hence complies with the following expressions \nm{\forall \ \mathcal{X}}, \nm{\mathcal{Y} \in \mathcal{G}}, \nm{\forall \ a, b \in \mathbb{R}}, \nm{\forall \ \vec \tau^{\mathcal{X}\wedge}, \vec \sigma^{\mathcal{X}\wedge} \in T_{\mathcal X}\mathcal G}, and \nm{\forall \ \vec \tau^{\mathcal{Y}\wedge} \in T_{\mathcal Y}\mathcal G}:
\begin{eqnarray}
\nm{\vec{Ad}_{\mathcal X}\lrp{a \, \vec \tau^{\mathcal{X}\wedge} + b \, \vec \sigma^{\mathcal{X}\wedge}}} & = & \nm{a \, \vec{Ad}_{\mathcal X}\lrp{\vec \tau^{\mathcal{X}\wedge}} + b \, \vec{Ad}_{\mathcal X}\lrp{\vec \sigma^{\mathcal{X}\wedge}}} \label{eq:algebra_adjoint_linear} \\
\nm{\vec{Ad}_{\mathcal X}\lrp{\vec{Ad}_{\mathcal Y}\lrp{\vec \tau^{\mathcal{Y}\wedge}}}} & = & \nm{\vec{Ad}_{\mathcal X \circ \mathcal Y}\lrp{\vec \tau^{\mathcal{Y}\wedge}}} \label{eq:algebra_adjoint_homo}
\end{eqnarray}

As the adjoint is a linear transform, it is always possible to obtain an equivalent matrix operator, the \emph{adjoint matrix} \nm{\lrb{\vec{Ad}: \mathcal{G} \times \mathbb{R}^m \rightarrow \mathbb{R}^m \ | \ \vec{Ad}_{\mathcal X} \cdot \vec \tau = \lrp{\mathcal X \, \vec \tau^{\wedge} \, \mathcal{X}^{-1}}^{\vee}, \ \vec{Ad}_{\mathcal X} \in \mathbb{R}^{mxm}}}, that maps the Cartesian tangent space vectors instead of the Lie algebra elements \cite{Sola2017}. Both maps share the same symbols but are easily distinguished by context.
\neweq{\vec \tau^{\mathcal{E}} = \vec{Ad}_{\mathcal X} \cdot \vec \tau^{\mathcal{X}} = \lrp{\mathcal X \, \vec \tau^{{\mathcal X}\wedge} \, \mathcal{X}^{-1}}^{\vee}  } {eq:algebra_adjoint_matrix}

The adjoint matrix complies with the following properties:
\begin{eqnarray}
\nm{\mathcal{X} \oplus \vec \tau^{\mathcal X}} & = & \nm{\lrp{\vec{Ad}_{\mathcal X} \cdot \vec \tau^{\mathcal{X}}} \boxplus \mathcal{X}} \label{eq:algebra_adjoint_matrix_general} \\
\nm{\vec{Ad}_{\mathcal X^{-1}}} & = & \nm{\vec{Ad}_{\mathcal X}^{-1}} \label{eq:algebra_adjoint_matrix_inverse} \\
\nm{\vec{Ad}_{\mathcal X \circ \mathcal Y}} & = & \nm{\vec{Ad}_{\mathcal X} \, \vec{Ad}_{\mathcal Y}} \label{eq:algebra_adjoint_matrix_product}
\end{eqnarray}

%%%%%%%%%%%%%%%%%%%%%%%%%%%%%%%%%%%%%%%%%%%%%%%%%%%%%%%%%%%%%%%%%%%%%%%%
%%%%%%%%%%%%%%%%%%%%%%%%%%%%%%%%%%%%%%%%%%%%%%%%%%%%%%%%%%%%%%%%%%%%%%%%
% SUBSECTION      RIGHT AND LEFT LIE GROUP DERIVATIVES
%%%%%%%%%%%%%%%%%%%%%%%%%%%%%%%%%%%%%%%%%%%%%%%%%%%%%%%%%%%%%%%%%%%%%%%%
%%%%%%%%%%%%%%%%%%%%%%%%%%%%%%%%%%%%%%%%%%%%%%%%%%%%%%%%%%%%%%%%%%%%%%%%

\subsection{Right and Left Lie Group Derivatives}\label{subsec:algebra_lie_derivatives}

Given a function \nm{\lrb{f: \mathcal{G} \rightarrow \mathcal {H} \ | \ \mathcal {Y} = f\lrp{\mathcal {X}} \in \mathcal {H}, \, \forall \mathcal {X} \in \mathcal {G}}} that maps together two manifolds or Lie groups of dimensions \emph{m} and \emph{n} respectively, it is possible to make use of the plus and minus operators to establish right and left derivatives (Jacobians) that linearly map their respective Lie algebras or tangent spaces, either locally \nm{\lrp{T_{\mathcal X}\mathcal G \rightarrow T_{f\lrp{\mathcal X}}\mathcal H}} if employing the right \nm{\oplus} and \nm{\ominus} operators, or globally \nm{\lrp{T_{\mathcal E}\mathcal G \rightarrow T_{\mathcal E}\mathcal H}} when using the left \nm{\boxplus} and \nm{\boxminus} operators \cite{Sola2018}. As a tangent space can always be identified to a Euclidean space of the same dimension, this enables the application of the concepts of random vectors, stochastic processes, and their correlation (section \ref{sec:Error_Random}) to Lie algebras, leading to the section \ref{subsec:algebra_lie_covariance} expressions for Lie covariances. In addition, these derivatives constitute the basis for the construction of the Lie group Jacobians in section \ref{sec:algebra_lie_jacobians}, which in turn are indispensable for the establishment of rigorous solutions for the gradient descent minimization (optimization) and state estimation in Lie groups, as described in sections \ref{sec:algebra_gradient_descent} and \ref{sec:algebra_SS}, respectively.

The \emph{right Jacobian} of \nm{f\lrp{\mathcal X}} is defined as the derivative of \nm{f\lrp{\mathcal X}} with respect to \nm{\mathcal X} when the increments are viewed in their respective local tangent spaces, this is, tangent respectively at \nm{\mathcal X \in \mathcal G} and \nm{f\lrp{\mathcal X} \in \mathcal H} \cite{Sola2018}. The \emph{left Jacobian} of \nm{f\lrp{\mathcal X}} is defined similarly, but with the increments viewed in the global tangent spaces for \nm{\mathcal G} and \nm{\mathcal H} respectively:
\begin{eqnarray}
\nm{\vec J_{\ds{\oplus \; \mathcal X}}^{\ds{\ominus \; f(\mathcal X)}}} & = & \nm{\lim_{\Delta \vec \tau^{\mathcal X} \to \vec 0} \dfrac{f\lrp{\mathcal X \oplus \Delta \vec \tau^{\mathcal X}} \ominus f(\mathcal X)}{\Delta \vec \tau^{\mathcal X}} = \lim_{\Delta \vec \tau^{\mathcal X} \to \vec 0} \dfrac{Log\lrsb{f^{-1}(\mathcal X) \circ f\Big(\mathcal X \circ Exp\lrp{\Delta \vec \tau^{\mathcal X}}\Big)}}{\Delta \vec \tau^{\mathcal X}} \in \mathbb{R}^{nxm}} \label{eq:algebra_lie_derivative_right} \\ 
\nm{\vec J_{\ds{\boxplus \; \mathcal X}}^{\ds{\boxminus \; f(\mathcal X)}}} & = & \nm{\lim_{\Delta \vec \tau^{\mathcal E} \to \vec 0} \dfrac{f\lrp{\Delta \vec \tau^{\mathcal E} \boxplus \mathcal X} \boxminus f\lrp{\mathcal X}}{\Delta \vec \tau^{\mathcal E}} = \lim_{\Delta \vec \tau^{\mathcal E} \to \vec 0} \dfrac{Log\lrsb{f\Big(Exp\lrp{\Delta \vec \tau^{\mathcal E} \circ \mathcal X}\Big) \circ f^{-1}\lrp{\mathcal X}}}{\Delta \vec \tau^{\mathcal E}} \in \mathbb{R}^{nxm}} \label{eq:algebra_lie_derivative_left}
\end{eqnarray}

The following first order Taylor expansions can then be directly established:
\begin{eqnarray}
\nm{f\lrp{\mathcal X \oplus \Delta \vec \tau^{\mathcal X}}} & \nm{\approx} & \nm{f\lrp{\mathcal X} \oplus \lrsb{\vec J_{\ds{\oplus \; \mathcal X}}^{\ds{\ominus \; f\lrp{\mathcal X}}} \, \Delta \vec \tau^{\mathcal X}} = \mathcal Y \oplus \Delta \vec \tau^{\mathcal Y} \ \ \in \mathcal H} \label{eq:algebra_lie_derivative_right_taylor} \\
\nm{f\lrp{\Delta \vec \tau^{\mathcal E_{\mathcal G}} \boxplus \mathcal X}} & \nm{\approx} & \nm{\lrsb{\vec J_{\ds{\boxplus \; \mathcal X}}^{\ds{\boxminus \; f\lrp{\mathcal X}}} \, \Delta \vec \tau^{\mathcal E_{\mathcal G}}} \boxplus f\lrp{\mathcal X} = \Delta \vec \tau^{\mathcal E_{\mathcal H}} \boxplus \mathcal Y \ \ \in \mathcal H} \label{eq:algebra_lie_derivative_left_taylor} 
\end{eqnarray}

Note that the \nm{\oplus} or \nm{\boxplus} symbols that appear as Jacobian subindexes in (\ref{eq:algebra_lie_derivative_right}) and (\ref{eq:algebra_lie_derivative_left}) indicate that the domain \nm{\mathcal G} is indeed a Lie group, and should be replaced by a standard \nm{+} operator if this is not the case and the function \emph{f} domain is in fact a real or Euclidean space. If this is the case, the expressions \nm{\lrp{\mathcal X \oplus \Delta \vec \tau^{\mathcal X}}} and \nm{\lrp{\Delta \vec \tau^{\mathcal E_{\mathcal G}} \boxplus \mathcal X}} within the equations (\ref{eq:algebra_lie_derivative_right}) through (\ref{eq:algebra_lie_derivative_left_taylor}) shall both be replaced by \nm{\lrp{\mathcal X + \Delta \vec \tau}}. 

Similarly, the \nm{\ominus} and \nm{\boxminus} symbols that appear as Jacobian superindexes indicate that the function \emph{f} image or codomain \nm{\mathcal H} is also a Lie group, and should otherwise be replaced by \nm{-} if the codomain is a Euclidean space. If this is the case, the \nm{\ominus} and \nm{\boxminus} operators within (\ref{eq:algebra_lie_derivative_right}) and (\ref{eq:algebra_lie_derivative_left}) shall be replaced by the standard \nm{-} operator, and the \nm{f\lrp{\mathcal X} \oplus} and \nm{\boxplus \, f\lrp{\mathcal X}} expressions within (\ref{eq:algebra_lie_derivative_right_taylor}) and (\ref{eq:algebra_lie_derivative_left_taylor}) shall both be replaced by \nm{f\lrp{\mathcal X} +}. 

Equations (\ref{eq:algebra_lie_derivative_right_taylor}) and (\ref{eq:algebra_lie_derivative_left_taylor}) lead to the following expressions for the function \emph{f} propagation of the tangent spaces: 
\begin{eqnarray}
\nm{\Delta \vec \tau^{\mathcal Y}} & = & \nm{\Delta \vec \tau^{f\lrp{\mathcal X}} = \vec J_{\ds{\oplus \; \mathcal X}}^{\ds{\ominus \; f\lrp{\mathcal X}}} \, \Delta \vec \tau^{\mathcal X}} \label{eq:algebra_lie_derivative_right_equiv} \\
\nm{\Delta \vec \tau^{\mathcal E_{\mathcal H}}} & = & \nm{\vec J_{\ds{\boxplus \; \mathcal X}}^{\ds{\boxminus \; f\lrp{\mathcal X}}} \, \Delta \vec \tau^{\mathcal E_{\mathcal G}}} \label{eq:algebra_lie_derivative_left_equiv} 
\end{eqnarray}

In addition, (\ref{eq:algebra_adjoint_matrix_general}) enables establishing a relationship between the right and left Jacobians of \nm{f\lrp{\mathcal X}}:
\neweq{\vec J_{\ds{\boxplus \; \mathcal X}}^{\ds{\boxminus \; f\lrp{\mathcal X}}} = \vec{Ad}_{f\lrp{\mathcal X}} \ \vec J_{\ds{\oplus \; \mathcal X}}^{\ds{\ominus \; f\lrp{\mathcal X}}} \ \vec{Ad}_{\mathcal X}^{-1}} {eq:algebra_lie_derivative_relationship}

%%%%%%%%%%%%%%%%%%%%%%%%%%%%%%%%%%%%%%%%%%%%%%%%%%%%%%%%%%%%%%%%%%%%%%%%
%%%%%%%%%%%%%%%%%%%%%%%%%%%%%%%%%%%%%%%%%%%%%%%%%%%%%%%%%%%%%%%%%%%%%%%%
% SUBSECTION      LIE GROUPS UNCERTAINTY AND COVARIANCE
%%%%%%%%%%%%%%%%%%%%%%%%%%%%%%%%%%%%%%%%%%%%%%%%%%%%%%%%%%%%%%%%%%%%%%%%
%%%%%%%%%%%%%%%%%%%%%%%%%%%%%%%%%%%%%%%%%%%%%%%%%%%%%%%%%%%%%%%%%%%%%%%%

\subsection{Lie Groups Uncertainty and Covariance}\label{subsec:algebra_lie_covariance}

As the \nm{\oplus} and \nm{\ominus} operators can be employed to define perturbations in the local tangent space \nm{\lrp{T_{\vec \mu_{\mathcal X}}\mathcal G}} around a nominal or expected point \nm{E\lrsb{\mathcal X} = \vec \mu_{\mathcal X} \in \mathcal G}, it is possible to employ a \emph{local autocovariance} definition similar to the one used for Euclidean spaces (\ref{eq:Error_rvec_autoCovarianceMatrix}), leading to the definition of stochastic variables (vectors) on manifolds \nm{\mathcal X \sim \lrp{\vec \mu_{\mathcal X}, \ \vec C_{\mathcal X \mathcal X}^{\mathcal X}}} \cite{Sola2018}:
\begin{eqnarray}
\nm{\mathcal X} & = & \nm{\vec \mu_{\mathcal X} \oplus \Delta \vec \tau^{\mathcal X} \ \  \in \mathcal G} \label{eq:algebra_lie_covariance_right_plus} \\
\nm{\Delta \vec \tau^{\mathcal X}} & = & \nm{\mathcal X \ominus \vec \mu_{\mathcal X} \ \ \in T_{\vec \mu_{\mathcal X}}\mathcal G}\label{eq:algebra_lie_covariance_right_minus} \\
\nm{\vec C_{\mathcal X \mathcal X}^{\mathcal X}} & = & \nm{E\lrsb{\Delta \vec \tau^{\mathcal X} \,  \Delta \vec \tau^{{\mathcal X,T}}} = E\lrsb{\lrp{\mathcal X \ominus \vec \mu_{\mathcal X}}\lrp{\mathcal X \ominus \vec \mu_{\mathcal X}}^T} \ \ \in \mathbb{R}^{mxm}}\label{eq:algebra_lie_covariance_right_def}
\end{eqnarray}

The different types of correlation matrices for stochastic processes introduced in section \ref{subsec:Error_StochasticProcesses} can also be defined accordingly. Note that although the notation of \nm{\vec C_{\mathcal X \mathcal X}^{\mathcal X}} refers to the covariance of the manifold or Lie group \nm{\mathcal X \in \mathcal G}, the definition in fact refers to the covariance of the nominal point local tangent space \nm{\Delta \vec \tau^{\mathcal X} \in T_{\vec \mu \mathcal X}\mathcal G}, with its dimension matching the number of degrees of freedom of the manifold.

A similar process employing the \nm{\boxplus} and \nm{\boxminus} operators leads to the \emph{global autocovariance}:
\begin{eqnarray}
\nm{\vec C_{\mathcal X \mathcal X}^{\mathcal E}} & = & \nm{E\lrsb{\Delta \vec \tau^{\mathcal E} \,  \Delta \vec \tau^{{\mathcal E},T}} = E\lrsb{\lrp{\mathcal X \boxminus \vec \mu_{\mathcal X}}\lrp{\mathcal X \boxminus \vec \mu_{\mathcal X}}^T} \ \ \in \mathbb{R}^{mxm}}\label{eq:algebra_lie_covariance_left_def} \\
\nm{\vec C_{\mathcal X \mathcal X}^{\mathcal E}} & = & \nm{E\lrsb{\vec{Ad}_{\mathcal X} \, \Delta \vec \tau^{\mathcal X} \, \Delta \vec \tau^{{\mathcal X},T} \, \vec{Ad}_{\mathcal X}^T} = \vec{Ad}_{\mathcal X} \ \vec C_{\mathcal X \mathcal X}^{\mathcal X} \ \vec{Ad}_{\mathcal X}^T}\label{eq:algebra_lie_covariance_left_relationship} \\
\nm{\vec C_{\mathcal X \mathcal X}^{\mathcal X}} & = & \nm{E\lrsb{\vec{Ad}_{\mathcal X}^{-1} \, \Delta \vec \tau^{\mathcal E} \, \Delta \vec \tau^{{\mathcal E},T} \, \vec{Ad}_{\mathcal X}^{-T}} = \vec{Ad}_{\mathcal X}^{-1} \ \vec C_{\mathcal X \mathcal X}^{\mathcal E} \ \vec{Ad}_{\mathcal X}^{-T}}\label{eq:algebra_lie_covariance_right_relationship}
\end{eqnarray}

Given a function \nm{\lrb{f: \mathcal{G} \rightarrow \mathcal {H} \ | \ \mathcal {Y} = f\lrp{\mathcal {X}} \in \mathcal {H}, \, \forall \mathcal {X} \in \mathcal {G}}} as that introduced in section \ref{subsec:algebra_lie_derivatives}, which maps together two manifolds or Lie groups of dimensions \emph{m} and \emph{n} respectively, it is possible to propagate the covariances \nm{\vec C_{\mathcal X \mathcal X}^{\mathcal X}} and \nm{\vec C_{\mathcal X \mathcal X}^{\mathcal E}} from the domain manifold \nm{\mathcal G} to the image one \nm{\mathcal H}:
\begin{eqnarray}
\nm{\vec C_{\mathcal Y \mathcal Y}^{\mathcal Y}} & = & \nm{E\lrsb{\Delta \vec \tau^{\mathcal Y} \,  \Delta \vec \tau^{\mathcal Y,T}} = E\lrsb{\Delta \vec \tau^{f\lrp{\mathcal X}} \,  \Delta \vec \tau^{f\lrp{\mathcal X},T}}} \nonumber \\
& = & \nm{E\lrsb{\vec J_{\ds{\oplus \; \mathcal X}}^{\ds{\ominus \; f\lrp{\mathcal X}}} \, \Delta \vec \tau^{\mathcal X} \, \Delta \vec \tau^{{\mathcal X},T} \, \vec J_{\ds{\oplus \; \mathcal X}}^{{\ds{\ominus \; f\lrp{\mathcal X}}},T}} = \vec J_{\ds{\oplus \; \mathcal X}}^{\ds{\ominus \; f\lrp{\mathcal X}}} \ \vec C_{\mathcal X \mathcal X}^{\mathcal X} \ \vec J_{\ds{\oplus \; \mathcal X}}^{{\ds{\ominus \; f\lrp{\mathcal X}}},T} \ \ \ \ \in \mathbb{R}^{nxn}} \label{eq:algebra_lie_covariance_right_propagation} \\
\nm{\vec C_{\mathcal Y \mathcal Y}^{\mathcal E}} & = & \nm{E\lrsb{\Delta \vec \tau^{\mathcal E_{\mathcal H}} \, \Delta \vec \tau^{\mathcal E_{\mathcal H},T}} = \vec J_{\ds{\boxplus \; \mathcal X}}^{\ds{\boxminus \; f\lrp{\mathcal X}}} \ \vec C_{\mathcal X \mathcal X}^{\mathcal E} \ \vec J_{\ds{\boxplus \; \mathcal X}}^{{\ds{\boxminus \; f\lrp{\mathcal X}}},T} \ \ \ \ \ \ \ \ \ \ \ \ \ \ \ \ \ \ \ \ \ \ \ \ \ \ \ \ \ \in \mathbb{R}^{nxn}} \label{eq:algebra_lie_covariance_left_propagation}
\end{eqnarray}

The establishment and propagation of covariance matrices of the proper dimensions is key for the application of state estimation techniques such as Kalman filtering when some of the state vector components belong to Lie manifolds and their tangent spaces, as described in section \ref{sec:algebra_SS}.

%%%%%%%%%%%%%%%%%%%%%%%%%%%%%%%%%%%%%%%%%%%%%%%%%%%%%%%%%%%%%%%%%%%%%%%%
%%%%%%%%%%%%%%%%%%%%%%%%%%%%%%%%%%%%%%%%%%%%%%%%%%%%%%%%%%%%%%%%%%%%%%%%
%%%%%%%%%%%%%%%%%%%%%%%%%%%%%%%%%%%%%%%%%%%%%%%%%%%%%%%%%%%%%%%%%%%%%%%%
% SECTION      EUCLIDEAN AND LIE JACOBIANS
%%%%%%%%%%%%%%%%%%%%%%%%%%%%%%%%%%%%%%%%%%%%%%%%%%%%%%%%%%%%%%%%%%%%%%%%
%%%%%%%%%%%%%%%%%%%%%%%%%%%%%%%%%%%%%%%%%%%%%%%%%%%%%%%%%%%%%%%%%%%%%%%%
%%%%%%%%%%%%%%%%%%%%%%%%%%%%%%%%%%%%%%%%%%%%%%%%%%%%%%%%%%%%%%%%%%%%%%%%

\section{Euclidean and Lie Jacobians}\label{sec:algebra_lie_jacobians}

The definition of the proper derivative matrices or Jacobians is indispensable for all calculus techniques that rely on linearization, such as optimization by means of the gradient descent method (section \ref{sec:algebra_gradient_descent}) or state estimation through Kalman filtering (section \ref{sec:algebra_SS}). Given a function \nm{\lrb{\vec f\lrp{\vec x}: \mathbb{R}^m \rightarrow \mathbb{R}^n}}, its (Euclidean) Jacobian \nm{\vec J_{\ds{+ \; \vec x}}^{\ds{- \; f(\vec x)}} \in \mathbb{R}^{nxm}} stacks the partial derivatives of each component of the output space with respect to those of the input space:
\neweq{\vec J_{{\ds{+ \; \vec x}}, ij}^{\ds{- \; \vec f(\vec x)}} = \lim_{\Delta x_j\to 0} \dfrac{f_i\lrp{x_j + \Delta x_j} - f_i\lrp{x_j}}{\Delta x_j}}{eq:algebra_jacobian_euclidean}

This section relies on the right and left Lie group derivatives introduced in section \ref{subsec:algebra_lie_derivatives} to properly define Jacobians when either the input or output spaces (or both) are not Euclidean but Lie groups. The various Jacobians listed in table \ref{tab:algebra_lie_jacobians} have been obtained by means of the expressions that appear on section \ref{sec:algebra_lie} together with the chain rule, and include instances in which both the domain and codomain of the \nm{f\lrp{\mathcal X}} function are either Euclidean or Lie groups. Some are generic, while others, which rely on group actions, depend on the specific set on which the action is applied and hence can only be established for a specific Lie group, such as rotational or rigid body motions (chapters \ref{cha:Rotate} and \ref{cha:Motion}). 
\renewcommand{\arraystretch}{1.5} % increase row height
\begin{center}
\begin{tabular}{lcp{0.2cm}rcll}
	\hline
	Jacobian & Result & & \multicolumn{3}{c}{Taylor Expansion} & \\
	\hline
	\nm{\vec J_{\ds{\oplus \; \mathcal X}}^{\ds{\ominus \; \mathcal X}^{-1}}}       				& \nm{- \vec{Ad}_{\mathcal X}} 		    	& & \nm{\lrp{\mathcal X \oplus \Delta \vec \tau}^{-1}} & \nm{\approx} & \nm{\mathcal X^{-1} \oplus \lrsb{\vec J_{\ds{\oplus \; \mathcal X}}^{\ds{\ominus \; \mathcal X}^{-1}} \, \Delta \vec \tau}} & \nm{\in \mathcal G} \\ 
	\nm{\vec J_{\ds{\boxplus \; \mathcal X}}^{\ds{\boxminus \; \mathcal X}^{-1}}}         		& \nm{- \vec{Ad}_{\mathcal X}^{-1}}			& & \nm{\lrp{\Delta \vec \tau \boxplus \mathcal X}^{-1}} & \nm{\approx} & \nm{\lrsb{\vec J_{\ds{\boxplus \; \mathcal X}}^{\ds{\boxminus \; \mathcal X}^{-1}} \, \Delta \vec \tau} \boxplus \mathcal X^{-1}} & \nm{\in \mathcal G} \\ 

	\nm{\vec J_{\ds{\oplus \; \mathcal X}}^{\ds{\ominus \; \mathcal X \circ \mathcal Y}}}		& \nm{\vec{Ad}_{\mathcal Y}^{-1}}	    	& & \nm{\lrp{\mathcal X \oplus \Delta \vec \tau} \circ \mathcal Y} & \nm{\approx} & \nm{\lrp{\mathcal X \circ \mathcal Y} \oplus \lrsb{\vec J_{\ds{\oplus \; \mathcal X}}^{\ds{\ominus \; \mathcal X \circ \mathcal Y}} \, \Delta \vec \tau}} & \nm{\in \mathcal G} \\ 
	\nm{\vec J_{\ds{\boxplus \; \mathcal X}}^{\ds{\boxminus \; \mathcal X \circ \mathcal Y}}}	& \nm{\vec{I}_{mxm}}	 					& & \nm{\lrp{\Delta \vec \tau \boxplus \mathcal X} \circ \mathcal Y} & \nm{\approx} & \nm{\lrsb{\vec J_{\ds{\boxplus \; \mathcal X}}^{\ds{\boxminus \; \mathcal X \circ \mathcal Y}} \, \Delta \vec \tau} \boxplus \lrp{\mathcal X \circ \mathcal Y}} & \nm{\in \mathcal G} \\ 

	\nm{\vec J_{\ds{\oplus \; \mathcal Y}}^{\ds{\ominus \; \mathcal X \circ \mathcal Y}}}		& \nm{\vec{I}_{mxm}}						& & \nm{\mathcal X \circ \lrp{\mathcal Y \oplus \Delta \vec \tau}} & \nm{\approx} & \nm{\lrp{\mathcal X \circ \mathcal Y} \oplus \lrsb{\vec J_{\ds{\oplus \; \mathcal Y}}^{\ds{\ominus \; \mathcal X \circ \mathcal Y}} \, \Delta \vec \tau}} & \nm{\in \mathcal G} \\ 
	\nm{\vec J_{\ds{\boxplus \; \mathcal Y}}^{\ds{\boxminus \; \mathcal X \circ \mathcal Y}}}	& \nm{\vec{Ad}_{\mathcal X}}				& & \nm{\mathcal X \circ \lrp{\Delta \vec \tau \boxplus \mathcal Y}} & \nm{\approx} & \nm{\lrsb{\vec J_{\ds{\boxplus \; \mathcal Y}}^{\ds{\boxminus \; \mathcal X \circ \mathcal Y}} \, \Delta \vec \tau} \boxplus \lrp{\mathcal X \circ \mathcal Y}} & \nm{\in \mathcal G} \\ 
		
	\nm{\vec J_{\ds{+ \; \vec \tau}}^{\ds{\ominus \; Exp\lrp{\vec \tau}}}}						& \nm{\vec J_R\lrp{\vec \tau}}\footnotemark		& & \nm{Exp\lrp{\vec \tau + \Delta \vec \tau}} & \nm{\approx} & \nm{Exp\lrp{\vec \tau} \oplus \lrsb{\vec J_R\lrp{\vec \tau} \, \Delta \vec \tau}} & \nm{\in \mathcal G} \\ 
	\nm{\vec J_R^{-1}\lrp{\vec \tau}} 																&											& & \nm{\vec \tau + \vec J_R^{-1}\lrp{\vec \tau} \, \Delta \vec \tau} & \nm{\approx}\footnotemark & \nm{Log\Big(Exp\lrp{\vec \tau} \oplus \Delta \vec \tau\Big)} & \nm{\in \mathbb{R}^m} \\ 
	\nm{\vec J_{\ds{+ \; \vec \tau}}^{\ds{\boxminus \; Exp\lrp{\vec \tau}}}} 					& \nm{\vec J_L\lrp{\vec \tau}}\footnotemark		& & \nm{Exp\lrp{\vec \tau + \Delta \vec \tau}} & \nm{\approx} & \nm{\lrsb{\vec J_L\lrp{\vec \tau} \, \Delta \vec \tau} \boxplus Exp\lrp{\vec \tau}} & \nm{\in \mathcal G} \\ 
	\nm{\vec J_L^{-1}\lrp{\vec \tau}}																& 											& & \nm{\vec \tau + \vec J_L^{-1}\lrp{\vec \tau} \, \Delta \vec \tau} & \nm{\approx}\footnotemark & \nm{Log\Big(\Delta \vec \tau \boxplus Exp\lrp{\vec \tau}\Big)} & \nm{\in \mathbb{R}^m} \\ 
		
	\nm{\vec J_{\ds{\oplus \; \mathcal X}}^{\ds{- \; Log\lrp{\mathcal X}}}}        				& \nm{\vec J_R^{-1}\big(Log\lrp{\mathcal X}\big)} & & \nm{Log\lrp{\mathcal X \oplus \Delta \vec \tau}} & \nm{\approx} & \nm{Log\lrp{\mathcal X} + \lrsb{\vec J_{\ds{\oplus \; \mathcal X}}^{\ds{- \; Log\lrp{\mathcal X}}} \, \Delta \vec \tau}} & \nm{\in \mathbb{R}^m} \\  
	\nm{\vec J_{\ds{\boxplus \; \mathcal X}}^{\ds{- \; Log\lrp{\mathcal X}}}}     				& \nm{\vec J_L^{-1}\big(Log\lrp{\mathcal X}\big)} & & \nm{Log\lrp{\Delta \vec \tau \boxplus \mathcal X}} & \nm{\approx} & \nm{Log\lrp{\mathcal X} + \lrsb{\vec J_{\ds{\boxplus \; \mathcal X}}^{\ds{- \; Log\lrp{\mathcal X}}} \, \Delta \vec \tau}} & \nm{\in \mathbb{R}^m} \\  	
		
	\nm{\vec J_{\ds{\oplus \; \mathcal X}}^{\ds{\ominus \; \mathcal X \oplus \vec \tau}}}     	& \nm{\vec{Ad}_{Exp\lrp{\vec \tau}}^{-1}}	& & \nm{\lrp{\mathcal X \oplus \Delta \vec \tau} \oplus \vec \tau} & \nm{\approx} & \nm{\lrp{\mathcal X \oplus \vec \tau} \oplus \lrsb{\vec J_{\ds{\oplus \; \mathcal X}}^{\ds{\ominus \; \mathcal X \oplus \vec \tau}} \, \Delta \vec \tau}} & \nm{\in \mathcal G} \\ 
	\nm{\vec J_{\ds{\boxplus \; \mathcal X}}^{\ds{\boxminus \; \vec \tau \boxplus \mathcal X}}}  & \nm{\vec{Ad}_{Exp\lrp{\vec \tau}}}		& & \nm{\vec \tau \boxplus \lrp{\Delta \vec \tau \boxplus \mathcal X}} & \nm{\approx} & \nm{\lrsb{\vec J_{\ds{\boxplus \; \mathcal X}}^{\ds{\boxminus \; \vec \tau \boxplus\mathcal X}} \, \Delta \vec \tau} \boxplus \lrp{\vec \tau \boxplus \mathcal X}} & \nm{\in \mathcal G} \\ 
	\nm{\vec J_{\ds{+ \; \vec \tau}}^{\ds{\ominus \; \mathcal X \oplus \vec \tau}}}         		& \nm{\vec J_R\lrp{\vec \tau}}           	   	& & \nm{\mathcal X \oplus \lrp{\vec \tau + \Delta \vec \tau}} & \nm{\approx} & \nm{\lrp{\mathcal X \oplus \vec \tau} \oplus \lrsb{\vec J_{\ds{+ \; \vec \tau}}^{\ds{\ominus \; \mathcal X \oplus \vec \tau}} \, \Delta \vec \tau}} & \nm{\in \mathcal G} \\ 	
	\nm{\vec J_{\ds{+ \; \vec \tau}}^{\ds{\boxminus \; \vec \tau \boxplus \mathcal X}}}     		& \nm{\vec J_L\lrp{\vec \tau}}           	   	& & \nm{\lrp{\vec \tau + \Delta \vec \tau} \boxplus \mathcal X} & \nm{\approx} & \nm{\lrsb{\vec J_{\ds{+ \; \vec \tau}}^{\ds{\boxminus \; \vec \tau \boxplus \mathcal X}} \, \Delta \vec \tau} \boxplus \lrp{\vec \tau \boxplus \mathcal X}} & \nm{\in \mathcal G} \\ 		
	
	\nm{\vec J_{\ds{\oplus \; \mathcal X}}^{\ds{- \; \mathcal Y \ominus \mathcal X}}} 	      	& \nm{- \vec J_L^{-1}\lrp{\mathcal Y \ominus \mathcal X}}	& & \nm{\mathcal Y \ominus \lrp{\mathcal X \oplus \Delta \vec \tau}} & \nm{\approx} & \nm{\lrp{\mathcal Y \ominus \mathcal X} + \lrsb{\vec J_{\ds{\oplus \; \mathcal X}}^{\ds{- \; \mathcal Y \ominus \mathcal X}} \, \Delta \vec \tau}} & \nm{\in \mathbb{R}^m} \\ 		
	\hline
\end{tabular}
\end{center}
\renewcommand{\arraystretch}{1.0} % reset row height
\addtocounter{footnote}{-3}	
\footnotetext{Obtained in sections \ref{sec:RigidBody_rotation_calculus_jacobians} and \ref{sec:RigidBody_motion_calculus_jacobians} for rotational and rigid body motion, respectively.}
\addtocounter{footnote}{1}	
\footnotetext{Obtained by replacing \nm{\Delta \vec \tau} by \nm{\vec J_R^{-1} \ \Delta \vec \tau} in the expression above.}
\addtocounter{footnote}{1}	
\footnotetext{Obtained in sections \ref{sec:RigidBody_rotation_calculus_jacobians} and \ref{sec:RigidBody_motion_calculus_jacobians} for rotational and rigid body motion, respectively.}
\addtocounter{footnote}{1}	
\footnotetext{Obtained by replacing \nm{\Delta \vec \tau} by \nm{\vec J_L^{-1} \ \Delta \vec \tau} in the expression above.}

\renewcommand{\arraystretch}{1.5} % increase row height
\begin{center}
\begin{tabular}{lcp{0.2cm}rcll}
	\hline
	Jacobian & Result & & \multicolumn{3}{c}{Taylor Expansion} & \\
	\hline
	\nm{\vec J_{\ds{\boxplus \; \mathcal X}}^{\ds{- \; \mathcal Y \boxminus \mathcal X}}} 	   	& \nm{- \vec J_R^{-1}\lrp{\mathcal Y \boxminus \mathcal X}}	& & \nm{\mathcal Y \boxminus \lrp{\Delta \vec \tau \boxplus \mathcal X}} & \nm{\approx} & \nm{\lrp{\mathcal Y \boxminus \mathcal X} + \lrsb{\vec J_{\ds{\boxplus \; \mathcal X}}^{\ds{- \; \mathcal Y \boxminus \mathcal X}} \, \Delta \vec \tau}} & \nm{\in \mathbb{R}^m} \\ 	
	\nm{\vec J_{\ds{\oplus \; \mathcal Y}}^{\ds{- \; \mathcal Y \ominus \mathcal X}}}    	    & \nm{\vec J_R^{-1}\lrp{\mathcal Y \ominus \mathcal X}}		& & \nm{\lrp{\mathcal Y \oplus \Delta \vec \tau} \ominus \mathcal X} & \nm{\approx} & \nm{\lrp{\mathcal Y \ominus \mathcal X} + \lrsb{\vec J_{\ds{\oplus \; \mathcal Y}}^{\ds{- \; \mathcal Y \ominus \mathcal X}} \, \Delta \vec \tau}} & \nm{\in \mathbb{R}^m} \\
	\nm{\vec J_{\ds{\boxplus \; \mathcal Y}}^{\ds{- \; \mathcal Y \boxminus \mathcal X}}}  	    & \nm{\vec J_L^{-1}\lrp{\mathcal Y \boxminus \mathcal X}}	& & \nm{\lrp{\Delta \vec \tau \boxplus \mathcal Y} \boxminus \mathcal X} & \nm{\approx} & \nm{\lrp{\mathcal Y \boxminus \mathcal X} + \lrsb{\vec J_{\ds{\boxplus \; \mathcal Y}}^{\ds{- \; \mathcal Y \boxminus \mathcal X}} \, \Delta \vec \tau}} & \nm{\in \mathbb{R}^m} \\  	
	
	\nm{\vec J_{\ds{\oplus \; \mathcal X}}^{\ds{- \; g_{\mathcal X}(\vec u)}}}          	 	& 									 		& & \nm{g_{\mathcal X \oplus \Delta \vec \tau}\lrp{\vec u}} & \nm{\approx} & \nm{g_{\mathcal X}\lrp{\vec u} + \lrsb{\vec J_{\ds{\oplus \; \mathcal X}}^{\ds{- \; g_{\mathcal X}(\vec u)}} \, \Delta \vec \tau}} & \nm{\in \mathbb{R}^u} \\ 
	\nm{\vec J_{\ds{\boxplus \; \mathcal X}}^{\ds{- \; g_{\mathcal X}(\vec u)}}}      			& Tables							 		& & \nm{g_{\Delta \vec \tau \boxplus \mathcal X}\lrp{\vec u}} & \nm{\approx} & \nm{g_{\mathcal X}\lrp{\vec u} + \lrsb{\vec J_{\ds{\boxplus \; \mathcal X}}^{\ds{- \; g_{\mathcal X}(\vec u)}} \, \Delta \vec \tau}} & \nm{\in \mathbb{R}^u} \\ 
	\nm{\vec J_{\ds{+ \; \vec u}}^{\ds{- \; g_{\mathcal X}(\vec u)}}}               			& \ref{tab:RigidBody_rotation_jacobians}	& & \nm{g_{\mathcal X}\lrp{\vec u + \Delta \vec u}} & \nm{\approx} & \nm{g_{\mathcal X}\lrp{\vec u} + \lrsb{\vec J_{\ds{+ \; \vec u}}^{\ds{- \; g_{\mathcal X}(\vec u)}} \, \Delta \vec u}} & \nm{\in \mathbb{R}^u} \\  
		    
	\nm{\vec J_{\ds{\oplus \; \mathcal X}}^{\ds{- \; g_{\mathcal X}^{-1}(\vec u)}}}          	& \&								 		& & \nm{g_{\mathcal X \oplus \Delta \vec \tau}^{-1}\lrp{\vec u}} & \nm{\approx} & \nm{g_{\mathcal X}^{-1}\lrp{\vec u} + \lrsb{\vec J_{\ds{\oplus \; \mathcal X}}^{\ds{- \; g_{\mathcal X}^{-1}(\vec u)}} \, \Delta \vec \tau}} & \nm{\in \mathbb{R}^u} \\ 
	\nm{\vec J_{\ds{\boxplus \; \mathcal X}}^{\ds{- \; g_{\mathcal X}^{-1}(\vec u)}}}  			& \ref{tab:RigidBody_motion_jacobians} 		& & \nm{g_{\Delta \vec \tau \boxplus \mathcal X}^{-1}\lrp{\vec u}} & \nm{\approx} & \nm{g_{\mathcal X}^{-1}\lrp{\vec u} + \lrsb{\vec J_{\ds{\boxplus \; \mathcal X}}^{\ds{- \; g_{\mathcal X}^{-1}(\vec u)}} \, \Delta \vec \tau}} & \nm{\in \mathbb{R}^u} \\ 
	\nm{\vec J_{\ds{+ \; \vec u}}^{\ds{- \; g_{\mathcal X}^{-1}(\vec u)}}}             			& 											& & \nm{g_{\mathcal X}^{-1}\lrp{\vec u + \Delta \vec u}} & \nm{\approx} & \nm{g_{\mathcal X}^{-1}\lrp{\vec u} + \lrsb{\vec J_{\ds{+ \; \vec u}}^{\ds{- \; g_{\mathcal X}^{-1}(\vec u)}} \, \Delta \vec u}} & \nm{\in \mathbb{R}^u} \\  
	
	\nm{\vec J_{\ds{\oplus \; \mathcal X}}^{\ds{- \; \vec{Ad}_{\mathcal X}(\vec v)}}} 			& 											& & \nm{\vec{Ad}_{\mathcal X \oplus \Delta \vec \tau}\lrp{\vec v}} & \nm{\approx} & \nm{\vec{Ad}_{\mathcal X}\lrp{\vec v} + \lrsb{\vec J_{\ds{\oplus \; \mathcal X}}^{\ds{- \; \vec{Ad}_{\mathcal X}(\vec v)}} \, \Delta \vec \tau}} & \nm{\in \mathbb{R}^m} \\ 
	\nm{\vec J_{\ds{\boxplus \; \mathcal X}}^{\ds{- \; \vec{Ad}_{\mathcal X}(\vec v)}}}			& 											& & \nm{\vec{Ad}_{\Delta \vec \tau \boxplus \mathcal X}\lrp{\vec v}} & \nm{\approx} & \nm{\vec{Ad}_{\mathcal X}\lrp{\vec v} + \lrsb{\vec J_{\ds{\boxplus \; \mathcal X}}^{\ds{- \; \vec{Ad}_{\mathcal X}(\vec v)}} \, \Delta \vec \tau}} & \nm{\in \mathbb{R}^m} \\ 
	\nm{\vec J_{\ds{+ \; \vec v}}^{\ds{- \; \vec{Ad}_{\mathcal X}(\vec v)}}}					& 											& & \nm{\vec{Ad}_{\mathcal X}\lrp{\vec v + \Delta \vec v}} & \nm{\approx} & \nm{\vec{Ad}_{\mathcal X}\lrp{\vec v} + \lrsb{\vec J_{\ds{+ \; \vec v}}^{\ds{- \; \vec{Ad}_{\mathcal X}(\vec v)}} \, \Delta \vec v}} & \nm{\in \mathbb{R}^m} \\ 
	    
	\nm{\vec J_{\ds{\oplus \; \mathcal X}}^{\ds{- \; \vec{Ad}_{\mathcal X}^{-1}(\vec v)}}} 		& Tables									& & \nm{\vec{Ad}_{\mathcal X \oplus \Delta \vec \tau}^{-1}\lrp{\vec v}} & \nm{\approx} & \nm{\vec{Ad}_{\mathcal X}^{-1}\lrp{\vec v} + \lrsb{\vec J_{\ds{\oplus \; \mathcal X}}^{\ds{- \; \vec{Ad}_{\mathcal X}^{-1}(\vec v)}} \, \Delta \vec \tau}} & \nm{\in \mathbb{R}^m} \\ 
	\nm{\vec J_{\ds{\boxplus \; \mathcal X}}^{\ds{- \; \vec{Ad}_{\mathcal X}^{-1}(\vec v)}}}	& \ref{tab:RigidBody_rotation_jacobians}	& & \nm{\vec{Ad}_{\Delta \vec \tau \boxplus \mathcal X}^{-1}\lrp{\vec v}} & \nm{\approx} & \nm{\vec{Ad}_{\mathcal X}^{-1}\lrp{\vec v} + \lrsb{\vec J_{\ds{\boxplus \; \mathcal X}}^{\ds{- \; \vec{Ad}_{\mathcal X}^{-1}(\vec v)}} \, \Delta \vec \tau}} & \nm{\in \mathbb{R}^m} \\ 
	\nm{\vec J_{\ds{+ \; \vec v}}^{\ds{- \; \vec{Ad}_{\mathcal X}^{-1}(\vec v)}}}				& \&										& & \nm{\vec{Ad}_{\mathcal X}^{-1}\lrp{\vec v + \Delta \vec v}} & \nm{\approx} & \nm{\vec{Ad}_{\mathcal X}^{-1}\lrp{\vec v} + \lrsb{\vec J_{\ds{+ \; \vec v}}^{\ds{- \; \vec{Ad}_{\mathcal X}^{-1}(\vec v)}} \, \Delta \vec v}} & \nm{\in \mathbb{R}^m} \\ 	
	    
	\nm{\vec J_{\ds{+ \; \vec \tau}}^{\ds{- \; g_{Exp\lrp{\vec \tau}}(\vec u)}}}				& \ref{tab:RigidBody_motion_jacobians}		& & \nm{g_{Exp\lrp{\vec \tau + \Delta \vec \tau}}\lrp{\vec u}} & \nm{\approx} & \nm{g_{Exp\lrp{\vec \tau}}\lrp{\vec u} + \lrsb{\vec J_{\ds{+ \; \vec \tau}}^{\ds{- \; g_{Exp\lrp{\vec \tau}}(\vec u)}} \, \Delta \vec \tau}} & \nm{\in \mathbb{R}^u} \\ 
	\nm{\vec J_{\ds{+ \; \vec \tau}}^{\ds{- \; g_{Exp\lrp{\vec \tau}}^{-1}(\vec u)}}}			& 										 	& & \nm{g_{Exp\lrp{\vec \tau + \Delta \vec \tau}}^{-1}\lrp{\vec u}} & \nm{\approx} & \nm{g_{Exp\lrp{\vec \tau}}^{-1}\lrp{\vec u} + \lrsb{\vec J_{\ds{+ \; \vec \tau}}^{\ds{- \; g_{Exp\lrp{\vec \tau}}^{-1}(\vec u)}} \, \Delta \vec \tau}} & \nm{\in \mathbb{R}^u} \\ 
        
	\hline
\end{tabular}
\end{center}
\captionof{table}{Lie Jacobians} \label{tab:algebra_lie_jacobians}
\renewcommand{\arraystretch}{1.0} % reset row height

There are two constructions that appear repeatedly within table \ref{tab:algebra_lie_jacobians}. The first is the adjoint matrix (\ref{eq:algebra_adjoint_matrix}), which maps the local and global tangent spaces at a given point on a manifold or Lie group, while the second are the right and left Jacobians of the capitalized exponential function, also known as simply the \emph{right Jacobian} \nm{\vec J_R\lrp{\vec \tau}} and the \emph{left Jacobian} \nm{\vec J_L\lrp{\vec \tau}}. These compare variations in the tangent space of the output \nm{Exp\lrp{\vec \tau}} map (locally for \nm{\vec J_R} and globally for \nm{\vec J_L}) with (Euclidean) variations in the input argument \nm{\vec \tau}, and are obtained in sections \ref{sec:RigidBody_rotation_calculus_jacobians} and \ref{sec:RigidBody_motion_calculus_jacobians} for the specific cases of rotational and rigid body motions, respectively.
\begin{eqnarray}
\nm{\vec{Ad}_{Exp\lrp{\vec \tau}}} & = & \nm{\vec J_L\lrp{\vec \tau} \, \vec J_R^{-1}\lrp{\vec \tau}} \label{eq:algebra_lie_jacobians_right_left1} \\
\nm{\vec J_R\lrp{- \vec \tau}} & = & \nm{\vec J_L\lrp{\vec \tau}} \label{eq:algebra_lie_jacobians_right_left2}
\end{eqnarray}

Being located in the tangent spaces, the right and left Jacobians can be related by means of the adjoint, resulting in (\ref{eq:algebra_lie_jacobians_right_left1}). Expression (\ref{eq:algebra_lie_jacobians_right_left2}) in turn can be obtained by means of the chain rule.

%%%%%%%%%%%%%%%%%%%%%%%%%%%%%%%%%%%%%%%%%%%%%%%%%%%%%%%%%%%%%%%%%%%%%%%%
%%%%%%%%%%%%%%%%%%%%%%%%%%%%%%%%%%%%%%%%%%%%%%%%%%%%%%%%%%%%%%%%%%%%%%%%
%%%%%%%%%%%%%%%%%%%%%%%%%%%%%%%%%%%%%%%%%%%%%%%%%%%%%%%%%%%%%%%%%%%%%%%%
% SECTION      DISCRETE INTEGRATION IN LIE GROUPS
%%%%%%%%%%%%%%%%%%%%%%%%%%%%%%%%%%%%%%%%%%%%%%%%%%%%%%%%%%%%%%%%%%%%%%%%
%%%%%%%%%%%%%%%%%%%%%%%%%%%%%%%%%%%%%%%%%%%%%%%%%%%%%%%%%%%%%%%%%%%%%%%%
%%%%%%%%%%%%%%%%%%%%%%%%%%%%%%%%%%%%%%%%%%%%%%%%%%%%%%%%%%%%%%%%%%%%%%%%

\section{Discrete Integration in Lie Groups}\label{sec:algebra_integration}

Following on the discrete integration in Euclidean spaces described in section \ref{sec:euclidean_integration}, consider now that the state system is composed by a vector \nm{\vec y \in \mathbb{R}^n}, an element of a Lie group \nm{\mathcal X \in \mathcal G}, and a vector \nm{\vec v^{\mathcal X} \in \mathbb{R}^m} representing the velocity of \nm{\mathcal X} as it moves along its manifold, contained in the local or body tangent space \nm{T_{\mathcal X}\mathcal G}. Neither \nm{\mathcal X} nor its components are Euclidean, and hence the section \ref{sec:euclidean_integration} integration schemes are not applicable. If treated so, the resulting element \nm{\mathcal X_{k+1}} would not be located on the manifold as it would not comply with the Lie group constraints, and it would be necessary to reproject it back to it, incurring in errors that although small for a single integration step may become significant when accumulated.

Group these states into a composite state vector made up by \emph{n} plus \emph{m} components of an Euclidean space plus an element of a Lie group. As in the Euclidean case, the initial composite vector value is known:
\begin{eqnarray}
\nm{\vec x} & = & \nm{\lrsb{\vec y \ \ \vec v^{\mathcal X} \ \ \mathcal X}^T} \label{eq:algebra_integration_comp} \\
\nm{\vec x_k} & = & \nm{\vec x\lrp{t_k} = \lrsb{\vec y_k \ \ \vec v_k^{\mathcal X} \ \ \mathcal X_k}^T} \label{eq:algebra_integration_comp_initial}
\end{eqnarray}

As in the Euclidean case, the objective is the determination of the composite state vector value at a time \nm{\vec x_{k+1} = \vec x\lrp{t_{k+1}} = \vec x\lrp{t_k + \Delta t}} by relying on evaluations of the \nm{\vec y} and \nm{\vec v^{\mathcal X}} time derivatives:
\begin{eqnarray}
\nm{\vec {\dot y}\lrp{t}} & = & \nm{\vec f_y\big(\yvec\lrp{t}, \, \vec v^{\mathcal X}\lrp{t}, \, \mathcal X\lrp{t}, \, t\big)} \label{eq:algebra_integration_comp_x_deriv} \\
\nm{\vec {\dot v}^{\mathcal X}\lrp{t}} & = & \nm{\vec f_v\big(\yvec\lrp{t}, \, \vec v^{\mathcal X}\lrp{t}, \, \mathcal X\lrp{t}, \, t\big)} \label{eq:algebra_integration_comp_v_deriv} 
\end{eqnarray}

The solution consists on employing the Euclidean integration method of choice to obtain \nm{\vec y_{k+1}} and \nm{\vec v_{k+1}^{\mathcal X}}, but rely on the right plus operator and the capitalized exponential map of the Lie group to determine \nm{\mathcal X_{k+1}}. In case of Euler's method, the solution is the following:
\begin{eqnarray}
\nm{\vec y_{k+1}} & \nm{\approx} & \nm{\vec y_k + \Delta t \ \vec{\dot y}(\vec y_k, \, \vec v_k^{\mathcal X}, \, \mathcal X_k, \, t_k)} \label{eq:algebra_integration_comp_y_euler} \\
\nm{\vec v_{k+1}^{\mathcal X}} & \nm{\approx} & \nm{\vec v_k^{\mathcal X} + \Delta t \ \vec{\dot v}^{\mathcal X}(\vec y_k, \, \vec v_k^{\mathcal X}, \, \mathcal X_k, \, t_k)} \label{eq:algebra_integration_comp_v_euler} \\
\nm{\mathcal X_{k+1}} & \nm{\approx} & \nm{\mathcal X_k \oplus \lrsb{\Delta t \ \vec v_k^{\mathcal X}} = \mathcal X_k \circ Exp\lrp{\Delta t \ \vec v_k^{\mathcal X}}} \label{eq:algebra_integration_comp_X_euler} \\
\nm{\vec x_{k+1}} & = & \nm{\vec x\lrp{t_{k+1}} \approx \lrsb{\vec y_{k+1} \ \ \vec v_{k+1}^{\mathcal X} \ \ \mathcal X_{k+1}}^T} \label{eq:algebra_integration_comp_final}
\end{eqnarray}

In case the state vector velocity \nm{\vec v^{\mathcal E}} is that of the global or space tangent space \nm{T_{\mathcal E}\mathcal G}, it is necessary to modify (\ref{eq:algebra_integration_comp_X_euler}) to employ the left plus operator:
\neweq{\mathcal X_{k+1} \approx \lrsb{\Delta t \ \vec v_k^{\mathcal E}} \boxplus \mathcal X_k = Exp\lrp{\Delta t \ \vec v_k^{\mathcal E}} \circ \mathcal X_k} {eq:algebra_integration_comp_X_euler_left}

It is necessary to proceed in a similar manner if a different integration method is selected. In the case of Heun's method with local velocity \nm{\vec v^{\mathcal X}}, the modified integration scheme is the following:
\begin{eqnarray}
\nm{\vec {\dot y}_1} & = & \nm{\vec{\dot y}(\vec y_k, \, \vec v_k^{\mathcal X}, \, \mathcal X_k, \, t_k)} \label{eq:algebra_integration_comp_y_heun_one} \\
\nm{\vec {\dot y}_2} & = & \nm{\vec{\dot y}(\vec y_k + \Delta t \ \vec {\dot y}_1, \vec v_k^{\mathcal X} + \Delta t \ \vec {\dot v}_1^{\mathcal X}, \mathcal{X}_k \oplus \Delta t \ \vec v_k^{\mathcal X},t_k + \Delta t)} \label{eq:algebra_integration_comp_v_heun_one} \\
\nm{\vec {\dot v}_1^{\mathcal X}} & = & \nm{\vec{\dot v}^{\mathcal X}(\vec y_k, \, \vec v_k^{\mathcal X}, \, \mathcal X_k, \, t_k)} \label{eq:algebra_integration_comp_y_heun_two} \\
\nm{\vec {\dot v}_2^{\mathcal X}} & = & \nm{\vec{\dot v}^{\mathcal X}(\vec y_k + \Delta t \ \vec {\dot y}_1, \vec v_k^{\mathcal X} + \Delta t \ \vec {\dot v}_1^{\mathcal X}, \mathcal{X}_k \oplus \Delta t \ \vec v_k^{\mathcal X},t_k + \Delta t)} \label{eq:algebra_integration_comp_v_heun_two} \\
\nm{\vec y_{k+1}} & \nm{\approx} & \nm{\vec y_k + \dfrac{\Delta t}{2} \ \lrsb{\vec {\dot y}_1 + \vec {\dot y}_2}} \label{eq:algebra_integration_comp_y_heun} \\
\nm{\vec v_{k+1}^{\mathcal X}} & \nm{\approx} & \nm{\vec v_k^{\mathcal X} + \dfrac{\Delta t}{2} \ \lrsb{\vec {\dot v}_1^{\mathcal X} + \vec {\dot v}_2^{\mathcal X}}} \label{eq:algebra_integration_comp_v_heun} \\
\nm{\mathcal X_{k+1}} & \nm{\approx} & \nm{\mathcal X_k \oplus \lrsb{\dfrac{\Delta t}{2} \ \lrsb{\vec v_k^{\mathcal X} + \lrp{\vec v_k^{\mathcal X} + \Delta t \ \vec {\dot v}_1^{\mathcal X}}}} = \mathcal X_k \oplus \lrsb{\Delta t \ \vec v_k^{\mathcal X} + \dfrac{\Delta t^2}{2} \ \vec {\dot v}_1^{\mathcal X}}} \nonumber \\
& = & \nm{\mathcal X_k \circ Exp\lrp{\Delta t \ \vec v_k^{\mathcal X} + \dfrac{\Delta t^2}{2} \ \vec {\dot v}_1^{\mathcal X}}} \label{eq:algebra_integration_comp_X_heun} 
\end{eqnarray}

In case of the \nm{4^{th}} order Runge-Kutta integration scheme, it results in the following:
\begin{eqnarray}
\nm{\vec {\dot y}_1} & = & \nm{\vec{\dot y}\lrp{\vec y_k, \, \vec v_k^{\mathcal X}, \, \mathcal X_k, \, t_k}} \label{eq:algebra_integration_comp_y_rk4th_one} \\
\nm{\vec {\dot v}_1^{\mathcal X}} & = & \nm{\vec{\dot v}^{\mathcal X}\lrp{\vec y_k, \, \vec v_k^{\mathcal X}, \, \mathcal X_k, \, t_k}} \label{eq:algebra_integration_comp_vv_rk4th_one} \\
\nm{\vec {\dot y}_2} & = & \nm{\vec{\dot y}\lrp{\vec y_k + \dfrac{\Delta t}{2} \ \vec {\dot y}_1, \vec v_k^{\mathcal X} + \dfrac{\Delta t}{2} \ \vec {\dot v}_1^{\mathcal X}, \mathcal{X}_k \oplus \dfrac{\Delta t}{2} \ \vec v_k^{\mathcal X},t_k + \dfrac{\Delta t}{2}}} \label{eq:algebra_integration_comp_y_rk4th_two} \\
\nm{\vec {\dot v}_2^{\mathcal X}} & = & \nm{\vec{\dot v}^{\mathcal X}\lrp{\vec y_k + \frac{\Delta t}{2} \ \vec {\dot y}_1, \vec v_k^{\mathcal X} + \dfrac{\Delta t}{2} \ \vec {\dot v}_1^{\mathcal X}, \mathcal{X}_k \oplus \dfrac{\Delta t}{2} \ \vec v_k^{\mathcal X},t_k + \dfrac{\Delta t}{2}}} \label{eq:algebra_integration_comp_vv_rk4th_two} \\
\nm{\vec {\dot y}_3} & = & \nm{\vec{\dot y}\lrp{\vec y_k + \dfrac{\Delta t}{2} \ \vec {\dot y}_2, \vec v_k^{\mathcal X} + \dfrac{\Delta t}{2} \ \vec {\dot v}_2^{\mathcal X}, \mathcal{X}_k \oplus \dfrac{\Delta t}{2} \ \lrsb{\vec v_k^{\mathcal X} + \dfrac{\Delta t}{2} \ \vec {\dot v}_1^{\mathcal X}},t_k + \dfrac{\Delta t}{2}}} \label{eq:algebra_integration_comp_y_rk4th_three} \\
\nm{\vec {\dot v}_3^{\mathcal X}} & = & \nm{\vec{\dot v}^{\mathcal X}\lrp{\vec y_k + \frac{\Delta t}{2} \ \vec {\dot y}_2, \vec v_k^{\mathcal X} + \dfrac{\Delta t}{2} \ \vec {\dot v}_2^{\mathcal X}, \mathcal{X}_k \oplus \dfrac{\Delta t}{2} \ \lrsb{\vec v_k^{\mathcal X} + \dfrac{\Delta t}{2} \ \vec {\dot v}_1^{\mathcal X}},t_k + \dfrac{\Delta t}{2}}} \label{eq:algebra_integration_comp_vv_rk4th_three} \\
\nm{\vec {\dot y}_4} & = & \nm{\vec{\dot y}\lrp{\vec y_k + \Delta t \ \vec {\dot y}_3, \vec v_k^{\mathcal X} + \Delta t \ \vec {\dot v}_3^{\mathcal X}, \mathcal{X}_k \oplus \Delta t \ \lrsb{\vec v_k^{\mathcal X} + \dfrac{\Delta t}{2} \ \vec {\dot v}_2^{\mathcal X}}, t_k + \Delta t}} \label{eq:algebra_integration_comp_y_rk4th_four} \\
\nm{\vec {\dot v}_4^{\mathcal X}} & = & \nm{\vec{\dot v}^{\mathcal X}\lrp{\vec y_k + \Delta t \ \vec {\dot y}_3, \vec v_k^{\mathcal X} + \Delta t \ \vec {\dot v}_3^{\mathcal X}, \mathcal{X}_k \oplus \Delta t \ \lrsb{\vec v_k^{\mathcal X} + \dfrac{\Delta t}{2} \ \vec {\dot v}_2^{\mathcal X}}, t_k + \Delta t}} \label{eq:algebra_integration_comp_vv_rk4th_four} \\
\nm{\vec y_{k+1}} & \nm{\approx} & \nm{\vec y_k + \Delta t \ \lrsb{\dfrac{\vec {\dot y}_1}{6} + \dfrac{\vec {\dot y}_2}{3} + \dfrac{\vec {\dot y}_3}{3} + \dfrac{\vec {\dot y}_4}{6}}} \label{eq:algebra_integration_comp_y_rk4th} \\
\nm{\vec v_{k+1}^{\mathcal X}} & \nm{\approx} & \nm{\vec v_k^{\mathcal X} + \Delta t \ \lrsb{\dfrac{\vec {\dot v}_1^{\mathcal X}}{6} + \dfrac{\vec {\dot v}_2^{\mathcal X}}{3} + \dfrac{\vec {\dot v}_3^{\mathcal X}}{3} + \dfrac{\vec {\dot v}_4^{\mathcal X}}{6}}} \label{eq:algebra_integration_comp_vv_rk4th} \\
\nm{\mathcal X_{k+1}} & \nm{\approx} & \nm{\mathcal X_k \oplus \lrsb{\dfrac{\Delta t}{6} \ \vec v_k^{\mathcal X} + \dfrac{\Delta t}{3} \lrp{\vec v_k^{\mathcal X} + \dfrac{\Delta t}{2} \ \vec {\dot v}_1^{\mathcal X}} + \dfrac{\Delta t}{3} \lrp{\vec v_k^{\mathcal X} + \dfrac{\Delta t}{2} \ \vec {\dot v}_2^{\mathcal X}} + \dfrac{\Delta t}{6} \lrp{\vec v_k^{\mathcal X} + \Delta t \ \vec {\dot v}_3^{\mathcal X}}}} \nonumber \\
& = & \nm{\mathcal X_k \oplus \lrsb{\Delta t \ \vec v_k^{\mathcal X} + \dfrac{\Delta t^2}{6} \ \vec {\dot v}_1^{\mathcal X} + \dfrac{\Delta t^2}{6} \ \vec {\dot v}_2^{\mathcal X} + \dfrac{\Delta t^2}{6} \ \vec {\dot v}_3^{\mathcal X}}} \nonumber \\
& = & \nm{\mathcal X_k \circ Exp\lrp{\Delta t \ \vec v_k^{\mathcal X} + \dfrac{\Delta t^2}{6} \ \vec {\dot v}_1^{\mathcal X} + \dfrac{\Delta t^2}{6} \ \vec {\dot v}_2^{\mathcal X} + \dfrac{\Delta t^2}{6} \ \vec {\dot v}_3^{\mathcal X}}} \label{eq:algebra_integration_comp_X_rk4th} 
\end{eqnarray}

Similar modifications to that of (\ref{eq:algebra_integration_comp_X_euler_left}) are required if the tangent space velocity \nm{\vec v^{\mathcal E}} is viewed in the global space.

%%%%%%%%%%%%%%%%%%%%%%%%%%%%%%%%%%%%%%%%%%%%%%%%%%%%%%%%%%%%%%%%%%%%%%%%
%%%%%%%%%%%%%%%%%%%%%%%%%%%%%%%%%%%%%%%%%%%%%%%%%%%%%%%%%%%%%%%%%%%%%%%%
%%%%%%%%%%%%%%%%%%%%%%%%%%%%%%%%%%%%%%%%%%%%%%%%%%%%%%%%%%%%%%%%%%%%%%%%
% SECTION      GRADIENT DESCENT OPTIMIZATION IN LIE GROUPS
%%%%%%%%%%%%%%%%%%%%%%%%%%%%%%%%%%%%%%%%%%%%%%%%%%%%%%%%%%%%%%%%%%%%%%%%
%%%%%%%%%%%%%%%%%%%%%%%%%%%%%%%%%%%%%%%%%%%%%%%%%%%%%%%%%%%%%%%%%%%%%%%%
%%%%%%%%%%%%%%%%%%%%%%%%%%%%%%%%%%%%%%%%%%%%%%%%%%%%%%%%%%%%%%%%%%%%%%%%

\section{Gradient Descent Optimization in Lie Groups}\label{sec:algebra_gradient_descent}

The Gauss-Newton implementation of the gradient descent optimization method is described in section \ref{sec:euclidean_gradient_descent} for the case of Euclidean spaces. Consider now a Lie group element \nm{\mathcal X \in \mathcal G} and a nonlinear map based on its tangent space \nm{\lrb{\vec f: \mathbb{R}^m \rightarrow \mathbb{R}^n \ | \ \vec f\lrp{\vec \tau} \in \mathbb{R}^n, \vec \tau = Log\lrp{\mathcal X} \in \mathbb{R}^m, \vec \tau^\wedge = log\lrp{\mathcal X} \in \mathfrak{m}, \forall \, \mathcal X \in \mathcal G}}. As in the Euclidean case, it is possible to evaluate its Jacobian \nm{\lrb{\vec J: \mathbb{R}^m \rightarrow \mathbb{R}^{nxm} \ | \ \vec J\lrp{\vec \tau} = \partial{\vec f\lrp{\vec \tau}} / \partial{\vec \tau} \in \mathbb{R}^{nxm}, \forall \ \vec \tau \in \mathbb{R}^m}}, and the error function is defined as \nm{\vec{\mathcal E}\lrp{\vec \tau} = \vec{\mathcal E}\big(Log(\mathcal X)) = \vec f\lrp{\vec \tau} - \vec f_T}.
 
Given an initial state \nm{\mathcal X_0 = Exp\lrp{\vec \tau_0}}, the objective is to determine a Lie group element \nm{\mathcal X = Exp\lrp{\vec \tau} = \Delta \vec \tau^{\mathcal E} \boxplus \mathcal X_0 = \Delta \vec \tau^{\mathcal E} \circ Exp\lrp{\vec \tau_0}} in the vicinity of \nm{\mathcal X_0} for which the cost function norm \nm{\| \vec{\mathcal E}\lrp{\vec \tau} \| \in \mathbb{R}} holds a local minimum. As the \nm{\vec \tau^{{\mathcal E}\wedge}} perturbation belongs to the spatial tangent space \nm{T_{\mathcal E}\mathcal G}, all gradient descent methods advance the solution by means of (\ref{eq:algebra_gradient_descent_iterative_lie_left}):
\neweq{\mathcal X_{k+1} \longleftarrow \Delta \vec \tau_k^{\mathcal E} \boxplus \mathcal X_k = \Delta \vec \tau_k^{\mathcal E} \circ Exp\lrp{\vec \tau_k}}{eq:algebra_gradient_descent_iterative_lie_left}

A robust formulation is necessary to ensure that the transform vector \nm{\tau}, which is treated as Euclidean by the functions \nm{\vec{\mathcal E}} and \nm{\vec f}, is advanced as a member of the tangent space that adheres to the Lie group constraints, as guaranteed by the use of the left Jacobian inverse \nm{\vec J_L^{-1}\lrp{\vec \tau}} (table \ref{tab:algebra_lie_jacobians}). The Gauss-Newton method first step hence consists of two first order Taylor expansions, one for the logarithmic map and a second for the function \nm{\vec f}:
\begin{eqnarray}
\nm{\vec{\mathcal E}_{k+1}} & = & \nm{\vec f_{k+1} - \vec f_T = \vec f\Big(Log\lrp{\Delta \vec \tau_k^{\mathcal E} \circ Exp\lrp{\vec \tau_k}}\Big) - \vec f_T \approx \vec f\Big(\vec \tau_k + \vec J_L^{-1}\Bigr\rvert_{\ds{Exp(\vec \tau_k)}} \ \Delta \vec \tau_k^{\mathcal E}\Big) - \vec f_T} \nonumber \\
& \nm{\approx} & \nm{\vec f_k + \vec J_k \, \vec J_L^{-1}\Bigr\rvert_{\ds{Exp(\vec \tau_k)}} \ \Delta \vec \tau_k^{\mathcal E} - \vec f_T = \vec{\mathcal E}_k + \vec J_k \, \vec J_{Lk}^{-1} \ \Delta \vec \tau_k^{\mathcal E}} \label{eq:algebra_gradient_descent_taylor_lie_left}
\end{eqnarray}    

The remaining Gauss-Newton steps are modified accordingly:
\begin{eqnarray}
\nm{\| \vec{\mathcal E}_{k+1} \|} & = & \nm{\vec{\mathcal E}_{k+1}^T \, \vec{\mathcal E}_{k+1} = \vec{\mathcal E}_k^T \, \vec{\mathcal E}_k + \Delta \vec \tau_k^{{\mathcal E}T} \, \lrsb{\vec J_k \, \vec J_{Lk}^{-1}}^T \, \lrsb{\vec J_k \, \vec J_{Lk}^{-1}} \, \Delta \vec \tau_k^{\mathcal E} + 2 \, \Delta \vec \tau_k^{{\mathcal E}T} \, \lrsb{\vec J_k \, \vec J_{Lk}^{-1}}^T \, \vec{\mathcal E}_k} \label{eq:algebra_gradient_descent_norm_lie_left} \\
\nm{\pderpar{\| \vec{\mathcal E}_{k+1} \|}{\Delta \vec \tau_k^{\mathcal E}}} & = & \nm{0 \ \longrightarrow \ \Delta \vec \tau_k^{\mathcal E} = - \lrsb{\lrsb{\vec J_k \, \vec J_{Lk}^{-1}}^T \, \lrsb{\vec J_k \, \vec J_{Lk}^{-1}}}^{-1} \, \lrsb{\vec J_k \, \vec J_{Lk}^{-1}}^T \, \vec{\mathcal E}_k \ \longrightarrow} \nonumber \\
\nm{\Delta \vec \tau_k^{\mathcal E}} & = & \nm{- \lrsb{\vec J_{Lk}^{-T} \, \vec J_k^T \, \vec J_k \, \vec J_{Lk}^{-1}}^{-1} \, \vec J_{Lk}^{-T} \, \vec J_k^T \, \vec{\mathcal E}_k} \label{eq:algebra_gradient_descent_solution_lie_left} 
\end{eqnarray}    

If the perturbation \nm{\vec \tau^{{\mathcal X}\wedge}} instead belongs to the local tangent space \nm{T_{\mathcal X}\mathcal G}, it is necessary to employ the right Jacobian instead of the left one, resulting in:
\begin{eqnarray}
\nm{\mathcal X_{k+1}} & \nm{\longleftarrow} & \nm{\mathcal X_k \oplus \Delta \vec \tau_k^{\mathcal X} = Exp\lrp{\vec \tau_k} \circ \Delta \vec \tau_k^{\mathcal X}} \label{eq:algebra_gradient_descent_iterative_lie_right} \\
\nm{\Delta \vec \tau_k^{\mathcal X}} & = & \nm{- \lrsb{\vec J_{Rk}^{-T} \, \vec J_k^T \, \vec J_k \, \vec J_{Rk}^{-1}}^{-1} \, \vec J_{Rk}^{-T} \, \vec J_k^T \, \vec{\mathcal E}_k} \label{eq:algebra_gradient_descent_solution_lie_right} 
\end{eqnarray}    

%%%%%%%%%%%%%%%%%%%%%%%%%%%%%%%%%%%%%%%%%%%%%%%%%%%%%%%%%%%%%%%%%%%%%%%%
%%%%%%%%%%%%%%%%%%%%%%%%%%%%%%%%%%%%%%%%%%%%%%%%%%%%%%%%%%%%%%%%%%%%%%%%
%%%%%%%%%%%%%%%%%%%%%%%%%%%%%%%%%%%%%%%%%%%%%%%%%%%%%%%%%%%%%%%%%%%%%%%%
% SECTION		STATE ESTIMATION IN LIE GROUPS
%%%%%%%%%%%%%%%%%%%%%%%%%%%%%%%%%%%%%%%%%%%%%%%%%%%%%%%%%%%%%%%%%%%%%%%%
%%%%%%%%%%%%%%%%%%%%%%%%%%%%%%%%%%%%%%%%%%%%%%%%%%%%%%%%%%%%%%%%%%%%%%%%
%%%%%%%%%%%%%%%%%%%%%%%%%%%%%%%%%%%%%%%%%%%%%%%%%%%%%%%%%%%%%%%%%%%%%%%%

\section{State Estimation in Lie Groups}\label{sec:algebra_SS}

The state estimation \hypertt{EKF} discussion of section \ref{sec:SS} states that given a continuous time nonlinear Euclidean state system (\ref{eq:SS_cont_time_system}) with process noise provided by (\ref{eq:SS_cont_time_system_noise1}) and (\ref{eq:SS_cont_time_system_noise2}), together with a series of discrete time nonlinear observations (\ref{eq:SS_measur_nonlinear}) with measurement noise given by (\ref{eq:SS_measur_nonlinear_noise1}) and (\ref{eq:SS_measur_nonlinear_noise2}), and considering no correlation between both noises (\ref{eq:SS_measur_nonlinear_noise3}), it is possible to compute estimations of the Euclidean state and its covariance at the same time points at which the observations are provided, in such a way that the estimation errors (difference with respect to the true state) are zero mean. The estimations are obtained by means of (\ref{eq:SS_EKF_x0_initial_state_FINAL}) through (\ref{eq:SS_EKF_P_plus_propagate_FINAL}).

Instead of the Euclidean space time varying state vector \nm{\vec x\lrp{t} \in \mathbb{R}^m} considered in section \ref{sec:SS}, consider now that the continuous time nonlinear state system is composed by a vector \nm{\vec z\lrp{t} \in \mathbb{R}^n}, an element of a Lie group \nm{\mathcal X\lrp{t} \in \mathcal G}, and a vector \nm{\vec v^{\mathcal X}\lrp{t} \in \mathbb{R}^m} representing the velocity of \nm{\mathcal X} as it moves along its manifold, contained in the local or body tangent space \nm{T_{\mathcal X}\mathcal G}. As \nm{\mathcal X} is not Euclidean, the direct application of the \hypertt{EKF} state estimation scheme of section \ref{sec:SS} would result in the need to continuously reproject the estimated Lie group elements \nm{\mathcal X_k} back to the manifold as otherwise the estimated states would not comply with the Lie group constraints. The repeated deviations and reprojections from and to the manifold may result in a significant degradation in the estimation accuracy.

The most rigorous and precise way to adapt the \hypertt{EKF} scheme is to exclude the Lie group element \nm{\mathcal X \in \mathcal G} from the state system, replacing it by a local tangent space perturbation \nm{\Delta \vec \tau^{\mathcal X} \in T_{\mathcal X}\mathcal G}. Each filter step now consists on estimating the Lie group element \nm{\hat{\mathcal X}_k^+ = \hat{\mathcal X}^+\lrp{t_k} \in \mathcal G}, the state vector \nm{\xvecest_k^+ = \xvecest^+\lrp{t_k} = \lrsb{\Delta \hat{\vec \tau}_k^{{\mathcal X}+} \ \ \hat{\vec v}_k^{{\mathcal X}+} \ \ \hat{\vec z}_k^+}^T \in \mathbb{R}^{2 \, m + n}}, and its covariance \nm{\Pvec_k^+ = \Pvec^+\lrp{t_k} \in \mathbb{R}^{(2 \, m + n)x(2 \, m + n)}}, based on their values at \nm{t_{k-1} = \lrp{k-1} \, \Deltat}. For clarity purposes the Lie velocity \nm{\vec v^{\mathcal X}} and the state vector \nm{\vec z} can be grouped into a bigger Euclidean state vector \nm{\vec p = \lrsb{\vec v^{\mathcal X} \ \ \vec z}^T \in \mathbb{R}^{m+n}}, so \nm{\xvecest_k^+ = \lrsb{\Delta \hat{\vec \tau}_k^{\mathcal X} \ \ \hat{\vec p}_k}^T}.

Considering with no loss of generality that the local tangent state perturbation \nm{\Delta \vec \tau^{\mathcal X}} is located on the first positions of the combined state vector \nm{\xvec}, the definition of the covariance matrix is a combination of that of its Euclidean components (\ref{eq:SS_discr_time_system_state_cov}) and its local Lie counterparts (\ref{eq:algebra_lie_covariance_right_def}), with additional combined members:
\begin{eqnarray}
\nm{\Pvec_k} & = & \nm{\begin{bmatrix} \nm{\vec C_{{\mathcal {XX}},k}^{\mathcal X}} & \nm{\vec C_{{\mathcal X}p,k}^{\mathcal X}} \\ \nm{\vec C_{p{\mathcal X},k}^{\mathcal X}} & \nm{\vec C_{pp,k}} \end{bmatrix} \ \ \ \ \ \ \ \ \ \ \ \ \ \ \ \ \ \ \ \ \ \ \ \ \ \ \ \ \ \ \ \ \ \ \ \ \ \ \ \ \ \ \ \ \ \ \ \ \ \ \ \ \ \ \ \in \mathbb{R}^{(2 \, m + n)x(2 \, m + n)}} \label{eq:algebra_SS_EKF_P_generic} \\
\nm{\vec C_{{\mathcal {XX}},k}^{\mathcal X}} & = & \nm{E\lrsb{\Delta \vec \tau_k^{\mathcal X} \, \Delta \vec \tau_k^{{\mathcal X,T}}} = E\lrsb{\lrp{\mathcal X_k \ominus \vec \mu_{{\mathcal X},k}} \, \lrp{\mathcal X_k \ominus \vec \mu_{{\mathcal X},k}}^T} \ \ \ \ \ \ \ \ \ \ \ \ \in \mathbb{R}^{mxm}} \label{eq:algebra_SS_EKF_P_lie} \\
\nm{\vec C_{{\mathcal X}p,k}^{\mathcal X}} & = & \nm{E\lrsb{\Delta \vec \tau_k^{\mathcal X} \, \lrp{\vec p_k - \vec \mu_{p,k}}^T} = E\lrsb{\lrp{\mathcal X_k \ominus \vec \mu_{{\mathcal X},k}} \, \lrp{\vec p_k - \vec \mu_{p,k}}^T} \ \ \ \ \in \mathbb{R}^{mx(m+n)}} \label{eq:algebra_SS_EKF_P_lieeuc} \\
\nm{\vec C_{p{\mathcal X},k}^{\mathcal X}} & = & \nm{E\lrsb{\lrp{\vec p_k - \vec \mu_{p,k}} \, \Delta \vec \tau_k^{{\mathcal X},T}} = E\lrsb{\lrp{\vec p_k - \vec \mu_{p,k}} \, \lrp{\mathcal X_k \ominus \vec \mu_{{\mathcal X},k}}^T} \ \ \ \in \mathbb{R}^{(m+n)xm}} \label{eq:algebra_SS_EKF_P_euclie} \\
\nm{\vec C_{pp,k}} & = & \nm{E\lrsb{\lrp{\vec p_k - \vec \mu_{p,k}} \, \lrp{\vec p_k - \vec \mu_{p,k}}^T} \ \ \ \ \ \ \ \ \ \ \ \ \ \ \ \ \ \ \ \ \ \ \ \ \ \ \ \ \ \ \ \ \ \ \ \ \ \ \  \in \mathbb{R}^{(m+n)x(m+n)}} \label{eq:algebra_SS_EKF_P_euc} 
\end{eqnarray}

The following paragraphs do not describe the full state estimation process, but only the changes with respect to the Euclidean case described in section \ref{sec:SS}:
\begin{itemize}
\item \textbf{Initialization}. The Lie group element \nm{\hat{\mathcal X}_0^+}, Lie velocity \nm{\hat{\vec v}_0^{{\mathcal X}+}}, and Euclidean state vector \nm{\hat{\vec z}_0^+} are initialized with their expected values, while the local tangent space perturbation \nm{\Delta \hat{\vec \tau}_0^{{\mathcal X}+}} is initialized to zero. The covariance of the initial estimation error \nm{\Pvec_0^+} represents the uncertainty in the initial estimation \nm{\xvecest_0^+}.
\begin{eqnarray}
\nm{\hat{\mathcal X}_0^+} & = & \nm{\vec \mu_{\mathcal X,0} = E\lrsb{\mathcal X_0}} \label{eq:algebra_SS_EKF_man0_initial_manifold} \\ 
\nm{\xvecest_0^+}         & = & \nm{\vec \mu_{x,0} = E\lrsb{\xvec_0} = \lrsb{\Delta \hat{\vec \tau}_0^{{\mathcal X}+} \ \hat{\vec v}_0^{{\mathcal X}+} \ \hat{\vec z}_0^+}^T = \lrsb{\Delta \hat{\vec \tau}_0^{{\mathcal X}+} \ \hat{\vec p}_0^+}^T = E\Big[\lrsb{\vec{0}_m \ \ \vec v_0^{\mathcal X} \ \ \vec z_0}^T\Big]} \label{eq:algebra_SS_EKF_x0_initial_state} \\ 
\nm{\Pvec_0^+}            & = & \nm{\begin{bmatrix} \nm{E\lrsb{\lrp{\mathcal X_0 \ominus \vec \mu_{{\mathcal X},0}} \, \lrp{\mathcal X_0 \ominus \vec \mu_{{\mathcal X},0}}^T}} & \nm{E\lrsb{\lrp{\mathcal X_0 \ominus \vec \mu_{{\mathcal X},0}} \, \lrp{\vec p_0 - \vec \mu_{p,0}}^T}} \\ \nm{E\lrsb{\lrp{\vec p_0 - \vec \mu_{p,0}} \, \lrp{\mathcal X_0 \ominus \vec \mu_{{\mathcal X},0}}^T}} & \nm{E\lrsb{\lrp{\vec p_0 - \vec \mu_{p,0}} \, \lrp{\vec p_0 - \vec \mu_{p,0}}^T}} \end{bmatrix}} \label{eq:algebra_SS_EKF_P0_initial_covariance} 
\end{eqnarray}

\item \textbf{Time Update Equations}. The first \hypertt{EKF} step propagates the state estimation without the use of any observations, and is similar to that described in section \ref{sec:SS}. The (\ref{eq:SS_cont_time_system}) continuous time nonlinear state system is however replaced by (\ref{eq:algebra_SS_cont_time_system}):
\begin{eqnarray}
\nm{\xvecdot\lrp{t}} & = & \nm{\lrsb{\Delta \vec{\dot \tau}^{\mathcal X} \ \ \vec{\dot v}^{\mathcal X} \ \ \vec {\dot z}}^T = \vec f\big(\mathcal X\lrp{t} \oplus \Delta \vec \tau^{\mathcal X}\lrp{t}, \, \vec v^{\mathcal X}\lrp{t}, \, \vec z\lrp{t}, \, \uvec\lrp{t}, \ \wvec\lrp{t}, \, t\big)} \label{eq:algebra_SS_cont_time_system} \\
\nm{\Delta \vec{\dot \tau}^{\mathcal X}} & = & \nm{\vec v^{\mathcal X}} \label{eq:algebra_SS_cont_time_system_tau} \\
\nm{\vec{\dot p}} & = & \nm{\lrsb{\vec{\dot v}^{\mathcal X} \ \ \vec {\dot z}}^T = \vec f_p\big(\mathcal X\lrp{t} \oplus \Delta \vec \tau^{\mathcal X}\lrp{t}, \, \vec v^{\mathcal X}\lrp{t}, \, \vec z\lrp{t}, \, \uvec\lrp{t}, \ \wvec\lrp{t}, \, t\big)} \label{eq:algebra_SS_cont_time_system_y}
\end{eqnarray}

Linearization of the continuous time system results in the (\ref{eq:SS_cont_time_system_linear_system_matrix}) system matrix \nm{\Avec\lrp{t} \in \mathbb{R}^{(2\,m+n)x(2\,m+n)}}, where various blocks are likely to be based on the \nm{\ominus} and \nm{-} Jacobians of table \ref{tab:algebra_lie_jacobians}. Its discretization by means of (\ref{eq:SS_discr_time_system_linear_system_matrix}) leads to the system state transition matrix \nm{\Fvec_k \in \mathbb{R}^{(2\,m+n)x(2\,m+n)}}. With no other differences with respect to the section \ref{sec:SS} Euclidean process, the manifold element is left unchanged, while the a priori state vector and error covariance are obtained by means of (\ref{eq:algebra_SS_xest_minus_propagate}) and (\ref{eq:algebra_SS_P_minus_propagate}):
\begin{eqnarray}
\nm{\hat{\mathcal X}_k^-} & = & \nm{\hat{\mathcal X}_{k-1}^+} \label{eq:algebra_SS_man_minus_propagate} \\
\nm{\xvecest_k^-}         & = & \nm{\xvecest_{k-1}^+ + \Deltat \cdot \vec f \, \big(\hat{\mathcal X}_{k-1}^+ \oplus \Delta \hat{\vec \tau}_{k-1}^{{\mathcal X}+}, \, \hat{\vec v}_{k-1}^{{\mathcal X}+}, \, \hat{\vec z}_{k-1}^+, \, \uvec_{k-1}, \ \vec 0, \, t_{k-1}\big)}\label{eq:algebra_SS_xest_minus_propagate} \\
\nm{\Pvec_k^-}            & = & \nm{\Fvec_{k-1} \, \Pvec_{k-1}^+ \, \Fvec_{k-1}^T + \Qtilde_{d,k-1}}\label{eq:algebra_SS_P_minus_propagate}
\end{eqnarray}

\item \textbf{Measurement Update Equations}. The second \hypertt{EKF} step updates the estimations by means of the observations, and is very similar to that described in section \ref{sec:SS}. The (\ref{eq:SS_measur_nonlinear}) discrete time nonlinear observation system is however replaced by (\ref{eq:algebra_SS_measur_nonlinear}):
\neweq{\yvec_k = \vec h\lrp{\mathcal X_k \oplus \Delta \vec \tau_k^{\mathcal X}, \, \vec v_k^{\mathcal X}, \, \vec z_k, \, \, \vvec_k, \, t_k}}{eq:algebra_SS_measur_nonlinear}

Its linearization results in the (\ref{eq:SS_measur_linear_output_matrix}) output matrix \nm{\Hvec_k \in \mathbb{R}^{qx(2\,m+n)}}, where it is also likely for various blocks to be based on the \nm{\ominus} and \nm{-} Jacobians of table \ref{tab:algebra_lie_jacobians}. With no other differences with respect to the section \ref{sec:SS} Euclidean process, the manifold element is again left unchanged, while the a posteriori state vector and error covariance are obtained by means of (\ref{eq:algebra_SS_EKF_xest_plus_propagate}) and (\ref{eq:algebra_SS_EKF_P_plus_propagate}):
\begin{eqnarray}
\nm{\Kvec_k}              & = & \nm{\Pvec_k^- \, \Hvec_k^T \lrp{\Hvec_k \, \Pvec_k^- \, \Hvec_k^T + \Rtilde_k}^{-1}}\label{eq:algebra_SS_EKF_kalman_gain} \\
\nm{\hat{\mathcal X}_k^+} & = & \nm{\hat{\mathcal X}_k^-} \label{eq:algebra_SS_man_plus_propagate} \\
\nm{\xvecest_k^+}         & = & \nm{\xvecest_k^- + \Kvec_k \, \lrsb{\yvec_k - h\lrp{\hat{\mathcal X}_k^- \oplus \Delta \hat{\vec \tau}_k^{{\mathcal X}-}, \, \hat{\vec v}_k^{{\mathcal X}-}, \, \hat{\vec z}_k^-, \, \vec 0, \, t_k}}} \label{eq:algebra_SS_EKF_xest_plus_propagate} \\
\nm{\Pvec_k^+}            & = & \nm{\lrp{\Ivec - \Kvec_k \, \Hvec_k}\, \Pvec_k^- \, \lrp{\Ivec - \Kvec_k \, \Hvec_k}^T + \Kvec_k \, \Rtilde_k \, \Kvec_k^T}\label{eq:algebra_SS_EKF_P_plus_propagate}
\end{eqnarray}

\item \textbf{Reset Equations}. The third \hypertt{EKF} step, which is not necessary for purely Euclidean systems, resets the a posteriori estimation of the tangent space perturbation \nm{\Delta \hat{\vec \tau}_k^{{\mathcal X}+}} to zero while modifying the a posteriori estimations for the Lie group element \nm{\hat{\mathcal X}_k^+} and the error covariance \nm{\Pvec_k^+} accordingly\footnote{The a posteriori estimations for the Lie velocity \nm{\hat{\vec v}_k^{{\mathcal X}+}} and the Euclidean components \nm{\hat{\vec z}_k^+} are not modified in this step.}. Note that the accuracy of the linearizations present in the previous two steps, which result in the \nm{\vec A\lrp{t}} and \nm{\vec H_k} system and output matrices, is based on the first order Taylor expansions present in table \ref{tab:algebra_lie_jacobians}, which are directly related to the size of the tangent space perturbations. Although it is not strictly necessary to execute this step in every \hypertt{EKF} cycle, the accuracy of the whole state estimation process is improved by maintaining the perturbations as small as possible, so it is recommended to never bypass the reset step.

Taking into account that the Lie group element is going to be updated per (\ref{eq:algebra_SS_man_reset}), the error covariance is propagated to the new Lie group element per (\ref{eq:algebra_lie_covariance_right_propagation}) as follows:
\begin{eqnarray}
\nm{\Pvec_k^+} & \nm{\longleftarrow} & \nm{\vec D \, \Pvec_k^+ \, \vec D^T} \label{eq:algebra_SS_P_reset} \\
\nm{\vec D} & = & \nm{\begin{bmatrix} \nm{\vec J_{\ds{\oplus \; \mathcal X}}^{\ds{\ominus \; \hat{\mathcal X}_k^+ \oplus \Delta \hat{\vec \tau}_k^{{\mathcal X}+}}}} & \nm{\vec{0}_{mx(m+n)}} \\ \nm{\vec{0}_{(m+n)xm}} & \nm{\vec {I}_{(m+n)x(m+n)}} \end{bmatrix} \ \ \ \ \in \mathbb{R}^{(2\,m+n)x(2\,m+n)}} \label{eq:algebra_SS_D_reset}
\end{eqnarray}

Once propagated, the Lie group element is updated with the local tangent space perturbation, which is itself reset to zero:
\begin{eqnarray}
\nm{\hat{\mathcal X}_k^+} & \nm{\longleftarrow} & \nm{\hat{\mathcal X}_k^+ \oplus \Delta \hat{\vec \tau}_k^{{\mathcal X}+}} \label{eq:algebra_SS_man_reset} \\
\nm{\Delta \hat{\vec \tau}_k^{{\mathcal X}+}} & \nm{\longleftarrow} & \nm{\vec{0}_m} \label{eq:algebra_SS_pertur_reset} 
\end{eqnarray}
\end{itemize}

Although the above process has been described making use of a local tangent space perturbation \nm{\Delta \vec \tau^{\mathcal X} \in T_{\mathcal X}\mathcal G} and a vector \nm{\vec v^{\mathcal X} \in \mathbb{R}^m} also viewed in the local tangent space that represents the velocity of \nm{\mathcal X \in \mathcal G} as it moves along its manifold, there exists an equivalent formulation that employs perturbations and velocities viewed in the manifold global tangent space, this is, \nm{\Delta \vec \tau^{\mathcal E} \in T_{\mathcal E}\mathcal G} and \nm{\vec v^{\mathcal E} \in \mathbb{R}^m}. In this case, the filter estimates the Lie group element \nm{\hat{\mathcal X}_k^+ \in \mathcal G}, the state vector \nm{\xvecest_k^+ = \lrsb{\Delta \hat{\vec \tau}_k^{{\mathcal E}+} \ \ \hat{\vec v}_k^{{\mathcal E}+} \ \ \hat{\vec z}_k^+}^T \in \mathbb{R}^{2 \, m + n}}, and the error covariance \nm{\Pvec_k^+}:
\begin{eqnarray}
\nm{\Pvec_k} & = & \nm{\begin{bmatrix} \nm{\vec C_{{\mathcal {XX}},k}^{\mathcal E}} & \nm{\vec C_{{\mathcal X}p,k}^{\mathcal E}} \\ \nm{\vec C_{p{\mathcal X},k}^{\mathcal E}} & \nm{\vec C_{pp,k}} \end{bmatrix} \ \ \ \ \ \ \ \ \ \ \ \ \ \ \ \ \ \ \ \ \ \ \ \ \ \ \ \ \ \ \ \ \ \ \ \ \ \ \ \ \ \ \ \ \ \ \ \ \ \ \ \ \ \ \in \mathbb{R}^{(2 \, m + n)x(2 \, m + n)}} \label{eq:algebra_SS_EKF_P_generic_global} \\
\nm{\vec C_{{\mathcal {XX}},k}^{\mathcal E}} & = & \nm{E\lrsb{\Delta \vec \tau_k^{\mathcal E} \, \Delta \vec \tau_k^{{\mathcal E,T}}} = E\lrsb{\lrp{\mathcal X_k \boxminus \vec \mu_{{\mathcal X},k}} \, \lrp{\mathcal X_k \boxminus \vec \mu_{{\mathcal X},k}}^T} \ \ \ \ \ \ \ \ \ \ \ \ \in \mathbb{R}^{mxm}} \label{eq:algebra_SS_EKF_P_lie_global} \\
\nm{\vec C_{{\mathcal X}p,k}^{\mathcal E}} & = & \nm{E\lrsb{\Delta \vec \tau_k^{\mathcal E} \, \lrp{\vec p_k - \vec \mu_{p,k}}^T} = E\lrsb{\lrp{\mathcal X_k \boxminus \vec \mu_{{\mathcal X},k}} \, \lrp{\vec p_k - \vec \mu_{p,k}}^T} \ \ \ \ \in \mathbb{R}^{mx(m+n)}} \label{eq:algebra_SS_EKF_P_lieeuc_global} \\
\nm{\vec C_{p{\mathcal X},k}^{\mathcal E}} & = & \nm{E\lrsb{\lrp{\vec p_k - \vec \mu_{p,k}} \, \Delta \vec \tau_k^{{\mathcal E},T}} = E\lrsb{\lrp{\vec p_k - \vec \mu_{p,k}} \, \lrp{\mathcal X_k \boxminus \vec \mu_{{\mathcal X},k}}^T} \ \ \ \in \mathbb{R}^{(m+n)xm}} \label{eq:algebra_SS_EKF_P_euclie_global} 
\end{eqnarray}

The only additional changes are the use of (\nm{\Delta \vec \tau^{\mathcal E} \boxplus \mathcal X}) instead of (\nm{\mathcal X \oplus \Delta \vec \tau^{\mathcal X}}), the fact that the blocks of the \nm{\vec A\lrp{t}} and \nm{\vec H_k} matrices are now based on the \nm{\boxminus} and \nm{-} Jacobians of table \ref{tab:algebra_lie_jacobians}, and the use of \nm{\vec J_{\ds{\boxplus \; \mathcal X}}^{\ds{\boxminus \; \vec \tau \boxplus \mathcal X}} \Bigr\rvert_{\ds{\Delta \hat{\vec \tau}_k^{{\mathcal E}+} \boxplus \hat{\mathcal X}_k^+}}} to propagate the covariance.

%% file: files_arxiv/ch04_rotate.tex
\chapter{Rotation of Rigid Bodies} \label{cha:Rotate}

A \emph{rigid body} is an object in which the distance between any two of its points is constant, as is the orientation between any two of its vectors (refer to section \ref{sec:algebra_points_and_vectors} for the definition of points and vectors in \nm{\mathbb{R}^3}). Rotational rigid body motion is that in which one of its points, named the \emph{center of rotation} \nm{\vec O_{\sss {CR}}}, does not move. Rigid body rotations do not comply with the axioms of an Euclidean space (section \ref{sec:algebra_structures}) but with those of Lie groups (section \ref{sec:algebra_lie}), and hence this chapter heavily relies on the concepts of Lie theory discussed in sections \ref{sec:algebra_lie} and \ref{sec:algebra_lie_jacobians}. Table \ref{tab:Rotate_lie_comparison} provides a comparison between the generic nomenclature employed in chapter \ref{cha:Algebra} and their rotation equivalents. The different representations discussed in this chapter are summarized in table \ref{tab:Rotate_summary}.
\begin{center}
\begin{tabular}{lcclcc}
	\hline
	Concept				& Lie Theory	  & Rotation & Concept & Lie Theory & Rotation \\
	\hline
	Lie group           & \nm{\mathcal G} & \nm{\mathbb{SO}\lrp{3}}		   & Lie group elements    & \nm{\mathcal X, \, \mathcal Y}       & \nm{\mathcal R, \, \mathcal S} \\ 
	Concatenation       & \nm{\circ}      & \nm{\circ}                 & Lie algebra           & \nm{\mathfrak{m}}     & \nm{\mathfrak{so}\lrp{3}} \\ 
	Identity            & \nm{\mathcal E} & \nm{\mathcal {I_R}}        & Inverse               & \nm{\mathcal X^{-1}}  & \nm{\mathcal R^{-1}} \\
	Velocity            & \nm{\vec v}     & \nm{\vec \omega}           & Tangent element       & \nm{\vec \tau}        & \nm{\vec r} \\
	Local frame         & \nm{\mathcal X} & B                          & Global frame          & \nm{\mathcal E}       & N \\
	Point action        & \nm{g_{\mathcal X}()} & \nm{\vec g_{\mathcal R}(\vec p)} & Vector action      & \nm{g_{\mathcal X}()} & \nm{\vec g_{\mathcal R*}(\vec v)} \\
	Adjoint             & \nm{\vec{Ad}_{\mathcal X}\lrp{\vec \tau^{\wedge}}} & \nm{\vec{Ad}_{\mathcal R}\lrp{\vec r^{\wedge}}} & Adjoint matrix & \nm{\vec{Ad}_{\mathcal X}\, \vec \tau} & \nm{\vec{Ad}_{\mathcal R} \, \vec r} \\
	\hline
\end{tabular}
\end{center}
\captionof{table}{Comparison between generic Lie elements and those of rigid body rotations} \label{tab:Rotate_lie_comparison}

This chapter begins with an introduction to rotational motion in section \ref{sec:RigidBody_bases}, followed by a description of the different rotation Lie group representations: the rotation matrix (section \ref{sec:RigidBody_rotation_dcm}), the rotation vector (section \ref{sec:RigidBody_rotation_rotv}), the unit quaternion (section \ref{sec:RigidBody_rotation_rodrigues}), the half rotation vector (section \ref{sec:RigidBody_rotation_halfrotv}), and the Euler angles (section \ref{sec:RigidBody_rotation_euler}). Algebraic operations on rigid body rotations, such as powers, linear interpolation, and the plus and minus operators, are introduced in section \ref{sec:RigidBody_rotation_algebra}. Section \ref{sec:RigidBody_rotation_calculus_derivatives} presents the rotation time derivative that leads to the definition of the angular velocity in the tangent space. The velocity of the rigid body points is discussed in section \ref{sec:RigidBody_rotation_velocity}, followed by the adjoint map in section \ref{sec:RigidBody_rotation_adjoint}, which transforms elements of the tangent space while the rotation advances on its manifold, and by an analysis of uncertainty and covariances applied to rotational motion (section \ref{sec:RigidBody_rotation_covariance}). An extensive analysis of the rotation Jacobians is presented in section \ref{sec:RigidBody_rotation_calculus_jacobians}. Sections \ref{sec:RigidBody_rotation_integration}, \ref{sec:RigidBody_rotation_gauss_newton}, and \ref{sec:RigidBody_rotation_SS} apply the discrete integration of Lie groups, the Gauss-Newton optimization of Lie group functions, and the state estimation of Lie groups contained in sections \ref{sec:algebra_integration}, \ref{sec:algebra_gradient_descent}, and \ref{sec:algebra_SS} to the case of rotations. Finally, the advantages and disadvantages of each rotation representation are discussed in section \ref{sec:RigidBody_rotation_applications}.

%%%%%%%%%%%%%%%%%%%%%%%%%%%%%%%%%%%%%%%%%%%%%%%%%%%%%%%%%%%%%%%%%%%%%%%%
%%%%%%%%%%%%%%%%%%%%%%%%%%%%%%%%%%%%%%%%%%%%%%%%%%%%%%%%%%%%%%%%%%%%%%%%
%%%%%%%%%%%%%%%%%%%%%%%%%%%%%%%%%%%%%%%%%%%%%%%%%%%%%%%%%%%%%%%%%%%%%%%%
% SECTION      SPECIAL ORTHOGONAL (LIE) GROUP
%%%%%%%%%%%%%%%%%%%%%%%%%%%%%%%%%%%%%%%%%%%%%%%%%%%%%%%%%%%%%%%%%%%%%%%%
%%%%%%%%%%%%%%%%%%%%%%%%%%%%%%%%%%%%%%%%%%%%%%%%%%%%%%%%%%%%%%%%%%%%%%%%
%%%%%%%%%%%%%%%%%%%%%%%%%%%%%%%%%%%%%%%%%%%%%%%%%%%%%%%%%%%%%%%%%%%%%%%%

\section{Special Orthogonal (Lie) Group}\label{sec:RigidBody_bases}

A rigid body can be represented with a Cartesian frame attached to any of its points (the origin), with the basis vectors \nm{\vec e_1}, \nm{\vec e_2}, and \nm{\vec e_3} being simply unit vectors along the main axes. It can be assumed with no loss of generality that the frame origin coincides with the center of rotation. Rigid body rotations can be combined and reversed, complying with the algebraic concept of group, but are not endowed with a metric, so they are not part of a metric or Euclidean space (section \ref{sec:algebra_structures}). They do however comply with the axioms of a Lie group (section \ref{sec:algebra_lie}), and hence the set of rigid body rotations together with the operation of rotation concatenation comprises \nm{\langle \mathbb{SO}\lrp{3}, \circ \rangle}, known as the \emph{rotation group} or \emph{special orthogonal group} of \nm{\mathbb{R}^3} \cite{Sola2017}, where its elements are denoted by \nm{\mathcal R}, the identify rotation by \nm{\mathcal {I_R}}, and the inverse by \nm{\mathcal R^{-1}}. The rotation group has two main actions, which are the rotation of points \nm{\lrb{\vec g() : \mathbb{SO}\lrp{3} \times \mathbb{R}^3 \rightarrow \mathbb{R}^3 \ | \ \vec p \rightarrow \vec g_{\mathcal R}\lrp{\vec p}}} and that of vectors \nm{\lrb{\vec g_*() : \mathbb{SO}\lrp{3} \times \mathbb{R}^3 \rightarrow \mathbb{R}^3 \ | \ \vec v \rightarrow \vec g_{\mathcal R*}\lrp{\vec v}}}.

Based on the rigid body definition above, its motion corresponds to an \emph{orthogonal transformation}, this is, one that preserves the norm (maintaining distances between points as well as angles between vectors) and the cross product (maintaining orientation)\footnote{An orthogonal transformation can also be defined as one that preserves both the inner and cross products.} \cite{Soatto2001}. These are called \emph{orthogonality} and \emph{handedness} \cite{Sola2017}:
\begin{itemize} 
\item Norm: \nm{\|\vec g_{\mathcal R*}\lrp{\vec v}\| = \|\vec v\|, \forall \, \vec v \in \mathbb{R}^3}
\item Cross product: \nm{\vec g_{\mathcal R*}\lrp{\vec u} \times \vec g_{\mathcal R*}\lrp{\vec v} = \vec g_{\mathcal R*}\lrp{\vec u \times \vec v}, \forall \, \vec u, \vec v \in \mathbb{R}^3}
\end{itemize}

Noting that a vector represents the difference between two points, \nm{\vec g_{\mathcal R*}\lrp{\vec v} = \vec g_{\mathcal R*}\lrp{\vec q - \vec p} = \vec g_{\mathcal R}\lrp{\vec q} - \vec g_{\mathcal R}\lrp{\vec p}}, and considering the possibility that one of the points may be the origin, \nm{\vec g_{\mathcal R}\lrp{\vec p} = \vec g_{\mathcal R}\lrp{\vec 0} = \vec 0}, results in the equivalence between the point and vector rotation maps:
\neweq{\vec g_{\mathcal R*}\lrp{\vec q - \vec 0} = \vec g_{\mathcal R*}\lrp{\vec q} = \vec g_{\mathcal R}\lrp{\vec q} - \vec g_{\mathcal R}\lrp{\vec 0} = \vec g_{\mathcal R}\lrp{\vec q} \rightarrow \vec g_{\mathcal R}() = \vec g_{\mathcal R*}()} {eq:SO3_equivalence}
\begin{center}
\begin{tabular}{lccc}
	\hline
	Representation			         		& Symbol		                                   & Structure    & Space \\
	\hline
	Rotation matrix                        	& \nm{\vec R}                                       & orthogonal 3x3 matrix & \nm{\mathbb{SO}\lrp{3}}                  \\
	Angular velocity						& \nm{\vec \omega^\wedge = \omegaskew}              & skew-symmetric matrix & \nm{\mathfrak{so}\lrp{3}} \\
	                                        & \nm{\vec \omega}                                  & free 3-vector         &                                      \\
	Rotation vector					        & \nm{\vec r^\wedge = \rskew}                       & skew-symmetric matrix & \nm{\mathbb{SO}\lrp{3}} \& \nm{\mathfrak{so}\lrp{3}} \\
	                                	    & \nm{\vec r = \vec \omega \, t = \vec n \, \phi}   & free 3-vector        	&                                      \\
	Unit quaternion                     	& \nm{\vec q}                                       & unit quaternion       & \nm{\mathbb{SO}\lrp{3}}                  \\
	Half angular velocity					& \nm{\vec \Omega^\wedge}                           & pure quaternion       & \nm{\mathfrak{so}\lrp{3}} \\
	                                       	& \nm{\vec \Omega = \vec \omega / 2}                & free 3-vector         &                                      \\
	Half rotation vector					& \nm{\vec h^\wedge}                                & pure quaternion       & \nm{\mathbb{SO}\lrp{3}} \& \nm{\mathfrak{so}\lrp{3}} \\
	                                       	& \nm{\vec h = \vec \Omega \, t = \vec n \, \theta = \vec r / 2} & free 3-vector           &                                      \\
	Euler angles                           	& \nm{\vec \phi}                                    & 3 angles              & \nm{\mathbb{SO}\lrp{3}}                  \\
	\hline
\end{tabular}
\end{center}
\captionof{table}{Summary of rotational motion representations} \label{tab:Rotate_summary}

The \nm{\mathbb{SO}\lrp{3}} analysis below adopts the convention introduced in section \ref{sec:algebra_lie}, in which all actions, including concatenation \nm{\lrb{\circ : \mathbb{SO}\lrp{3} \times \mathbb{SO}\lrp{3} \rightarrow \mathbb{SO}\lrp{3}}}, transform elements viewed in the local or body frame \nm{F_{\sss B} = \{\OB, \vec b_1, \vec b_2, \vec b_3\}} into elements viewed in the global or spatial frame \nm{F_{\sss N} = \{\ON, \vec n_1, \vec n_2, \vec n_3\} = \{\vec g_{\mathcal R}\lrp{\OB}, \vec g_{\mathcal R*}\lrp{\vec b_1}, \vec g_{\mathcal R*}\lrp{\vec b_2}, \vec g_{\mathcal R*}\lrp{\vec b_3}\}}\footnote{The spatial and local bases are denoted N and B as they usually correspond to the \hypertt{NED} and body frames, respectively. The North - East - Down or \hypertt{NED} frame is centered at the object center of mass, with axes aligned with the geodetic North, East, and Down directions. The body frame is also centered at the object center of mass, with \nm{\iBi} contained in its plane of symmetry pointing forward along a fixed direction, \nm{\iBiii} also located in the object plane of symmetry, normal to \nm{\iBi} and pointing downward, and \nm{\iBii} orthogonal to both in such a way that they form a right hand system.}:
\begin{eqnarray}
\nm{\pN} & = & \nm{\vec g_{\mathcal R_{NB}}\lrp{\pB}} \label{eq:Rotate_point_action} \\
\nm{\vN} & = & \nm{\vec g_{\mathcal R*_{NB}}\lrp{\vB}} \label{eq:Rotate_vector_action} 
\end{eqnarray}

%%%%%%%%%%%%%%%%%%%%%%%%%%%%%%%%%%%%%%%%%%%%%%%%%%%%%%%%%%%%%%%%%%%%%%%%
%%%%%%%%%%%%%%%%%%%%%%%%%%%%%%%%%%%%%%%%%%%%%%%%%%%%%%%%%%%%%%%%%%%%%%%%
%%%%%%%%%%%%%%%%%%%%%%%%%%%%%%%%%%%%%%%%%%%%%%%%%%%%%%%%%%%%%%%%%%%%%%%%
% SECTION      ROTATION MATRIX
%%%%%%%%%%%%%%%%%%%%%%%%%%%%%%%%%%%%%%%%%%%%%%%%%%%%%%%%%%%%%%%%%%%%%%%%
%%%%%%%%%%%%%%%%%%%%%%%%%%%%%%%%%%%%%%%%%%%%%%%%%%%%%%%%%%%%%%%%%%%%%%%%
%%%%%%%%%%%%%%%%%%%%%%%%%%%%%%%%%%%%%%%%%%%%%%%%%%%%%%%%%%%%%%%%%%%%%%%%

\section{Rotation Matrix}\label{sec:RigidBody_rotation_dcm}

The three basis vectors of the output frame can be stacked side by side into a matrix \nm{\vec R = \RNB = \lrsb{\vec n_1 \ \vec n_2 \ \vec n_3} \in \mathbb{R}^{3x3}}, called the \emph{rotation matrix}. Since its columns form a right handed orthonormal basis, it complies with the orthogonality and handedness conditions, and it can be proven that the rotation matrix \nm{\vec R} is an special orthogonal matrix\footnote{Orthogonal means that the transpose equals the inverse, while special or proper means that the determinant is positive one.}. Rotation matrices hence represent rigid body rotations, and their space \nm{\mathbb{SO}\lrp{3} = \{\vec R \in \mathbb{R}^{3x3} \ | \ {\vec R}^T \vec R = \vec I_3, \ det\lrp{\vec R} = +1\}} has group structure under matrix multiplication \nm{\{\mathbb{R}^{3x3} \times \mathbb{R}^{3x3} \rightarrow \mathbb{R}^{3x3} \ | \ \vec R_a \, \vec R_b \in \mathbb{R}^{3x3}, \forall \ \vec R_a, \ \vec R_b \in \mathbb{R}^{3x3}\}} \cite{Pinter1990}. While having dimension nine, the special orthogonal group \nm{\mathbb{SO}\lrp{3}} defined by means of rotation matrices constitutes a three dimensional manifold to Euclidean space \nm{\mathbb{E}^3}. Note that in this group the identity element is given by the identity matrix \nm{\lrp{\vec I = \vec I_3}}, and the inverse coincides with the transpose \nm{\lrp{\vec R^{-1} = \vec R^T}}.
\begin{center}
\begin{tabular}{lcclcc}
	\hline
	Concept			    & \nm{\mathbb{SO}\lrp{3}} & Rotation Matrix & Concept & \nm{\mathbb{SO}\lrp{3}} & Rotation Matrix \\
	\hline
	Lie group element   & \nm{\mathcal R} & \nm{\vec R}                    & Concatenation         & \nm{\circ}      & Matrix product \\
	Identity            & \nm{\mathcal {I_R}} & \nm{\vec I_3}            & Inverse               & \nm{\mathcal R^{-1}} & \nm{\vec R^T} \\
	Point rotation      & \nm{\vec g_{\mathcal R}(\vec p)} & \nm{\vec R \, \vec p}    & Vector rotation  & \nm{\vec g_{\mathcal R*}(\vec v)} & \nm{\vec R \, \vec v} \\
	\hline
\end{tabular}
\end{center}
\captionof{table}{Comparison between generic \nm{\mathbb{SO}\lrp{3}} and rotation matrix} \label{tab:Rotate_lie_dcm}

The rotation matrix \nm{\vec R} represents the actual coordinate transformation from the local to the global frame:
\neweq{\vec g_{\mathcal R*}\lrp{\vec v} = \vec R \; \vec v} {eq:SO3_dcm_transform} 

The inverse rotation (from the global to the local frame), is simply the transpose:
\neweq{\vec R^{-1} = \vec R^T} {eq:SO3_dcm_inverse}

The concatenation of rotations is also straight forward as it coincides with matrix multiplication. Note that \nm{\mathbb{SO}\lrp{3}} as defined above is not an abelian group, so the order of the factors is important. 
\neweq{\vec R_{\sss EB} = \vec R_{\sss EN} \, \vec R_{\sss NB}} {eq:SO3_dcm_concatenation}

%%%%%%%%%%%%%%%%%%%%%%%%%%%%%%%%%%%%%%%%%%%%%%%%%%%%%%%%%%%%%%%%%%%%%%%%
%%%%%%%%%%%%%%%%%%%%%%%%%%%%%%%%%%%%%%%%%%%%%%%%%%%%%%%%%%%%%%%%%%%%%%%%
%%%%%%%%%%%%%%%%%%%%%%%%%%%%%%%%%%%%%%%%%%%%%%%%%%%%%%%%%%%%%%%%%%%%%%%%
% SECTION      ROTATION VECTOR as TANGENT SPACE
%%%%%%%%%%%%%%%%%%%%%%%%%%%%%%%%%%%%%%%%%%%%%%%%%%%%%%%%%%%%%%%%%%%%%%%%
%%%%%%%%%%%%%%%%%%%%%%%%%%%%%%%%%%%%%%%%%%%%%%%%%%%%%%%%%%%%%%%%%%%%%%%%
%%%%%%%%%%%%%%%%%%%%%%%%%%%%%%%%%%%%%%%%%%%%%%%%%%%%%%%%%%%%%%%%%%%%%%%%

\section{Rotation Vector as Tangent Space}\label{sec:RigidBody_rotation_rotv}

As discussed in section \ref{subsec:algebra_lie_velocities}, the structure of the Lie algebra associated to \nm{\mathbb{SO}\lrp{3}} can be obtained by time derivating the Lie group inverse constraint, \nm{\vec R^T\lrp{t} \, \vec R\lrp{t} = \vec R\lrp{t} \, \vec R^T\lrp{t} =  \vec I_3}, resulting in the following particularizations of (\ref{eq:algebra_vE}) and (\ref{eq:algebra_vX}): 
\begin{eqnarray}
\nm{\vec \omega_{\sss {NB}}^{\sss N\wedge} = \wNBNskew} & = & \nm{\RNBdot \; \RNBtrans = - \RNB \; \RNBdottrans} \label{eq:SO3_dcm_omega_space} \\
\nm{\vec \omega_{\sss {NB}}^{\sss B\wedge} = \wNBBskew} & = & \nm{\RNBtrans \; \RNBdot = - \RNBdottrans \; \RNB} \label{eq:SO3_dcm_omega_body} 
\end{eqnarray}

The Lie algebra velocity \nm{\vec v^\wedge} of \nm{\mathbb{SO}\lrp{3}} is known as the \emph{angular velocity} \nm{\vec \omega^\wedge}, and as shown in (\ref{eq:SO3_dcm_omega_space}) and (\ref{eq:SO3_dcm_omega_body}), has the structure of a skew-symmetric matrix because its negative coincides with its transpose, so it is generally denoted as \nm{\widehat{\vec \omega}}. An alternative definition of the angular velocity is presented in section \ref{sec:RigidBody_rotation_calculus_derivatives}. Inverting the previous equations results in the rotation matrix time derivative, which is linear:
\neweq{\RNBdot = \wNBNskew \; \RNB = \RNB \; \wNBBskew} {eq:SO3_dcm_dot}

Notice that if \nm{\vec R\lrp{t_0} = \vec I_3}, then \nm{\vec{\dot R}\lrp{t_0} = \omegaskew \lrp{t_0}}, and hence the skew-symmetric matrix \nm{\omegaskew\lrp{t_0}} provides a first order approximation of the rotation matrix around the identity matrix \nm{\vec I_3}:
\neweq{\vec R\lrp{t_0 + \Deltat} \approx \vec I_3 + \omegaskew \lrp{t_0} \, \Deltat}{eq:SO3_rotv_taylor}

The \emph{space of skew-symmetric matrices} \nm{\mathfrak{so}\lrp{3} = \{\omegaskew \in \mathbb{R}^{3x3} \ | \ \vec \omega \in \mathbb{R}^3, \ - \omegaskew = \omegaskew^T\}} is hence the \emph{tangent space} of \nm{\mathbb{SO}\lrp{3}} at the identity \nm{\vec I_3} \cite{Soatto2001}, denoted as \nm{T_{\vec I_3}{\mathcal R}}. The \emph{hat} \nm{\lrb{\cdot^\wedge: \mathbb{R}^3 \rightarrow \mathfrak{so}\lrp{3} \ | \ \vec \omega \rightarrow \vec \omega^\wedge = \widehat{\vec \omega}}} and \emph{vee} \nm{\lrb{\cdot^\vee: \mathfrak{so}\lrp{3} \rightarrow \mathbb{R}^3 \ | \ \lrp{\widehat{\vec \omega}^\vee \rightarrow \vec \omega}}} operators convert the Cartesian vector form of the angular velocity into its skew-symmetric form, and vice versa.

If \nm{\vec R\lrp{t_0} \neq \vec I_3}, the tangent space needs to be transported right multiplying by \nm{\RNB\lrp{t_0}} (in the case of space angular velocity), or left multiplying for the local based velocity:
\begin{eqnarray}
\nm{\RNB\lrp{t_0 + \Deltat}} & \nm{\approx} & \nm{\RNB\lrp{t_0} + \lrsb{\wNBNskew \lrp{t_0} \, \Deltat} \, \RNB\lrp{t_0} = \lrsb{\vec I_3 + \wNBNskew \lrp{t_0} \, \Deltat} \, \RNB\lrp{t_0}    }\label{eq:SO3_rotv_taylor_space} \\
\nm{\RNB\lrp{t_0 + \Deltat}} & \nm{\approx} & \nm{\RNB\lrp{t_0} + \RNB\lrp{t_0} \, \lrsb{\wNBBskew \lrp{t_0} \, \Deltat} = \RNB\lrp{t_0} \, \lrsb{\vec I_3 + \wNBBskew \lrp{t_0} \, \Deltat}}\label{eq:SO3_rotv_taylor_body}
\end{eqnarray}

Note that the solution to the ordinary differential equation \nm{\vec{\dot x}\lrp{t} = \vec x\lrp{t} \, \omegaskew, \ \vec x\lrp{t} \in \mathbb{R}^3}, where \nm{\omegaskew} is constant, is \nm{\vec x\lrp{t} = \vec x\lrp{0} \, e^{\ds{\omegaskew t}}}. Based on it, assuming \nm{\vec R\lrp{0} = \vec I_3} as initial condition, and considering for the time being that \nm{\omegaskew} is constant,
\neweq{\vec R\lrp{t} = e^{\ds{\omegaskew t}} = \vec I_3 + \omegaskew t + \frac{\lrp{\omegaskew t}^2}{2!} + \dots + \frac{\lrp{\omegaskew t}^n}{n!} + \dots}{eq:SO3_rotv_exponential3}

which is indeed a rotation matrix as it complies with the \nm{\mathbb{SO}\lrp{3}} conditions of orthogonality and handedness \cite{Soatto2001}.
\begin{center}
\begin{tabular}{lcc}
	\hline
	Concept    & Lie Theory & \nm{\mathbb{SO}\lrp{3}} \\
	\hline
	Tangent space element & \nm{\vec \tau^\wedge} & \nm{\vec r^\wedge = \widehat{\vec r}} \\
	Velocity element      & \nm{\vec v^\wedge} & \nm{\vec \omega^\wedge = \widehat{\vec \omega}} \\
	Structure             & \nm{\wedge} & skew symmetric matrix \\
	\hline
\end{tabular}
\end{center}
\captionof{table}{Comparison between generic \nm{\mathbb{SO}\lrp{3}} and rotation vector as tangent space} \label{tab:Rotate_lie_rotv}

Remembering that so far \nm{\omegaskew} is constant, (\ref{eq:SO3_rotv_exponential3}) means that any rotation \nm{\vec R\lrp{t} = e^{\ds{\omegaskew t}}} can be realized by maintaining a constant angular velocity \nm{\vec \omega} for a given time \emph{t}. This is analogous to stating that any angular displacement \nm{\vec R\lrp{\phi} = e^{\ds{\nskew \phi}}} can be achieved by rotating an angle \nm{\phi} about a fixed unitary rotation axis \nm{\vec n}, which enables the definition of the \emph{rotation vector} \nm{\vec r}, also known as the \emph{exponential coordinates} of the \nm{\mathcal R} rotation, as
\neweq{\vec r = \vec \omega \, t = \vec n \, \phi \in \mathbb{R}^3}{eq:SO3_rotv_definition}

Note that the rotation vector \nm{\vec r} belongs to the tangent space as it is a multiple of the angular velocity \nm{\vec \omega \in \mathfrak{so}\lrp{3}}, and hence tends to coincide with it as time tends to zero. The \emph{exponential map} \nm{\lrb{exp\lrp{} : \mathfrak{so}\lrp{3} \rightarrow \mathbb{SO}\lrp{3} \ | \ \mathcal R = exp\lrp{\vec r^\wedge}}} and its capitalized form \nm{\lrb{Exp\lrp{} : \mathbb{R}^3 \rightarrow \mathbb{SO}\lrp{3} \ | \ \mathcal R = Exp\lrp{\vec r}}} wrap the rotation vector around the rotation group. In the case of the rotation matrix, the exponential map can be obtained from (\ref{eq:SO3_rotv_exponential3}) based on the fact that all skew-symmetric matrices verify that \nm{\rskew^2 = \vec r \, \vec r^T - \vec I_3} and \nm{\rskew^3 = - \rskew}, converting skew symmetric matrices into orthogonal ones:
\neweq{\vec R\lrp{\vec r} = exp\lrp{\rskew} = Exp\lrp{\vec r} = e^{\rskew} = \vec I_3 + \frac{\rskew}{\|\vec r\|} \sin \| \vec r\| + \frac{\rskew^2}{\|\vec r\|^2} \lrp{1 - \cos \|\vec r\|}}{eq:SO3_rotv_exponential2}

Geometrically, the skew symmetric matrix corresponds to an axis of rotation (via the mapping \nm{\vec n \rightarrow \nskew}) and the exponential map generates the rotation corresponding to rotating about that axis by an amount \nm{\phi} \cite{Murray1994}.

The angular velocity \nm{\vec \omega} however is in fact not required to be constant. Given a rotation matrix \nm{\vec R \in \mathbb{SO}\lrp{3}}, it can be proven that there exists a not necessarily unique vector \nm{\vec r \in \mathbb{R}^3} such that \nm{\vec R = e^{\rskew}}. The \emph{logarithmic map} \nm{\lrb{log\lrp{} : \mathbb{SO}\lrp{3} \rightarrow \mathfrak{so}\lrp{3} \ | \ \vec r^\wedge = log\lrp{\mathcal R}}} and its capitalized version \nm{\lrb{Log\lrp{} : \mathbb{SO}\lrp{3} \rightarrow \mathbb{R}^3 \ | \ \vec r = Log\lrp{\mathcal R}}} hence convert rigid body rotations into rotation vectors.
\neweq{\vec R = \RNB = \begin{bmatrix} \nm{R_{\sss11}} & \nm{R_{\sss12}} & \nm{R_{\sss13}} \\ \nm{R_{\sss21}} & \nm{R_{\sss22}} & \nm{R_{\sss23}} \\ \nm{R_{\sss31}} & \nm{R_{\sss32}} & \nm{R_{\sss33}} \end{bmatrix} \ \rightarrow \ \|\vec r \| = \arccos\lrp{\frac{trace\lrp{\vec R} - 1}{2}}, \ \vec r= \frac{\| \vec r\|}{2 \sin \| \vec r \|} \begin{bmatrix} \nm{R_{\sss32} - R_{\sss23}} \\ \nm{R_{\sss13} - R_{\sss31}} \\ \nm{R_{\sss21} - R_{\sss12}} \end{bmatrix}}{eq:SO3_rotv_logarithm}

Any rotation matrix can hence be realized by rotating a certain angle about a given axis, as indicated in (\ref{eq:SO3_rotv_definition}). The vector \nm{\vec n = \vec r / \|\vec r\|} indicates the rotation direction while \nm{\phi = \|\vec r\|} represents the turn angle. The exponential map described by (\ref{eq:SO3_rotv_exponential2}) is thus surjective (there is at least one rotation vector for every rotation matrix) but not injective, as a rotation of \nm{\lrp{\|\vec r\| + 2 \, k \, \pi} \forall \ k \in \mathbb{Z}} about \nm{\nm{\vec r / \|\vec r\|}} or a rotation of \nm{\lrp{- \|\vec r\| + 2 \, k \, \pi}} about \nm{-\vec r / \|\vec r\|} produce exactly the same rotation matrix.

Although inverting the rotation by means of the rotation vector is straightforward,
\neweq{\vec r_{\sss BN} = {\vec r_{\sss NB}}^{-1} = - \vec r_{\sss NB}}{eq:SO_rotv_inversion} 

the different \nm{\mathbb{SO}\lrp{3}} actions (concatenation, point rotation, vector rotation), as well as the relationship between the rotation vector derivative with time and the angular velocities, are complex and rarely used.

%%%%%%%%%%%%%%%%%%%%%%%%%%%%%%%%%%%%%%%%%%%%%%%%%%%%%%%%%%%%%%%%%%%%%%%%
%%%%%%%%%%%%%%%%%%%%%%%%%%%%%%%%%%%%%%%%%%%%%%%%%%%%%%%%%%%%%%%%%%%%%%%%
%%%%%%%%%%%%%%%%%%%%%%%%%%%%%%%%%%%%%%%%%%%%%%%%%%%%%%%%%%%%%%%%%%%%%%%%
% SECTION      UNIT QUATERNION
%%%%%%%%%%%%%%%%%%%%%%%%%%%%%%%%%%%%%%%%%%%%%%%%%%%%%%%%%%%%%%%%%%%%%%%%
%%%%%%%%%%%%%%%%%%%%%%%%%%%%%%%%%%%%%%%%%%%%%%%%%%%%%%%%%%%%%%%%%%%%%%%%
%%%%%%%%%%%%%%%%%%%%%%%%%%%%%%%%%%%%%%%%%%%%%%%%%%%%%%%%%%%%%%%%%%%%%%%%

\section{Unit Quaternion}\label{sec:RigidBody_rotation_rodrigues}

The quaternions with unity norm, known as unit quaternions, comprise an additional representation of the rotation group \nm{\mathbb{SO}\lrp{3}}, as shown below. Quaternions in turn are generalizations of complex numbers in the same way that these are generalizations of real ones \cite{Soatto2001}. It is hence necessary to first describe the complex numbers in section \ref{subsec:RigidBody_rotation_rodrigues_complex} and the quaternions in section \ref{subsec:RigidBody_rotation_rodrigues_quat} before focusing on the unit quaternions in section \ref{subsec:RigidBody_rotation_rodrigues_unit_quat}.

%%%%%%%%%%%%%%%%%%%%%%%%%%%%%%%%%%%%%%%%%%%%%%%%%%%%%%%%%%%%%%%%%%%%%%%%
%%%%%%%%%%%%%%%%%%%%%%%%%%%%%%%%%%%%%%%%%%%%%%%%%%%%%%%%%%%%%%%%%%%%%%%%
% SUBSECTION      COMPLEX NUMBERS
%%%%%%%%%%%%%%%%%%%%%%%%%%%%%%%%%%%%%%%%%%%%%%%%%%%%%%%%%%%%%%%%%%%%%%%%
%%%%%%%%%%%%%%%%%%%%%%%%%%%%%%%%%%%%%%%%%%%%%%%%%%%%%%%%%%%%%%%%%%%%%%%%

\subsection{Complex Numbers}\label{subsec:RigidBody_rotation_rodrigues_complex}

The set of \emph{complex numbers} \nm{\mathbb{C}} is composed of two real numbers \nm{\lrb{\mathbb{C} = \mathbb{R} + \mathbb{R} \, i \ | \ i^2 = i \cdot i = - 1}}. Given two complex numbers \nm{c_1 = x_1 + y_1 \, i \in \mathbb{C}, c_2 = x_2 + y_2 \, i \in \mathbb{C}, \forall \ x_1, y_1, x_2, y_2 \in \mathbb{R}}, it is possible to define the operations of addition \nm{\lrb{+ : \mathbb{C} \times  \mathbb{C} \rightarrow \mathbb{C}}} and  multiplication \nm{\lrb{\cdot : \mathbb{C} \times  \mathbb{C} \rightarrow \mathbb{C}}}. 
\begin{eqnarray}
\nm{c_1 + c_2} & = & \nm{\lrp{x_1 + y_1 \, i} + \lrp{x_2 + y_2 \, i} = \lrp{x_1 + x_2} + \lrp{y_1 + y_2} \, i}\label{eq:SO3_complex_addition} \\
\nm{c_1 \cdot c_2} & = & \nm{c_1 \, c_2 = \lrp{x_1 + y_1 \, i} \cdot \lrp{x_2 + y_2 \, i} = \lrp{x_1 x_2 - y_1 y_2} + \lrp{x_1 y_2 + y_1 x_2} \, i}\label{eq:SO3_complex_multiplication}
\end{eqnarray}

The conjugate is defined as \nm{c^{\ast} = x - y \, i \in \mathbb{C}} and verifies that \nm{\lrp{c_1 \cdot c_2}^{\ast} = c_1^{\ast} \cdot c_2^{\ast}}, while the norm \nm{\|c\| = \sqrt{c \cdot c^{\ast}} = \sqrt{c^{\ast} \cdot c} = \sqrt{x^2 + y^2} \in \mathbb{R}} satisfies that \nm{\|c_1 \cdot c_2\| = \|c_1\| \cdot \|c_2\|}. The set of complex numbers \nm{\mathbb{C}} endowed with the operations of addition \nm{+} and multiplication \nm{\cdot} forms a field (not ordered), known as the field of complex numbers \nm{\langle \mathbb{C}, +, \cdot \rangle}, nearly always abbreviated to simply \nm{\mathbb{C}}. The additive identity is \nm{0 = 0 + 0 \, i} and the inverse \nm{- c = - x - y \, i}, while the multiplication identity is \nm{1 = 1 + 0 \, i} and the inverse \nm{c^{-1} = c^{\ast} / \| c \|^2}. 

Complex numbers can always be written in polar form \nm{\lrp{c = r \lrp{\cos \phi + \sin \phi \, i} = r \, e^{\ds{i \, \phi}}}}, and as such are valid representations of the circle group or plane rotations group \nm{\mathbb{SO}\lrp{2}}, similarly to the case of rotation vectors in \nm{\mathbb{SO}\lrp{3}} described in section \ref{sec:RigidBody_rotation_rotv}.

%%%%%%%%%%%%%%%%%%%%%%%%%%%%%%%%%%%%%%%%%%%%%%%%%%%%%%%%%%%%%%%%%%%%%%%%
%%%%%%%%%%%%%%%%%%%%%%%%%%%%%%%%%%%%%%%%%%%%%%%%%%%%%%%%%%%%%%%%%%%%%%%%
% SUBSECTION      QUATERNIONS
%%%%%%%%%%%%%%%%%%%%%%%%%%%%%%%%%%%%%%%%%%%%%%%%%%%%%%%%%%%%%%%%%%%%%%%%
%%%%%%%%%%%%%%%%%%%%%%%%%%%%%%%%%%%%%%%%%%%%%%%%%%%%%%%%%%%%%%%%%%%%%%%%

\subsection{Quaternions}\label{subsec:RigidBody_rotation_rodrigues_quat}

The set of \emph{quaternions} \nm{\mathbb{H}} is defined as \nm{\{\mathbb{H} = \mathbb{C} + \mathbb{C} \, j \ | \ j^2 = -1, \ i \cdot j = - j \cdot i\}}. A quaternion \nm{\vec q \in \mathbb{H}} has the form \nm{\vec q = q_0 + q_1 \, i + q_2 \, j + q_3 \, i \, j}, with \nm{q_i \in \mathbb{R}}. \emph{Pure quaternions} \nm{\vec q = q_1 \, i + q_2 \, j + q_3 \, i \, j \in \mathbb{H}_p} are those defined in the tridimensional imaginary subspace of \nm{\mathbb{H}}, and verify that \nm{\vec q = - \, \qast}. 

There are many different conventions for the quaternion found in the literature \cite{Sola2017}. This document adopts the \emph{Hamilton convention}, characterized by locating the real part first (instead of last), being right handed (left), passive (rotates frames and not vectors as in active), and local to global rotations (global to local). Any variation to these choices would result in different expressions below, although the physical concepts do not vary.

The real plus imaginary notation \nm{\{1, i, j, i \, j\}}  is not always the most convenient. A quaternion can also be expressed as the sum of a scalar plus a vector in the form \nm{\vec q = q_0 + \vec q_v}, where \nm{q_0} is the real or scalar part and \nm{\vec q_v = q_1 \, i + q_2 \, j + q_3 \, i \, j} is the imaginary or vector part. Quaternions are however mostly represented as 4-vectors \nm{\vec q = \lrsb{q_0 \ \ \vec q_v}^T = \lrsb{q_0 \ \ q_1 \ \ q_2 \ \ q_3}^T}, which enables the usage of matrix algebra for quaternion operations. It is also convenient to abuse the equal operator by combining general, real, and pure quaternions as in \nm{\vec q = q_0 + \vec q_v}, where \nm{q_0 = \lrsb{q_0 \ \ \vec 0_v }^T} and \nm{\vec q_v = \lrsb{0 \ \ \vec q_v }^T}.

The following expressions define the addition \nm{\lrb{+ : \mathbb{H} \times \mathbb{H} \rightarrow \mathbb{H}}} and inner product \nm{\lrb{\langle \cdot \, , \cdot \rangle: \mathbb{H} \times \mathbb{H} \rightarrow \mathbb{R}}} of two quaternions, which commute, as well as the scalar multiplication \nm{\lrb{\cdot : \mathbb{R} \times \mathbb{H} \rightarrow \mathbb{H}}}:
\begin{eqnarray}
\nm{\vec q + \vec p} & = & \nm{\lrsb{q_0 \ \ \vec q_v}^T + \lrsb{p_0 \ \ \vec p_v}^T = \lrsb{q_0 + p_0 \ \ \ q_1 + p_1 \ \ \ q_2 + p_2 \ \ \ q_3 + p_3}^T = \lrsb{q_0 + p_0 \ \ \ \vec q_v + \vec p_v}^T}\label{eq:SO3_quat_addition} \\
\nm{\langle \vec q , \vec p \rangle} & = & \nm{\vec q \cdot \vec p = {\vec q}^T \, \vec p = q_{\sss 0} \, p_{\sss 0} + q_{\sss 1} \, p_{\sss 1} + q_{\sss 2} \, p_{\sss 2} +  q_{\sss 3} \, p_{\sss 3}} \label{eq:SO3_quat_inner_product} \\
\nm{a \cdot \vec q} & = & \nm{a \cdot \lrsb{q_0 \ \ \vec q_v}^T = \lrsb{a \, q_0 \ \ \ a \cdot \vec q_v}^T} \label{eq:SO3_quat_scalar_product}
\end{eqnarray}

The multiplication of quaternions \nm{\{\otimes : \mathbb{H} \times \mathbb{H} \rightarrow \mathbb{H}\}} is not commutative as it includes the cross product:
\neweq{\vec q \otimes \vec p = \begin{bmatrix}
		\nm{q_0 \cdot p_0 - q_1 \cdot p_1 - q_2 \cdot p_2 - q_3 \cdot p_3} \\
		\nm{q_1 \cdot p_0 + q_0 \cdot p_1 - q_3 \cdot p_2 + q_2 \cdot p_3} \\
		\nm{q_2 \cdot p_0 + q_3 \cdot p_1 + q_0 \cdot p_2 - q_1 \cdot p_3} \\
		\nm{q_3 \cdot p_0 - q_2 \cdot p_1 + q_1 \cdot p_2 + q_0 \cdot p_3} \end{bmatrix} =
	     \begin{bmatrix}
		 \nm{q_0 \, p_0 - {\vec q_v}^T \, \vec p_v} \\
		 \nm{q_0 \, \vec p_v + p_0 \, \vec q_v + \widehat{\vec q}_v \, \vec p_v} \end{bmatrix} } {eq:SO3_quat_product}

It is also bilinear \cite{Sola2017}:
\neweq{\vec q \otimes \vec p = [\vec q]_L \, \vec p = \begin{bmatrix}
		\nm{+ q_0} & \nm{- q_1} & \nm{- q_2} & \nm{- q_3} \\
		\nm{+ q_1} & \nm{+ q_0} & \nm{- q_3} & \nm{+ q_2} \\
		\nm{+ q_2} & \nm{+ q_3} & \nm{+ q_0} & \nm{- q_1} \\
		\nm{+ q_3} & \nm{- q_2} & \nm{+ q_1} & \nm{+ q_0} \end{bmatrix} \,
		\begin{bmatrix} \nm{p_0} \\ \nm{p_1} \\ \nm{p_2} \\ \nm{p_3} \end{bmatrix} = 
       [\vec p]_R \, \vec q = \begin{bmatrix}
		\nm{+ p_0} & \nm{- p_1} & \nm{- p_2} & \nm{- p_3} \\
		\nm{+ p_1} & \nm{+ p_0} & \nm{+ p_3} & \nm{- p_2} \\
		\nm{+ p_2} & \nm{- p_3} & \nm{+ p_0} & \nm{+ p_1} \\
		\nm{+ p_3} & \nm{+ p_2} & \nm{- p_1} & \nm{+ p_0} \end{bmatrix} \,
		\begin{bmatrix} \nm{q_0} \\ \nm{q_1} \\ \nm{q_2} \\ \nm{q_3} \end{bmatrix}} {eq:SO3_quat_product_matrices}

These expressions can be simplified for the common case of the multiplication of a quaternion \nm{\vec q} by a pure quaternion \nm{\vec p_v}, this is, \nm{\{\otimes : \mathbb{H} \times \mathbb{H}_p \rightarrow \mathbb{H}\}}:
\begin{eqnarray}
\nm{\vec q \otimes \vec p_v} & = & \nm{\begin{bmatrix}
		\nm{- q_1 \cdot p_1 - q_2 \cdot p_2 - q_3 \cdot p_3} \\
		\nm{+ q_0 \cdot p_1 - q_3 \cdot p_2 + q_2 \cdot p_3} \\
		\nm{+ q_3 \cdot p_1 + q_0 \cdot p_2 - q_1 \cdot p_3} \\
		\nm{- q_2 \cdot p_1 + q_1 \cdot p_2 + q_0 \cdot p_3} \end{bmatrix} =
	     \begin{bmatrix}
		 \nm{- {\vec q_v}^T \, \vec p_v} \\
		 \nm{q_0 \, \vec p_v + \widehat{\vec q}_v \, \vec p_v} \end{bmatrix} } \label{eq:SO3_quat_product_pure} \\
\nm{\vec q \otimes \vec p_v} & = & \nm{[\vec q]_{L3} \, \vec p_v = \begin{bmatrix}
		\nm{- q_1} & \nm{- q_2} & \nm{- q_3} \\
		\nm{+ q_0} & \nm{- q_3} & \nm{+ q_2} \\
		\nm{+ q_3} & \nm{+ q_0} & \nm{- q_1} \\
		\nm{- q_2} & \nm{+ q_1} & \nm{+ q_0} \end{bmatrix} \,
		\begin{bmatrix} \nm{p_1} \\ \nm{p_2} \\ \nm{p_3} \end{bmatrix} = 
       [\vec p_v]_{R3} \, \vec q = \begin{bmatrix}
		\nm{0}           & \nm{- p_1} & \nm{- p_2} & \nm{- p_3} \\
		\nm{+ p_1} & \nm{0}           & \nm{+ p_3} & \nm{- p_2} \\
		\nm{+ p_2} & \nm{- p_3} & \nm{0}           & \nm{+ p_1} \\
		\nm{+ p_3} & \nm{+ p_2} & \nm{- p_1} & \nm{0} \end{bmatrix} \,
		\begin{bmatrix} \nm{q_0} \\ \nm{q_1} \\ \nm{q_2} \\ \nm{q_3} \end{bmatrix}} \label{eq:SO3_quat_product_pure_matrices}
\end{eqnarray}

The conjugate quaternion is defined as \nm{\qast = q_0 - \vec q_v \in \mathbb{H}} and verifies that \nm{\lrp{\vec q \otimes \vec p}^{\ast} = \vec p^{\ast} \otimes \qast}, while the quaternion norm \nm{\|\vec q\| = \sqrt{\langle \vec q \ , \ \qast\rangle} = \sqrt{\langle \qast \ , \ \vec q\rangle} = \sqrt{\vec q \otimes \qast}  = \sqrt{\qast \otimes \vec q} \in \mathbb{R}} satisfies that \nm{\|\vec q \otimes \vec p\| = \|\vec p \otimes \vec q\| = \|\vec q\| \, \|\vec p\|}. Quaternions endowed with \nm{\otimes} form the non-commutative group \nm{\langle \mathbb{H}, \otimes \rangle}, where \nm{\vec{q_1} = 1 + \vec0_v} is the identity and \nm{\vec q^{-1} = \nicefrac{\qast}{\| \vec q \|^2}} the inverse \cite{Sola2017}. Additionally, quaternions endowed with addition \nm{+} and multiplication \nm{\otimes} form the ring \nm{\langle \mathbb{H}, +, \otimes \rangle} where \nm{\vec{q_0} = 0 + \vec0_v} is the addition identity and \nm{- \vec q} the addition inverse or negative.

The quaternion rotation operator or \emph{double product} is defined as \nm{\{\mathbb{H} \times \mathbb{R}^3 \rightarrow \mathbb{R}^3 \ | \ \vec q \in \mathbb{H},} \nm{\vec v \in \mathbb{R}^3 \rightarrow \vec q \otimes \vec v \otimes \qast \in \mathbb{R}^3\}}\footnote{It is easily proven that the double quaternion product results in \nm{\mathbb{R}^3} and not \nm{\mathbb{R}^4}.}, while the natural power of a quaternion \nm{\vec q^n, n \in \mathbb{N}} is obtained by multiplying the quaternion by itself \nm{n-1} times.

%%%%%%%%%%%%%%%%%%%%%%%%%%%%%%%%%%%%%%%%%%%%%%%%%%%%%%%%%%%%%%%%%%%%%%%%
%%%%%%%%%%%%%%%%%%%%%%%%%%%%%%%%%%%%%%%%%%%%%%%%%%%%%%%%%%%%%%%%%%%%%%%%
% SUBSECTION      UNIT QUATERNIONS
%%%%%%%%%%%%%%%%%%%%%%%%%%%%%%%%%%%%%%%%%%%%%%%%%%%%%%%%%%%%%%%%%%%%%%%%
%%%%%%%%%%%%%%%%%%%%%%%%%%%%%%%%%%%%%%%%%%%%%%%%%%%%%%%%%%%%%%%%%%%%%%%%

\subsection{Unit Quaternion}\label{subsec:RigidBody_rotation_rodrigues_unit_quat}

\emph{Unit quaternions} verify that \nm{\|\vec q\| = 1}, which implies that \nm{\vec q^{-1} = \qast}. They can always be written as 
\neweq{\vec q = \cos \theta + \vec n \,  \sin \theta}{eq:SO3_quat_unit}

where \nm{\vec n} is a unit vector and \nm{\theta} is a scalar. The exponential of a quaternion \nm{e^{\vec q}} is defined analogously to that of real numbers \cite{Sola2017}. For pure quaternions \nm{\vec q = \vec q_v}, if abusing notation with \nm{\vec q_v = \vec v = \vec n \; \theta} where \nm{\theta = \|\vec v\|} and \nm{\|\vec n \| = 1}, it can be proven that \nm{e^{\vec q_v} = \cos \theta + \vec n \, \sin \theta}, which can be considered an extension of the \nm{e^{\ds{i \theta}} = \cos \theta + i \, \sin \theta} expression for complex numbers introduced in section \ref{subsec:RigidBody_rotation_rodrigues_complex} \cite{Sola2017}. Notice that since \nm{\|e^{\vec q_v}\| = 1}, the exponential of a pure quaternion is a unit quaternion. If \nm{\|\vec q\| = 1}, it is easy to verify that \nm{log\lrp{\vec q} = log\lrp{e^{\vec n \; \theta}} = \vec q_v}, so the logarithm of a unit quaternion is a pure quaternion \cite{Sola2017}.

Unit quaternions endowed with \nm{\otimes} constitute a subgroup that represents a 3-sphere, this is, the three dimensional surface of the unit sphere of \nm{\mathbb{R}^4}, and is commonly noted as \nm{\mathbb{S}^3}. They comply with the orthogonality and handedness conditions required in section \ref{sec:RigidBody_bases} for rigid body rotations, and hence their space \nm{\mathbb{SO}\lrp{3} = \{\vec q \in \mathbb{S}^3 \ | \ \qast \otimes \vec q = \vec q \otimes \qast = \vec{q_1}\}} possesses group structure under quaternion multiplication \nm{\{\otimes : \mathbb{S}^3 \times \mathbb{S}^3 \rightarrow \mathbb{S}^3 \ | \ \vec q_a \otimes \vec q_b \in \mathbb{S}^3, \forall \ \vec q_a, \ \vec q_b \in \mathbb{S}^3\}} \cite{Pinter1990}. While having dimension four, the special orthogonal group \nm{\mathbb{SO}\lrp{3}} defined by means of unit quaternions constitutes a three dimensional manifold to Euclidean space \nm{\mathbb{E}^3}. Note that in this group \nm{\vec{q_1}} constitutes the identity and \nm{\qast} the inverse.
\begin{center}
\begin{tabular}{lcclcc}
	\hline
	Concept    & \nm{\mathbb{SO}\lrp{3}} & \nm{\mathbb{S}^3} & Concept    & \nm{\mathbb{SO}\lrp{3}} & \nm{\mathbb{S}^3} \\
	\hline
	Lie group element   & \nm{\mathcal R} & \nm{\vec q}                    & Concatenation         & \nm{\circ}      & \nm{\otimes} \\
	Identity            & \nm{\mathcal {I_R}} & \nm{\vec{q_1}}            & Inverse               & \nm{\mathcal R^{-1}} & \nm{\qast} \\
	Point rotation      & \nm{\vec g_{\mathcal R}(\vec p)} & \nm{\vec q \otimes \vec p \otimes \qast}    & Vector rotation  & \nm{\vec g_{\mathcal R*}(\vec v)} & \nm{\vec q \otimes \vec v \otimes \qast} \\
	\hline
\end{tabular}
\end{center}
\captionof{table}{Comparison between generic \nm{\mathbb{SO}\lrp{3}} and unit quaternion} \label{tab:Rotate_lie_unit_quat}

Coordinate transformation (point or vector rotation) and rotation concatenation are both linear:
\begin{eqnarray}
\nm{\vec g_{\mathcal R*}\lrp{\vec v}} & = & \nm{\vec q \otimes \vec v \otimes \qast} \label{eq::SO3_quat_transform} \\ 
\nm{\vec q_{\sss EB}} & = & \nm{\vec q_{\sss EN} \otimes \vec q_{\sss NB}} \label{eq:SO3_quat_concatenation} 
\end{eqnarray}

%%%%%%%%%%%%%%%%%%%%%%%%%%%%%%%%%%%%%%%%%%%%%%%%%%%%%%%%%%%%%%%%%%%%%%%%
%%%%%%%%%%%%%%%%%%%%%%%%%%%%%%%%%%%%%%%%%%%%%%%%%%%%%%%%%%%%%%%%%%%%%%%%
%%%%%%%%%%%%%%%%%%%%%%%%%%%%%%%%%%%%%%%%%%%%%%%%%%%%%%%%%%%%%%%%%%%%%%%%
% SECTION      HALF ROTATION VECTOR as TANGENT SPACE
%%%%%%%%%%%%%%%%%%%%%%%%%%%%%%%%%%%%%%%%%%%%%%%%%%%%%%%%%%%%%%%%%%%%%%%%
%%%%%%%%%%%%%%%%%%%%%%%%%%%%%%%%%%%%%%%%%%%%%%%%%%%%%%%%%%%%%%%%%%%%%%%%
%%%%%%%%%%%%%%%%%%%%%%%%%%%%%%%%%%%%%%%%%%%%%%%%%%%%%%%%%%%%%%%%%%%%%%%%

\section{Half Rotation Vector as Tangent Space}\label{sec:RigidBody_rotation_halfrotv}

It is interesting to point out that in the case of rotation matrices (section \ref{sec:RigidBody_rotation_dcm}), the orthogonality (\nm{\vec R^T \, \vec R = \vec I_3}) and handedness \nm{\lrp{det\lrp{\vec R} = +1}} constraints constitute two different expressions, while in the case of quaternions both are contained within \nm{\lrp{\qast \otimes \vec q = \vec q \otimes \qast = \vec{q_1}}}. As in other Lie groups, the time derivation of this constraint results in the structure of the Lie algebra. Derivating leads to \nm{\qast \otimes \vec{\dot q} = - \lrp{\qast \otimes \vec{\dot q}}^{\ast}}, which indicates that \nm{\qast \otimes \vec{\dot q}} is in fact a pure quaternion, as is \nm{\vec{\dot q} \otimes \qast}. This results in the following particularizations of (\ref{eq:algebra_vE}) and (\ref{eq:algebra_vX}): 
\begin{eqnarray}
\nm{\vec \Omega_{\sss {NB}}^{\sss N\wedge}} & = & \nm{\vec{\dot q}_{\sss {NB}} \otimes \qNB^{\ast} = - \qNB \otimes \vec{\dot q}_{\sss NB}^{\ast}} \label{eq:SO3_quat_Omega_space} \\
\nm{\vec \Omega_{\sss {NB}}^{\sss B\wedge}} & = & \nm{\qNB^{\ast} \otimes \qNBdot = - \vec{\dot q}_{\sss NB}^{\ast} \otimes \qNB} \label{eq:SO3_quat_Omega_body} 
\end{eqnarray}

The Lie algebra velocity \nm{\vec v^\wedge} of \nm{\mathbb{S}^3} is known as the \emph{half angular velocity} \nm{\vec \Omega^\wedge} \cite{Sola2017}, and as shown in (\ref{eq:SO3_quat_Omega_space}) and (\ref{eq:SO3_quat_Omega_body}), has the structure of a pure quaternion because its negative coincides with its conjugate:
\neweq{\vec \Omega^\wedge\lrp{t} = \lrsb{0, \vec \Omega\lrp{t}}^T \in \mathbb{H}_p}{eq:SO3_quat_pure}

Inverting the previous equations results in the unit quaternion time derivative, which is linear:
\neweq{\qNBdot = \vec \Omega_{\sss {NB}}^{\sss N\wedge} \otimes \qNB = \qNB \otimes \vec \Omega_{\sss {NB}}^{\sss B\wedge}} {eq::SO3_quat_Omega_dot}

Notice that if \nm{\vec q\lrp{t_0} = \vec{q_1}}, then \nm{\vec{\dot q }\lrp{t_0} = \vec \Omega\lrp{t_0}}, and hence the pure quaternion \nm{\vec \Omega^\wedge\lrp{t_0}} provides a first order approximation of the unit quaternion around the identity \nm{\vec{q_1}}:
\neweq{\vec q\lrp{t_0 + \Deltat} \approx \vec q_1 +\vec \Omega^\wedge\lrp{t_0} \, \Deltat}{eq:SO3_quat_taylor}

The \emph{space of pure quaternions} \nm{\mathfrak{so}\lrp{3} = \{\vec \Omega^\wedge \in \mathbb{H}_p \ | \ \vec \Omega \in \mathbb{R}^3\}} is hence the \emph{tangent space} of the unit sphere \nm{\mathbb{S}^3} of quaternions at the identity \nm{\vec q_1}, denoted as \nm{T_{\vec q_1}{\mathcal R}}. The \emph{hat} \nm{\lrb{\cdot^\wedge: \mathbb{R}^3 \rightarrow \mathfrak{so}\lrp{3} \ | \ \vec \Omega \rightarrow \vec \Omega^\wedge}} and \emph{vee} \nm{\lrb{\cdot^\vee: \mathfrak{so}\lrp{3} \rightarrow \mathbb{R}^3 \ | \ \lrp{\vec \Omega^\wedge}^\vee \rightarrow \vec \Omega}} operators convert the half angular velocity vector into its pure quaternion form, and vice versa.

If \nm{\vec q\lrp{t_0} \neq \vec q_1}, the tangent space needs to be transported right multiplying by \nm{\qNB\lrp{t_0}} (in the case of space tangent space), or left multiplying for the local space:
\begin{eqnarray}
\nm{\qNB\lrp{t_0 + \Deltat}} & \nm{\approx} & \nm{\qNB\lrp{t_0} + \lrsb{\vec \Omega_{\sss NB}^{\sss N\wedge} \lrp{t_0} \, \Deltat} \otimes \qNB\lrp{t_0} = \lrsb{\vec q_1 + \vec \Omega_{\sss NB}^{\sss N\wedge} \lrp{t_0} \, \Deltat} \otimes \qNB\lrp{t_0}    }\label{eq:SO3_quat_taylor_space} \\
\nm{\qNB\lrp{t_0 + \Deltat}} & \nm{\approx} & \nm{\qNB\lrp{t_0} + \qNB\lrp{t_0} \otimes \lrsb{\vec \Omega_{\sss NB}^{\sss B\wedge} \lrp{t_0} \, \Deltat} = \qNB\lrp{t_0} \otimes \lrsb{\vec q_1 + \vec \Omega_{\sss NB}^{\sss B\wedge} \lrp{t_0} \, \Deltat}}\label{eq:SO3_quat_taylor_body}
\end{eqnarray}

Note that the solution to the ordinary differential equation \nm{\vec{\dot x}\lrp{t} = \vec x\lrp{t} \otimes \vec \Omega^\wedge, \ \vec x\lrp{t} \in \mathbb{R}^4}, where \nm{\vec \Omega^\wedge} is constant, is \nm{\vec x\lrp{t} = \vec x\lrp{0} \, e^{\ds{\vec \Omega^\wedge t}}}. Based on it, assuming \nm{\vec q\lrp{0} = \vec{q_1}} as initial condition, and considering for the time being that \nm{\vec \Omega} is constant,
\neweq{\vec q\lrp{t} = e^{\ds{\vec \Omega^\wedge t}} = \vec{q_1} + \vec \Omega^\wedge t + \frac{\lrp{\vec \Omega^\wedge t}^2}{2!} + \dots + \frac{\lrp{\vec \Omega^\wedge t}^n}{n!} + \dots}{eq:SO3_quat_exponential3}

which is indeed a unit quaternion \cite{Sola2017}.
\begin{center}
\begin{tabular}{lcc}
	\hline
	Concept    & Lie Theory & \nm{\mathbb{SO}\lrp{3}} \\
	\hline
	Tangent space element & \nm{\vec \tau^\wedge} & \nm{\vec h^\wedge = \lrsb{0 \ \ \vec h}^T} \\
	Velocity element      & \nm{\vec v^\wedge} & \nm{\vec \Omega^\wedge = \lrsb{0 \ \ \vec \Omega}^T} \\
	Structure             & \nm{\wedge} & pure quaternion \\
	\hline
\end{tabular}
\end{center}
\captionof{table}{Comparison between generic \nm{\mathbb{SO}\lrp{3}} and half rotation vector as tangent space} \label{tab:Rotate_lie_halfrotv}

Remembering that so far \nm{\vec \Omega^\wedge} is constant, (\ref{eq:SO3_quat_exponential3}) means that any rotation \nm{\vec q\lrp{t} = e^{\ds{\vec \Omega^\wedge t}}} can be realized by maintaining a constant half angular velocity \nm{\vec \Omega^\wedge} in \nm{\mathbb{H}_p} for a given time \emph{t}. This is analogous to stating that any angular displacement \nm{\vec q\lrp{\theta} = e^{\ds{\vec n^\wedge \theta}}} can be achieved by rotating an angle \nm{\theta} about a fixed unitary rotation axis \nm{\vec n^\wedge \in \mathbb{H}_p}. It is customary to absorb \emph{t} into \nm{\vec \Omega} or \nm{\theta} into \nm{\vec n}, resulting in \emph{the half rotation vector} \nm{\vec h}: 
\neweq{\vec h = \vec \Omega \, t = \vec n \, \theta \ \ \in \mathbb{R}^3}{eq:SO3_quat_half_velocity}

Expression (\ref{eq:SO3_quat_exponential3}) represents the \emph{exponential map} \nm{\{exp : \mathfrak{so}\lrp{3} \rightarrow \mathbb{SO}\lrp{3} | \ \vec h^\wedge \in \mathbb{H}_p \rightarrow exp\lrp{\vec h^\wedge} \in \mathbb{S}^3\}} \cite{Sola2017}, which transforms pure quaternions into unit quaternions. Before continuing, it is possible to compare the similarities and differences between the exponential maps when applied to rotation matrices versus quaternions, as both represent maps between the tangent space \nm{\mathfrak{so}\lrp{3}} and the special orthogonal group \nm{\mathbb{SO}\lrp{3}}. In the case of rotation matrices, this translates to a map between skew-symmetric matrices and orthogonal ones, while for quaternions the exponential map converts pure quaternions into unitary ones. There is one additional difference, however. In the case of rotation matrices, the map encodes through the rotation vector \nm{\vec r = \vec \omega \, t = \vec n \, \phi}, this is, the axis of rotation \nm{\vec n} and the rotated angle \nm{\phi}. In the case of quaternions, the encoding is through the half rotation vector \nm{\vec h = \vec \Omega \, t = \vec n \, \theta}. Since the rotation is accomplished by the double product \nm{\vec q \otimes \vec v \otimes \qast} as noted in (\ref{eq::SO3_quat_transform}), the vector \nm{\vec v} experiences a rotation that is twice that encoded in \nm{\vec q}, which means that \nm{\vec q} encodes half the intended rotation on \nm{\vec v}. This implies that the space of unit quaternions \nm{\mathbb{S}^3} is in fact a double covering of \nm{\mathbb{SO}\lrp{3}}, not \nm{\mathbb{SO}\lrp{3}} itself \cite{Sola2017}.
\begin{eqnarray}
\nm{\vec r} & = & \nm{2 \cdot \vec h} \label{eq:SO3_quat_equiv_r} \\
\nm{\omega} & = & \nm{2 \cdot \Omega} \label{eq:SO3_quat_equiv_omega} \\
\nm{\phi} & = & \nm{2 \cdot \theta} \label{eq:SO3_quat_equiv_phi}
\end{eqnarray}

Taking these relationships into consideration, it is possible to obtain a more practical expression for the exponential map \nm{\lrb{exp\lrp{} : \mathfrak{so}\lrp{3} \rightarrow \mathbb{SO}\lrp{3} \ | \ \mathcal R = exp\lrp{\vec h^\wedge} = exp\lrp{\vec r^\wedge / 2}}} and its capitalized form \nm{\lrb{Exp\lrp{} : \mathbb{R}^3 \rightarrow \mathbb{SO}\lrp{3} \ | \ \mathcal R = Exp\lrp{\vec h} = Exp\lrp{\vec r / 2}}} than (\ref{eq:SO3_quat_exponential3}) \cite{Sola2017}:
\neweq{\vec q\lrp{\vec h} = \vec q\lrp{\vec r / 2} = Exp\lrp{\vec h} = Exp\lrp{\vec r / 2} =  e^{\ds{\vec \Omega^{\wedge} t}} = e^{\ds{\vec n^\wedge \theta}} = e^{\ds{\vec n^{\wedge} \phi / 2}} = \cos \dfrac{\phi}{2} + \vec n \, \sin \dfrac{\phi}{2}}{eq:SO3_quat_exponential2}

When regarded as a pure quaternion in \nm{\mathbb{H}_p}, the angle \nm{\theta} between a unit quaternion \nm{\vec q} and the identity \nm{\vec{q_1}} is \nm{\cos \theta = \vec{q_1}^T \, \vec q = q_0}. At the same time, the angle \nm{\phi} rotated by the unit quaternion \nm{\vec q} on objects in \nm{\mathbb{R}^3} satisfies (\ref{eq:SO3_quat_exponential2}), so the angle between a unit quaternion vector and the identify in \nm{\mathbb{H}_p} is half the angle rotated by the unit quaternion in \nm{\mathbb{R}^3} space, as depicted by (\ref{eq:SO3_quat_equiv_phi}).

So far the exponential map has been obtained based on the assumption of constant angular velocity, but this does not need to be the case. Given a unit quaternion \nm{\vec q \in \mathbb{S}^3}, there exists a not necessarily unique rotation vector \nm{\vec r = 2 \cdot \vec h \in \mathbb{R}^3} such that \nm{\vec q = Exp\lrp{\vec r / 2} = Exp\lrp{\vec n \, \phi / 2}}:
\begin{eqnarray}
\nm{\phi} & = & \nm{2 \, \arctan \lrp{\dfrac{\| \vec{q_v} \|}{q_0} }} \label{eq:SO3_theta_from_quat} \\
\nm{\vec n} & = & \nm{\dfrac{\vec{q_v}}{\| \vec{q_v} \|}} \label{eq:SO3_rotvec_from_quat}
\end{eqnarray}

This expression represents the capitalized \emph{logarithmic map} \nm{\{log : \mathbb{SO}\lrp{3} \rightarrow \mathfrak{so}\lrp{3} \ | \ \vec q \in \mathbb{S}^3 \rightarrow \vec h = \vec r / 2 \in \mathbb{R}^3\}} \cite{Sola2017}. As in the case of the rotation matrix described in section \ref{sec:RigidBody_rotation_rotv}, the exponential map described by (\ref{eq:SO3_quat_exponential2}) is surjective but not injective, as a rotation of \nm{\lrp{\|\vec r\| + 2 \, k \, \pi} \forall \ k \in \mathbb{Z}} about \nm{\nm{\vec r / \|\vec r\|}} produces exactly the same unit quaternion. In contrast with the case of rotation matrices, a rotation of \nm{\lrp{- \|\vec r\| + 2 \, k \, \pi}} about \nm{-\vec r / \|\vec r\|} produces the opposite (negative) unit quaternion, although both represent the same rotation.

This shows that the map from rotation matrix to unit quaternion \nm{\{\mathbb{SO}\lrp{3} \rightarrow \mathbb{SO}\lrp{3} | \ \vec R \in \mathbb{R}^{3x3} \rightarrow \vec q \in \mathbb{S}^3\}} is also surjective but not injective, as there are two and only two quaternions corresponding to the same rotation matrix \nm{\lrp{\vec R\lrp{\vec q} = \vec R\lrp{- \vec q}}}. The reason is again the double covering of \nm{\mathbb{SO}\lrp{3}} by the unit quaternion \cite{Sola2017}. One quaternion induces a rotation in \nm{\mathbb{R}^3} that follows the shortest direction to the final angle \nm{\lrp{\phi < \pi / 2}}, while the opposite quaternion rotates the opposite way reaching the same final angle after rotating \nm{\lrp{\phi > \pi / 2}}.

As the vector \nm{\vec \Omega} represents half the angular velocity \nm{\vec \omega}, it is possible to adjust expressions (\ref{eq:SO3_quat_Omega_space}), (\ref{eq:SO3_quat_Omega_body}), and (\ref{eq::SO3_quat_Omega_dot}):
\begin{eqnarray}
\nm{\vec \omega_{\sss {NB}}^{\sss N\wedge}} & = & \nm{2 \ \vec{\dot q}_{\sss {NB}} \otimes \qNB^{\ast}} \label{eq:SO3_quat_omega_space} \\
\nm{\vec \omega_{\sss {NB}}^{\sss B\wedge}} & = & \nm{2 \ \qNB^{\ast} \otimes \qNBdot} \label{eq:SO3_quat_omega_body} \\
\nm{\qNBdot} & = & \nm{\dfrac{1}{2} \; \vec \omega_{\sss {NB}}^{\sss N\wedge} \otimes \qNB = \dfrac{1}{2} \; \qNB \otimes \vec \omega_{\sss {NB}}^{\sss B\wedge}} \label{eq::SO3_quat_omega_dot}
\end{eqnarray}

%%%%%%%%%%%%%%%%%%%%%%%%%%%%%%%%%%%%%%%%%%%%%%%%%%%%%%%%%%%%%%%%%%%%%%%%
%%%%%%%%%%%%%%%%%%%%%%%%%%%%%%%%%%%%%%%%%%%%%%%%%%%%%%%%%%%%%%%%%%%%%%%%
%%%%%%%%%%%%%%%%%%%%%%%%%%%%%%%%%%%%%%%%%%%%%%%%%%%%%%%%%%%%%%%%%%%%%%%%
% SECTION      EULER ANGLES
%%%%%%%%%%%%%%%%%%%%%%%%%%%%%%%%%%%%%%%%%%%%%%%%%%%%%%%%%%%%%%%%%%%%%%%%
%%%%%%%%%%%%%%%%%%%%%%%%%%%%%%%%%%%%%%%%%%%%%%%%%%%%%%%%%%%%%%%%%%%%%%%%
%%%%%%%%%%%%%%%%%%%%%%%%%%%%%%%%%%%%%%%%%%%%%%%%%%%%%%%%%%%%%%%%%%%%%%%%

\section{Euler Angles}\label{sec:RigidBody_rotation_euler}

All the previous representations have some type of redundancy as their dimension is higher than three. It is always possible, however, to pick three unitary vectors \nm{\lrp{\vec n_1, \vec n_2, \vec n_3}} forming a basis\footnote{The base vectors do not need to be orthogonal, just linearly independent.} and perform three consecutive rotations to define a map \nm{\{\mathbb{R}^3 \rightarrow \mathbb{SO}\lrp{3} \ | \ \lrp{\beta_1, \beta_2, \beta_3} \in \mathbb{R}^3 \rightarrow R = e^{\ds{\widehat {\vec n}_1 \beta_1}} \ e^{\ds{\widehat {\vec n}_2 \beta_2}} \ e^{\ds{\widehat {\vec n}_3 \beta_3}} \in \mathbb{SO}\lrp{3}\}} \cite{Soatto2001}.

In case the selected basis is orthonormal there are only twelve possible combinations or Euler angles, of which \nm{3-2-1} is the one employed in this document\footnote{This means to first rotate about the \third\ axis, then about the resulting \second\ axis, and finally about the ensuing \first\ axis.}. The rotation from the spatial or global frame \nm{F_{\sss N} = \{\vec O_{\sss {CR}}, \vec n_1, \vec n_2, \vec n_3\}} to the local or body frame \nm{F_{\sss B} = \{\vec O_{\sss {CR}}, \vec b_1, \vec b_2, \vec b_3\}} is performed by first rotating a \emph{yaw} angle (y) about \nm{\vec n_3}, followed by rotating a \emph{pitch} angle (p) about \nm{\vec n_2'}, and finally rotating a \emph{bank} or \emph{roll} angle (r) about \nm{\vec n_1''}, where {\nm{\vec n_2'} is the result of applying the first rotation to \nm{\vec n_2} and \nm{\vec n_1''}, which coincides with \nm{\vec b_1}, that of applying the first two rotations to \nm{\vec n_1}.
\neweq{ \vec R_1(r) = e^{\ds{\widehat {\vec n}_1 \ttt r}} = \begin{bmatrix} 1 & 0 & 0 \\ 0 & \nm{cr} & \nm{-sr}\\ 0 & \nm{sr} & \nm{cr} \end{bmatrix} \
        \vec R_2(p) = e^{\ds{\widehat {\vec n}_2 \ttt p}} = \begin{bmatrix} \nm{cp} & 0 & \nm{sp} \\ 0 & 1 & 0 \\ \nm{-sp} & 0 & \nm{cp} \end{bmatrix} \
        \vec R_3(y) = e^{\ds{\widehat {\vec n}_3 \ttt y}} = \begin{bmatrix} \nm{cy} & \nm{-sy} & 0 \\ \nm{sy} & \nm{cy} & 0 \\ 0 & 0 & 1 \end{bmatrix}} {eq:SO3_euler_individual}

where \emph{s} and \emph{c} stand for sine and cosine respectively. The complete map for the three rotations is:
\neweq{\RNB = \vec R_3(y) \; \vec R_2(p) \; \vec R_1(r) =  \begin{bmatrix}
				\nm{+ cp \cdot cy} &
				\nm{- cr \cdot sy + sr \cdot sp \cdot cy} &
				\nm{+ sr \cdot sy + cr \cdot sp \cdot cy} \\
				\nm{+ cp \cdot sy} &
				\nm{+ cr \cdot cy + sr \cdot sp \cdot sy} &
				\nm{- sr \cdot cy + cr \cdot sp \cdot sy} \\
				\nm{- sp} &
				\nm{+ sr \cdot cp} &
				\nm{+ cr \cdot cp} \end{bmatrix}} {eq:SO3euler_to_R}

In this document the Euler angles are denoted by \nm{\phiNB = \lrsb{y, \, p, \, r}^T}. They can also be obtained from the rotation matrix, but there are singular instances (\nm{p = \pm \pi/2}) where the angles can not be uniquely determined. 
\neweq{y = \arctan \frac{R_{\sss{21}}}{R_{\sss{11}}} \ \ \ \ p = \arcsin\lrp{- R_{\sss{31}}} \ \ \ \ r = \arctan \frac{R_{\sss{32}}}{R_{\sss{33}}}} {eq:SO3_euler_from_R}

%%%%%%%%%%%%%%%%%%%%%%%%%%%%%%%%%%%%%%%%%%%%%%%%%%%%%%%%%%%%%%%%%%%%%%%%
%%%%%%%%%%%%%%%%%%%%%%%%%%%%%%%%%%%%%%%%%%%%%%%%%%%%%%%%%%%%%%%%%%%%%%%%
%%%%%%%%%%%%%%%%%%%%%%%%%%%%%%%%%%%%%%%%%%%%%%%%%%%%%%%%%%%%%%%%%%%%%%%%
% SECTION      ROTATIONAL MOTION ALGEBRAIC OPERATIONS
%%%%%%%%%%%%%%%%%%%%%%%%%%%%%%%%%%%%%%%%%%%%%%%%%%%%%%%%%%%%%%%%%%%%%%%%
%%%%%%%%%%%%%%%%%%%%%%%%%%%%%%%%%%%%%%%%%%%%%%%%%%%%%%%%%%%%%%%%%%%%%%%%
%%%%%%%%%%%%%%%%%%%%%%%%%%%%%%%%%%%%%%%%%%%%%%%%%%%%%%%%%%%%%%%%%%%%%%%%

\section{Rotational Motion Algebraic Operations}\label{sec:RigidBody_rotation_algebra}

The basic algebraic operations of addition, subtraction, multiplication, division, and exponentiation are not defined for objects of the special orthogonal group \nm{\mathbb{SO}\lrp{3}}, no matter if they are represented by a rotation matrix \nm{\vec R \in \mathbb{R}^{3x3}}, a rotation vector \nm{\vec r \in \mathbb{R}^3}, or a unit quaternion \nm{\vec q \in \mathbb{S}^3}. However, as members of the special orthogonal group \nm{\mathbb{SO}\lrp{3}}, all rotation representations are closed under a given operation that represents the concatenation of rotations, and define not only an identity rotation that represents the lack of rotation, but also an inverse operation representing the opposite rotation. The concatenation of rotations and the identity and inverse operations enable the definition of the power, exponential and logarithmic operators (section \ref{subsec:RigidBody_rotation_algebra_exp_log}), the spherical linear interpolation (section \ref{subsec:RigidBody_rotation_algebra_slerp}), and the perturbations together with the plus and minus operators (section \ref{subsec:RigidBody_rotation_algebra_plus_minus}).

%%%%%%%%%%%%%%%%%%%%%%%%%%%%%%%%%%%%%%%%%%%%%%%%%%%%%%%%%%%%%%%%%%%%%%%%
%%%%%%%%%%%%%%%%%%%%%%%%%%%%%%%%%%%%%%%%%%%%%%%%%%%%%%%%%%%%%%%%%%%%%%%%
% SUBSECTION      POWERS, EXPONENTIALS AND LOGARITHMS
%%%%%%%%%%%%%%%%%%%%%%%%%%%%%%%%%%%%%%%%%%%%%%%%%%%%%%%%%%%%%%%%%%%%%%%%
%%%%%%%%%%%%%%%%%%%%%%%%%%%%%%%%%%%%%%%%%%%%%%%%%%%%%%%%%%%%%%%%%%%%%%%%

\subsection{Powers, Exponentials and Logarithms}\label{subsec:RigidBody_rotation_algebra_exp_log}

Any rotation can be executed by rotating a given angle \nm{\phi} about a fixed rotation axis \nm{\vec n}, resulting in the rotation vector \nm{\vec r = \vec n \, \phi} (section \ref{sec:RigidBody_rotation_rotv}) or its half \nm{\vec h = \vec n \, \phi / 2} (section \ref{sec:RigidBody_rotation_halfrotv}). Taking a multiple or a fraction of a rotation vector is hence straightforward, as \nm{t \, \vec r = t \, \vec n \, \phi = \vec n \, \lrp{t \, \phi} \, \forall \, t \in \mathbb{R}, \vec r \in \mathfrak{so}\lrp{3}}. The exponential maps defined in (\ref{eq:SO3_rotv_exponential2}) and (\ref{eq:SO3_quat_exponential2}) are named that way because they comply with the behavior of the real exponential function \nm{exp^b\lrp{a} = exp\lrp{a \cdot b}\, \forall \, a, \, b \in \mathbb{R}}. As such, the exponential function \nm{\{exp(): \mathbb{R}^3 \times \mathbb{R} \rightarrow \mathbb{SO}\lrp{3} \ | \ \vec r \in \mathbb{R}^3, t \in \mathbb{R} \rightarrow {\mathcal R}^t = Exp\lrp{t \, \vec r} \in \mathbb{SO}\lrp{3}\}} is defined as:
\begin{eqnarray}
\nm{\vec R^{\ds t}\lrp{\vec r}} & = & \nm{\vec R\lrp{t \, \vec r} = Exp\lrp{t \, \vec r} = \vec I_3 + \frac{\rskew}{\|\vec r\|} \sin \|t \, \vec r\| + \frac{\rskew^2}{\|\vec r\|^2} \lrp{1 - \cos \|t \, \vec r\|}}\label{eq:SO3_rotv_exponential_fraction} \\
\nm{\vec q^{\ds t}\lrp{\vec h = \vec r / 2}} & = & \nm{\vec q\lrp{t \, \vec h = t \, \vec r / 2} = Exp\lrp{t \, \vec r / 2} = \cos \dfrac{t \, \phi}{2} + \vec n \, \sin \dfrac{t \, \phi}{2}}\label{eq:SO3_quat_exponential_fraction}
\end{eqnarray}

In a similar way, the logarithmic maps defined in (\ref{eq:SO3_rotv_logarithm}), (\ref{eq:SO3_theta_from_quat}), and (\ref{eq:SO3_rotvec_from_quat}) also comply with the behavior of the real logarithmic function \nm{b \cdot log\lrp{a} = log\lrp{a^b} \, \forall \, a, \, b \in \mathbb{R}}. As such, the logarithmic function \nm{\{log(): \mathbb{SO}\lrp{3} \times \mathbb{R} \rightarrow \mathbb{R}^3 \ | \ \mathcal R \in \mathbb{SO}\lrp{3}, t \in \mathbb{R} \rightarrow t \, \vec r = Log\lrp{{\mathcal R}^t} \in \mathbb{R}^3\}} is defined as:
\begin{eqnarray}
\nm{log\lrp{\vec R^{\ds t}\lrp{\vec r}}} & = & \nm{Log\lrp{Exp\lrp{t \, \vec r}} = t \, Log\lrp{\vec R\lrp{\vec r}} =  t \, Log\lrp{Exp\lrp{\vec r}} = t \, \vec r}\label{eq:SO3_rotv_logarithmic_fraction} \\
\nm{log\lrp{\vec q^{\ds t}\lrp{\vec h = \vec r / 2}}} & = & \nm{Log\lrp{Exp\lrp{t \, \vec r / 2}} = t \, Log\lrp{\vec q\lrp{\vec r / 2}} = t \, Log\lrp{Exp\lrp{\vec r / 2}} = t \, \vec r / 2}\label{eq:SO3_quat_logarithmic_fraction} 
\end{eqnarray}

%%%%%%%%%%%%%%%%%%%%%%%%%%%%%%%%%%%%%%%%%%%%%%%%%%%%%%%%%%%%%%%%%%%%%%%%
%%%%%%%%%%%%%%%%%%%%%%%%%%%%%%%%%%%%%%%%%%%%%%%%%%%%%%%%%%%%%%%%%%%%%%%%
% SUBSECTION      SPHERICAL LINEAR INTERPOLATION
%%%%%%%%%%%%%%%%%%%%%%%%%%%%%%%%%%%%%%%%%%%%%%%%%%%%%%%%%%%%%%%%%%%%%%%%
%%%%%%%%%%%%%%%%%%%%%%%%%%%%%%%%%%%%%%%%%%%%%%%%%%%%%%%%%%%%%%%%%%%%%%%%

\subsection{Spherical Linear Interpolation}\label{subsec:RigidBody_rotation_algebra_slerp}

Given two rotations \nm{\mathcal R_0, \, \mathcal R_1 \in \mathbb{SO}\lrp{3}}, \emph{spherical linear interpolation} (\hypertt{SLERP}) seeks to obtain a rotation function \nm{\mathcal R\lrp{t}, \, t \in \mathbb{R}} that linearly interpolates from \nm{\mathcal R\lrp{0} = \mathcal R_0} to \nm{\mathcal R\lrp{1} = \mathcal R_1} in such a way that the rotation occurs at constant angular velocity along a fixed axis \cite{Sola2017}.

If employing unit quaternions, \nm{\Delta \vec q} is according to (\ref{eq:SO3_quat_concatenation}) the full rotation required to go from \nm{\vec q_0} to \nm{\vec q_1}, such that \nm{\vec q_1 = \vec q_0 \otimes \Delta \vec q}, from where \nm{\Delta \vec q = \vec q_0^{\ast} \otimes \vec q_1}. The corresponding rotation vector is then \nm{\Delta \vec r = \vec n \, \Delta \phi = 2 \, Log\lrp{\Delta \vec q}}. The quaternion corresponding to a fraction of the full rotation \nm{\delta \phi = t \, \Delta \phi} is the following:
\begin{eqnarray}
\nm{\delta \vec q} & = & \nm{Exp\lrp{\dfrac{\vec n \, \delta \phi}{2}} = Exp\lrp{t \, \dfrac{\vec n \, \Delta \phi}{2}} = Exp\lrp{t \, \dfrac{\Delta \vec r}{2}}} \nonumber \\
& = & \nm{Exp\big(t \, Log\lrp{\Delta \vec q}\big) = Exp\big(t \, Log\lrp{\vec q_0^{\ast} \otimes \vec q_1}\big) = \lrp{\vec q_0^{\ast} \otimes \vec q_1}^{\ds t}} \label{eq:SO3_interp_quat_partial}
\end{eqnarray}

The interpolated unit quaternion is hence the following:
\neweq{\vec q\lrp{t} = \vec q_0 \otimes \lrp{\vec q_0^{\ast} \otimes \vec q_1}^{\ds t}}{eq:SO3_interp_quat}

Because of the double covering of \nm{\mathbb{SO}\lrp{3}} by the quaternion, only the interpolation between quaternions at acute angles \nm{\lrp{\Delta \theta = \Delta \phi / 2 \leq \pi / 2}}, is executed following the shortest path, which occurs if \nm{\cos \Delta \theta = {\vec q_0}^T \, \vec q_1 < 0}. If this is not the case, just replace \nm{\vec q_1} by \nm{- \vec q_1} and repeat the process.

A similar result is obtained when employing rotation matrices instead of unit quaternions:
\neweq{\vec R\lrp{t} = \vec R_0 \, \lrp{\vec R_0^T \, \vec R_1}^{\ds t}}{eq:SO3_interp_dcm}

%%%%%%%%%%%%%%%%%%%%%%%%%%%%%%%%%%%%%%%%%%%%%%%%%%%%%%%%%%%%%%%%%%%%%%%%
%%%%%%%%%%%%%%%%%%%%%%%%%%%%%%%%%%%%%%%%%%%%%%%%%%%%%%%%%%%%%%%%%%%%%%%%
% SUBSECTION      PLUS AND MINUS OPERATORS
%%%%%%%%%%%%%%%%%%%%%%%%%%%%%%%%%%%%%%%%%%%%%%%%%%%%%%%%%%%%%%%%%%%%%%%%
%%%%%%%%%%%%%%%%%%%%%%%%%%%%%%%%%%%%%%%%%%%%%%%%%%%%%%%%%%%%%%%%%%%%%%%%

\subsection{Plus and Minus Operators}\label{subsec:RigidBody_rotation_algebra_plus_minus}

A perturbed rigid body rotation \nm{\widetilde{\mathcal R} \in \mathbb{SO}\lrp{3}} can always be expressed as the composition of the unperturbed rotation \nm{\mathcal R} with a (usually) small perturbation \nm{\Delta \mathcal R}. Perturbations can be specified either at the local or body frame \nm{\FB}, this is, at the local vector space tangent to \nm{\mathbb{SO}\lrp{3}} at the actual orientation, in which case they are known as \emph{local perturbations}. They can also be specified at the global frame \nm{\FN}, which coincides with the vector space tangent to \nm{\mathbb{SO}\lrp{3}} at the origin; in this case they are known as \emph{global perturbations} \cite{Sola2017}. Local perturbations appear on the right hand side of the rotation composition, resulting in \nm{\widetilde{\mathcal R} = \mathcal R \circ \Delta\mathcal{R}^{\sss B}}, while global ones appear to the left, hence \nm{\widetilde{\mathcal R} = \Delta\mathcal{R}^{\sss N} \circ \mathcal R}.

The \emph{plus} and \emph{minus operators} are introduced in section \ref{subsec:algebra_lie_exp_plus} and enable operating with increments of the nonlinear \nm{\mathbb{SO}\lrp{3}} manifold expressed in the linear tangent vector space \nm{\mathfrak{so}\lrp{3}}. There exist right (\nm{\oplus, \, \ominus}) or left (\nm{\boxplus, \, \boxminus}) versions depending on whether the increments are viewed in the local frame (right) or the global one (left). It is important to remark that although perturbations and the plus and left operators are best suited to work with small rotation changes (perturbations), the expressions below are generic and work just the same no matter the size of the perturbation.

The right plus operator \nm{\{\oplus : \mathbb{SO}\lrp{3} \times \mathfrak{so}\lrp{3} \rightarrow \mathbb{SO}\lrp{3} \, | \, \widetilde{\mathcal R} = \mathcal R \oplus \ \Delta \vec r^{\sss B} = \mathcal R \circ Exp\lrp{\Delta \vec r^{\sss B}}\}} produces a rotation element \nm{\widetilde{\mathcal R}} resulting from the composition of a reference rotation \nm{\mathcal R} with an often small rotation \nm{\Delta \vec r^{\sss B} = \vec n^{\sss B} \, \Delta \phi}, contained in the tangent space to the reference rotation \nm{\mathcal R}, this is, in the local space \cite{Sola2017}. The left plus operator \nm{\{\boxplus : \mathfrak{so}\lrp{3} \times \mathbb{SO}\lrp{3} \rightarrow \mathbb{SO}\lrp{3} \, | \, \widetilde{\mathcal R} = \Delta \vec r^{\sss N} \boxplus \mathcal R = Exp\lrp{\Delta \vec r^{\sss N}} \circ \mathcal R\}} is similar but the often small rotation \nm{\Delta \vec r^{\sss N} = \vec n^{\sss N} \,  \Delta \phi} is contained in the tangent space at the identify or global space. The expressions shown below are valid up to the first coverage of \nm{\mathbb{SO}\lrp{3}}, this is, \nm{\phi < \pi}. In the cases of rotation matrix and unit quaternion, the plus operator is defined as:
\begin{eqnarray}
\nm{\widetilde{\vec R}} & = & \nm{\vec R \oplus \Delta \vec r^{\sss B} = \vec R \ Exp\lrp{\Delta \vec r^{\sss B}} = \vec R \ \Delta \vec R^{\sss B}}\label{eq:SO3_dcm_plus} \\
\nm{\widetilde{\vec q}} & = & \nm{\vec q \oplus \Delta \vec r^{\sss B} = \vec q \otimes Exp\lrp{\Delta \vec r^{\sss B} / 2} = \vec q \otimes \Delta \vec q^{\sss B}}\label{eq:SO3_quat_plus} \\
\nm{\widetilde{\vec R}} & = & \nm{\Delta \vec r^{\sss N} \boxplus \vec R = Exp\lrp{\Delta \vec r^{\sss N}} \ \vec R = \Delta \vec R^{\sss N} \ \vec R}\label{eq:SO3_dcm_plus_left} \\
\nm{\widetilde{\vec q}} & = & \nm{\Delta \vec r^{\sss N} \boxplus \vec q = Exp\lrp{\Delta \vec r^{\sss N} / 2} \otimes \vec q = \Delta \vec q^{\sss N} \otimes \vec q}\label{eq:SO3_quat_plus_left}
\end{eqnarray}

The right minus operator \nm{\{\ominus : \mathbb{SO}\lrp{3} \times \mathbb{SO}\lrp{3} \rightarrow \mathfrak{so}\lrp{3} \, | \, \Delta \vec r^{\sss B} = \widetilde{\mathcal R} \ominus \mathcal R = Log\big(\mathcal R^{-1} \circ \widetilde{\mathcal R}\big)\}}, as well as the left \nm{\{\boxminus : \mathbb{SO}\lrp{3} \times \mathbb{SO}\lrp{3} \rightarrow \mathfrak{so}\lrp{3} \, | \, \Delta \vec r^{\sss N} = \widetilde{\mathcal R} \boxminus \mathcal R = Log\big(\widetilde{\mathcal R} \circ \mathcal R^{-1}\big)\}}, represent the inverse operations, returning the rotation vector difference \nm{\Delta \vec r} between two rotations \nm{\mathcal R} and \nm{\widetilde{\mathcal R}} expressed in either the local or global tangent spaces to \nm{\mathcal R}.
\begin{eqnarray}
\nm{\Delta \vec r^{\sss B} } & = & \nm{\widetilde{\vec R} \ominus \vec R = Log\lrp{\vec R^T \ \widetilde{\vec R}} = Log\lrp{\Delta \vec R^{\sss B}}}\label{eq:SO3_dcm_minus} \\
\nm{\Delta \vec r^{\sss B} } & = & \nm{\widetilde{\vec q} \ominus \vec q = 2 \ Log\lrp{\qast \otimes \widetilde{\vec q}} = 2 \ Log\lrp{\Delta \vec q^{\sss B}}}\label{eq:SO3_quat_minus} \\
\nm{\Delta \vec r^{\sss N} } & = & \nm{\widetilde{\vec R} \boxminus \vec R = Log\lrp{\widetilde{\vec R} \ \vec R^T} = Log\lrp{\Delta \vec R^{\sss N}}}\label{eq:SO3_dcm_minus_left} \\
\nm{\Delta \vec r^{\sss N} } & = & \nm{\widetilde{\vec q} \boxminus \vec q = 2 \ Log\lrp{\widetilde{\vec q} \otimes \qast} = 2 \ Log\lrp{\Delta \vec q^{\sss N}}}\label{eq:SO3_quat_minus_left}
\end{eqnarray}

If the \nm{\Delta \vec r} perturbation is small, the (\ref{eq:SO3_rotv_exponential3}) and (\ref{eq:SO3_quat_exponential3}) Taylor expansions can be truncated, resulting in the following expressions, valid for both the body frame (\nm{\Delta \vec r^{\sss B}}) or the global one (\nm{\Delta \vec r^{\sss N}}):
\begin{eqnarray}
\nm{\Delta \vec R = Exp\lrp{\Delta \vec r}} & \nm{\approx} & \nm{\vec I_3 + \Delta \vec r^\wedge = \vec I_3 + \Delta \phi \, \nskew}\label{eq:SO3_perturbation_dcm_truncated_local} \\
\nm{\Delta \vec q = Exp\lrp{\Delta \vec r /2}} & \nm{\approx} & \nm{\vec {q_1} + \Delta \vec r^\wedge / 2 = \lrsb{1, \vec n \, \Delta \phi / 2}^T}\label{eq:SO3_perturbation_quat_truncated_local} 
\end{eqnarray}

%%%%%%%%%%%%%%%%%%%%%%%%%%%%%%%%%%%%%%%%%%%%%%%%%%%%%%%%%%%%%%%%%%%%%%%%
%%%%%%%%%%%%%%%%%%%%%%%%%%%%%%%%%%%%%%%%%%%%%%%%%%%%%%%%%%%%%%%%%%%%%%%%
%%%%%%%%%%%%%%%%%%%%%%%%%%%%%%%%%%%%%%%%%%%%%%%%%%%%%%%%%%%%%%%%%%%%%%%%
% SECTION      ROTATIONAL MOTION TIME DERIVATIVE AND ANGULAR VELOCITY
%%%%%%%%%%%%%%%%%%%%%%%%%%%%%%%%%%%%%%%%%%%%%%%%%%%%%%%%%%%%%%%%%%%%%%%%
%%%%%%%%%%%%%%%%%%%%%%%%%%%%%%%%%%%%%%%%%%%%%%%%%%%%%%%%%%%%%%%%%%%%%%%%
%%%%%%%%%%%%%%%%%%%%%%%%%%%%%%%%%%%%%%%%%%%%%%%%%%%%%%%%%%%%%%%%%%%%%%%%

\section{Rotational Motion Time Derivative and Angular Velocity}\label{sec:RigidBody_rotation_calculus_derivatives}

Let \nm{\mathcal R\lrp{t} \in \mathbb{SO}\lrp{3}, t \in \mathbb{R}} be a rotating rigid body, and compute its derivative with time, which belongs to neither \nm{\mathbb{SO}\lrp{3}} nor \nm{\mathfrak{so}\lrp{3}} but to the Euclidean space of the chosen rotation representation, \nm{\mathbb{R}^{3x3}} for the rotation matrix and \nm{\mathbb{H}} for the unit quaternion:
\neweq{\dot{\mathcal{R}}\lrp{t} = \lim\limits_{\Delta t \to 0} \dfrac{\mathcal{R}\lrp{t + \Delta t} - \mathcal{R}\lrp{t}}{\Delta t}}{eq:SO3_time_derivative_def}

Considering the time modified rotation \nm{\mathcal{R}\lrp{t + \Delta t}} as the perturbed state (section \ref{subsec:RigidBody_rotation_algebra_plus_minus}), the resulting time derivatives for the rotation matrix and unit quaternion representations are the following:
\begin{eqnarray}
\nm{\vec {\dot R}\lrp{t}} & = & \nm{\lim\limits_{\Delta t \to 0} \dfrac{\vec R \, \Delta \vec R^{\sss B} - \vec R}{\Delta t} \ \nm{\approx} \lim\limits_{\Delta t \to 0}\dfrac{\vec R \, \lrsb{\lrp{\vec I_3 + \Delta \phi \ \nskew^{\sss B}} - \vec I_3}}{\Delta t} = \vec R \, \lim\limits_{\Delta t \to 0}\dfrac{\Delta \phi \, \nskew^{\sss B}}{\Delta t}}\label{eq:SO3_time_derivative_R1b} \\
\nm{\vec {\dot q}\lrp{t}} & = & \nm{\lim\limits_{\Delta t \to 0} \dfrac{\vec q \otimes \Delta \vec q^{\sss B} - \vec q}{\Delta t} \ \nm{\approx} \lim\limits_{\Delta t \to 0}\dfrac{\vec q \otimes \lrsb{\lrsb{1 \ \ \ \vec n^{\sss B} \, \Delta \phi / 2}^T - \vec{q_1}}}{\Delta t} = \vec q \otimes \lim\limits_{\Delta t \to 0}\dfrac{\lrsb{0 \ \ \ \vec n^{\sss B} \, \Delta \phi / 2}^T}{\Delta t}}\label{eq:SO3_time_derivative_q1b} 
\end{eqnarray}

Similar expressions based on \nm{\vec r^{\sss N} = \Delta \phi \, \vec n^{\sss N}} can be obtained if left multiplying by the perturbation instead of right multiplying. The \nm{\vec {\dot R}\lrp{t}} and \nm{\vec {\dot q}\lrp{t}} expressions (\ref{eq:SO3_dcm_dot}) and (\ref{eq::SO3_quat_omega_dot}) are then directly obtained when defining the \emph{body angular velocity} \nm{\wNBB} as the time derivative of the rotation vector \nm{\vec r^{\sss B} = \vec n^{\sss B} \, \phi} when viewed in local or body frame \nm{\FB}, and the \emph{spatial angular velocity} \nm{\wNBN} as the time derivative of the rotation vector \nm{\vec r^{\sss N} = \vec n^{\sss N} \, \phi} when viewed in global or spatial frame \nm{\FN}:
\begin{eqnarray}
\nm{\wNBB\lrp{t}} & = & \nm{\Delta \vec{\dot r}^{\sss B}\lrp{t} = \lim\limits_{\Deltat \to 0} \frac{\Delta \vec r^{\sss B}}{\Deltat} = \lim\limits_{\Deltat \to 0} \frac{\vec n^{\sss B} \, \Delta \phi}{\Deltat}}\label{eq:SO3_time_derivative_wNBB} \\
\nm{\wNBN\lrp{t}}   & = & \nm{\Delta \vec{\dot r}^{\sss N}\lrp{t} = \lim\limits_{\Deltat \to 0} \frac{\Delta \vec r^{\sss N}}{\Deltat} = \lim\limits_{\Deltat \to 0} \frac{\vec n^{\sss N} \, \Delta \phi}{\Deltat}}\label{eq:SO3_time_derivative_wNBN}  
\end{eqnarray}

Note that the rotation of the angular velocity (relationship between \nm{\wNBN} and \nm{\wNBB}) is not given by the rotation action \nm{\vec g_{\mathcal R*}} (\ref{eq:Rotate_vector_action}) but by the adjoint map \nm{\vec{Ad}_{\mathcal R}} described in section \ref{sec:RigidBody_rotation_adjoint}, although in the case of the \nm{\mathbb{SO}\lrp{3}} rotation group, both maps coincide.

%%%%%%%%%%%%%%%%%%%%%%%%%%%%%%%%%%%%%%%%%%%%%%%%%%%%%%%%%%%%%%%%%%%%%%%%
%%%%%%%%%%%%%%%%%%%%%%%%%%%%%%%%%%%%%%%%%%%%%%%%%%%%%%%%%%%%%%%%%%%%%%%%
%%%%%%%%%%%%%%%%%%%%%%%%%%%%%%%%%%%%%%%%%%%%%%%%%%%%%%%%%%%%%%%%%%%%%%%%
% SECTION      ROTATIONAL MOTION POINT VELOCITY
%%%%%%%%%%%%%%%%%%%%%%%%%%%%%%%%%%%%%%%%%%%%%%%%%%%%%%%%%%%%%%%%%%%%%%%%
%%%%%%%%%%%%%%%%%%%%%%%%%%%%%%%%%%%%%%%%%%%%%%%%%%%%%%%%%%%%%%%%%%%%%%%%
%%%%%%%%%%%%%%%%%%%%%%%%%%%%%%%%%%%%%%%%%%%%%%%%%%%%%%%%%%%%%%%%%%%%%%%%

\section{Rotational Motion Point Velocity}\label{sec:RigidBody_rotation_velocity}

There exists a direct relationship between the velocity of a point belonging to a rigid body and the elements of its tangent space, this is, the angular velocity in \nm{\mathfrak{so}\lrp{3}} in the case of rotational motion. This relationship is independent of the \nm{\mathbb{SO}\lrp{3}} representation, although the rotation matrix is employed in the expressions below. As discussed in section \ref{sec:RigidBody_bases}, the rotation actions have the same form for points as for vectors (\ref{eq:SO3_equivalence}). Hence, if \nm{\pB} are the fixed coordinates of a point belonging to the \nm{\FB} rigid body, the point spatial coordinates \nm{\pN} can be obtained by means of (\ref{eq:SO3_dcm_transform}):
\neweq{\pN\lrp{t} = \vec g_{\mathcal R_{NB}(t)}\lrp{\pB} = \RNB\lrp{t} \; \pB}{eq:SO3_velocity1} 

The velocity of a point is the time derivative of its spatial or global coordinates. As \nm{\vec p} is fixed to \nm{F_{\sss B}}, its time derivative is zero \nm{\lrp{\vec {\dot p}^{\sss B} = \vec 0}}, so its velocity viewed in the spatial frame responds to:
\neweq{\vec v_{\sss p}^{\sss N}\lrp{t} = \vec {\dot p}^{\sss N}\lrp{t} = \vec {\dot R}_{\sss NB}\lrp{t} \; \pB}{eq:SO3_velocity2}

Although \nm{\RNBdot} maps the point body coordinates to its spatial velocity per (\ref{eq:SO3_velocity2}), its high dimension makes it inefficient \cite{Murray1994}. By making use of the spatial and body instantaneous angular velocities (\nm{\wNBNskew, \, \wNBBskew}) introduced in (\ref{eq:SO3_dcm_dot}), the velocity of a point \nm{\pB} viewed in \nm{\FN} can be obtained as follows:
\begin{eqnarray}
\nm{\vec v_{\sss p}^{\sss N}\lrp{t}} & = & \nm{\wNBNskew\lrp{t} \; \RNB\lrp{t} \; \pB = \wNBNskew\lrp{t} \; \pN\lrp{t}}\label{eq:SO3_velocity_n} \\
\nm{\vec v_{\sss p}^{\sss N}\lrp{t}} & = & \nm{\RNB\lrp{t} \; \wNBBskew\lrp{t} \; \pB}\label{eq:SO3_velocity_n_bis} 
\end{eqnarray}

The velocity of \nm{\pB} viewed in \nm{\FB} can then be obtained by means of the vector action map:
\neweq{\vec v_{\sss p}^{\sss B}\lrp{t} = \vec g_{\mathcal R_{NB(t)}*}^{-1} \big(\vec v_{\sss p}^{\sss N}\lrp{t}\big) = \RNBtrans\lrp{t} \; \vec v_{\sss p}^{\sss N}\lrp{t} = \wNBBskew\lrp{t} \; \pB} {eq:SO3_velocity_b} 

The point velocity is hence the result of the cross product between the angular velocity and the point coordinates (\ref{eq:SO3_velocity_n_bis}, \ref{eq:SO3_velocity_b}). Similar expressions are obtained if employing the unit quaternion (\nm{\vec v_{\sss p}^{\sss N}\lrp{t} = \vec \omega_{\sss NB}^{{\sss N}\wedge} \otimes \pN\lrp{t}}, \nm{\vec v_{\sss p}^{\sss B}\lrp{t} = \vec \omega_{\sss NB}^{{\sss B}\wedge} \otimes \pB}). 

%%%%%%%%%%%%%%%%%%%%%%%%%%%%%%%%%%%%%%%%%%%%%%%%%%%%%%%%%%%%%%%%%%%%%%%%
%%%%%%%%%%%%%%%%%%%%%%%%%%%%%%%%%%%%%%%%%%%%%%%%%%%%%%%%%%%%%%%%%%%%%%%%
%%%%%%%%%%%%%%%%%%%%%%%%%%%%%%%%%%%%%%%%%%%%%%%%%%%%%%%%%%%%%%%%%%%%%%%%
% SECTION      ROTATIONAL MOTION ADJOINT
%%%%%%%%%%%%%%%%%%%%%%%%%%%%%%%%%%%%%%%%%%%%%%%%%%%%%%%%%%%%%%%%%%%%%%%%
%%%%%%%%%%%%%%%%%%%%%%%%%%%%%%%%%%%%%%%%%%%%%%%%%%%%%%%%%%%%%%%%%%%%%%%%
%%%%%%%%%%%%%%%%%%%%%%%%%%%%%%%%%%%%%%%%%%%%%%%%%%%%%%%%%%%%%%%%%%%%%%%%

\section{Rotational Motion Adjoint}\label{sec:RigidBody_rotation_adjoint}

The \emph{adjoint map} of a Lie group is defined in section \ref{subsec:algebra_lie_adjoint} as an action of the Lie group on its own Lie algebra that converts between the local tangent space and that at the identity. In the case of rotational motion, both the rotation vector and the angular velocity belong to the tangent space, so \nm{\lrb{\vec{Ad}\lrp{}: \mathbb{SO}\lrp{3} \times \mathfrak{so}\lrp{3} \rightarrow \mathfrak{so}\lrp{3} \ | \ \vec{Ad}_{\mathcal R}\lrp{\vec r^{\wedge}} = \mathcal R \circ \vec r^{\wedge} \circ \mathcal{R}^{-1}, \ \vec{Ad}_{\mathcal R}\lrp{\vec \omega^{\wedge}} = \mathcal R \circ \vec \omega^{\wedge} \circ \mathcal{R}^{-1}}}. This is equivalent to \nm{\vec q \otimes \vec \omega^{\wedge} \otimes \qast} for unit quaternions or \nm{\vec R \; \omegaskew \; \vec R^T} for rotation matrices, which represents the congruency transformation\footnote{Two square matrices \nm{\vec A} and \nm{\vec B} are called congruent if \nm{\vec B = {\vec P}^T \; \vec A \; \vec P} for some invertible square matrix \nm{\vec P}.} between the spatial and body angular velocities \nm{\wNBNskew} and \nm{\wNBBskew}:
\neweq{\wNBNskew = \RNB \; \wNBBskew \; \RNBtrans}{eq:SO3_dcm_velocity5} 

The application of the vee operator results in the adjoint matrix coinciding with the rotation matrix itself \nm{\lrb{\vec{Ad}_{\mathcal R} \, \vec \omega = \vec R \, \vec \omega}}\footnote{The adjoint matrix is generic and applicable to all \nm{\mathbb{SO}\lrp{3}} representations.}, implying that elements of the \nm{\mathbb{SO}\lrp{3}} tangent space (both rotation vectors and angular velocities) can be transformed by means of the rotation action as any other free vector. Note that this result only applies to rotational motion, as for example the vector action and adjoint matrix of \nm{\mathbb{SE}\lrp{3}} discussed in chapter \ref{cha:Motion} do not coincide:
\neweq{\wNBN = \vec{Ad}_{\mathcal R_{NB}} \; \wNBB = \RNB \; \wNBB}{eq:SO3_dcm_velocity6} 

A similar process leads to the inverse adjoint matrix (\nm{\vec{Ad}_{\mathcal R}^{-1} \, \vec \omega = \vec{Ad}_{\mathcal R^{-1}} \, \vec \omega = \vec R^T \, \vec \omega}):
\neweq{\wNBB = \vec{Ad}_{\mathcal R_{NB}}^{-1} \; \wNBN = \RNB^T \; \wNBN}{eq:SO3_dcm_velocity7} 

%%%%%%%%%%%%%%%%%%%%%%%%%%%%%%%%%%%%%%%%%%%%%%%%%%%%%%%%%%%%%%%%%%%%%%%%
%%%%%%%%%%%%%%%%%%%%%%%%%%%%%%%%%%%%%%%%%%%%%%%%%%%%%%%%%%%%%%%%%%%%%%%%
%%%%%%%%%%%%%%%%%%%%%%%%%%%%%%%%%%%%%%%%%%%%%%%%%%%%%%%%%%%%%%%%%%%%%%%%
% SECTION      ROTATIONAL MOTION UNCERTAINTY AND COVARIANCE
%%%%%%%%%%%%%%%%%%%%%%%%%%%%%%%%%%%%%%%%%%%%%%%%%%%%%%%%%%%%%%%%%%%%%%%%
%%%%%%%%%%%%%%%%%%%%%%%%%%%%%%%%%%%%%%%%%%%%%%%%%%%%%%%%%%%%%%%%%%%%%%%%
%%%%%%%%%%%%%%%%%%%%%%%%%%%%%%%%%%%%%%%%%%%%%%%%%%%%%%%%%%%%%%%%%%%%%%%%

\section{Rotational Motion Uncertainty and Covariance}\label{sec:RigidBody_rotation_covariance}

Following the analysis of uncertainty on Lie groups presented in section \ref{subsec:algebra_lie_covariance}, the definitions of local and global autocovariances for \nm{\mathbb{SO}\lrp{3}} elements around a nominal or expected rotation \nm{E\lrsb{\mathcal R} = \vec \mu_{\mathcal R} \in \mathbb{SO}\lrp{3}} are the following:
\begin{eqnarray}
\nm{\vec C_{\mathcal R \mathcal R}^{\sss B}} & = & \nm{E\lrsb{\Delta \vec r^{\sss B} \,  \Delta \vec r^{{\sss B}T}} = E\lrsb{\lrp{\mathcal R \ominus \vec \mu_{\mathcal R}}\lrp{\mathcal R \ominus \vec \mu_{\mathcal R}}^T} \ \ \in \mathbb{R}^{3x3}}\label{eq:SO3_covariance_right_def} \\
\nm{\vec C_{\mathcal R \mathcal R}^{\sss N}} & = & \nm{E\lrsb{\Delta \vec r^{\sss N} \,  \Delta \vec r^{{\sss N}T}} = E\lrsb{\lrp{\mathcal R \boxminus \vec \mu_{\mathcal R}}\lrp{\mathcal R \boxminus \vec \mu_{\mathcal R}}^T} \ \ \in \mathbb{R}^{3x3}}\label{eq:SO3_covariance_left_def} 
\end{eqnarray}

Note that although the notation refers to the covariance of the rotation manifold \nm{\mathcal R \in \mathbb{SO}\lrp{3}}, the definition in fact refers to the covariance of the rotation vectors \nm{\Delta \vec r^{\sss B}} or \nm{\Delta \vec r^{\sss N}} located in the tangent space, with its dimension (3) matching the number of degrees of freedom of the \nm{\mathbb{SO}\lrp{3}} manifold. The relationship between the local and global autocovariances responds to:
\neweq{\vec C_{\mathcal R \mathcal R}^{\sss N} = \vec{Ad}_{\mathcal R_{NB}} \ \vec C_{\mathcal R \mathcal R}^{\sss B} \ \vec{Ad}_{\mathcal R_{NB}}^T = \RNB \, \vec C_{\mathcal R \mathcal R}^{\sss B} \, \RNB^T} {eq:SO3_covariance_left_relationship}

Given a function \nm{\lrb{f: \mathcal{R} \rightarrow \mathcal {S} \ | \ \mathcal {S} = f\lrp{\mathcal {R}} \in \mathbb{SO}\lrp{3}, \, \forall \mathcal {R} \in \mathbb{SO}\lrp{3}}} between two rotations, the covariances are propagated as follows:
\begin{eqnarray}
\nm{\vec C_{\mathcal S \mathcal S}^{\sss B}} & = & \nm{\vec J_{\ds{\oplus \; \mathcal R}}^{\ds{\ominus \; f\lrp{\mathcal R}}} \ \vec C_{\mathcal R \mathcal R}^{\sss B} \ \vec J_{\ds{\oplus \; \mathcal R}}^{{\ds{\ominus \; f\lrp{\mathcal R}}},T} \ \ \ \ \ \ \ \in \mathbb{R}^{3x3}} \label{eq:SO3_covariance_right_propagation} \\
\nm{\vec C_{\mathcal S \mathcal S}^{\sss N}} & = & \nm{\vec J_{\ds{\boxplus \; \mathcal R}}^{\ds{\boxminus \; f\lrp{\mathcal R}}} \ \vec C_{\mathcal R \mathcal R}^{\sss N} \ \vec J_{\ds{\boxplus \; \mathcal R}}^{{\ds{\boxminus \; f\lrp{\mathcal R}}},T} \ \ \ \ \ \ \ \in \mathbb{R}^{3x3}} \label{eq:SO3_covariance_left_propagation}
\end{eqnarray}

%%%%%%%%%%%%%%%%%%%%%%%%%%%%%%%%%%%%%%%%%%%%%%%%%%%%%%%%%%%%%%%%%%%%%%%%
%%%%%%%%%%%%%%%%%%%%%%%%%%%%%%%%%%%%%%%%%%%%%%%%%%%%%%%%%%%%%%%%%%%%%%%%
%%%%%%%%%%%%%%%%%%%%%%%%%%%%%%%%%%%%%%%%%%%%%%%%%%%%%%%%%%%%%%%%%%%%%%%%
% SECTION      ROTATIONAL MOTION JACOBIANS
%%%%%%%%%%%%%%%%%%%%%%%%%%%%%%%%%%%%%%%%%%%%%%%%%%%%%%%%%%%%%%%%%%%%%%%%
%%%%%%%%%%%%%%%%%%%%%%%%%%%%%%%%%%%%%%%%%%%%%%%%%%%%%%%%%%%%%%%%%%%%%%%%
%%%%%%%%%%%%%%%%%%%%%%%%%%%%%%%%%%%%%%%%%%%%%%%%%%%%%%%%%%%%%%%%%%%%%%%%

\section{Rotational Motion Jacobians}\label{sec:RigidBody_rotation_calculus_jacobians}

Lie group Jacobians are introduced in section \ref{sec:algebra_lie_jacobians} based on the right and left Lie group derivatives of section \ref{subsec:algebra_lie_derivatives}, and in this section are customized for the \nm{\mathbb{SO}\lrp{3}} case, with table \ref{tab:RigidBody_rotation_jacobians} representing the particularization of table \ref{tab:algebra_lie_jacobians} to the case of rigid body rotations. The various Jacobians listed in table \ref{tab:RigidBody_rotation_jacobians} have been obtained by means of the chain rule, the expressions already introduced in this document, and those of section \ref{sec:algebra_lie}. Note that although in many cases the results include the rotation matrix, all Jacobians are generic and do not depend on the specific \nm{\mathbb{SO}\lrp{3}} parameterization.

In addition to the adjoint matrix, two other Jacobians are of particular importance as they appear repeatedly in table \ref{tab:RigidBody_rotation_jacobians}. These are the right and left Jacobians of the capitalized exponential function, also known as simply the \emph{right Jacobian} \nm{\vec J_R\lrp{\vec r}} and the \emph{left Jacobian} \nm{\vec J_L\lrp{\vec r}}, and they evaluate the variation of the \nm{\mathfrak{so}\lrp{3}} tangent space provided by the output of the \nm{Exp\lrp{\vec r}} map (locally for \nm{\vec J_R} and globally for \nm{\vec J_L}) while moving along the \nm{\mathbb{SO}\lrp{3}} manifold with respect to the (Euclidean) variations within the original tangent space provided by \nm{\vec r}. Their closed forms as well as those of their inverses are included in table \ref{tab:RigidBody_rotation_jacobians}, and have been obtained from \cite{Chirikjian2012}; they verify that \nm{\vec J_L\lrp{\vec r} = \vec J_R^T\lrp{\vec r}}, and \nm{\vec J_L^{-1}\lrp{\vec r} = \vec J_R^{-T}\lrp{\vec r}}.

It is also worth noting the special importance of the \nm{\vec J_{\ds{+ \; \vec r}}^{\ds{- \; g_{Exp\lrp{\vec r}*}(\vec v)}}} Jacobian present at the bottom of table \ref{tab:RigidBody_rotation_jacobians}, which represents the derivative of a rotated vector with respect to perturbations in the Euclidean tangent space (not on the curved manifold) that generates the rotation, as it enables tangent space optimization by calculus methods designed exclusively for Euclidean spaces. 
\renewcommand{\arraystretch}{1.5} % increase row height
\begin{center}
\begin{tabular}{lcccll}
	\hline
	Jacobian & & Table \ref{tab:algebra_lie_jacobians} & & \multicolumn{1}{c}{Expression} & Size \\
	\hline
	\nm{\vec J_{\ds{\oplus \; \mathcal R}}^{\ds{\ominus \; \mathcal R}^{-1}}}       				& = & \nm{- \vec{Ad}_{\mathcal R}} 			& = & \nm{- \vec R}				 	& \nm{\in \mathbb R^{3x3}} \\ 
	\nm{\vec J_{\ds{\boxplus \; \mathcal R}}^{\ds{\boxminus \; \mathcal R}^{-1}}}         		& = & \nm{- \vec{Ad}_{\mathcal R}^{-1}}		& = & \nm{- \vec R^T}				& \nm{\in \mathbb R^{3x3}} \\ 

	\nm{\vec J_{\ds{\oplus \; \mathcal R}}^{\ds{\ominus \; \mathcal R \circ \mathcal S}}}		& = & \nm{\vec{Ad}_{\mathcal S}^{-1}}	   	& = & \nm{\vec R_{\mathcal S}^T} 	& \nm{\in \mathbb R^{3x3}} \\ 
	\nm{\vec J_{\ds{\boxplus \; \mathcal R}}^{\ds{\boxminus \; \mathcal R \circ \mathcal S}}}	& = & \nm{\vec I}	 						& = & \nm{\vec{I}_{3x3}}		 	& \nm{\in \mathbb R^{3x3}} \\ 
	\nm{\vec J_{\ds{\oplus \; \mathcal S}}^{\ds{\ominus \; \mathcal R \circ \mathcal S}}}		& = & \nm{\vec I}							& = & \nm{\vec{I}_{3x3}}           	& \nm{\in \mathbb R^{3x3}} \\ 
	\nm{\vec J_{\ds{\boxplus \; \mathcal S}}^{\ds{\boxminus \; \mathcal R \circ \mathcal S}}}	& = & \nm{\vec{Ad}_{\mathcal R}}			& = & \nm{\vec R_{\mathcal R}}     	& \nm{\in \mathbb R^{3x3}} \\ 

	\nm{\vec J_{\ds{+ \; \vec r}}^{\ds{\ominus \; Exp\lrp{\vec r}}}}								& = & \nm{\vec J_R\lrp{\vec r}} 				& = & \nm{\vec I_3 - \frac{1 - \cos \|\vec r\|}{\|\vec r\|^2} \ \rskew + \frac{\|\vec r\| - \sin \|\vec r\|}{\|\vec r\|^3} \ \rskew^2} & \nm{\in \mathbb R^{3x3}} \\ 
	\nm{\vec J_R^{-1}\lrp{\vec r}}																	&   & 										& = & \nm{\vec I_3 + \frac{\rskew}{2} + \lrp{\frac{1}{\|\vec r\|^2} - \frac{1 + \cos \|\vec r\|}{2 \, \|\vec r\| \, \sin \|\vec r\|}} \, \rskew^2} &\nm{\in \mathbb R^{3x3}} \\ 
	\nm{\vec J_{\ds{+ \; \vec r}}^{\ds{\boxminus \; Exp\lrp{\vec r}}}} 							& = & \nm{\vec J_L\lrp{\vec r}} 					& = & \nm{\vec I_3 + \frac{1 - \cos \|\vec r\|}{\|\vec r\|^2} \ \rskew + \frac{\|\vec r\| - \sin \|\vec r\|}{\|\vec r\|^3} \ \rskew^2} & \nm{\in \mathbb R^{3x3}} \\ 
	\nm{\vec J_L^{-1}\lrp{\vec r}}																	&   & 										& = & \nm{\vec I_3 - \frac{\rskew}{2} + \lrp{\frac{1}{\|\vec r\|^2} - \frac{1 + \cos \|\vec r\|}{2 \, \|\vec r\| \, \sin \|\vec r\|}} \, \rskew^2} & \nm{\in \mathbb R^{3x3}} \\ 

	\nm{\vec J_{\ds{\oplus \; \mathcal R}}^{\ds{- \; Log\lrp{\mathcal R}}}}        				& = & \nm{\vec J_R^{-1}\big(Log\lrp{\mathcal R}\big)} & & & \nm{\in \mathbb R^{3x3}} \\  
	\nm{\vec J_{\ds{\boxplus \; \mathcal R}}^{\ds{- \; Log\lrp{\mathcal R}}}}     				& = & \nm{\vec J_L^{-1}\big(Log\lrp{\mathcal R}\big)} & & & \nm{\in \mathbb R^{3x3}} \\  
	
	\nm{\vec J_{\ds{\oplus \; \mathcal R}}^{\ds{\ominus \; \mathcal R \oplus \vec r}}}				& = & \nm{\vec{Ad}_{Exp\lrp{\vec r}}^{-1}}	& = & \nm{\vec R^T\lrp{\vec r}} & \nm{\in \mathbb R^{3x3}} \\ 
	\nm{\vec J_{\ds{\boxplus \; \mathcal R}}^{\ds{\boxminus \; \vec r \boxplus \mathcal R}}}			& = & \nm{\vec{Ad}_{Exp\lrp{\vec r}}}		& = & \nm{\vec R\lrp{\vec r}}   & \nm{\in \mathbb R^{3x3}} \\ 
	\nm{\vec J_{\ds{+ \; \vec r}}^{\ds{\ominus \; \mathcal R \oplus \vec r}}}         				& = & \nm{\vec J_R\lrp{\vec r}}           	   	& & & \nm{\in \mathbb R^{3x3}} \\ 	
	\nm{\vec J_{\ds{+ \; \vec r}}^{\ds{\boxminus \; \vec r \boxplus \mathcal R}}}     				& = & \nm{\vec J_L\lrp{\vec r}}           	   	& & & \nm{\in \mathbb R^{3x3}} \\ 	

	\nm{\vec J_{\ds{\oplus \; \mathcal R}}^{\ds{- \; \mathcal S \ominus \mathcal R}}} 	      		& = & \nm{- \vec J_L^{-1}\lrp{\mathcal S \ominus \mathcal R}}	& & & \nm{\in \mathbb R^{3x3}} \\ 	
	\nm{\vec J_{\ds{\boxplus \; \mathcal R}}^{\ds{- \; \mathcal S \boxminus \mathcal R}}} 	   		& = & \nm{- \vec J_R^{-1}\lrp{\mathcal S \boxminus \mathcal R}}	& & & \nm{\in \mathbb R^{3x3}} \\ 	
	\nm{\vec J_{\ds{\oplus \; \mathcal S}}^{\ds{- \; \mathcal S \ominus \mathcal R}}}    	    	& = & \nm{\vec J_R^{-1}\lrp{\mathcal S \ominus \mathcal R}}		& & & \nm{\in \mathbb R^{3x3}} \\
	\nm{\vec J_{\ds{\boxplus \; \mathcal S}}^{\ds{- \; \mathcal S \boxminus \mathcal R}}}  	    	& = & \nm{\vec J_L^{-1}\lrp{\mathcal S \boxminus \mathcal R}}	& & & \nm{\in \mathbb R^{3x3}} \\  	
	
	\nm{\vec J_{\ds{\oplus \; \mathcal R}}^{\ds{- \; g_{\mathcal R*}(\vec v)}}}          	 		&   & 								 	& = & \nm{- \vec R \, \widehat{\vec v}}    & \nm{\in \mathbb R^{3x3}} \\ 
	\nm{\vec J_{\ds{\boxplus \; \mathcal R}}^{\ds{- \; g_{\mathcal R*}(\vec v)}}}      				&   & 								 	& = & \nm{- \lrp{\vec R \, \vec v}^\wedge} & \nm{\in \mathbb R^{3x3}} \\ 
	\nm{\vec J_{\ds{+ \; \vec v}}^{\ds{- \; g_{\mathcal R*}(\vec v)}}}               				&   & 									& = & \nm{\vec R}                          & \nm{\in \mathbb R^{3x3}} \\  
	
	\nm{\vec J_{\ds{\oplus \; \mathcal R}}^{\ds{- \; g_{\mathcal R*}^{-1}(\vec v)}}}          		&   & 								 	& = & \nm{\lrp{\vec R^T \, \vec v}^\wedge} & \nm{\in \mathbb R^{3x3}} \\ 
	\nm{\vec J_{\ds{\boxplus \; \mathcal R}}^{\ds{- \; g_{\mathcal R*}^{-1}(\vec v)}}}  			&   & 								 	& = & \nm{\vec R^T \, \vec v}              & \nm{\in \mathbb R^{3x3}} \\ 
	\nm{\vec J_{\ds{+ \; \vec v}}^{\ds{- \; g_{\mathcal R*}^{-1}(\vec v)}}}             			&   & 									& = & \nm{\vec R^T} 	                   & \nm{\in \mathbb R^{3x3}} \\  	
	\hline
\end{tabular}
\end{center}
	
\begin{center}
\begin{tabular}{lcccll}
	\hline
	Jacobian & & Table \ref{tab:algebra_lie_jacobians} & & \multicolumn{1}{c}{Expression} & Size \\
	\hline
	\nm{\vec J_{\ds{\oplus \; \mathcal R}}^{\ds{- \; \vec{Ad}_{\mathcal R}(\vec \omega)}}} 			&   & 									& = & \nm{- \vec R \, \widehat{\vec \omega}}    & \nm{\in \mathbb R^{3x3}} \\ 
	\nm{\vec J_{\ds{\boxplus \; \mathcal R}}^{\ds{- \; \vec{Ad}_{\mathcal R}(\vec \omega)}}}		&   & 									& = & \nm{- \lrp{\vec R \, \vec \omega}^\wedge} & \nm{\in \mathbb R^{3x3}} \\ 
	\nm{\vec J_{\ds{+ \; \vec \omega}}^{\ds{- \; \vec{Ad}_{\mathcal R}(\vec \omega)}}}				& = & \nm{\vec{Ad}_{\mathcal R}}		& = & \nm{\vec R}                               & \nm{\in \mathbb R^{3x3}} \\ 

	\nm{\vec J_{\ds{\oplus \; \mathcal R}}^{\ds{- \; \vec{Ad}_{\mathcal R}^{-1}(\vec \omega)}}} 	&   & 									& = & \nm{\lrp{\vec R^T \, \vec \omega}^\wedge} & \nm{\in \mathbb R^{3x3}} \\ 
	\nm{\vec J_{\ds{\boxplus \; \mathcal R}}^{\ds{- \; \vec{Ad}_{\mathcal R}^{-1}(\vec \omega)}}}	&   & 									& = & \nm{\vec R^T \, \vec \omega}              & \nm{\in \mathbb R^{3x3}} \\ 	
	
	\nm{\vec J_{\ds{+ \; \vec \omega}}^{\ds{- \; \vec{Ad}_{\mathcal R}^{-1}(\vec \omega)}}}			& = & \nm{\vec{Ad}_{\mathcal R}^{-1}}	& = & \nm{\vec R^T} 							& \nm{\in \mathbb R^{3x3}} \\ 	
	
	\nm{\vec J_{\ds{+ \; \vec r}}^{\ds{- \; g_{Exp\lrp{\vec r}*}(\vec v)}}}							&   & 									& = & \nm{- \vec R\lrp{\vec r} \, \widehat{\vec v} \, \vec J_R\lrp{\vec r}} = \nm{- \big(\vec R\lrp{\vec r} \, \vec v\big)^\wedge \, \vec J_L\lrp{\vec r}} & \nm{\in \mathbb R^{3x3}} \\ 
	\nm{\vec J_{\ds{+ \; \vec r}}^{\ds{- \; g_{Exp\lrp{\vec r}*}^{-1}(\vec v)}}}					&   & 									& = & \nm{- \vec R^T\lrp{\vec r} \, \widehat{\vec v} \, \vec J_L\lrp{\vec r}} = \nm{\big(\vec R^T\lrp{\vec r} \, \vec v\big)^\wedge \, \vec J_R\lrp{\vec r}} & \nm{\in \mathbb R^{3x3}} \\ 
	\hline
\end{tabular}
\end{center}
\captionof{table}{Rotational motion Jacobians} \label{tab:RigidBody_rotation_jacobians}
\renewcommand{\arraystretch}{1.0} % reset row height

%%%%%%%%%%%%%%%%%%%%%%%%%%%%%%%%%%%%%%%%%%%%%%%%%%%%%%%%%%%%%%%%%%%%%%%%
%%%%%%%%%%%%%%%%%%%%%%%%%%%%%%%%%%%%%%%%%%%%%%%%%%%%%%%%%%%%%%%%%%%%%%%%
%%%%%%%%%%%%%%%%%%%%%%%%%%%%%%%%%%%%%%%%%%%%%%%%%%%%%%%%%%%%%%%%%%%%%%%%
% SECTION      ROTATIONAL MOTION DISCRETE INTEGRATION
%%%%%%%%%%%%%%%%%%%%%%%%%%%%%%%%%%%%%%%%%%%%%%%%%%%%%%%%%%%%%%%%%%%%%%%%
%%%%%%%%%%%%%%%%%%%%%%%%%%%%%%%%%%%%%%%%%%%%%%%%%%%%%%%%%%%%%%%%%%%%%%%%
%%%%%%%%%%%%%%%%%%%%%%%%%%%%%%%%%%%%%%%%%%%%%%%%%%%%%%%%%%%%%%%%%%%%%%%%

\section{Rotational Motion Discrete Integration}\label{sec:RigidBody_rotation_integration}

The discrete integration with time of an element of a Lie group based on its Lie algebra is discussed in detail in section \ref{sec:algebra_integration}, which includes expressions for the Euler, Heun and Runge-Kutta methods. In the case of rotational motion, the state vector includes the rotation element \nm{\mathcal R \in \mathbb{SO}\lrp{3}} and its angular velocity \nm{\vec \omega \in \mathbb{R}^3} contained in the tangent space, viewed either in the local (\nm{\wNBB}) or global (\nm{\wNBN}) frames. The Euler method expressions equivalent to (\ref{eq:algebra_integration_comp_X_euler}) and (\ref{eq:algebra_integration_comp_X_euler_left}) are shown below. Expressions for other integration schemes can easily be derived from those in section \ref{sec:algebra_integration}:
\begin{eqnarray}
\nm{\mathcal R_{k+1}} & \nm{\approx} & \nm{\mathcal R_k \oplus \lrsb{\Delta t \ \vec \omega_{{\sss NB}k}^{\sss B}} = \mathcal R_k \circ Exp\lrp{\Delta t \ \vec \omega_{{\sss NB}k}^{\sss B}}} \label{eq:SO3_integration_comp_X_euler} \\
\nm{\mathcal R_{k+1}} & \nm{\approx} & \nm{\lrsb{\Delta t \ \vec \omega_{{\sss NB}k}^{\sss N}} \boxplus \mathcal R_k = Exp\lrp{\Delta t \ \vec \omega_{{\sss NB}k}^{\sss N}} \circ \mathcal R_k} \label{eq:SO3_integration_comp_X_euler_left}
\end{eqnarray}

%%%%%%%%%%%%%%%%%%%%%%%%%%%%%%%%%%%%%%%%%%%%%%%%%%%%%%%%%%%%%%%%%%%%%%%%
%%%%%%%%%%%%%%%%%%%%%%%%%%%%%%%%%%%%%%%%%%%%%%%%%%%%%%%%%%%%%%%%%%%%%%%%
%%%%%%%%%%%%%%%%%%%%%%%%%%%%%%%%%%%%%%%%%%%%%%%%%%%%%%%%%%%%%%%%%%%%%%%%
% SECTION      ROTATIONAL MOTION GAUSS-NEWTON OPTIMIZATION
%%%%%%%%%%%%%%%%%%%%%%%%%%%%%%%%%%%%%%%%%%%%%%%%%%%%%%%%%%%%%%%%%%%%%%%%
%%%%%%%%%%%%%%%%%%%%%%%%%%%%%%%%%%%%%%%%%%%%%%%%%%%%%%%%%%%%%%%%%%%%%%%%
%%%%%%%%%%%%%%%%%%%%%%%%%%%%%%%%%%%%%%%%%%%%%%%%%%%%%%%%%%%%%%%%%%%%%%%%

\section{Rotational Motion Gauss-Newton Optimization}\label{sec:RigidBody_rotation_gauss_newton}

The minimization by means of the Gauss-Newton iterative method of the Euclidean norm of a nonlinear function whose input is a Lie group element is presented in section \ref{sec:algebra_gradient_descent}. In the case of rotational motion, the resulting expressions for perturbations \nm{\Delta \vec r_{\sss NB}^{\sss N} \in \mathfrak{so}\lrp{3}} to an input rotation \nm{\mathcal R \in \mathbb{SO}\lrp{3}} viewed in the global frame \nm{\FN} are shown in (\ref{eq:SO3_gauss_newton_iterative_left}) and (\ref{eq:SO3_gauss_newton_solution_left}), which are equivalent to the generic (\ref{eq:algebra_gradient_descent_iterative_lie_left}) and (\ref{eq:algebra_gradient_descent_solution_lie_left}). Refer to section \ref{sec:algebra_gradient_descent} for the meaning of the function Jacobian \nm{\vec J} and to section \ref{sec:RigidBody_rotation_calculus_jacobians} for that of the left Jacobian \nm{\vec J_L}.
\begin{eqnarray}
\nm{\mathcal R_{k+1}} & \nm{\longleftarrow} & \nm{\Delta \vec r_{{\sss NB}k}^{\sss N} \boxplus \mathcal R_k = \Delta \vec r_{{\sss NB}k}^{\sss N} \circ Exp\lrp{\vec r_{{\sss NB}k}}} \label{eq:SO3_gauss_newton_iterative_left} \\
\nm{\Delta \vec r_{{\sss NB}k}^{\sss N}} & = & \nm{- \lrsb{\vec J_{Lk}^{-T} \, \vec J_k^T \, \vec J_k \, \vec J_{Lk}^{-1}}^{-1} \, \vec J_{Lk}^{-T} \, \vec J_k^T \, \vec{\mathcal E}_k} \label{eq:SO3_gauss_newton_solution_left} 
\end{eqnarray}    

If the perturbation is viewed in the local frame \nm{\FB}, (\ref{eq:algebra_gradient_descent_iterative_lie_right}) and (\ref{eq:algebra_gradient_descent_solution_lie_right}) are customized as follows, making use of the right Jacobian \nm{\vec J_R} defined in section \ref{sec:RigidBody_rotation_calculus_jacobians}:
\begin{eqnarray}
\nm{\mathcal R_{k+1}} & \nm{\longleftarrow} & \nm{\mathcal R_k \oplus \Delta \vec r_{{\sss NB}k}^{\sss B} = Exp\lrp{\vec r_{{\sss NB}k}} \circ \Delta \vec r_{{\sss NB}k}^{\sss B}} \label{eq:SO3_gauss_newton_iterative_right} \\
\nm{\Delta \vec r_{{\sss NB}k}^{\sss B}} & = & \nm{- \lrsb{\vec J_{Rk}^{-T} \, \vec J_k^T \, \vec J_k \, \vec J_{Rk}^{-1}}^{-1} \, \vec J_{Rk}^{-T} \, \vec J_k^T \, \vec{\mathcal E}_k} \label{eq:SO3_gauss_newton_solution_right} 
\end{eqnarray}    

%%%%%%%%%%%%%%%%%%%%%%%%%%%%%%%%%%%%%%%%%%%%%%%%%%%%%%%%%%%%%%%%%%%%%%%%
%%%%%%%%%%%%%%%%%%%%%%%%%%%%%%%%%%%%%%%%%%%%%%%%%%%%%%%%%%%%%%%%%%%%%%%%
%%%%%%%%%%%%%%%%%%%%%%%%%%%%%%%%%%%%%%%%%%%%%%%%%%%%%%%%%%%%%%%%%%%%%%%%
% SECTION      ROTATIONAL MOTION STATE ESTIMATION
%%%%%%%%%%%%%%%%%%%%%%%%%%%%%%%%%%%%%%%%%%%%%%%%%%%%%%%%%%%%%%%%%%%%%%%%
%%%%%%%%%%%%%%%%%%%%%%%%%%%%%%%%%%%%%%%%%%%%%%%%%%%%%%%%%%%%%%%%%%%%%%%%
%%%%%%%%%%%%%%%%%%%%%%%%%%%%%%%%%%%%%%%%%%%%%%%%%%%%%%%%%%%%%%%%%%%%%%%%

\section{Rotational Motion State Estimation}\label{sec:RigidBody_rotation_SS}

The adaptation of the \hypertt{EKF} state estimation introduced in section \ref{sec:SS} to the case in which Lie group elements and their velocities are present is discussed in detail in section \ref{sec:algebra_SS}. For rotational motion with local perturbations, it is necessary to replace \nm{\mathcal X \in \mathcal G} by \nm{\mathcal R \in \mathbb{SO}\lrp{3}}, \nm{\Delta \vec \tau^{\mathcal X} \in T_{\mathcal X}\mathcal G} by \nm{\Delta \vec r^{\sss B} \in \ \mathfrak{so}\lrp{3}}, \nm{\vec v^{\mathcal X} \in \mathbb{R}^m} by \nm{\vec \omega^{\sss B} \in \mathbb{R}^3}, \nm{\vec C_{{\mathcal {XX}}}^{\mathcal X} \in \mathbb{R}^{mxm}} by \nm{\vec C_{{\mathcal {RR}}}^{\sss B} \in \mathbb{R}^{3x3}}, and \nm{\vec J_{\ds{\oplus \; \mathcal X}}^{\ds{\ominus \; \mathcal X \oplus \vec \tau}}} by \nm{\vec J_{\ds{\oplus \; \mathcal R}}^{\ds{\ominus \; \mathcal R \oplus \vec r}}}. The particularizations for global perturbations are similar.

%%%%%%%%%%%%%%%%%%%%%%%%%%%%%%%%%%%%%%%%%%%%%%%%%%%%%%%%%%%%%%%%%%%%%%%%
%%%%%%%%%%%%%%%%%%%%%%%%%%%%%%%%%%%%%%%%%%%%%%%%%%%%%%%%%%%%%%%%%%%%%%%%
%%%%%%%%%%%%%%%%%%%%%%%%%%%%%%%%%%%%%%%%%%%%%%%%%%%%%%%%%%%%%%%%%%%%%%%%
% SECTION     APPLICATIONS OF THE VARIOUS ROTATION MOTION REPRESENTATIONS
%%%%%%%%%%%%%%%%%%%%%%%%%%%%%%%%%%%%%%%%%%%%%%%%%%%%%%%%%%%%%%%%%%%%%%%%
%%%%%%%%%%%%%%%%%%%%%%%%%%%%%%%%%%%%%%%%%%%%%%%%%%%%%%%%%%%%%%%%%%%%%%%%
%%%%%%%%%%%%%%%%%%%%%%%%%%%%%%%%%%%%%%%%%%%%%%%%%%%%%%%%%%%%%%%%%%%%%%%%

\section{Applications of the Various Rotation Representations}\label{sec:RigidBody_rotation_applications}

This chapter discusses five different representations of the rotation or special orthogonal group \nm{\mathbb{SO}\lrp{3}}: the rotation matrix, the rotation vector, the unit quaternion, the half rotation vector, and the Euler angles. Although in theory all of them can be employed for each of the purposes described in this chapter, and the required expressions derived, each parameterization has its own advantages and disadvantages, being suited for certain purposes but not recommended for others. 
\begin{itemize}
\item The rotation matrix \nm{\vec R} is the most natural parameterization, possesses an easy to obtain inverse, and linear expressions for composition and rotation. It provides a clear connection with the tangent space, together with the exponential and logarithmic maps, \hypertt{SLERP}, and plus and minus operators, which are not complex. Its main drawbacks are the storage costs associated with its high dimension (9) and the expense involved in maintaining orthogonality if allowed to deviate from the manifold \cite{Shuster1993, Baritzhack1969}. Its high cost precludes its use to track the rotation over its manifold, although most implementations continuously compute it if the adjoint matrix or the Jacobian blocks are required.

\item The unit quaternion \nm{\vec q} is the preferred representation to track the rotation over its manifold, even if it is necessary to obtain the rotation matrix for the adjoint and Jacobian blocks. Its advantages with respect to the rotation matrix are its small dimension (4) and ease to maintain unitary if allowed to deviate from the manifold. Unit quaternions are the least natural of the rotation parameterizations, being necessary to convert to a different \nm{\mathbb{SO}\lrp{3}} representation for visualization \cite{Shuster1993}. While the inverse and concatenation are linear, the rotation action is bilinear, which presents a disadvantage with respect to the rotation matrix. Unit quaternion expressions are slightly more complex than those of the rotation matrix, and present a slightly less obvious connection with the tangent space because in fact they represent a double covering of \nm{\mathbb{SO}\lrp{3}} instead of \nm{\mathbb{SO}\lrp{3}} itself. 

\item The main advantage of the rotation vector \nm{\vec r} is that it belongs to the \nm{\mathfrak{so}\lrp{3}} tangent space while simultaneously being an \nm{\mathbb{SO}\lrp{3}} parameterization. It is hence indicated for those uses related with incremental rotation changes by means of the exponential map together with the plus and minus operators (periodically adding the perturbations to the unit quaternion tracking the rotation), such as discrete integration, optimization, and state estimation. The rotation vector norm is the most adequate metric for evaluating the rotated distance (or estimation error) between two rigid bodies. Although it benefits from its straightforward inverse, its geometric appeal, and its small dimension (3+1)\footnote{Strictly speaking the dimension is 3, although any usage requires computing the norm, which is often stored to accelerate the transformations.}, its usage for other applications is discouraged by its complex nonlinear kinematics, rotation action, and composition \cite{Shuster1993}, which are not shown in this chapter. 

\item The half rotation vector \nm{\vec h} is so similar (half) to the rotation vector that its usage is not recommended in order to avoid confusion. Its only real application as the tangent space of the unit quaternion is in practice solved by dividing the rotation vector by two when necessary.

\item The Euler angles \nm{\vec \phi} have a long history and a clear physical meaning, which makes them the best choice for attitude visualization, and constitute the only representation in which its dimension (3) coincides with that of the manifold. However, they are not recommended for any other usage because of the presence of discontinuities, together with complex and nonlinear expressions for inversion, composition, and rotation action \cite{Shuster1993}.
\end{itemize}

%% file: files_arxiv/ch05_motion.tex
\chapter{Motion of Rigid Bodies} \label{cha:Motion}

This chapter can be considered as a continuation of the analysis of the rotational motion of rigid bodies contained in chapter \ref{cha:Rotate}, in which its center of rotation \nm{\vec O_{\sss {CR}}} is not stationary but moves in the Euclidean space \nm{\mathbb{E}^3}. It follows a similar scheme, relying on Lie theory concepts discussed in sections \ref{sec:algebra_lie} and \ref{sec:algebra_lie_jacobians}. Table \ref{tab:Motion_lie_comparison} provides a comparison between the generic nomenclature employed in chapter \ref{cha:Algebra} and their rigid body motion equivalents. The different representations discussed in this chapter are summarized in Table \ref{tab:Motion_summary}.
\begin{center}
\begin{tabular}{lcclcc}
	\hline
	Concept    & Lie Theory & Motion & Concept    & Lie Theory & Motion \\
	\hline
	Lie group           & \nm{\mathcal G} & \nm{\mathbb{SE}\lrp{3}}		   	& Lie group element     & \nm{\mathcal X, \, \mathcal Y}       & \nm{\mathcal M, \, \mathcal N} \\ 
	Concatenation       & \nm{\circ}      & \nm{\circ}                	& Lie algebra           & \nm{\mathfrak{m}}     & \nm{\mathfrak{se}\lrp{3}} \\ 
	Identity            & \nm{\mathcal E} & \nm{\mathcal {I_M}}         & Inverse               & \nm{\mathcal X^{-1}}  & \nm{\mathcal M^{-1}} \\
	Velocity            & \nm{\vec v}     & \nm{\vec \xi}               & Tangent element       & \nm{\vec \tau}        & \nm{\vec \tau} \\
	Local frame         & \nm{\mathcal X} & B                           & Global frame          & \nm{\mathcal E}       & E \\
	Point action        & \nm{g_{\mathcal X}()} & \nm{\vec g_{\mathcal M}(\vec p)} & Vector action      & \nm{g_{\mathcal X}()} & \nm{\vec g_{\mathcal M*}(\vec v)} \\
	Adjoint             & \nm{\vec{Ad}_{\mathcal X}\lrp{\vec \tau^{\wedge}}} & \nm{\vec{Ad}_{\mathcal M}\lrp{\vec \tau^{\wedge}}} & Adjoint matrix & \nm{\vec{Ad}_{\mathcal X} \, \vec \tau} & \nm{\vec{Ad}_{\mathcal M} \, \vec \tau} \\
	\hline
\end{tabular}
\end{center}
\captionof{table}{Comparison between generic Lie elements and those of rigid body motions} \label{tab:Motion_lie_comparison}

This section begins with an introduction to rigid body motion in section \ref{sec:RigidBody_motion}, followed by a description of the different rigid body motion Lie group representations: the affine representation (section \ref{sec:RigidBody_motion_affine}), the homogeneous matrix (section \ref{sec:RigidBody_motion_homogeneous}), the transform vector (section \ref{sec:RigidBody_motion_transform_vector}), the unit dual quaternion (section \ref{sec:RigidBody_motion_unit_dual_quaternion}), the half transform vector (section \ref{sec:RigidBody_motion_halftransform_vector}), and the screw (section \ref{sec:RigidBody_motion_screw}). Algebraic operations on rigid body motions are introduced in section \ref{sec:RigidBody_motion_algebra}, such as powers, linear interpolation, and the plus and minus operators. Section \ref{sec:RigidBody_motion_calculus_derivatives} presents the motion time derivative that leads to the definition of the twist or motion velocity in the tangent space. The velocity of the rigid body points is discussed in section \ref{sec:RigidBody_motion_velocity}, followed by the adjoint map in section \ref{sec:RigidBody_motion_adjoint}, which transforms elements of the tangent space while the motion progresses on its manifold, and by an analysis of uncertainty and covariances applied to rigid body motion (section \ref{sec:RigidBody_motion_covariance}). An extensive analysis of the rigid body motion Jacobians is presented in section \ref{sec:RigidBody_motion_calculus_jacobians}. Sections \ref{sec:RigidBody_motion_integration}, \ref{sec:RigidBody_motion_gauss_newton}, and \ref{sec:RigidBody_motion_SS} apply the discrete integration of Lie groups, the Gauss-Newton optimization of Lie group functions, and the state estimation of Lie groups contained in sections \ref{sec:algebra_integration}, \ref{sec:algebra_gradient_descent}, and \ref{sec:algebra_SS} to the case of rigid body motions. Finally, the advantages and disadvantages of each motion representation are discussed in section \ref{sec:RigidBody_motion_applications}.

%%%%%%%%%%%%%%%%%%%%%%%%%%%%%%%%%%%%%%%%%%%%%%%%%%%%%%%%%%%%%%%%%%%%%%%%
%%%%%%%%%%%%%%%%%%%%%%%%%%%%%%%%%%%%%%%%%%%%%%%%%%%%%%%%%%%%%%%%%%%%%%%%
%%%%%%%%%%%%%%%%%%%%%%%%%%%%%%%%%%%%%%%%%%%%%%%%%%%%%%%%%%%%%%%%%%%%%%%%
% SECTION      SPECIAL EUCLIDEAN (LIE) GROUP
%%%%%%%%%%%%%%%%%%%%%%%%%%%%%%%%%%%%%%%%%%%%%%%%%%%%%%%%%%%%%%%%%%%%%%%%
%%%%%%%%%%%%%%%%%%%%%%%%%%%%%%%%%%%%%%%%%%%%%%%%%%%%%%%%%%%%%%%%%%%%%%%%
%%%%%%%%%%%%%%%%%%%%%%%%%%%%%%%%%%%%%%%%%%%%%%%%%%%%%%%%%%%%%%%%%%%%%%%%

\section{Special Euclidean (Lie) Group}\label{sec:RigidBody_motion}

A rigid body can be represented with a Cartesian frame attached to any of its points (the origin), with the basis vectors \nm{\vec e_1}, \nm{\vec e_2}, and \nm{\vec e_3} being simply unit vectors along the main axes. Rigid body motions can be combined and reversed, complying with the algebraic concept of group, but are not endowed with a metric, so they are not part of a metric or Euclidean space (section \ref{sec:algebra_structures}). They do however comply with the axioms of a Lie group (section \ref{sec:algebra_lie}), and hence the set of rigid body motions together with the operation of motion concatenation comprises \nm{\langle \mathbb{SE}\lrp{3}, \circ \rangle}, known as the \emph{special Euclidean group} of \nm{\mathbb{R}^3} \cite{Soatto2001}, where its elements are denoted by \nm{\mathcal M}, the identify motion by \nm{\mathcal {I_M}}, and the inverse by \nm{\mathcal M^{-1}}. The group has two main actions, which are the motion of points \nm{\lrb{\vec g() : \mathbb{SE}^3 \times \mathbb{R}^3 \rightarrow \mathbb{R}^3 \ | \ \vec p \rightarrow \vec g_{\mathcal M}\lrp{\vec p}}} and that of vectors \nm{\lrb{\vec g_*() : \mathbb{SE}^3 \times \mathbb{R}^3 \rightarrow \mathbb{R}^3 \ | \ \vec v \rightarrow \vec g_{\mathcal M*}\lrp{\vec v}}}.
\begin{center}
\begin{tabular}{lccc}
	\hline
	Representation & Symbol                                   & Structure    & Space \\
	\hline
	Affine representation   & \nm{(\mathcal R, \, \vec T)}                      & \nm{\mathbb{SO}\lrp{3}} \& free 3-vector & \nm{\mathbb{SE}\lrp{3}}  \\
	Homogeneous matrix      & \nm{\vec M}                                       & 4x4 matrix (\ref{eq:SE3_homogeneous_SE3}) & \nm{\mathbb{SE}\lrp{3}} \\
	Twist					& \nm{\vec \xi^\wedge}                              & 4x4 matrix (\ref{eq:SE3_twist_expression2}) & \nm{\mathfrak{se}\lrp{3}} \\
	                        & \nm{\vec \xi = \lrsb{\vec \nu \ \ \vec \omega}^T} & free 6-vector           &             \\
	Transform vector		& \nm{\vec \tau^\wedge}                             & 4x4 matrix (\ref{eq:SE3_twist_expression2}) & \nm{\mathbb{SE}\lrp{3}} \& \nm{\mathfrak{se}\lrp{3}} \\
	                        & \nm{\vec \tau = \vec \xi \, t = \lrsb{\vec s \ \ \vec r}^T = \lrsb{ \vec k \, \rho \ \ \vec n \, \phi}^T} & free 6-vector   &    \\
	Unit dual quaternion    & \nm{\vec \zeta}                                   & unit dual quaternion  & \nm{\mathbb{SE}\lrp{3}}     \\
	Half twist			    & \nm{\vec \Upsilon^\wedge}                         & pure dual quaternion & \nm{\mathfrak{se}\lrp{3}} \\
	                        & \nm{\vec \Upsilon = \vec \xi / 2}                 & free 6-vector        &   \\
	Half transform vector	& \nm{\vec \Psi^\wedge}                             & pure dual quaternion & \nm{\mathbb{SE}\lrp{3}} \& \nm{\mathfrak{se}\lrp{3}} \\
	                        & \nm{\vec \Psi = \vec \Upsilon \, t =  \vec \xi \, t / 2 = \vec \tau / 2} & free 6-vector    &           \\
	Screw					& \nm{\vec S^\wedge = (\phi^\diamond / 2, \, \vec{nm}^\diamond)} & dual number \& dual vector & \nm{\mathbb{SE}\lrp{3}} \& \nm{\mathfrak{se}\lrp{3}} \\
	                        & \nm{\vec S = \lrsb{\vec n \ \ \vec m \ \ h \ \ \phi}^T} & 8-vector & \\
	\hline
\end{tabular}
\end{center}
\captionof{table}{Summary of rigid body motion representations} \label{tab:Motion_summary}

The movements of rigid bodies are introduced in section \ref{sec:RigidBody_bases} as orthogonal transformations, this is, those that preserve orthogonality and handedness. 
\begin{itemize} 
\item Norm: \nm{\|\vec g_{\mathcal M*}\lrp{\vec v}\| = \|\vec v\|, \forall \, \vec v \in \mathbb{R}^3}
\item Cross product: \nm{\vec g_{\mathcal M*}\lrp{\vec u} \times \vec g_{\mathcal M*}\lrp{\vec v} = \vec g_{\mathcal M*}\lrp{\vec u \times \vec v}, \forall \, \vec u, \vec v \in \mathbb{R}^3}
\end{itemize}

It is also worth noting the relationship between the motions of vectors and points:
\neweq{\vec g_{\mathcal M*}\lrp{\vec v} = \vec g_{\mathcal M*}\lrp{\vec q - \vec p} = \vec g_{\mathcal M}\lrp{\vec q} - \vec g_{\mathcal M}\lrp{\vec p}}{eq:Motion_maps}

The \nm{\mathbb{SE}\lrp{3}} analysis below adopts the convention introduced in section \ref{sec:algebra_lie}, in which all actions, including concatenation \nm{\lrb{\circ : \mathbb{SE}\lrp{3} \times \mathbb{SE}\lrp{3} \rightarrow \mathbb{SE}\lrp{3}}}, transform elements viewed in the local or body frame \nm{F_{\sss B} = \{\OB, \vec b_1, \vec b_2, \vec b_3\}} into elements viewed in the global or spatial frame \nm{\FE = \{\OECEF, \vec e_1, \vec e_2, \vec e_3\}} \nm{= \{\vec g_{\mathcal M}\lrp{\OB}, \vec g_{\mathcal M*}\lrp{\vec b_1}, \vec g_{\mathcal M*}\lrp{\vec b_2}, \vec g_{\mathcal M*}\lrp{\vec b_3}\}}\footnote{In contrast with the case of rotational motion described in chapter \ref{cha:Rotate}, the spatial frame is now named E as it usually corresponds to the \hypertt{ECEF} frame. The \hypertt{NED} case does not apply to this case as it shares origin with the body frame. The Earth Centered Earth Fixed or \hypertt{ECEF} frame is centered at the Earth center of mass, with \nm{\iEiii} pointing towards the geodetic North along the Earth rotation axis, \nm{\iEi} contained in both the Equator and zero longitude planes, and \nm{\iEii} orthogonal to \nm{\iEi} and \nm{\iEiii} forming a right handed system.}, which overlap each other before the motion takes place.

%%%%%%%%%%%%%%%%%%%%%%%%%%%%%%%%%%%%%%%%%%%%%%%%%%%%%%%%%%%%%%%%%%%%%%%%
%%%%%%%%%%%%%%%%%%%%%%%%%%%%%%%%%%%%%%%%%%%%%%%%%%%%%%%%%%%%%%%%%%%%%%%%
%%%%%%%%%%%%%%%%%%%%%%%%%%%%%%%%%%%%%%%%%%%%%%%%%%%%%%%%%%%%%%%%%%%%%%%%
% SECTION      AFFINE REPRESENTATION
%%%%%%%%%%%%%%%%%%%%%%%%%%%%%%%%%%%%%%%%%%%%%%%%%%%%%%%%%%%%%%%%%%%%%%%%
%%%%%%%%%%%%%%%%%%%%%%%%%%%%%%%%%%%%%%%%%%%%%%%%%%%%%%%%%%%%%%%%%%%%%%%%
%%%%%%%%%%%%%%%%%%%%%%%%%%%%%%%%%%%%%%%%%%%%%%%%%%%%%%%%%%%%%%%%%%%%%%%%

\section{Affine Representation}\label{sec:RigidBody_motion_affine}

The motion of a rigid body can always be divided into a rotation plus a translation, in which the point motion action responds to:
\neweq{\vec g_{\mathcal M}\lrp{\vec p} = \vec g_{\mathcal R}\lrp{\vec p} + \vec T} {eq:SE3_affine_transform} 

where \nm{\vec g_{\mathcal R}} is the point rotation action discussed in chapter \ref{cha:Rotate}, the point \nm{\vec p} is viewed in the local or body frame, and \nm{\vec T} represents the vector going from the origin of the global or spatial frame to that of the body frame, viewed in the global frame. Any \nm{\mathbb{SO}\lrp{3}} representation can be employed for the above expression, but the rotation matrix (section \ref{sec:RigidBody_rotation_dcm}) and the unit quaternion (section \ref{sec:RigidBody_rotation_rodrigues}) are the most common:
\begin{eqnarray}
\nm{\pE} & = & \nm{\REB \, \pB + \TEBE} \label{eq:SE3_affine_dcm_transform} \\ 
\nm{\pE} & = & \nm{\qEB \otimes \pB \otimes \qEBast + \TEBE} \label{eq:SE3_affine_quat_transform}
\end{eqnarray}

The set of all possible rigid body motions \nm{\mathbb{SE}\lrp{3} = \lrb{\mathcal M = \lrp{\mathcal R, \ \vec T} | \ \mathcal R \in \mathbb{SO}\lrp{3}, \ \vec T \in \mathbb{R}^3}}, coupled with the motion concatenation defined below, is a valid representation of the special Euclidean group. The inverse motion, as well as the concatenation operation, get slightly more complex because of the affine nature of the point action:
\begin{eqnarray}
\nm{\lrp{\mathcal R, \, \vec T}^{-1}} & = & \nm{\lrp{\mathcal R^{-1}, \, - g_{\mathcal R*}^{-1}\lrp{\vec T}}} \label{eq:SE3_affine_inverse} \\
\nm{\lrp{\mathcal R_{\sss EB}, \, \vec T_{\sss EB}^{\sss E}}} & = & \nm{\lrp{\mathcal R_{\sss EN}, \, \vec T_{\sss EN}^{\sss E}} \circ \lrp{\mathcal R_{\sss NB}, \, \vec T_{\sss NB}^{\sss N}} = \lrp{\mathcal R_{\sss EN} \circ \mathcal R_{\sss NB}, \,  \vec g_{\mathcal{R}_{EN}*}\lrp{\vec T_{\sss NB}^{\sss N}} + \vec T_{\sss EN}^{\sss E}}} \label{eq:SE3_affine_concatenation} 
\end{eqnarray}

\begin{center}
\begin{tabular}{lcclcc}
	\hline
	Concept    & \nm{\mathbb{SE}\lrp{3}} & Affine & Concept    & \nm{\mathbb{SE}\lrp{3}} & Affine \\
	\hline
	Lie group element   & \nm{\mathcal M} & \nm{\lrp{\mathcal R, \, \vec T}}         & Concatenation         & \nm{\circ}      & \nm{\circ}         \\
	Identity            & \nm{\mathcal {I_M}} & \nm{\lrp{\mathcal {I_R}, \, \vec 0}}  & Inverse               & \nm{\mathcal M^{-1}} & \nm{\lrp{\mathcal R, \, \vec T}^{-1}}  \\
	Point motion        & \nm{\vec g_{\mathcal M}(\vec p)} & \nm{\vec g_{\mathcal R}\lrp{\vec p} + \vec T}   & Vector motion  & \nm{\vec g_{\mathcal M*}(\vec v)} & \nm{\vec g_{\mathcal R*}\lrp{\vec v}} \\
	\hline
\end{tabular}
\end{center}
\captionof{table}{Comparison between generic \nm{\mathbb{SE}\lrp{3}} and motion affine representation} \label{tab:RigidBody_motion_lie_affine}

As indicated in (\ref{eq:Motion_maps}), the effect of a rigid body motion on a vector requires a different map than that of points because the translational part has the same influence on both the vector initial and final points:
\neweq{\vec g_{\mathcal M*}\lrp{\vec v} = \vec g_{\mathcal M}\lrp{\vec q} - \vec g_{\mathcal M}\lrp{\vec p} = \lrp{\vec g_{\mathcal R}\lrp{\vec q} + \vec T} - \lrp{ \vec g_{\mathcal R}\lrp{\vec p} + \vec T} = \vec g_{\mathcal R}\lrp{\vec q - \vec p} = \vec g_{\mathcal R*}\lrp{\vec v} \neq \vec g_{\mathcal M}\lrp{\vec v}} {eq:SE3_affine_vector}

%%%%%%%%%%%%%%%%%%%%%%%%%%%%%%%%%%%%%%%%%%%%%%%%%%%%%%%%%%%%%%%%%%%%%%%%
%%%%%%%%%%%%%%%%%%%%%%%%%%%%%%%%%%%%%%%%%%%%%%%%%%%%%%%%%%%%%%%%%%%%%%%%
%%%%%%%%%%%%%%%%%%%%%%%%%%%%%%%%%%%%%%%%%%%%%%%%%%%%%%%%%%%%%%%%%%%%%%%%
% SECTION      HOMOGENEOUS MATRIX
%%%%%%%%%%%%%%%%%%%%%%%%%%%%%%%%%%%%%%%%%%%%%%%%%%%%%%%%%%%%%%%%%%%%%%%%
%%%%%%%%%%%%%%%%%%%%%%%%%%%%%%%%%%%%%%%%%%%%%%%%%%%%%%%%%%%%%%%%%%%%%%%%
%%%%%%%%%%%%%%%%%%%%%%%%%%%%%%%%%%%%%%%%%%%%%%%%%%%%%%%%%%%%%%%%%%%%%%%%

\section{Homogeneous Matrix}\label{sec:RigidBody_motion_homogeneous}

\emph{Homogeneous coordinates} are introduced with the objective of replacing the affine transformation (\ref{eq:SE3_affine_dcm_transform}) representing the rigid body motion with a linear transformation. Given a point \nm{\vec p = \lrsb{p_1 \ \ p_2 \ \ p_3}^T \in \mathbb{R}^3}, its homogeneous representation is obtained adding a ``1'' as a fourth coordinate, so that \nm{\pbar = \lrsb{\vec p \ \ 1 }^T = \lrsb{p_1 \ \ p_2 \ \ p_3 \ \ 1 }^T \in \mathbb{R}^4}. In the case of a vector \nm{\vec v = \vec q - \vec p \in \mathbb{R}^3}, its homogeneous coordinates are \nm{\vbar = \lrsb{\vec v \ \ 0 }^T = \qbar - \pbar = \lrsb{v_1 \ \ v_2 \ \ v_3 \ \ 0 }^T \in \mathbb{R}^4}. The affine coordinate transformation (\ref{eq:SE3_affine_dcm_transform}) can then be converted into a linear transformation:
\neweq{\vec g_{\mathcal M}\lrp{\pbar} = \begin{bmatrix} \nm{\vec g_{\mathcal M}\lrp{\vec p}} \\ 1 \end{bmatrix} = \begin{bmatrix} \nm{\vec R} & \nm{\vec T} \\ \nm{\vec 0_3^T} & 1 \end{bmatrix} \ \begin{bmatrix} \nm{\vec p} \\ 1 \end{bmatrix} = \vec M \; \pbar} {eq:SE3_homogeneous_transform} 

where \nm{\vec M \in \mathbb{R}^{4x4}} is the homogeneous representation of \nm{\lrp{\vec R, \, \vec T} \in \mathbb{SE}\lrp{3}}.
\begin{center}
\begin{tabular}{lcclcc}
	\hline
	Concept    & \nm{\mathbb{SE}\lrp{3}} & Homogeneous & Concept    & \nm{\mathbb{SE}\lrp{3}} & Homogeneous \\
	\hline
	Lie group element   & \nm{\mathcal M} & \nm{\vec M}         & Concatenation         & \nm{\circ}      & Matrix product \\
	Identity            & \nm{\mathcal {I_M}} & \nm{\vec I_4} & Inverse               & \nm{\mathcal M^{-1}} & \nm{\vec M^{-1}} \\
	Point motion        & \nm{\vec g_{\mathcal M}(\vec p)} & \nm{\vec M \, \pbar}   & Vector motion  & \nm{\vec g_{\mathcal M*}(\vec v)} & \nm{\vec M \, \vbar} \\
	\hline
\end{tabular}
\end{center}
\captionof{table}{Comparison between generic \nm{\mathbb{SE}\lrp{3}} and homogeneous matrix} \label{tab:RigidBody_motion_lie_homogeneous}

This enables a natural matrix representation of the special Euclidean group \cite{Soatto2001}:
\neweq{\mathbb{SE}\lrp{3} = \Bigg\{\vec M = \begin{bmatrix} \nm{\vec R} & \nm{\vec T} \\ \nm{\vec 0_3^T} & 1 \end{bmatrix} \ \Bigg| \ \vec R \in \mathbb{R}^{3x3}, \ \vec  T \in \mathbb{R}^3\Bigg\} \subset \mathbb{R}^{4x4}} {eq:SE3_homogeneous_SE3}

which has group structure under matrix multiplication \nm{\{\mathbb{R}^{4x4} \times \mathbb{R}^{4x4} \rightarrow \mathbb{R}^{4x4} \ | \ \Ma \, \Mb \in \mathbb{R}^{4x4}, \forall \ \Ma, \ \Mb \in \mathbb{R}^{4x4}\}} \cite{Pinter1990}. While having dimension sixteen, the special Euclidean group \nm{\mathbb{SE}\lrp{3}} defined by means of homogeneous matrices constitutes a six dimensional manifold to Euclidean space \nm{\mathbb{E}^6}. Note that in this group the identity element is given by the identity matrix \nm{\lrp{\vec I = \vec I_4}}.

The inversion and concatenation of transformations are linear when using homogeneous coordinates \cite{Soatto2001}:
\begin{eqnarray}
\nm{\vec M^{-1}} & \nm{\!\! =} & \nm{\!\! {\begin{bmatrix} \nm{\vec R} & \nm{\vec T} \\ \nm{\vec 0_3^T} & 1 \end{bmatrix}}^{-1} = \begin{bmatrix} \nm{\vec R^T} & \nm{- \vec R^T \, \vec T} \\ \nm{\vec 0_3^T} & 1 \end{bmatrix}} \label{eq:SE3_homogeneous_inverse} \\
\nm{\MEB} & \nm{\!\! =} & \nm{\!\! \begin{bmatrix} \nm{\vec R_{\sss EB}} & \nm{\vec T_{\sss EB}^{\sss E}} \\ \nm{\vec 0_3^T} & 1 \end{bmatrix} = \MEN \, \MNB = \begin{bmatrix} \nm{\vec R_{\sss EN}} & \nm{\vec T_{\sss EN}^{\sss E}} \\ \nm{\vec 0_3^T} & 1 \end{bmatrix} \!\! \begin{bmatrix} \nm{\vec R_{\sss NB}} & \nm{\vec T_{\sss NB}^{\sss N}} \\ \nm{\vec 0_3^T} & 1 \end{bmatrix} = \begin{bmatrix} \nm{\vec R_{\sss EN} \vec R_{\sss NB}} & \nm{\vec R_{\sss EN} \vec T_{\sss NB}^{\sss N} + \vec T_{\sss EN}^{\sss E}} \\ \nm{\vec 0_3^T} & 1 \end{bmatrix}} \label{eq:SE3_homogeneous_concatenation} 
\end{eqnarray}

An advantage of the homogeneous representation is that the motion actions on points and vectors share the same expression:
\neweq{\vec g_{\mathcal M*}\lrp{\vbar} = \vec g_{\mathcal M}\lrp{\qbar} - \vec g_{\mathcal M}\lrp{\pbar} = \vec M \ \qbar - \vec M \ \pbar = \vec M \ \lrp{\qbar - \pbar} = \vec M \ \vbar = \vec g_{\mathcal M}\lrp{\vbar}} {eq:SE3_homogeneous_vector_linear}

%%%%%%%%%%%%%%%%%%%%%%%%%%%%%%%%%%%%%%%%%%%%%%%%%%%%%%%%%%%%%%%%%%%%%%%%
%%%%%%%%%%%%%%%%%%%%%%%%%%%%%%%%%%%%%%%%%%%%%%%%%%%%%%%%%%%%%%%%%%%%%%%%
%%%%%%%%%%%%%%%%%%%%%%%%%%%%%%%%%%%%%%%%%%%%%%%%%%%%%%%%%%%%%%%%%%%%%%%%
% SECTION      TRANSFORM VECTOR as TANGENT SPACE
%%%%%%%%%%%%%%%%%%%%%%%%%%%%%%%%%%%%%%%%%%%%%%%%%%%%%%%%%%%%%%%%%%%%%%%%
%%%%%%%%%%%%%%%%%%%%%%%%%%%%%%%%%%%%%%%%%%%%%%%%%%%%%%%%%%%%%%%%%%%%%%%%
%%%%%%%%%%%%%%%%%%%%%%%%%%%%%%%%%%%%%%%%%%%%%%%%%%%%%%%%%%%%%%%%%%%%%%%%

\section{Transform Vector as Tangent Space}\label{sec:RigidBody_motion_transform_vector}

As discussed in section \ref{subsec:algebra_lie_velocities}, the structure of the Lie algebra associated to \nm{\mathbb{SE}\lrp{3}} can be obtained by time derivating the Lie group inverse constraint, \nm{\vec M^{-1}\lrp{t} \, \vec M\lrp{t} = \vec M\lrp{t} \, \vec M^{-1}\lrp{t} =  \vec I_4}, resulting in the following particularizations of (\ref{eq:algebra_vE}) and (\ref{eq:algebra_vX}): 
\begin{eqnarray}
\nm{\xiEBEskew} & = & \nm{\MEBdot \; \MEBinv = - \MEB \; \MEBdotinv} \label{eq:SE3_homogeneous_twist_space} \\
\nm{\xiEBBskew} & = & \nm{\MEBinv \; \MEBdot = - \MEBdotinv \; \MEB} \label{eq:SE3_homogeneous_twist_body} 
\end{eqnarray}

The Lie algebra velocity \nm{\vec v^\wedge} of \nm{\mathbb{SE}\lrp{3}} is known as the \emph{twist} \nm{\vec \xi^\wedge}, and has the following structure, derived from (\ref{eq:SE3_homogeneous_twist_space}) and (\ref{eq:SE3_homogeneous_twist_body}): 
\neweq{\vec \xi^{\wedge}\lrp{t} = \begin{bmatrix} \nm{\omegaskew\lrp{t}} & \nm{\vec \nu\lrp{t}} \\ \nm{\vec 0_3^T} & 0 \end{bmatrix} \subset \mathbb{R}^{4x4}} {eq:SE3_twist_expression2}

The twist \nm{\vec \xi} represents the motion velocity and is composed by the angular velocity \nm{\vec \omega} defined in section \ref{sec:RigidBody_rotation_calculus_derivatives} and the \emph{linear velocity} \nm{\vec \nu}, defined in section \ref{sec:RigidBody_motion_calculus_derivatives}. Inverting (\ref{eq:SE3_homogeneous_twist_space}) and (\ref{eq:SE3_homogeneous_twist_body}) results in the homogeneous matrix time derivative, which is linear:
\neweq{\MEBdot = \xiEBEskew \; \MEB = \MEB \; \xiEBBskew} {eq:SE3_homogeneous_dot}

Notice that if \nm{\vec M\lrp{t_0} = \vec I_4}, then \nm{\vec{\dot M}\lrp{t_0} = \vec \xi^{\wedge} \lrp{t_0}}, and hence the twist matrix \nm{\vec \xi^{\wedge}\lrp{t_0}} provides a first order approximation of the homogeneous matrix around the identity matrix \nm{\vec I_4}:
\neweq{\vec M\lrp{t_0 + \Deltat} \approx \vec I_4 + \vec \xi^{\wedge} \lrp{t_0} \, \Deltat}{eq:SE3_twist_taylor}

The space of matrices with the (\ref{eq:SE3_twist_expression2}) structure, \nm{\mathfrak{se}\lrp{3} = \bigg\{\vec \xi^{\wedge} = \begin{bmatrix} \omegaskew & \vec \nu \\ \nm{\vec 0_3^T} & 0 \end{bmatrix} \subset \mathbb{R}^{4x4} \ | \ \omegaskew \in \mathfrak{so}\lrp{3}, \ \vec \nu \in \mathbb{R}^3\bigg\}} is hence the \emph{tangent space} of \nm{\mathbb{SE}\lrp{3}} at the identity \nm{\vec I_4} \cite{Soatto2001}, denoted as \nm{T_{\vec I_4}{\mathcal M}}. With the twist Cartesian coordinates defined as \nm{\vec \xi = \lrsb{\vec \nu \ \ \vec \omega}^T \in \mathbb{R}^6}, the \emph{hat} \nm{\lrb{\cdot^\wedge: \mathbb{R}^6 \rightarrow \mathfrak{se}\lrp{3} \ | \ \vec \xi \rightarrow \vec \xi^\wedge}} and \emph{vee} \nm{\lrb{\cdot^\vee: \mathfrak{se}\lrp{3} \rightarrow \mathbb{R}^6 \ | \ \lrp{\vec \xi^\wedge}^\vee \rightarrow \vec \xi}} operators convert the Cartesian vector form of the twist into its matrix form, and vice versa.

If \nm{\vec M\lrp{t_0} \neq \vec I_4}, the tangent space needs to be transported right multiplying by \nm{\MEB\lrp{t_0}} (in the case of space twist), or left multiplying for the local based twist:
\begin{eqnarray}
\nm{\MEB\lrp{t_0 + \Deltat}} & \nm{\approx} & \nm{\MEB\lrp{t_0} + \lrsb{\xiEBEskew \lrp{t_0} \, \Deltat} \, \MEB\lrp{t_0} = \lrsb{\vec I_4 + \xiEBEskew \lrp{t_0} \, \Deltat} \, \MEB\lrp{t_0} }\label{eq:SE3_twist_taylor_space} \\
\nm{\MEB\lrp{t_0 + \Deltat}} & \nm{\approx} & \nm{\MEB\lrp{t_0} + \MEB\lrp{t_0} \, \lrsb{\xiEBBskew \lrp{t_0} \, \Deltat} = \MEB\lrp{t_0} \, \lrsb{\vec I_4 + \xiEBBskew \lrp{t_0} \, \Deltat}}\label{eq:SE3_twist_taylor_body}
\end{eqnarray}

Note that the solution to the ordinary differential equation \nm{\vec{\dot x}\lrp{t} = \vec \xi^\wedge \, \vec x\lrp{t}, \ \vec x\lrp{t} \in \mathbb{R}^6}, where \nm{\vec \xi^\wedge} is constant, is \nm{\vec x\lrp{t} = e^{\vec \xi^\wedge t} \, \vec x\lrp{0}}. Based on it, assuming \nm{\vec M\lrp{0} = \vec I_4} as initial condition, and considering for the time being that \nm{\vec \xi^\wedge} is constant,
\neweq{\vec M\lrp{t} = e^{\vec \xi^\wedge t} = \vec I_4 + \vec \xi^\wedge t + \frac{\lrp{\vec \xi^\wedge t}^2}{2!} + \dots + \frac{\lrp{\vec \xi^\wedge t}^n}{n!} + \dots}{eq:SE3_twist_exponential3}

It can be proven by means of (\ref{eq:SO3_rotv_exponential2}) and the matrix exponential properties that (\ref{eq:SE3_twist_exponential3}) is indeed an homogeneous matrix that hence represents the special Euclidean \nm{\mathbb{SE}\lrp{3}} transformations.
\begin{center}
\begin{tabular}{lcc}
	\hline
	Concept    & Lie Theory & \nm{\mathbb{SE}\lrp{3}} \\
	\hline
	Tangent space element & \nm{\vec \tau^\wedge} & \nm{\vec \tau^\wedge = \begin{bmatrix} \rskew & \vec s \\ \nm{\vec 0_3^T} & 0 \end{bmatrix}} \\
	Velocity element      & \nm{\vec v^\wedge} & \nm{\vec \xi^\wedge = \begin{bmatrix} \omegaskew & \vec \nu \\ \nm{\vec 0_3^T} & 0\end{bmatrix}} \\
	Structure             & \nm{\wedge} & (\ref{eq:SE3_twist_expression2}) \\
	\hline
\end{tabular}
\end{center}
\captionof{table}{Comparison between generic \nm{\mathbb{SE}\lrp{3}} and transform vector as tangent space} \label{tab:Motion_lie_trfv}

Remembering that so far \nm{\vec \xi^\wedge} is constant, which is equivalent to both \nm{\omegaskew} and \nm{\vec \nu} within (\ref {eq:SE3_twist_expression2}) also being constant, (\ref{eq:SE3_twist_exponential3}) means that any rigid body motion \nm{\vec M\lrp{t} = e^{\ds{\vec \xi^\wedge t}}} can be realized by maintaining a constant twist \nm{\vec \xi^\wedge} for a given time \emph{t} \cite{Soatto2001}. The vectors \nm{\vec n = \vec \omega \, t / \| \vec \omega \, t \| = \vec r / \|\vec r\|} and \nm{\vec k = \vec \nu \, t / \| \vec \nu \, t \| = \vec s / \|\vec s\|} indicate the twist directions, while \nm{\phi = \|\vec r\|} and \nm{\rho = \|\vec s\|} represent the twist magnitudes, respectively. This enables the definition of the \emph{transform vector} \nm{\vec \tau}, also known as the \emph{exponential coordinates} of the \nm{\mathcal M} motion, as
\neweq{\vec \tau = \vec \xi \, t = \lrsb{\vec \nu \, t \ \ \vec \omega \, t}^T = \lrsb{\vec s \ \ \vec r}^T = \lrsb{ \vec k \, \rho \ \ \vec n \, \phi}^T \in \mathbb{R}^6}{eq:SE3_transform_vector_definition}

Note that the transform vector \nm{\vec \tau} belongs to the tangent space as it is a multiple of the twist \nm{\vec \xi \in \mathfrak{se}\lrp{3}}, and hence tends to coincide with it as time tends to zero. The \emph{exponential map} \nm{\lrb{exp\lrp{} : \mathfrak{se}\lrp{3} \rightarrow \mathbb{SE}\lrp{3} \ | \ \mathcal M = exp\lrp{\vec \tau^\wedge}}} and its capitalized form \nm{\lrb{Exp\lrp{} : \mathbb{R}^6 \rightarrow \mathbb{SE}\lrp{3} \ | \ \mathcal M = Exp\lrp{\vec \tau}}} wrap the transform vector around the special Euclidean group. However, the twist \nm{\vec \xi^\wedge \lrp{t}} in fact is not required to be constant. Given a rigid body motion represented by its homogeneous matrix \nm{\vec M \in \mathbb{SE}\lrp{3}}, it can be proven that there exists a not necessarily unique transform vector \nm{\vec \tau = \lrsb{\vec s \ \ \vec r}^T = \lrsb{\vec k \, \rho \ \ \vec n \, \phi}^T} such that \nm{\vec M = e^{\vec \tau^\wedge}} \cite{Soatto2001,Murray1994}. The exponential map has the following form \cite{Soatto2001}:
\begin{eqnarray}
\nm{\vec M\lrp{\vec \tau}} & = & \nm{exp\lrp{\vec \tau^\wedge} = \begin{bmatrix} \nm{exp\lrp{\rskew}} & \nm{\dfrac{\lrsb{\vec I_3 - exp\lrp{\rskew}} \, \rskew \, \vec s + \vec r \, {\vec r}^T \, \vec s}{{\|\vec r\|}^2}} \\ 0 & 1 \end{bmatrix} \ \ \ \ \ \ \ \ \ \ \ \ \vec r \neq \vec 0}\label{eq:SE3_twist_exponential4_a} \\
\nm{\vec M\lrp{\vec \tau}} & = & \nm{exp\lrp{\vec \tau^\wedge} = \begin{bmatrix} \nm{\vec I_3} & \nm{\vec s} \\ 0 & 1 \end{bmatrix} \ \ \ \ \ \ \ \ \ \ \ \ \ \ \ \ \ \ \ \ \ \ \ \ \ \ \ \ \ \ \ \ \ \ \ \ \ \ \ \ \ \ \ \ \ \ \ \ \ \ \vec r = \vec 0}\label{eq:SE3_twist_exponential4_b}
\end{eqnarray}

The exponential map described above is thus surjective but not injective, as in general there are infinitely many solutions to the map. The \emph{logarithmic map} \nm{\lrb{log\lrp{} : \mathbb{SE}\lrp{3} \rightarrow \mathfrak{se}\lrp{3} \ | \ \vec \tau^\wedge = log\lrp{\mathcal M}}} and its capitalized version \nm{\lrb{Log\lrp{} : \mathbb{SE}\lrp{3} \rightarrow \mathbb{R}^6 \ | \ \vec \tau = Log\lrp{\mathcal M}}} hence convert rigid body motions into transform vectors \cite{Soatto2001}. The rotation vector \nm{\vec r} is provided by (\ref{eq:SO3_rotv_logarithm}); if \nm{\vec r = \vec 0}, \nm{\vec s} coincides with \nm{\vec T}, while otherwise it is obtained by solving for \nm{\vec s} from the following linear system \cite{Soatto2001}, taken from (\ref{eq:SE3_twist_exponential4_a}):
\neweq{\lrsb{(\vec I_3 - e^{\rskew}) \, \rskew + \vec r \, {\vec r}^T} \, \vec s = \|\vec r\|^2 \, \vec T}{eq:SE3_twist_logarithm} 

Unlike the case of the rotation vector (\ref{eq:SO_rotv_inversion}), the transform vector inverse coincides with its negative only if the motion is very small. The different \nm{\mathbb{SE}\lrp{3}} actions (concatenation, point motion, vector motion), the inverse, and the relationship between the transform vector derivative with time and the twist, are complex and rarely used.

%%%%%%%%%%%%%%%%%%%%%%%%%%%%%%%%%%%%%%%%%%%%%%%%%%%%%%%%%%%%%%%%%%%%%%%%
%%%%%%%%%%%%%%%%%%%%%%%%%%%%%%%%%%%%%%%%%%%%%%%%%%%%%%%%%%%%%%%%%%%%%%%%
%%%%%%%%%%%%%%%%%%%%%%%%%%%%%%%%%%%%%%%%%%%%%%%%%%%%%%%%%%%%%%%%%%%%%%%%
% SECTION      UNIT DUAL QUATERNION
%%%%%%%%%%%%%%%%%%%%%%%%%%%%%%%%%%%%%%%%%%%%%%%%%%%%%%%%%%%%%%%%%%%%%%%%
%%%%%%%%%%%%%%%%%%%%%%%%%%%%%%%%%%%%%%%%%%%%%%%%%%%%%%%%%%%%%%%%%%%%%%%%
%%%%%%%%%%%%%%%%%%%%%%%%%%%%%%%%%%%%%%%%%%%%%%%%%%%%%%%%%%%%%%%%%%%%%%%%

\section{Unit Dual Quaternion}\label{sec:RigidBody_motion_unit_dual_quaternion}

The dual quaternions with unity norm, known as unit dual quaternions, comprise an additional representation of the special Euclidean group \nm{\mathbb{SE}\lrp{3}}, as shown below. Dual quaternions in turn are generalizations of quaternions in the same way as dual numbers are generalization or real ones \cite{Jia2013,Kenwright2012,Valverde2018}. For this reason, it is necessary to first describe the dual numbers and dual vectors in sections \ref{subsec:RigidBody_motion_dual_numbers} and \ref{subsec:RigidBody_motion_dual_vectors} before discussing the dual quaternions in section \ref{subsec:RigidBody_motion_dual_quat} and finally the unit dual quaternions in section \ref{subsec:RigidBody_motion_unit_dual_quat}.

%%%%%%%%%%%%%%%%%%%%%%%%%%%%%%%%%%%%%%%%%%%%%%%%%%%%%%%%%%%%%%%%%%%%%%%%
%%%%%%%%%%%%%%%%%%%%%%%%%%%%%%%%%%%%%%%%%%%%%%%%%%%%%%%%%%%%%%%%%%%%%%%%
% SUBSECTION      DUAL NUMBERS
%%%%%%%%%%%%%%%%%%%%%%%%%%%%%%%%%%%%%%%%%%%%%%%%%%%%%%%%%%%%%%%%%%%%%%%%
%%%%%%%%%%%%%%%%%%%%%%%%%%%%%%%%%%%%%%%%%%%%%%%%%%%%%%%%%%%%%%%%%%%%%%%%

\subsection{Dual Numbers}\label{subsec:RigidBody_motion_dual_numbers}

The set of \emph{dual numbers} \nm{\mathbb{D}} is defined as \nm{\lrb{\mathbb{D} = \mathbb{R} + \mathbb{R} \, \epsilon \ | \ \epsilon^2 = \epsilon \cdot \epsilon = 0}}. Given two dual numbers \nm{d_1^{\diamond} = x_1 + y_1 \, \epsilon \in \mathbb{D}, d_2^{\diamond} = x_2 + y_2 \, \epsilon \in \mathbb{D}, \forall \ x_1, y_1, x_2, y_2 \in \mathbb{R}}, it is possible to define the operations of addition \nm{\lrb{+ : \mathbb{D} \times  \mathbb{D} \rightarrow \mathbb{D}}} and  multiplication \nm{\lrb{\cdot : \mathbb{D} \times \mathbb{D} \rightarrow \mathbb{D}}} \cite{Jia2013}:
\begin{eqnarray}
\nm{d_1^{\diamond} + d_2^{\diamond}} & = & \nm{\lrp{x_1 + y_1 \, \epsilon} + \lrp{x_2 + y_2 \, \epsilon} = \lrp{x_1 + x_2} + \lrp{y_1 + y_2} \, \epsilon}\label{eq:SE3_dual_numbers_addition} \\
\nm{d_1^{\diamond} \cdot d_2^{\diamond}} & = & \nm{\lrp{x_1 + y_1 \, \epsilon} \cdot \lrp{x_2 + y_2 \, \epsilon} = \lrp{x_1 x_2} + \lrp{x_1 y_2 + y_1 x_2} \, \epsilon} \label{eq:SE3_dual_numbers_multiplication}
\end{eqnarray}

The set of dual numbers \nm{\mathbb{D}} endowed with the operations of addition \nm{+} and multiplication \nm{\cdot} forms a ring, known as the ring of dual numbers \nm{\langle \mathbb{D}, +, \cdot \rangle}, nearly always abbreviated to simply \nm{\mathbb{D}}. The additive identity is \nm{0^{\diamond} = 0 + 0 \, \epsilon} and the inverse \nm{- d^{\diamond} = - x - y \, \epsilon}, while the multiplication identity is \nm{1^{\diamond} = 1 + 0 \, \epsilon} and the inverse \nm{d^{\diamond-1} = 1 / x - y \, \epsilon / x^2}. Note that \nm{\mathbb{D}} is a ring instead of a field as the multiplicative inverse \nm{d^{\diamond-1}} is not defined when \nm{x = 0}. The conjugate of a dual number is obtained by switching the sign of its dual part (\nm{d^{\ast} = x - y \, \epsilon \in \mathbb D}).

The most useful property of dual numbers is the explicit relationship that exists between the value of any function evaluated at a dual number \nm{f\lrp{d^{\diamond}} = f\lrp{x + y \, \epsilon}} and its value when evaluated exclusively at its real part \nm{f\lrp{x}} \cite{Kenwright2012}. The Taylor expansion of \nm{f\lrp{x + y \, \epsilon}} around \emph{x} reads:
\neweq{f\lrp{d^{\diamond}} = f\lrp{x + y \, \epsilon} = f\lrp{x} + \pderpar{f}{d^{\diamond}}\lrp{x} \, \lrp{d^{\diamond} - \!x} + \frac{1}{2!} \, \dfrac{\partial^2{f}}{\partial{d^{\diamond 2}}}\lrp{x} \, \lrp{d^{\diamond} - \!x}^2 + \frac{1}{3!} \, \dfrac{\partial^3{f}}{\partial{d^{\diamond 3}}}\lrp{x} \, \lrp{d^{\diamond} - \!x}^3 + \cdots}{eq:SE3_dual_number_taylor1}

As \nm{\lrp{d^{\diamond} - x}^n = y^n \, \epsilon^n} is zero when \nm{n > 1}, this translates into:
\neweq{f\lrp{d^{\diamond}} = f\lrp{x + y \, \epsilon} = f\lrp{x} + \pderpar{f}{d^{\diamond}}\lrp{x} \, y \, \epsilon}{eq:SE3_dual_number_taylor}

%%%%%%%%%%%%%%%%%%%%%%%%%%%%%%%%%%%%%%%%%%%%%%%%%%%%%%%%%%%%%%%%%%%%%%%%
%%%%%%%%%%%%%%%%%%%%%%%%%%%%%%%%%%%%%%%%%%%%%%%%%%%%%%%%%%%%%%%%%%%%%%%%
% SUBSECTION      DUAL VECTORS
%%%%%%%%%%%%%%%%%%%%%%%%%%%%%%%%%%%%%%%%%%%%%%%%%%%%%%%%%%%%%%%%%%%%%%%%
%%%%%%%%%%%%%%%%%%%%%%%%%%%%%%%%%%%%%%%%%%%%%%%%%%%%%%%%%%%%%%%%%%%%%%%%

\subsection{Dual Vectors}\label{subsec:RigidBody_motion_dual_vectors}

Dual vectors in three dimensions are formed by grouping three dual numbers \nm{\{\vec d^{\diamond} = \lrsb{d_1^{\diamond} \ \ d_2^{\diamond} \ \ d_3^{\diamond}}^T \in \mathbb{D}^3,} \nm{\forall \ d_1^{\diamond}, \, d_2^{\diamond}, \, d_3^{\diamond} \in \mathbb{D}\}}. It is then possible to define, \nm{\forall \ d^{\diamond} \in \mathbb{D}, \vec d^{\diamond}, \vec e^{\diamond} \in \mathbb{D}^3}, the scalar multiplication of a double number by a double vector \nm{\lrb{\cdot : \mathbb{D} \times \mathbb{D}^3 \rightarrow \mathbb{D}^3}}, the inner product between two double vectors \nm{\lrb{\langle \cdot \, , \cdot \rangle: \mathbb{D}^3 \times \mathbb{D}^3 \rightarrow \mathbb{D}}}, and the cross product between two double vectors \nm{\lrb{\times : \mathbb{D}^3 \times \mathbb{D}^3 \rightarrow \mathbb{D}^3}}. The results are similar to those of real numbers shown in section \ref{sec:algebra_structures} \cite{Jia2013}:
\begin{eqnarray}
\nm{d^{\diamond} \cdot \vec d^{\diamond}} & = & \nm{\lrsb{d^{\diamond} \, d_1^{\diamond} \ \ \ d^{\diamond} \, d_2^{\diamond} \ \ \ d^{\diamond} \, d_3^{\diamond}}^T}\label{eq:SE3_dual_vector_multiplication} \\
\nm{\langle \vec d^{\diamond}, \vec e^{\diamond} \rangle} & = & \nm{\vec d^{\diamond} \cdot \vec e^{\diamond} = {\vec d^{\diamond}}^T \, \vec e^{\diamond} = \lrsb{d_1^{\diamond} \, e_1^{\diamond} \ \ \ \ d_2^{\diamond} \, e_2^{\diamond} \ \ \ \ d_3^{\diamond} \, e_3^{\diamond}}^T}\label{eq:SE3_dual_vector_scalar_product} \\
\nm{\vec d^{\diamond} \times \vec e^{\diamond}} & = & \nm{\widehat{\vec d^{\diamond}} \; \vec e^{\diamond} = \begin{bmatrix} \nm{0^{\diamond}} & \nm{- d_3^{\diamond}} & \nm{+ d_2^{\diamond}} \\ \nm{+ d_3^{\diamond}} & \nm{0^{\diamond}} & \nm{- d_{\sss 1}^{\diamond}} \\ \nm{- d_2^{\diamond}} & \nm{+ d_{\sss 1}^{\diamond}} & \nm{0^{\diamond}} \end{bmatrix} \begin{bmatrix} \nm{e_{\sss 1}^{\diamond}} \\ \nm{e_2^{\diamond}} \\ \nm{e_3^{\diamond}} \end{bmatrix} = \begin{bmatrix} \nm{d_2^{\diamond} \, e_3^{\diamond} - d_3^{\diamond} \, e_2^{\diamond}} \\ \nm{d_3^{\diamond} \, e_{\sss 1}^{\diamond} - d_{\sss 1}^{\diamond} \, e_3^{\diamond}} \\ \nm{d_{\sss 1}^{\diamond} \, e_2^{\diamond} - d_2^{\diamond} \, e_{\sss 1}^{\diamond}} \end{bmatrix} = - \vec e^{\diamond} \times \vec d^{\diamond} = - \widehat{\vec e^{\diamond}} \; \vec d^{\diamond}}\label{eq:SE3_dual_vector_cross_product} 
\end{eqnarray}

%%%%%%%%%%%%%%%%%%%%%%%%%%%%%%%%%%%%%%%%%%%%%%%%%%%%%%%%%%%%%%%%%%%%%%%%
%%%%%%%%%%%%%%%%%%%%%%%%%%%%%%%%%%%%%%%%%%%%%%%%%%%%%%%%%%%%%%%%%%%%%%%%
% SUBSECTION      DUAL QUATERNIONS
%%%%%%%%%%%%%%%%%%%%%%%%%%%%%%%%%%%%%%%%%%%%%%%%%%%%%%%%%%%%%%%%%%%%%%%%
%%%%%%%%%%%%%%%%%%%%%%%%%%%%%%%%%%%%%%%%%%%%%%%%%%%%%%%%%%%%%%%%%%%%%%%%

\subsection{Dual Quaternions}\label{subsec:RigidBody_motion_dual_quat}

The set of dual quaternions \nm{\mathbb{H}_d} is defined as \nm{\{\mathbb{H_d} = \mathbb{H} + \mathbb{H} \, \epsilon \ | \ \epsilon^2 = -1\}}. A dual quaternion \nm{\vec \zeta \in \mathbb{H}_d} has the form \nm{\vec \zeta = \qr + \qd \, \epsilon}, with \nm{\qr, \qd \in \mathbb{H}}. The real plus dual notation \nm{\lrb{1, \epsilon}} is not always the most convenient. A dual quaternion can also be expressed as the sum of a dual number plus a dual vector in the form \nm{\vec \zeta = d_0^{\diamond} + \vec d_v^{\diamond}}, where \nm{d_0^{\diamond}} is the scalar part and \nm{\vec d_v^{\diamond} = d_1^{\diamond} \, i + d_2^{\diamond} \, j + d_3^{\diamond} \, i \, j} is the vector part. Dual quaternions are however mostly represented as 8-vectors \nm{\vec \zeta = \lrsb{\qr \ \ \qd}^T = \lrsb{q_{0 r} \ \ \vec q_{vr} \ \ q_{0d} \ \ \vec q_{vd}}^T = \lrsb{q_{0 r} \ \ q_{1 r} \ \ q_{2 r} \ \ q_{3 r} \ \ q_{0 d} \ \ q_{1 d} \ \ q_{2 d} \ \ q_{3 d}}^T}, which enables the use of matrix algebra for quaternion operations \cite{Valverde2018}. It is also convenient to abuse the equal operator as required to combine general, real, and pure dual quaternions.

The dual quaternion addition \nm{\lrb{+ : \mathbb{H}_d \times \mathbb{H}_d \rightarrow \mathbb{H}_d}} and the scalar product \nm{\lrb{\cdot : \mathbb{D} \times \mathbb{H}_d \rightarrow \mathbb{H}_d}} are both straightforward and commutative:
\begin{eqnarray}
\nm{\vec \zeta_a + \vec \zeta_b} & = & \nm{\lrp{\vec q_{ra} + \vec q_{da} \, \epsilon} + \lrp{\vec q_{rb} + \vec q_{db} \, \epsilon} = \lrp{\vec q_{ra} + \vec q_{rb}} + \lrp{\vec q_{da} + \vec q_{db}} \, \epsilon = \begin{bmatrix} \nm{\vec q_{ra} + \vec q_{rb}} \\ \nm{\vec q_{da} + \vec q_{db}} \end{bmatrix}}\nonumber \\
& = & \nm{\lrp{d_{0a}^{\diamond} + \vec d_{va}^{\diamond}} + \lrp{d_{0b}^{\diamond} + \vec d_{vb}^{\diamond}} = \lrp{d_{0a}^{\diamond} + d_{0b}^{\diamond}} + \lrp{\vec d_{va}^{\diamond} + \vec d_{vb}^{\diamond}}}\label{eq:SE3_dual_quat_addition} \\
\nm{d^{\diamond} \cdot \vec \zeta} & = & \nm{\lrp{x + y \, \epsilon} \cdot \lrp{\qr + \qd \, \epsilon} = x \cdot \qr + \lrp{x \cdot \qd + y \cdot \qr} \, \epsilon}\label{eq:SE3_dual_quat_scalar} 
\end{eqnarray}

The multiplication of dual quaternions \nm{\{\otimes : \mathbb{H}_d \times \mathbb{H}_d \rightarrow \mathbb{H}_d\}} is not commutative as it includes the dual vector cross product (\ref{eq:SE3_dual_vector_cross_product}). Depending on how it is expressed, the similarities with the multiplication of dual numbers (\ref{eq:SE3_dual_numbers_multiplication}) or that of quaternions (\ref{eq:SO3_quat_product}) are obvious \cite{Jia2013, Valverde2018, Daniilidis1998}:
\begin{eqnarray}
\nm{\vec \zeta_a \otimes \vec \zeta_b} & = & \nm{\lrp{\vec q_{ra} + \vec q_{da} \, \epsilon} \otimes \lrp{\vec q_{rb} + \vec q_{db} \, \epsilon} = \lrp{\vec q_{ra} \otimes \vec q_{rb}} + \lrp{\vec q_{ra} \otimes \vec q_{db} + \vec q_{da} \otimes \vec q_{rb}} \, \epsilon}\nonumber \\
& = & \nm{\begin{bmatrix} \nm{\vec q_{ra} \otimes \vec q_{rb}} \\ \nm{\vec q_{ra} \otimes \vec q_{db} + \vec q_{da} \otimes \vec q_{rb}} \end{bmatrix} = \lrp{d_{0a}^{\diamond} + \vec d_{va}^{\diamond}} \otimes \lrp{d_{0b}^{\diamond} + \vec d_{vb}^{\diamond}}} \nonumber \\
& = & \nm{\lrp{d_{0a}^{\diamond} \, d_{0b}^{\diamond} - {\vec d_{va}^{\diamond}}^T \, \vec d_{vb}^{\diamond}} + \lrp{d_{0a}^{\diamond} \, \vec d_{vb}^{\diamond} + d_{0b}^{\diamond} \, \vec d_{va}^{\diamond} + \widehat{\vec d}_{va}^{\diamond} \, \vec d_{vb}^{\diamond}}}\label{eq:SE3_dual_quat_product}
\end{eqnarray}

Dual quaternion multiplication is also bilinear \cite{Valverde2018}, based on the operators defined in (\ref{eq:SO3_quat_product_matrices}):
\neweq{\vec \zeta_a \otimes \vec \zeta_b = [\vec \zeta_a]_L \, \vec \zeta_b = \begin{bmatrix}
		\nm{\lrsb{\vec a_r}_L} & \nm{\vec O_{4x4}} \\
		\nm{\lrsb{\vec a_d}_L} & \nm{\lrsb{\vec a_r}_L} \end{bmatrix} \,
		\begin{bmatrix} \nm{\vec b_r} \\ \nm{\vec b_d} \end{bmatrix} = 
       [\vec \zeta_b]_R \, \vec \zeta_a = \begin{bmatrix}
	    \nm{\lrsb{\vec b_r}_R} & \nm{\vec O_{4x4}} \\
		\nm{\lrsb{\vec b_d}_R} & \nm{\lrsb{\vec b_r}_R} \end{bmatrix} \,
		\begin{bmatrix} \nm{\vec a_r} \\ \nm{\vec a_d} \end{bmatrix}} {eq:SE3_dual_quat_product_matrices}

It is possible to define three different conjugates for a dual quaternion \cite{Jia2013}, based on whether it only switches the sign of the dual part as in the case of dual numbers (\ref{eq:SE3_dual_quat_conj1}), it employs the conjugates of the real and dual quaternion components (\ref{eq:SE3_dual_quat_conj2}), or a combination of both (\ref{eq:SE3_dual_quat_conj3}):
\begin{eqnarray}
\nm{\zetacirc = \qr - \qd \, \epsilon} & \nm{\rightarrow} & \nm{\lrp{\vec \zeta_a \otimes \vec \zeta_b}^{\circ} = \vec \zeta_a^{\circ} \otimes \vec \zeta_b^{\circ}}\label{eq:SE3_dual_quat_conj1} \\
\nm{\zetaast = \qrast + \qdast \, \epsilon} & \nm{\rightarrow} & \nm{\lrp{\vec \zeta_a \otimes \vec \zeta_b}^{\ast} = \vec \zeta_b^{\ast} \otimes \vec \zeta_a^{\ast}}\label{eq:SE3_dual_quat_conj2} \\
\nm{\zetabullet = \qrast - \qdast \, \epsilon} & \nm{\rightarrow} & \nm{\lrp{\vec \zeta_a \otimes \vec \zeta_b}^{\bullet} = \vec \zeta_b^{\bullet} \otimes \vec \zeta_a^{\bullet}}\label{eq:SE3_dual_quat_conj3}
\end{eqnarray}

\emph{Pure dual quaternions} \nm{\vec \zeta = 0^{\diamond} + \vec d_v^{\diamond} \in \mathbb{H}_{dp}} are those in which its dual number is zero (\nm{0^\diamond}), or in which both its real and dual parts are pure quaternions (\nm{\qr, \, \qd \in \mathbb{H}_p}), and verify that \nm{\vec \zeta = - \, \zetaast}. The dual quaternion norm is defined as \nm{\|\vec \zeta\| = \sqrt{\vec \zeta \otimes \zetaast} = \sqrt{\qr \otimes \qrast + \lrp{\qr \otimes \qdast + \qd \otimes \qrast} \, \epsilon} \in \mathbb{D}} \cite{Daniilidis1998}. Dual quaternions endowed with \nm{\otimes} do not form a group, because although \nm{\vec{\zeta_1} = \vec{q_1} + \vec0 \, \epsilon} is the identity, the inverse \nm{\vec \zeta^{-1} = {\qr}^{-1} - {\qr}^{-1} \otimes \qd \otimes {\qr}^{-1} \, \epsilon = {\qr}^{-1} \otimes \lrp{\vec q_1 - \qd \otimes {\qr}^{-1} \, \epsilon}} is not defined when \nm{\qr = \vec 0}. Dual quaternions endowed with addition \nm{+} and multiplication \nm{\otimes} however do form the non-abelian ring \nm{\langle \mathbb{H}_d, +, \otimes \rangle}. 

As in the case of quaternions described in section \ref{subsec:RigidBody_rotation_rodrigues_quat}, the natural power of a dual quaternion \nm{\vec \zeta^n, n \in \mathbb{N}} is obtained by multiplying the dual quaternion by itself \nm{n-1} times. The double product of a dual quaternion by a vector \nm{\{\mathbb{H}_d \times \mathbb{R}^3 \rightarrow \mathbb{R}^3\}} is defined as the product \nm{\otimes} of the dual quaternion by the vector by the dual quaternion conjugate, resulting in three different versions based on the conjugate definition (\ref{eq:SE3_dual_quat_conj1}, \ref{eq:SE3_dual_quat_conj2}, \ref{eq:SE3_dual_quat_conj3}).

%%%%%%%%%%%%%%%%%%%%%%%%%%%%%%%%%%%%%%%%%%%%%%%%%%%%%%%%%%%%%%%%%%%%%%%%
%%%%%%%%%%%%%%%%%%%%%%%%%%%%%%%%%%%%%%%%%%%%%%%%%%%%%%%%%%%%%%%%%%%%%%%%
% SUBSECTION      UNIT DUAL QUATERNIONS
%%%%%%%%%%%%%%%%%%%%%%%%%%%%%%%%%%%%%%%%%%%%%%%%%%%%%%%%%%%%%%%%%%%%%%%%
%%%%%%%%%%%%%%%%%%%%%%%%%%%%%%%%%%%%%%%%%%%%%%%%%%%%%%%%%%%%%%%%%%%%%%%%

\subsection{Unit Dual Quaternion}\label{subsec:RigidBody_motion_unit_dual_quat}

\emph{Unit dual quaternions} are those dual quaternions in which \nm{\vec \zeta \otimes \zetaast = \zetaast \otimes \vec \zeta = \vec{\zeta_1}}, which implies that the inverse and the conjugate coincide as \nm{\vec \zeta^{-1} = \qrast - \qrast \otimes \qd \otimes \qrast \epsilon = \qrast + \qdast \, \epsilon = \zetaast}. Note that the norm \nm{\|\vec \zeta\|} has a unity real part and a zero dual part \cite{Daniilidis1998}. Based on (\ref{eq:SE3_dual_quat_product}), this translates into the following two conditions:
\begin{eqnarray}
	\nm{\qr \otimes \qrast} & = & \nm{1 \rightarrow \|\qr\| = 1}\label{eq:SE3_unit_dual_quat_cond1} \\
	\nm{\qr \otimes \qdast + \qd \otimes \qrast} & = & \nm{0 \rightarrow \langle \qr, \, \qd \rangle = 0}\label{eq:SE3_unit_dual_quat_cond2} 
\end{eqnarray}

In other words, unit dual quaternions are those in which the real part \nm{\qr} is a unit quaternion that is also orthogonal to the dual part \nm{\qd}.

The rigid body motion between a body frame \nm{\FB} and a spatial frame \nm{\FE} represented by the unit quaternion \nm{\qEB} and the translation \nm{\TEBE} (section \ref{sec:RigidBody_motion_affine}) can always be represented by the following unit dual quaternion \nm{\zetaEB} \cite{Jia2013, Valverde2018}, where the notation is abused to consider the quaternion \nm{\TEBE = \lrsb{0 \ \ \TEBE}^T}:
\neweq{\zetaEB = \qEB + \dfrac{\epsilon}{2} \, \TEBE \otimes \qEB}{eq:SE3_unit_dual_quat_from_affine}

(\ref{eq:SE3_unit_dual_quat_from_affine}) is indeed a unit dual quaternion as \nm{\zetaEB \otimes \zetaEBast = \vec \zeta_1 = \vec q_1 = 1} based on \nm{\vec T_{\sss EB}^{{\sss E}*} = - \TEBE} as it is a pure quaternion. The opposite map providing the affine representation based on the unit dual quaternion is the following:
\begin{eqnarray}
	\nm{\qEB} & = & \nm{\vec \zeta_{{\sss EB}r}}\label{eq:SE3_unit_dual_quat_to_affine_q} \\
	\nm{\TEBE} & = & \nm{2 \, \vec \zeta_{{\sss EB}d} \otimes \vec \zeta_{{\sss EB}r}^{\ast}}\label{eq:SE3_unit_dual_quat_to_affine_T} 
\end{eqnarray}

\begin{center}
\begin{tabular}{lcclcc}
	\hline
	Concept    & \nm{\mathbb{SE}\lrp{3}} & \nm{\mathbb{H}_d} & Concept    & \nm{\mathbb{SE}\lrp{3}} & \nm{\mathbb{H}_d} \\
	\hline
	Lie group element   & \nm{\mathcal M} & \nm{\vec \zeta}         & Concatenation         & \nm{\circ}      & \nm{\otimes} \\
	Identity            & \nm{\mathcal {I_M}} &\nm{\vec{\zeta_1}} & Inverse               & \nm{\mathcal M^{-1}} & \nm{\zetaast} \\
	Point motion        & \nm{\vec g_{\mathcal M}(\vec p)} & \nm{\vec \zeta \otimes \vec \zeta_{\vec p} \otimes \zetabullet}   & Vector motion  & \nm{\vec g_{\mathcal M*}(\vec v)} & \nm{\vec \zeta \otimes \vec \zeta_{\vec v} \otimes \zetabullet} \\
	\hline
\end{tabular}
\end{center}
\captionof{table}{Comparison between generic \nm{\mathbb{SE}\lrp{3}} and unit dual quaternion} \label{tab:RigidBody_motion_lie_unit_dual_quat}

The unit dual quaternion endowed with the double product can be employed to transform both points and vectors, verifying that it complies with the rigid body motion orthogonality and handedness conditions described in section \ref{sec:RigidBody_bases}. Given a point \nm{\vec p = \lrsb{p_1 \ \ p_2 \ \ p_3}^T \in \mathbb{R}^3}, its dual quaternion representation \nm{\vec \zeta_{\vec p}} is obtained by combining the unit quaternion \nm{\vec q_1} as the real part and the point coordinates as the dual part, resulting in \nm{\vec \zeta_{\vec p} = \vec q_1 + \epsilon \, \vec p \in \mathbb{R}^8} \cite{Jia2013}. In the case of a vector \nm{\vec v = \vec q - \vec p \in \mathbb{R}^3}, its dual quaternion representation is \nm{\vec \zeta_{\vec v} = \vec \zeta_{\vec q} - \vec \zeta_{\vec p} = \lrp{\vec q_1 + \epsilon \, \vec q} - \lrp{\vec q_1 + \epsilon \, \vec p} = \epsilon \, \lrp{\vec q - \vec p} = \epsilon \, \vec v \in \mathbb{R}^8}. It is then possible, based on (\ref{eq:SE3_affine_quat_transform}) and (\ref{eq:SE3_affine_vector}), to employ the double product to transform both points and vectors between different frames: 
\begin{eqnarray}
\nm{\vec \zeta_{\pE} = \zetaEB \otimes \vec \zeta_{\pB} \otimes \zetaEBbullet} & = & \nm{\lrp{\qEB + \dfrac{\epsilon}{2} \, \TEBE \otimes \qEB} \otimes \lrp{\vec q_1 + \epsilon \, \pB} \otimes \lrp{\qEBast + \dfrac{\epsilon}{2} \, \qEBast \otimes \TEBE}}\nonumber \\
& = & \nm{\vec q_1 + \epsilon \, \lrp{\qEB \otimes \pB \otimes \qEBast + \TEBE} = \vec q_1 + \epsilon \, \pE}\label{eq:SE3_unit_dual_quat_transform_point} \\
\nm{\vec \zeta_{\pE} = \zetaEB \otimes \vec \zeta_{\pB} \otimes \zetaEBbullet} & = & \nm{\lrp{\qr + \qd \, \epsilon} \otimes \lrp{\vec q_1 + \epsilon \, \pB} \otimes \lrp{\qrast - \qdast \, \epsilon}}\nonumber \\
& = & \nm{\vec q_1 + \epsilon \, \lrp{\qr \otimes \pB \otimes \qrast + \qd \otimes \qrast - \qr \otimes \qdast} = \vec q_1 + \epsilon \, \pE}\label{eq:SE3_unit_dual_quat_transform_point2} \\
\nm{\vec \zeta_{\vE} = \zetaEB \otimes \vec \zeta_{\vB} \otimes \zetaEBbullet} & = & \nm{\lrp{\qEB + \dfrac{\epsilon}{2} \, \TEBE \otimes \qEB} \otimes \epsilon \, \vB \otimes \lrp{\qEBast + \dfrac{\epsilon}{2} \, \qEBast \otimes \TEBE}}\nonumber \\
& = & \nm{\epsilon \, \lrp{\qEB \otimes \vB \otimes \qEBast} = \epsilon \, \vE}\label{eq:SE3_unit_dual_quat_transform_vector} \\
\nm{\vec \zeta_{\vE} = \zetaEB \otimes \vec \zeta_{\vB} \otimes \zetaEBbullet} & = & \nm{\lrp{\qr + \qd \, \epsilon} \otimes \epsilon \, \vB \otimes \lrp{\qrast - \qdast \, \epsilon} = \epsilon \, \lrp{\qr \otimes \vB \otimes \qrast} = \epsilon \, \vE}\label{eq:SE3_unit_dual_quat_transform_vector2}
\end{eqnarray}

A disadvantage of the unit dual quaternion as an \nm{\mathbb{SE}\lrp{3}} representation is that a different expression is required for the inverse transformation:
\begin{eqnarray}
\nm{\vec \zeta_{\pB} = \zetaEBast \otimes \vec \zeta_{\pE} \otimes \zetaEBcirc} & = & \nm{\lrp{\qEBast - \dfrac{\epsilon}{2} \, \qEBast \otimes \TEBE} \otimes \lrp{\vec q_1 + \epsilon \, \pE} \otimes \lrp{\qEB - \dfrac{\epsilon}{2} \, \TEBE \otimes \qEB}}\nonumber \\
& = & \nm{\vec q_1 + \epsilon \, \lrp{\qEBast \otimes \pE \otimes \qEB - \qEBast \otimes \TEBE \otimes \qEB} = \vec q_1 + \epsilon \, \pB}\label{eq:SE3_unit_dual_quat_transform_inv_point} \\
\nm{\vec \zeta_{\pB} = \zetaEBast \otimes \vec \zeta_{\pE} \otimes \zetaEBcirc} & = & \nm{\lrp{\qrast + \qdast \, \epsilon} \otimes \lrp{\vec q_1 + \epsilon \, \pE} \otimes \lrp{\qr - \qd \, \epsilon}}\nonumber \\
& = & \nm{\vec q_1 + \epsilon \, \lrp{\qrast \otimes \pE \otimes \qr + \qdast \otimes \qr - \qrast \otimes \qd} = \vec q_1 + \epsilon \, \pB}\label{eq:SE3_unit_dual_quat_transform_inv_point2} \\
\nm{\vec \zeta_{\vB} = \zetaEBast \otimes \vec \zeta_{\vE} \otimes \zetaEBcirc} & = & \nm{\lrp{\qEBast - \dfrac{\epsilon}{2} \, \qEBast \otimes \TEBE} \otimes \epsilon \, \vE \otimes \lrp{\qEB - \dfrac{\epsilon}{2} \, \TEBE \otimes \qEB}}\nonumber \\
& = & \nm{\epsilon \, \lrp{\qEBast \otimes \vE \otimes \qEB} = \epsilon \, \vB}\label{eq:SE3_unit_dual_quat_transform_inv_vector} \\
\nm{\vec \zeta_{\vB} = \zetaEBast \otimes \vec \zeta_{\vE} \otimes \zetaEBcirc} & = & \nm{\lrp{\qrast + \qdast \, \epsilon} \otimes \epsilon \, \vE \otimes \lrp{\qr - \qd \, \epsilon} = \epsilon \, \lrp{\qrast \otimes \vE \otimes \qr} = \epsilon \, \vB}\label{eq:SE3_unit_dual_quat_transform_inv_vector2}
\end{eqnarray}

The inverse transformation coincides with the dual quaternion conjugate provided by (\ref{eq:SE3_dual_quat_conj2}):
\begin {eqnarray}
\nm{\zetaBE = \vec \zeta_{\sss EB}^{-1} = \zetaEBast} & = & \nm{\qBE + \dfrac{\epsilon}{2} \, \TBEB \otimes \qBE = \qEBast - \dfrac{\epsilon}{2} \, \lrp{\qEBast \otimes \TEBE \otimes \qEB} \otimes \qEBast}\nonumber \\
& = & \nm{\qEBast - \dfrac{\epsilon}{2} \, \qEBast \otimes \TEBE = \qrast + \qdast \, \epsilon}\label{eq:SE3_unit_dual_quat_inverse}
\end{eqnarray}

The concatenation of transformations is straightforward based on (\ref{eq:SE3_affine_concatenation}):
\begin{eqnarray}
\nm{\vec \zeta_{\sss EB} = \vec \zeta_{\sss EN} \otimes \vec \zeta_{\sss NB}} & = & \nm{\lrp{\vec q_{\sss EN} + \dfrac{\epsilon}{2} \, \vec T_{\sss EN}^{\sss E} \otimes \vec q_{\sss EN}} \otimes \lrp{\vec q_{\sss NB} + \dfrac{\epsilon}{2} \, \vec T_{\sss NB}^{\sss N} \otimes \vec q_{\sss NB}}}\nonumber \\
& = & \nm{\vec q_{\sss EN} \otimes \vec q_{\sss NB} + \dfrac{\epsilon}{2} \, \lrp{\vec q_{\sss EN} \otimes \vec T_{\sss NB}^{\sss N} \otimes \vec q_{\sss NB} + \vec T_{\sss EN}^{\sss E} \otimes \vec q_{\sss EN} \otimes \vec q_{\sss NB}}}\nonumber \\
& = & \nm{\vec q_{\sss EB} + \dfrac{\epsilon}{2} \, \lrp{\vec q_{\sss EN} \otimes \vec T_{\sss NB}^{\sss N} \otimes \vec q_{\sss EN}^{\ast} + \vec T_{\sss EN}^{\sss E}} \otimes \vec q_{\sss EB} = \vec q_{\sss EB} + \dfrac{\epsilon}{2} \, \vec T_{\sss EB}^{\sss E} \otimes \vec q_{\sss EB}}\nonumber \\
& = & \nm{\lrp{\vec q_{{\sss EN}r} + \vec q_{{\sss EN}d} \, \epsilon} \otimes \lrp{\vec q_{{\sss NB}r} + \vec q_{{\sss NB}d} \, \epsilon}}\nonumber \\
& = & \nm{\lrp{\vec q_{{\sss EN}r} \otimes \vec q_{{\sss NB}r}} + \lrp{\vec q_{{\sss EN}r} \otimes \vec q_{{\sss NB}d} + \vec q_{{\sss EN}d} \otimes \vec q_{{\sss NB}r}} \, \epsilon}\label{eq:SE3_unit_dual_quat_concatenation}
\end{eqnarray}

Unit dual quaternions comply with the orthogonality and handedness conditions required in section \ref{sec:RigidBody_bases} for rigid body motions, and hence their space \nm{\mathbb{SE}\lrp{3} = \{\vec \zeta = \qr + \qd \, \epsilon \in \mathbb{H}_d \ | \|\qr\| = 1, \langle \qr, \, \qd \rangle = 0\}} possesses group structure under dual quaternion multiplication \nm{\{\otimes : \mathbb{H}_d \times \mathbb{H}_d \rightarrow \mathbb{H}_d \ | \ \vec \zeta_a \otimes \vec \zeta_b \in \mathbb{H}_d, \forall \ \vec \zeta_a, \ \vec \zeta_b \in \mathbb{H}_d\}}. Because of the (\ref{eq:SE3_unit_dual_quat_cond1}) and (\ref{eq:SE3_unit_dual_quat_cond2}) constraints, although they have dimension eight, the special Euclidean group \nm{\mathbb{SE}\lrp{3}} defined by means of unit dual quaternions constitutes a six dimensional manifold to Euclidean space \nm{\mathbb{E}^6} called the \emph{image space of spatial displacements}, which can be visualized in \nm{\mathbb{R}^4} as follows. Expression (\ref{eq:SE3_unit_dual_quat_cond1}) defines a unit hypersphere of three dimensions, while (\ref{eq:SE3_unit_dual_quat_cond2}) defines the three dimensional hyperplane orthogonal to the normal at the point \nm{\qr} on the hypersphere. Thus, the image space consists of the hypersphere and all of its tangent spaces, which have been translated to contain the origin. Note that in this group \nm{\vec{\zeta_1}} constitutes the identity and \nm{\zetaast} the inverse.

The map from the affine representation (or homogeneous matrix) to the unit dual quaternion is surjective but not injective for the same reasons as that between the rotation matrix and the unit quaternion described in section \ref{subsec:RigidBody_rotation_rodrigues_unit_quat}, this is, the double covering of the \nm{\mathbb{SO}\lrp{3}} by the unit quaternion.

%%%%%%%%%%%%%%%%%%%%%%%%%%%%%%%%%%%%%%%%%%%%%%%%%%%%%%%%%%%%%%%%%%%%%%%%
%%%%%%%%%%%%%%%%%%%%%%%%%%%%%%%%%%%%%%%%%%%%%%%%%%%%%%%%%%%%%%%%%%%%%%%%
%%%%%%%%%%%%%%%%%%%%%%%%%%%%%%%%%%%%%%%%%%%%%%%%%%%%%%%%%%%%%%%%%%%%%%%%
% SECTION      HALF TRANSFORM VECTOR as TANGENT SPACE
%%%%%%%%%%%%%%%%%%%%%%%%%%%%%%%%%%%%%%%%%%%%%%%%%%%%%%%%%%%%%%%%%%%%%%%%
%%%%%%%%%%%%%%%%%%%%%%%%%%%%%%%%%%%%%%%%%%%%%%%%%%%%%%%%%%%%%%%%%%%%%%%%
%%%%%%%%%%%%%%%%%%%%%%%%%%%%%%%%%%%%%%%%%%%%%%%%%%%%%%%%%%%%%%%%%%%%%%%%

\section{Half Transform Vector as Tangent Space}\label{sec:RigidBody_motion_halftransform_vector}

As indicated in section \ref{subsec:algebra_lie_velocities}, the structure of the Lie algebra \nm{\mathfrak{se}\lrp{3}} can be obtained by time derivating its \nm{\mathbb{SE}\lrp{3}} Lie group constraint \nm{\vec \zeta \otimes \zetaast = \zetaast \otimes \vec \zeta = \vec{\zeta_1}}, leading to \nm{\zetaast \otimes \vec{\dot \zeta} = - \big(\zetaast \otimes \vec{\dot \zeta}\big)^{\ast}}, which indicates that \nm{\zetaast \otimes \vec{\dot \zeta}} is in fact a pure dual quaternion, as is \nm{\vec{\dot \zeta} \otimes \zetaast}. This results in the following particularizations of (\ref{eq:algebra_vE}) and (\ref{eq:algebra_vX}): 
\begin{eqnarray}
\nm{\vec \Upsilon_{\sss {EB}}^{\sss E\wedge}} & = & \nm{\vec{\dot \zeta}_{\sss {EB}} \otimes \zetaEB^{\ast} = - \zetaEB \otimes \vec{\dot \zeta}_{\sss EB}^{\ast}} \label{eq:SE3_Upsilon_space} \\
\nm{\vec \Upsilon_{\sss {EB}}^{\sss B\wedge}} & = & \nm{\zetaEB^{\ast} \otimes \vec{\dot \zeta}_{\sss {EB}} = - \vec{\dot \zeta}_{\sss EB}^{\ast} \otimes \zetaEB} \label{eq:SE3_Upsilon_body} 
\end{eqnarray}

The Lie algebra velocity \nm{\vec v^\wedge} is known as the \emph{half twist} \nm{\vec \Upsilon^\wedge}, and as shown in (\ref{eq:SE3_Upsilon_space}) and (\ref{eq:SE3_Upsilon_body}), has the structure of a pure dual quaternion because its negative coincides with its conjugate:
\neweq{\vec \Upsilon^\wedge\lrp{t} = \lrsb{0^\diamond + \vec\Upsilon_v^\diamond\lrp{t}} \in \mathbb{H}_{dp}}{eq:SE3_Upsilon_pure}

Inverting the previous equations results in the unit dual quaternion time derivative, which is linear:
\neweq{\zetaEBdot = \vec \Upsilon_{\sss {EB}}^{\sss E\wedge} \otimes \zetaEB = \zetaEB \otimes \vec \Upsilon_{\sss {EB}}^{\sss B\wedge}} {eq::SE3_Upsilon_dot}

Notice that if \nm{\vec \zeta\lrp{t_0} = \vec{\zeta_1}}, then \nm{\vec{\dot \zeta}\lrp{t_0} = \vec \Upsilon\lrp{t_0}}, and hence the pure dual quaternion \nm{\vec \Upsilon^\wedge\lrp{t_0}} provides a first order approximation of the unit dual quaternion around the identity \nm{\vec{\zeta_1}}:
\neweq{\vec \zeta\lrp{t_0 + \Deltat} \approx \vec \zeta_1 +\vec \Upsilon^\wedge\lrp{t_0} \, \Deltat}{eq:SE3_Upsilon_taylor}
\begin{center}
\begin{tabular}{lcc}
	\hline
	Concept    & Lie Theory & \nm{\mathbb{SE}\lrp{3}} \\
	\hline
	Tangent space element & \nm{\vec \tau^\wedge} & \nm{\vec \Psi^\wedge =  \lrsb{0^\diamond + \vec\Psi_v^\diamond}} \\
	Velocity element      & \nm{\vec v^\wedge} & \nm{\vec \Upsilon^\wedge =  \lrsb{0^\diamond + \vec\Upsilon_v^\diamond}} \\
	Structure             & \nm{\wedge} & pure dual quaternion \\
	\hline
\end{tabular}
\end{center}
\captionof{table}{Comparison between generic \nm{\mathbb{SE}\lrp{3}} and half transform vector as tangent space} \label{tab:Rotate_lie_halftransform_vector}

The \emph{space of pure dual quaternions} \nm{\mathfrak{se}\lrp{3} = \{\vec \Upsilon^\wedge \in \mathbb{H}_{dp} \ | \ \vec \Upsilon^\diamond \in \mathbb{D}^3\}} is hence the \emph{tangent space} of the unit dual quaternions at the identity \nm{\vec \zeta_1}, denoted as \nm{T_{\vec \zeta_1}{\mathcal M}}. The \emph{hat} \nm{\lrb{\cdot^\wedge: \mathbb{R}^6 \rightarrow \mathbb{D}^3 \rightarrow \mathfrak{se}\lrp{3} \ | \ \vec \Upsilon \rightarrow \vec \Upsilon^\wedge}} and \emph{vee} \nm{\lrb{\cdot^\vee: \mathfrak{se}\lrp{3} \rightarrow \mathbb{D}^3 \rightarrow \mathbb{R}^6 \ | \ \lrp{\vec \Upsilon^\wedge}^\vee \rightarrow \vec \Upsilon}} operators convert the half twist vector into its pure dual quaternion form, and vice versa.

If \nm{\vec \zeta\lrp{t_0} \neq \vec \zeta_1}, the tangent space needs to be transported right multiplying by \nm{\zetaEB\lrp{t_0}} (in the case of space tangent space), or left multiplying for the local space:
\begin{eqnarray}
\nm{\zetaEB\lrp{t_0 + \Deltat}} & \nm{\approx} & \nm{\zetaEB\lrp{t_0} + \lrsb{\vec \Upsilon_{\sss EB}^{\sss E\wedge} \lrp{t_0} \, \Deltat} \otimes \zetaEB\lrp{t_0} = \lrsb{\vec \zeta_1 + \vec \Upsilon_{\sss EB}^{\sss E\wedge} \lrp{t_0} \, \Deltat} \otimes \zetaEB\lrp{t_0}    }\label{eq:SE3_Upsilon_taylor_space} \\
\nm{\zetaEB\lrp{t_0 + \Deltat}} & \nm{\approx} & \nm{\zetaEB\lrp{t_0} + \zetaEB\lrp{t_0} \otimes \lrsb{\vec \Upsilon_{\sss EB}^{\sss B\wedge} \lrp{t_0} \, \Deltat} = \zetaEB\lrp{t_0} \otimes \lrsb{\vec \zeta_1 + \vec \Upsilon_{\sss EB}^{\sss B\wedge} \lrp{t_0} \, \Deltat}}\label{eq:SE3_Upsilon_taylor_body}
\end{eqnarray}

Note that the solution to the ordinary differential equation \nm{\vec{\dot x}\lrp{t} = \vec x\lrp{t} \otimes \vec \Upsilon^\wedge, \ \vec x\lrp{t} \in \mathbb{R}^8}, where \nm{\vec \Upsilon^\wedge} is constant, is \nm{\vec x\lrp{t} = \vec x\lrp{0} \, e^{\ds{\vec \Upsilon^\wedge t}}}. Based on it, assuming \nm{\vec \zeta\lrp{0} = \vec{\zeta_1}} as initial condition, and considering for the time being that \nm{\vec \Upsilon} is constant,
\neweq{\vec \zeta\lrp{t} = e^{\ds{\vec \Upsilon^\wedge t}} = \vec{\zeta_1} + \vec \Upsilon^\wedge t + \frac{\lrp{\vec \Upsilon^\wedge t}^2}{2!} + \dots + \frac{\lrp{\vec \Upsilon^\wedge t}^n}{n!} + \dots}{eq:SE3_Upsilon_exponential3}

which is indeed a pure dual quaternion and coincides with half the twist defined in section \ref{sec:RigidBody_motion_transform_vector}, as proven next based on (\ref{eq::SO3_quat_omega_dot}), (\ref{eq:SE3_unit_dual_quat_from_affine}), (\ref{eq:SE3_Upsilon_space}), (\ref{eq:SE3_time_derivative_twist_space}), and \nm{\TEBE} being a pure quaternion:
\begin{eqnarray}
\nm{\vec \Upsilon_{\sss {EB}}^{\sss E\wedge}} & = & \nm{\vec{\dot \zeta}_{\sss {EB}} \otimes \lrsb{\qEBast + \frac{\epsilon}{2} \, \lrp{\TEBE \otimes \qEB}^\ast} = \vec{\dot \zeta}_{\sss {EB}} \otimes \lrsb{\qEBast - \frac{\epsilon}{2} \, \qEBast \otimes \TEBE}} \nonumber \\
& = & \nm{\frac{\vec \omega_{\sss EB}^{\sss E\wedge}}{2} + \frac{\epsilon}{2} \, \lrsb{\vec {\dot T}_{\sss EB}^{\sss E} + \TEBE \otimes \frac{\vec \omega_{\sss EB}^{\sss E\wedge}}{2} - \frac{\vec \omega_{\sss EB}^{\sss E\wedge}}{2} \otimes \TEBE} = \frac{\vec \omega_{\sss EB}^{\sss E\wedge}}{2} + \frac{\epsilon}{2} \, \lrsb{\vec {\dot T}_{\sss EB}^{\sss E} - \frac{\vec \omega_{\sss EB}^{\sss E\wedge}}{2} \otimes \TEBE}} \nonumber \\
& = & \nm{\frac{1}{2} \, \lrp{\vec \omega_{\sss EB}^{\sss E\wedge} + \epsilon \ \vec \nu_{\sss EB}^{\sss E\wedge}} = \frac{\vec \xi_{\sss {EB}}^{\sss E\wedge}}{2}} \label{eq:SE3_Upsilon_space_proof}
\end{eqnarray}

A similar process employing (\ref{eq:SE3_Upsilon_body}) and (\ref{eq:SE3_time_derivative_twist_body}) leads to:
\neweq{\vec \Upsilon_{\sss {EB}}^{\sss B\wedge} = \frac{1}{2} \, \lrp{\vec \omega_{\sss EB}^{\sss B\wedge} + \epsilon \ \vec \nu_{\sss EB}^{\sss B\wedge}} = \frac{\vec \xi_{\sss {EB}}^{\sss B\wedge}}{2}}{eq:SE3_Upsilon_body_proof}

Remembering that so far \nm{\vec \Upsilon^\wedge} is constant, (\ref{eq:SE3_Upsilon_exponential3}) means that any rigid body motion \nm{\vec \zeta\lrp{t} = e^{\ds{\vec \Upsilon^\wedge t}}} can be realized by maintaining a constant half twist \nm{\vec \Upsilon^\wedge \in \mathbb{H}_{dp}} for a given time \emph{t}. The vectors \nm{\vec n = \vec \omega \, t / \| \vec \omega \, t \| = \vec r / \|\vec r\|} and \nm{\vec k = \vec \nu \, t / \| \vec \nu \, t \| = \vec s / \|\vec s\|} indicate the half twist directions, while \nm{\theta = \phi / 2 = \|\vec r\| / 2} and \nm{\rho / 2 = \|\vec s\| / 2} represent the half twist magnitudes, respectively. This enables the definition of the \emph{half transform vector} \nm{\vec \Psi}, also known as the \emph{exponential coordinates} of the \nm{\mathcal M} motion, as
\neweq{\vec \Psi = \vec \Upsilon \, t = \frac{1}{2} \, \lrsb{\vec \nu \, t \ \ \ \vec \omega \, t}^T = \frac{1}{2} \, \lrsb{\vec s \ \ \vec r}^T = \frac{1}{2} \,  \lrsb{ \vec k \, \rho \ \ \ \vec n \, \phi}^T = \frac{\vec \tau}{2} \in \mathbb{R}^6}{eq:SE3_halftransform_vector_definition}

Note that the half transform vector \nm{\vec \Psi} belongs to the tangent space as it is a multiple of the half twist \nm{\vec \Upsilon \in \mathfrak{se}\lrp{3}}, and hence tends to coincide with it as time tends to zero. The \emph{exponential map} \nm{\lrb{exp\lrp{} : \mathfrak{se}\lrp{3} \rightarrow \mathbb{SE}\lrp{3} \ | \ \mathcal M = exp\lrp{\vec \Psi^\wedge}}} and its capitalized form \nm{\lrb{Exp\lrp{} : \mathbb{R}^6 \rightarrow \mathbb{SE}\lrp{3} \ | \ \mathcal M = Exp\lrp{\vec \Psi}}} wrap the half transform vector around the special Euclidean group. However, the half twist \nm{\vec \Upsilon^\wedge \lrp{t}} in fact is not required to be constant. Given a rigid body motion represented by its unit dual quaternion \nm{\vec \zeta \in \mathbb{SE}\lrp{3}}, it can be proven that there exists a not necessarily unique half transform vector \nm{\vec \Psi = \vec \tau / 2 = \lrsb{\vec s \ \ \vec r}^T / 2 = \lrsb{\vec k \, \rho \ \ \ \vec n \, \phi}^T / 2} such that \nm{\vec \zeta = e^{\vec \Psi^\wedge}}. The exponential map is made up by a combination of (\ref{eq:SO3_quat_exponential2}), (\ref{eq:SE3_twist_exponential4_a}), (\ref{eq:SE3_twist_exponential4_b}), and (\ref{eq:SE3_unit_dual_quat_from_affine}).

The \emph{logarithmic map} \nm{\lrb{log\lrp{} : \mathbb{SE}\lrp{3} \rightarrow \mathfrak{se}\lrp{3} \ | \ \vec \Psi^\wedge = log\lrp{\mathcal M}}} and its capitalized version \nm{\lrb{Log\lrp{} : \mathbb{SE}\lrp{3} \rightarrow \mathbb{R}^6 \ | \ \vec \Psi = Log\lrp{\mathcal M}}} convert unit dual quaternions into half transform vectors. It is composed by (\ref{eq:SE3_unit_dual_quat_to_affine_q}), (\ref{eq:SO3_theta_from_quat}), and (\ref{eq:SO3_rotvec_from_quat}) for the rotation part, and (\ref{eq:SE3_unit_dual_quat_to_affine_T}) together with (\ref{eq:SE3_twist_logarithm}) for the translation part.

As the vector \nm{\vec \Upsilon} represents half the twist \nm{\vec \xi}, it is possible to adjust expressions (\ref{eq:SE3_Upsilon_space}), (\ref{eq:SE3_Upsilon_body}), and (\ref{eq::SE3_Upsilon_dot}):
\begin{eqnarray}
\nm{\vec \xi_{\sss {EB}}^{\sss E\wedge}} & = & \nm{2 \ \vec{\dot \zeta}_{\sss {EB}} \otimes \zetaEB^{\ast}} \label{eq:SE3_dual_quat_xi_space} \\
\nm{\vec \xi_{\sss {EB}}^{\sss B\wedge}} & = & \nm{2 \ \zetaEB^{\ast} \otimes \vec{\dot \zeta}_{\sss {EB}}} \label{eq:SE3_dual_quat_xi_body} \\
\nm{\zetaEBdot} & = & \nm{\dfrac{1}{2} \; \vec \xi_{\sss {EB}}^{\sss E\wedge} \otimes \zetaEB = \dfrac{1}{2} \; \zetaEB \otimes \vec \xi_{\sss {EB}}^{\sss B\wedge}} \label{eq::SE3_dual_quat_xi_dot} 
\end{eqnarray}

%%%%%%%%%%%%%%%%%%%%%%%%%%%%%%%%%%%%%%%%%%%%%%%%%%%%%%%%%%%%%%%%%%%%%%%%
%%%%%%%%%%%%%%%%%%%%%%%%%%%%%%%%%%%%%%%%%%%%%%%%%%%%%%%%%%%%%%%%%%%%%%%%
%%%%%%%%%%%%%%%%%%%%%%%%%%%%%%%%%%%%%%%%%%%%%%%%%%%%%%%%%%%%%%%%%%%%%%%%
% SECTION      SCREW as TANGENT SPACE
%%%%%%%%%%%%%%%%%%%%%%%%%%%%%%%%%%%%%%%%%%%%%%%%%%%%%%%%%%%%%%%%%%%%%%%%
%%%%%%%%%%%%%%%%%%%%%%%%%%%%%%%%%%%%%%%%%%%%%%%%%%%%%%%%%%%%%%%%%%%%%%%%
%%%%%%%%%%%%%%%%%%%%%%%%%%%%%%%%%%%%%%%%%%%%%%%%%%%%%%%%%%%%%%%%%%%%%%%%

\section{Screw as Tangent Space}\label{sec:RigidBody_motion_screw}

In the chapter \ref{cha:Rotate} analysis of rotational motion there exists two different representations for the tangent space \nm{\mathfrak{so}\lrp{3}}. The first is the skew-symmetric angular velocity \nm{\vec \omega^\wedge = \vec \omega^\wedge}, which converts into the rotation vector \nm{\vec r^\wedge} when applied during a certain amount of time (section \ref{sec:RigidBody_rotation_rotv}), and represents the origin of the exponential map \nm{exp\lrp{\vec r^\wedge}} (\ref{eq:SO3_rotv_exponential2}) that transforms it into the rotation matrix \nm{\vec R} (section \ref{sec:RigidBody_rotation_dcm}). The second is the pure quaternion half angular velocity \nm{\vec \Omega^\wedge = \lrsb{0 \ \ \vec \omega / 2}^T}, which converts into the half rotation vector \nm{\vec h^\wedge = \lrsb{0 \ \ \vec r / 2}^T} (section \ref{sec:RigidBody_rotation_halfrotv}), and represents the origin of the exponential map \nm{exp\lrp{\vec h^\wedge} = exp\lrp{\vec r^\wedge / 2}} (\ref{eq:SO3_quat_exponential2}) that transforms it into the unit quaternion \nm{\vec q} (section \ref{sec:RigidBody_rotation_rodrigues}). Note that both representations of the tangent space \nm{\mathfrak{so}\lrp{3}} are so similar that for all practical purposes they are considered the same, resulting in two versions of the exponential map that convert the rotation vector \nm{\vec r} into either the rotation matrix \nm{\vec R} or the unit quaternion \nm{\vec q}.

So far the rigid body motion looks similar. There are two \nm{\mathfrak{se}\lrp{3}} velocities, the twist \nm{\vec \xi^\wedge} and the pure dual quaternion half twist \nm{\vec \Upsilon^\wedge}, which convert into the transform vector \nm{\vec \tau^\wedge} and the half transform vector \nm{\vec \Psi^\wedge} (sections \ref{sec:RigidBody_motion_transform_vector} and \ref{sec:RigidBody_motion_halftransform_vector}) when applied during an amount of time, and constitute the origins of the exponential maps that transform them into the homogeneous matrix \nm{\vec M} (section \ref{sec:RigidBody_motion_homogeneous}), the affine representation (section \ref{sec:RigidBody_motion_affine}), or the unit dual quaternion \nm{\vec \zeta} (section \ref{sec:RigidBody_motion_unit_dual_quaternion}). As in the rotation case, both \nm{\mathfrak{se}\lrp{3}} representations are so similar that for all practical purposes they are considered the same and can be interchanged in the various exponential maps.

There exists however an additional \nm{\mathbb{SE}\lrp{3}} parameterization, which also belongs to its tangent space \nm{\mathfrak{se}\lrp{3}}, that enables the definition of a different exponential map into the unit dual quaternion that explicitly separates the influence of the motion direction from that of its magnitude, and is also indispensable for the rigid body motion powers, linear interpolation, and perturbations introduced in section \ref{sec:RigidBody_motion_algebra}.

The origin of the screw \nm{\vec S} lies in the fact that every rigid body motion can be realized by a rotation about an axis combined with a translation parallel to that same axis \cite{Murray1994}. The rotation and translation can be executed simultaneously or one after another without modifying the result. It is however necessary to remark that in this case the axis, defined in section \ref{sec:algebra_points_and_vectors}, does not necessarily pass though the origin of the frame characterizing the rigid body, in contrast to previous representations of rigid body motions in which the rotating axis, more appropriately called rotating direction, in all cases passed through the origin. Note that according to section \ref{sec:algebra_points_and_vectors}, an axis is represented by \nm{\lrp{\vec n, \, \vec m}} and has four degrees of freedom, while if restricted to passing through the origin \nm{\vec m = \vec 0} and the degrees of freedom are two. The line point closest to the origin responds to \nm{\vec p_{\perp} = \widehat{\vec n} \, \vec m}.

A \emph{screw} \nm{\{\vec S = \lrsb{\vec n \ \ \vec m \ \ h \ \ \phi}^T \in \mathbb{R}^8 \ | \ \vec n, \vec m \in \mathbb{R}^3, h, \phi \in \mathbb{R}\}} consists of an axis \nm{\lrp{\vec n, \, \vec m}}, a pitch \emph{h}, and a magnitude \nm{\phi} \cite{Murray1994}. As all other \nm{\mathbb{SE}\lrp{3}} representations, it contains six degrees of freedom as it shares the line redundancies described in section \ref{sec:algebra_points_and_vectors}, this is, \nm{\|\vec n\| = 1} and \nm{\vec n^T \, \vec m = 0}. It represents a rotation by an amount \nm{\phi} about the axis \nm{\lrp{\vec n, \, \vec m}} combined by a translation by an amount \nm{d = h \, \phi} parallel to axis \nm{\lrp{\vec n, \, \vec m}}. If \nm{h = \infty}, the corresponding screw motion consists of a pure translation along the axis of the screw by a distance \nm{\phi}. Note that \nm{\vec n} and \nm{\phi} are indeed the direction and magnitude of the rotation vector \nm{\vec r = \vec n \, \phi} defined by (\ref{eq:SO3_rotv_definition}).

Given a rigid body motion represented by the combination of rotation and translation vectors \nm{\lrp{\vec r = \vec n \, \phi, \, \vec T}}, the map converting it into a screw has two versions:
\begin{itemize}
\item \nm{\vec r \neq \vec 0}. If the motion contains both rotation and translation components:
\begin{eqnarray}
\nm{\phi}   & = & \nm{\|\vec r \| = \phi}\label{eq:SE_screw_magnitude} \\
\nm{\vec n} & = & \nm{\dfrac{\vec r}{\|\vec r\|}}\label{eq:SE3_screw_line} \\
\nm{\vec m} & = & \nm{\widehat{\vec p}_{\perp} \, \vec n = \frac{1}{2} \lrsb{\widehat{\vec T} \, \vec n + \cot \dfrac{\phi}{2} \, {\lrp{\widehat{\vec n} \, \vec T} \times \vec n}}}\label{eq:SE3_screw_moment} \\
\nm{d}      & = & \nm{\vec T^T \, \vec n}\label{eq:SE3_screw_displacement} \\
\nm{h}      & = & \nm{\dfrac{d}{\phi}}\label{eq:SE3_screw_pitch}  
\end{eqnarray}

\item \nm{\vec r = \vec 0}. If the motion is only a translation, the screw definition changes so it contains an \nm{\infty} pitch, a magnitude equal to the translation amount, and an axis in the direction of {\nm{\vec T}} that passes through the origin. The displacement definition does not change though.
\begin{eqnarray}
\nm{h}      & = & \nm{\infty}\label{eq:SE3_screw_pitch_norotation} \\
\nm{\phi}   & = & \nm{\|\vec T \|}\label{eq:SE_screw_magnitude_norotation} \\
\nm{\vec n} & = & \nm{\vec T / \phi} \label{eq:SE3_screw_line_norotation} \\
\nm{\vec m} & = & \nm{\vec 0}\label{eq:SE3_screw_moment_norotation} \\
\nm{d}      & = & \nm{\vec T^T \, \vec n} \label{eq:SE3_screw_displacement_norotation} 
\end{eqnarray}
\end{itemize}

The opposite map, which provides the rotation and translation vectors from a screw, also has two versions:
\begin{itemize}
\item \nm{h \neq \infty}. The screw contains both translation and rotation components:
\begin{eqnarray}
\nm{\vec T} & = & \nm{\vec p_{\perp} - \sin \phi \ \widehat{\vec n} \, \vec p_{\perp} - \cos \phi \, \vec p_{\perp} + d \, \vec n}\label{eq:SE3_screw_translation} \\
\nm{\vec r} & = & \nm{\vec n \, \phi} \label{eq:SE3_screw_rotation}
\end{eqnarray}

\item \nm{h = \infty}. The screw does not rotate:
\begin{eqnarray}
\nm{\vec T} & = & \nm{\vec n \, \phi} \label{eq:SE3_screw_translation_norotation} \\
\nm{\vec r} & = & \nm{\vec 0} \label{eq:SE3_screw_rotation_norotation} 
\end{eqnarray}
\end{itemize}

It is however the \emph{exponential map} between the screw and the unit dual quaternion the one that provides a different perspective to the motion of a rigid body. It is built based on the expressions for the axis moment (\ref{eq:SE3_screw_moment}) and the unit dual quaternion (\ref{eq:SE3_unit_dual_quat_from_affine}), first as the sum of two quaternions (\ref{eq:SE3_screw_dual_first}) and next as that of a dual number plus a dual vector (\ref{eq:SE3_screw_dual}):
\begin{eqnarray}
\nm{\vec \zeta} & = & \nm{\lrp{\cos \frac{\phi}{2} + \sin \frac{\phi}{2} \, \vec n} + \lrsb{- \frac{d}{2} \, \sin \frac{\phi}{2} + \sin \frac{\phi}{2} \, \vec m + \frac{d}{2} \, \cos \frac{\phi}{2} \, \vec n} \, \epsilon} \label{eq:SE3_screw_dual_first} \\
& = & \nm{\qr + \qd \, \epsilon = \lrsb{q_{0 r} \ \ \vec q_{vr}}^T + \lrsb{q_{0d} \ \ \vec q_{vd}}^T \epsilon} \nonumber \\
& = & \nm{\lrsb{\cos \frac{\phi}{2} - \frac{d}{2} \, \sin \frac{\phi}{2} \, \epsilon} + \lrsb{\sin \frac{\phi}{2} \, \vec n + \lrp{\sin \frac{\phi}{2} \, \vec m + \frac{d}{2} \, \cos \frac{\phi}{2} \, \vec n} \, \epsilon}}\label{eq:SE3_screw_dual}
\end{eqnarray}

This last expression can be modified based the application of the dual number Taylor expansion (\ref{eq:SE3_dual_number_taylor}) to the sine and cosine:
\begin{eqnarray}
\nm{\cos d^{\diamond}} & = & \nm{\cos \lrp{x + y \, \epsilon} = \cos x - y \, \epsilon \, \sin x}\label{eq:SE3_unit_dual_quat_cos} \\
\nm{\sin d^{\diamond}} & = & \nm{\sin \lrp{x + y \, \epsilon} = \sin x + y \, \epsilon \, \cos x}\label{eq:SE3_unit_dual_quat_sin}
\end{eqnarray}

This results in:
\neweq{\vec \zeta = exp\lrp{\vec S} = \cos \frac{\phi + d \, \epsilon}{2} + \lrp{\vec n + \vec m \, \epsilon} \cdot \sin \frac{\phi + d \, \epsilon}{2} = \cos \frac{\phi^{\diamond}}{2} + {\vec{nm}}^{\diamond} \cdot \sin \frac{\phi^{\diamond}}{2} = \cos \theta^{\diamond} + {\vec{nm}}^{\diamond} \cdot \sin \theta^{\diamond}}{eq:SE3_unit_dual_quat1}

where \nm{{\vec{nm}}^{\diamond} = \vec n + \vec m \, \epsilon} is a unit dual vector, this is, one in which its real part is a unit vector that its orthogonal to its dual part, and \nm{\theta^{\diamond} = \frac{\phi^\diamond}{2} = \frac{\phi + d \, \epsilon}{2}} is a dual number. In fact unit dual quaternions can always be written as (\ref{eq:SE3_unit_dual_quat1}), which is the equivalent to (\ref{eq:SO3_quat_unit}) for unit quaternions.

A process in some aspects similar to that described in section \ref{sec:RigidBody_rotation_halfrotv} proves that (\ref{eq:SE3_unit_dual_quat1}) is indeed the exponential map \nm{\{exp() : \mathfrak{se}\lrp{3} \rightarrow \mathbb{SE}\lrp{3} | \ \vec S \in \mathbb{R}^8 \rightarrow exp\lrp{\vec S} \in \mathbb{H}_d\}}, which transforms screws into unit dual quaternions. Note that to do so, it is first necessary to represent the screw as the combination of a unit dual vector \nm{{\vec{nm}}^{\diamond} = \vec n + \vec m \, \epsilon} representing the screw axis and a dual number \nm{\theta^{\diamond} = \frac{\phi^\diamond}{2} = \frac{\phi + d \, \epsilon}{2}} containing the rotation angle and translation distance about the screw axis. This map algebraically separates the line information of the screw axis from the pitch and angle values, where the dual vector \nm{\vec{nm}^{\diamond}} represents the axis of the screw motion with its direction vector and the dual angle \nm{\theta^\diamond = \frac{\phi^\diamond}{2}} contains both the translation length and the angle of rotation \cite{Busam2017}. If there is no rotation (\nm{h = \infty}), the resulting unit dual quaternion responds to \nm{\vec \zeta = \qr + \qd \, \epsilon = \vec q_1 + \frac{\phi}{2} \, \vec n \, \epsilon}.

Obtainment of the logarithmic map \nm{\{log() : \mathbb{SE}\lrp{3} \rightarrow \mathfrak{se}\lrp{3} | \ \vec \zeta \in \mathbb{H}_d \rightarrow \vec S \in \mathbb{R}^8\}} is now straightforward, resulting in expressions (\ref{eq:SE3_unit_dual_quat_log_phi}) through (\ref{eq:SE3_unit_dual_quat_log_m}) for the case of rotation plus translation (\nm{\qr \neq \vec q_1}):
\begin{eqnarray}
\nm{\phi} & = & \nm{2 \, \arctan\frac{\|\vec q_{vr}\|}{q_{0 r}}}\label{eq:SE3_unit_dual_quat_log_phi} \\
\nm{\vec n} & = & \nm{\frac{\vec q_{vr}}{\|\vec q_{vr}\|}}\label{eq:SE3_unit_dual_quat_log_l} \\
\nm{d} & = & \nm{- 2 \, \frac{q_{0 d}}{\|\vec q_{vr}\|}}\label{eq:SE3_unit_dual_quat_log_d} \\
\nm{\vec m} & = & \nm{\lrp{\vec q_{vd} - \frac{d \, \, q_{0 r}}{2} \, \vec n} \, {\|\vec q_{vr}\|}^{-1}} \label{eq:SE3_unit_dual_quat_log_m} 
\end{eqnarray}

In the no rotation case (\nm{\qr = \vec q_1}), the logarithmic map changes as described above. Note that both the exponential and logarithmic maps share the same surjective traits as those between the rotation vector and unit quaternion described in section \ref{subsec:RigidBody_rotation_rodrigues_unit_quat}.

Although inverting the motion by means of the screw is straightforward,
\neweq{\vec S^{-1} = \lrsb{\vec n \ \ \vec m \ \ h \ \ - \phi}^T}{eq:SE3_screw_inversion} 

the different \nm{\mathbb{SE}\lrp{3}} actions (concatenation, point rotation, vector rotation) are complex and rarely used.

%%%%%%%%%%%%%%%%%%%%%%%%%%%%%%%%%%%%%%%%%%%%%%%%%%%%%%%%%%%%%%%%%%%%%%%%
%%%%%%%%%%%%%%%%%%%%%%%%%%%%%%%%%%%%%%%%%%%%%%%%%%%%%%%%%%%%%%%%%%%%%%%%
%%%%%%%%%%%%%%%%%%%%%%%%%%%%%%%%%%%%%%%%%%%%%%%%%%%%%%%%%%%%%%%%%%%%%%%%
% SECTION      RIGID BODY MOTION ALGEBRAIC OPERATIONS
%%%%%%%%%%%%%%%%%%%%%%%%%%%%%%%%%%%%%%%%%%%%%%%%%%%%%%%%%%%%%%%%%%%%%%%%
%%%%%%%%%%%%%%%%%%%%%%%%%%%%%%%%%%%%%%%%%%%%%%%%%%%%%%%%%%%%%%%%%%%%%%%%
%%%%%%%%%%%%%%%%%%%%%%%%%%%%%%%%%%%%%%%%%%%%%%%%%%%%%%%%%%%%%%%%%%%%%%%%

\section{Rigid Body Motion Algebraic Operations}\label{sec:RigidBody_motion_algebra}

As in the case of pure rotations described in section \ref{sec:RigidBody_rotation_algebra}, the basic algebraic operations of addition, subtraction, multiplication, division, and exponentiation are not defined for objects of the special Euclidean group \nm{\mathbb{SE}\lrp{3}}. However, all rigid body motion representations are closed under a given operation that represents the concatenation of transformations, and define not only an identity transformation that represents the lack of motion, but also an inverse operation representing the opposite movement. The concatenation of transformations and the identity and inverse operations enable the definition of the power, exponential and logarithmic operators (section \ref{subsec:RigidBody_motion_algebra_exp_log}), the screw linear interpolation (section \ref{subsec:RigidBody_motion_algebra_sclerp}), and the perturbations together with the plus and minus operators (section \ref{subsec:RigidBody_motion_algebra_plus_minus}).

%%%%%%%%%%%%%%%%%%%%%%%%%%%%%%%%%%%%%%%%%%%%%%%%%%%%%%%%%%%%%%%%%%%%%%%%
%%%%%%%%%%%%%%%%%%%%%%%%%%%%%%%%%%%%%%%%%%%%%%%%%%%%%%%%%%%%%%%%%%%%%%%%
% SUBSECTION      POWERS, EXPONENTIALS AND LOGARITHMS
%%%%%%%%%%%%%%%%%%%%%%%%%%%%%%%%%%%%%%%%%%%%%%%%%%%%%%%%%%%%%%%%%%%%%%%%
%%%%%%%%%%%%%%%%%%%%%%%%%%%%%%%%%%%%%%%%%%%%%%%%%%%%%%%%%%%%%%%%%%%%%%%%

\subsection{Powers, Exponentials and Logarithms}\label{subsec:RigidBody_motion_algebra_exp_log}

Any rigid body motion can be executed by rotating an angle \nm{\phi} about a certain fixed axis \nm{\lrp{\vec n, \, \vec m}} combined with a translation of a distance \nm{d = h \cdot \phi} along that same axis, resulting in the screw \nm{\vec S = \lrsb{\vec n \ \ \vec m \ \ h \ \ \phi}^T} (section \ref{sec:RigidBody_motion_screw}). As the rotation and translation can be executed simultaneously or one after another, taking a fraction of a screw results in \nm{t \, \vec S = t \, \lrsb{\vec n \ \ \vec m \ \ h \ \ \phi}^T = \lrsb{\vec n \ \ \vec m \ \ h \ \ t \phi}^T \, \forall \, t \in \mathbb{R}, \vec S \in \mathfrak{se}\lrp{3}}.

The exponential map defined in (\ref{eq:SE3_unit_dual_quat1}) is named that way because it complies with the behavior of the real exponential function \nm{exp^b\lrp{a} = exp\lrp{a \cdot b} \, \forall \, a, \, b \in \mathbb{R}}. As such, the exponential function \nm{\{exp(): \mathfrak{se}\lrp{3} \times \mathbb{R} \rightarrow \mathbb{SE}\lrp{3} \ | \ \vec S \in \mathfrak{se}\lrp{3}, t \in \mathbb{R} \rightarrow {\mathcal M}^t = exp\lrp{t \, \vec S} \in \mathbb{SE}\lrp{3}\}} is defined as:
\neweq{\vec \zeta^t\lrp{\vec S} = \vec \zeta\lrp{t \, \vec S} = exp\lrp{t \, \vec S} = \cos \frac{t \, \phi^\diamond}{2} + {\vec{nm}}^{\diamond} \cdot \sin \frac{t \, \theta^\diamond}{2} = \cos \frac{t \, \phi + t \, d \, \epsilon}{2} + \lrp{\vec n + \vec m \, \epsilon} \cdot \sin \frac{t \, \phi + t \, d \, \epsilon}{2}}{eq:SE3_dual_exponential_fraction}

In a similar way, the logarithmic map defined in (\ref{eq:SE3_unit_dual_quat_log_phi}) through (\ref{eq:SE3_unit_dual_quat_log_m}) also complies with the behavior of the real logarithmic function \nm{b \cdot log\lrp{a} = log\lrp{a^b} \, \forall \, a, \, b \in \mathbb{R}}. As such, the logarithmic function \nm{\{log : \mathbb{SE}\lrp{3} \times \mathbb{R} \rightarrow \mathfrak{se}\lrp{3} \ | \ \mathcal M \in \mathbb{SE}\lrp{3}, t \in \mathbb{R} \rightarrow t \, \vec S = log\lrp{{\mathcal M}^t} \in \mathfrak{se}\lrp{3}\}} is defined as:
\neweq{log\lrp{\vec \zeta^t\lrp{\vec S}} = log\lrp{exp\lrp{t \, \vec S}} = t \, log\lrp{\vec \zeta\lrp{\vec S}} = t \, log\lrp{exp\lrp{\vec S}} = t \, \vec S}{eq:SE3_quat_logarithmic_fraction} 

It is important to remark that although other exponential maps have been defined, with inputs either the transform vector (\ref{eq:SE3_twist_exponential4_a}) or the half transform vector, they can not be employed in the exponential function, as the multiple of a transform vector \nm{t \, \vec \tau = \lrp{t \, \vec s, \, t \, \vec r}} does not result in a uniform movement and hence its associated motion does not coincide with that of the same multiple of the screw \nm{t \, \vec S}.

%%%%%%%%%%%%%%%%%%%%%%%%%%%%%%%%%%%%%%%%%%%%%%%%%%%%%%%%%%%%%%%%%%%%%%%%
%%%%%%%%%%%%%%%%%%%%%%%%%%%%%%%%%%%%%%%%%%%%%%%%%%%%%%%%%%%%%%%%%%%%%%%%
% SUBSECTION      SCREW LINEAR INTERPOLATION
%%%%%%%%%%%%%%%%%%%%%%%%%%%%%%%%%%%%%%%%%%%%%%%%%%%%%%%%%%%%%%%%%%%%%%%%
%%%%%%%%%%%%%%%%%%%%%%%%%%%%%%%%%%%%%%%%%%%%%%%%%%%%%%%%%%%%%%%%%%%%%%%%

\subsection{Screw Linear Interpolation}\label{subsec:RigidBody_motion_algebra_sclerp}

Given two rigid body motions \nm{\mathcal M_0, \, \mathcal M_1 \in \mathbb{SE}\lrp{3}}, \emph{screw linear interpolation} (\hypertt{ScLERP}) is an extension of \hypertt{SLERP} (section \ref{subsec:RigidBody_rotation_algebra_slerp}) that obtains a motion function \nm{\mathcal M\lrp{t}, \, t \in \mathbb{R}} that linearly interpolates from \nm{\mathcal M\lrp{0} = \mathcal M_0} to \nm{\mathcal M\lrp{1} = \mathcal M_1} in such a way that the motion occurs with constant rotation and translation velocities \cite{Jia2013}.

If employing unit dual quaternions, \nm{\Delta \vec \zeta} is according to (\ref{eq:SE3_unit_dual_quat_concatenation}) the full motion required to go from \nm{\vec \zeta_0} to \nm{\vec \zeta_1}, such that \nm{\vec \zeta_1 = \vec \zeta_0 \otimes \Delta \vec \zeta}, from where \nm{\Delta \vec \zeta = \vec \zeta_0^{\ast} \otimes \vec \zeta_1}. The corresponding screw is then \nm{\Delta \vec S = \lrsb{\vec n \ \ \vec m \ \ h \ \ \Delta \phi}^T = log\lrp{\Delta \vec \zeta}}. The unit dual quaternion corresponding to a fraction of the full screw magnitude \nm{\delta \phi = t \, \Delta \phi} is the following:
\begin{eqnarray}
\nm{\delta \vec \zeta} & = & \nm{exp\big(\vec S \lrp{\vec n, \, \vec m, \, h, \, \delta \phi}\big) = exp\big(t \cdot \vec S \lrp{\vec n, \, \vec m, \, h, \, \Delta \phi}\big) = exp\lrp{t \, \Delta \vec S}}\nonumber \\
& = & \nm{exp\big(t \log\lrp{\Delta \vec \zeta}\big) = exp\big(t \, log\lrp{\vec \zeta_0^{\ast} \otimes \vec \zeta_1}\big) = \lrp{\vec \zeta_0^{\ast} \otimes \vec \zeta_1}^t}\label{eq:SE3_interp_partial}
\end{eqnarray}

The interpolated unit dual quaternion is hence the following, which relies on (\ref{eq:SE3_dual_exponential_fraction}) for its solution:
\neweq{\vec \zeta\lrp{t} = \vec \zeta_0 \otimes \lrp{\vec \zeta_0^{\ast} \otimes \vec \zeta_1}^t}{eq:SE3_interp}

The restrictions described in section \ref{subsec:RigidBody_rotation_algebra_slerp} intended to ensure that the rotation is executed following the shortest path are also applicable in this case.

%%%%%%%%%%%%%%%%%%%%%%%%%%%%%%%%%%%%%%%%%%%%%%%%%%%%%%%%%%%%%%%%%%%%%%%%
%%%%%%%%%%%%%%%%%%%%%%%%%%%%%%%%%%%%%%%%%%%%%%%%%%%%%%%%%%%%%%%%%%%%%%%%
% SUBSECTION      PLUS AND MINUS OPERATORS
%%%%%%%%%%%%%%%%%%%%%%%%%%%%%%%%%%%%%%%%%%%%%%%%%%%%%%%%%%%%%%%%%%%%%%%%
%%%%%%%%%%%%%%%%%%%%%%%%%%%%%%%%%%%%%%%%%%%%%%%%%%%%%%%%%%%%%%%%%%%%%%%%

\subsection{Plus and Minus Operators}\label{subsec:RigidBody_motion_algebra_plus_minus}

A perturbed rigid body motion \nm{\widetilde{\mathcal{M}} \in \mathbb{SE}\lrp{3}} can always be expressed as the composition of the unperturbed motion \nm{\mathcal M} with a (usually) small perturbation \nm{\Delta \mathcal{M}}. Perturbations can be specified either at the local or body frame \nm{\FB}, this is, at the local vector space tangent to \nm{\mathbb{SE}\lrp{3}} at the actual pose, in which case they are known as \emph{local perturbations}. They can also be specified at the global frame \nm{\FE}, which coincides with the vector space tangent to \nm{\mathbb{SE}\lrp{3}} at the origin; in this case they are known as \emph{global perturbations}. Local perturbations appear on the right hand side of the motion composition, resulting in \nm{\widetilde{\mathcal M} = \mathcal M \circ \Delta \mathcal{M}^{\sss B}}, while global ones appear to the left, hence \nm{\widetilde{\mathcal M} = \Delta\mathcal{M}^{\sss E} \circ \mathcal M}.

The \emph{plus} and \emph{minus operators} are introduced in section \ref{sec:algebra_lie} and enable operating with increments of the nonlinear \nm{\mathbb{SE}\lrp{3}} manifold expressed in the linear tangent vector space \nm{\mathfrak{se}\lrp{3}}. There exist right (\nm{\oplus, \, \ominus}) or left (\nm{\boxplus, \, \boxminus}) versions depending on whether the increments are viewed in the local frame (right) or the global one (left). It is important to remark that although perturbations and the plus and left operators are best suited to work with small motion changes (perturbations), the expressions below are generic and work just the same no matter the size of the perturbation.

The right plus operator \nm{\{\oplus : \mathbb{SE}\lrp{3} \times \mathfrak{se}\lrp{3} \rightarrow \mathbb{SE}\lrp{3} \, | \, \widetilde{\mathcal M} = \mathcal M \oplus \ \Delta \vec \tau^{\sss B} = \mathcal M \circ Exp\lrp{\Delta \vec \tau^{\sss B}}\}} produces a motion element \nm{\widetilde{\mathcal M}} resulting from the composition of a reference motion \nm{\mathcal M} with an often small motion \nm{\Delta \vec \tau^{\sss B}}, contained in the tangent space to the reference motion \nm{\mathcal M}, this is, in the local space. The left plus operator \nm{\{\boxplus : \mathfrak{se}\lrp{3} \times \mathbb{SE}\lrp{3} \rightarrow \mathbb{SE}\lrp{3} \, | \, \widetilde{\mathcal M} = \Delta \vec \tau^{\sss E} \boxplus \mathcal M = Exp\lrp{\Delta \vec \tau^{\sss E}} \circ \mathcal M\}} is similar but the often small motion \nm{\Delta \vec \tau^{\sss E}} is contained in the tangent space at the identify or global space. The expressions shown below are valid up to the first coverage of \nm{\mathbb{SE}\lrp{3}}, this is, \nm{\phi < \pi}. In the cases of homogeneous matrix and unit dual quaternion, the plus operator is defined as:
\begin{eqnarray}
\nm{\widetilde{\vec M}}     & = & \nm{\vec M \oplus \Delta \vec \tau^{\sss B} = \vec M \ Exp\lrp{\Delta \vec \tau^{\sss B}} = \vec M \ \Delta \vec M^{\sss B}}\label{eq:SE3_homogeneous_plus} \\
\nm{\widetilde{\vec \zeta}} & = & \nm{\vec \zeta \oplus \Delta \vec \tau^{\sss B} = \vec \zeta \otimes Exp\lrp{\Delta \vec \tau^{\sss B} / 2} = \vec \zeta \otimes \Delta \vec \zeta^{\sss B}}\label{eq:SE3_dual_quat_plus} \\
\nm{\widetilde{\vec M}}     & = & \nm{\Delta \vec \tau^{\sss E} \boxplus \vec M = Exp\lrp{\Delta \vec \tau^{\sss E}} \ \vec M = \Delta \vec M^{\sss E} \ \vec M}\label{eq:SE3_homogeneous_plus_left} \\
\nm{\widetilde{\vec \zeta}} & = & \nm{\Delta \vec \tau^{\sss E} \boxplus \vec \zeta = Exp\lrp{\Delta \vec \tau^{\sss E} / 2} \otimes \vec \zeta = \Delta \vec \zeta^{\sss E} \otimes \vec \zeta}\label{eq:SE3_dual_quat_plus_left}
\end{eqnarray}

The right minus operator \nm{\{\ominus : \mathbb{SE}\lrp{3} \times \mathbb{SE}\lrp{3} \rightarrow \mathfrak{se}\lrp{3} \, | \, \Delta \vec \tau^{\sss B} = \widetilde{\mathcal M} \ominus \mathcal M = Log\big(\mathcal M^{-1} \circ \widetilde{\mathcal M}\big)\}}, as well as the left \nm{\{\boxminus : \mathbb{SE}\lrp{3} \times \mathbb{SE}\lrp{3} \rightarrow \mathfrak{se}\lrp{3} \, | \, \Delta \vec \tau^{\sss E} = \widetilde{\mathcal M} \boxminus \mathcal M = Log\big(\widetilde{\mathcal M} \circ \mathcal M^{-1}\big)\}}, represent the inverse operations, returning the transform vector difference \nm{\Delta \vec \tau} between two motions \nm{\mathcal M} and \nm{\widetilde{\mathcal M}} expressed in either the local or global tangent spaces to \nm{\mathcal M}.
\begin{eqnarray}
\nm{\Delta \vec \tau^{\sss B} } & = & \nm{\widetilde{\vec M} \ominus \vec M = Log\lrp{\vec M^{-1} \ \widetilde{\vec M}} = Log\lrp{\Delta \vec M^{\sss B}}}\label{eq:SE3_homogeneous_minus} \\
\nm{\Delta \vec \tau^{\sss B} } & = & \nm{\widetilde{\vec \zeta} \ominus \vec \zeta = 2 \ Log\lrp{\zetaast \otimes \widetilde{\vec \zeta}} = 2 \ Log\lrp{\Delta \vec \zeta^{\sss B}}}\label{eq:SE3_dual_quat_minus} \\
\nm{\Delta \vec \tau^{\sss E} } & = & \nm{\widetilde{\vec M} \boxminus \vec M = Log\lrp{\widetilde{\vec M} \ \vec M^{-1}} = Log\lrp{\Delta \vec M^{\sss E}}}\label{eq:SE3_homogeneous_minus_left} \\
\nm{\Delta \vec \tau^{\sss E} } & = & \nm{\widetilde{\vec \zeta} \boxminus \vec \zeta = 2 \ Log\lrp{\widetilde{\vec \zeta} \otimes \zetaast} = 2 \ Log\lrp{\Delta \vec \zeta^{\sss E}}}\label{eq:SE3_dual_quat_minus_left}
\end{eqnarray}

If the \nm{\Delta \vec \tau} perturbation is small, the (\ref{eq:SE3_twist_exponential3}) and (\ref{eq:SE3_Upsilon_exponential3}) Taylor expansions can be truncated, resulting in the following expressions, valid for both the body frame (\nm{\Delta \vec \tau^{\sss B}}) or the global one (\nm{\Delta \vec \tau^{\sss E}}):
\begin{eqnarray}
\nm{\Delta \vec M = Exp\lrp{\Delta \vec \tau}} & \nm{\approx} & \nm{\vec I_4 + \Delta \vec \tau^\wedge = \vec I_4 + \lrsb{\vec k \, \Delta \rho \ \ \ \vec n \, \Delta \phi}^\wedge}\label{eq:SE3_perturbation_homogeneous_truncated_local} \\
\nm{\Delta \vec \zeta = exp\lrp{\Delta \vec \tau /2}} & \nm{\approx} & \nm{\vec {\zeta_1} + \Delta \vec \tau^\wedge / 2 = \big[\lrsb{1 \ \ \ \vec n \, \Delta \phi / 2}^T + \epsilon \, \vec k \, \Delta \rho /2\big]^\wedge} \label{eq:SE3_perturbation_dual_quat_truncated_local} 
\end{eqnarray}

%%%%%%%%%%%%%%%%%%%%%%%%%%%%%%%%%%%%%%%%%%%%%%%%%%%%%%%%%%%%%%%%%%%%%%%%
%%%%%%%%%%%%%%%%%%%%%%%%%%%%%%%%%%%%%%%%%%%%%%%%%%%%%%%%%%%%%%%%%%%%%%%%
%%%%%%%%%%%%%%%%%%%%%%%%%%%%%%%%%%%%%%%%%%%%%%%%%%%%%%%%%%%%%%%%%%%%%%%%
% SECTION      RIGID BODY MOTION TIME DERIVATIVE AND TWIST
%%%%%%%%%%%%%%%%%%%%%%%%%%%%%%%%%%%%%%%%%%%%%%%%%%%%%%%%%%%%%%%%%%%%%%%%
%%%%%%%%%%%%%%%%%%%%%%%%%%%%%%%%%%%%%%%%%%%%%%%%%%%%%%%%%%%%%%%%%%%%%%%%
%%%%%%%%%%%%%%%%%%%%%%%%%%%%%%%%%%%%%%%%%%%%%%%%%%%%%%%%%%%%%%%%%%%%%%%%

\section{Rigid Body Motion Time Derivative and Twist}\label{sec:RigidBody_motion_calculus_derivatives}

Let \nm{\mathcal M\lrp{t} \in \mathbb{SE}\lrp{3}, t \in \mathbb{R}} be a moving rigid body and compute its derivative with time, which belongs to neither \nm{\mathbb{SE}\lrp{3}} nor \nm{\mathfrak{se}\lrp{3}} but to the Euclidean space of the chosen motion representation, \nm{\mathbb{R}^{4x4}} for the homogeneous matrix and \nm{\mathbb{H}_d} for the unit dual quaternion:
\neweq{\dot{\mathcal{M}}\lrp{t} = \lim\limits_{\Delta t \to 0} \dfrac{\mathcal{M}\lrp{t + \Delta t} - \mathcal{M}\lrp{t}}{\Delta t}}{eq:SE3_time_derivative_def}

Considering the time modified motion \nm{\mathcal{M}\lrp{t + \Delta t}} as the perturbed state (section \ref{subsec:RigidBody_motion_algebra_plus_minus}), the resulting time derivatives for the homogeneous matrix and unit dual quaternion representations are the following:
\begin{eqnarray}
\nm{\vec {\dot M}\lrp{t}} & = & \nm{\lim\limits_{\Delta t \to 0} \dfrac{\vec M \, \Delta \vec M^{\sss B} - \vec M}{\Delta t} \ \nm{\approx} \lim\limits_{\Delta t \to 0}\dfrac{\vec M \, \Big[\big(\vec I_4 + \lrsb{\vec k^{\sss B} \, \Delta \rho \ \ \ \vec n^{\sss B} \, \Delta \phi}^\wedge\big) - \vec I_4\Big]}{\Delta t}} \nonumber \\
& = & \nm{\vec M \, \lim\limits_{\Delta t \to 0}\dfrac{\lrsb{\Delta \rho \, \vec k^{\sss B} \ \ \ \Delta \phi \, \vec n^{\sss B}}^\wedge}{\Delta t}}\label{eq:SE3_time_derivative_M1b} \\
\nm{\vec {\dot \zeta}\lrp{t}} & = & \nm{\lim\limits_{\Delta t \to 0} \dfrac{\vec \zeta \otimes \Delta \vec \zeta^{\sss B} - \vec \zeta}{\Delta t} \ \nm{\approx} \lim\limits_{\Delta t \to 0}\dfrac{\vec \zeta \otimes \lrsb{\lrp{\lrsb{1 \ \ \ \vec n^{\sss B} \, \Delta \phi / 2}^T + \epsilon \, \vec k^{\sss B} \, \Delta \rho /2}^\wedge - \vec{\zeta_1}}}{\Delta t}} \nonumber \\
& = & \nm{\vec \zeta \otimes \lim\limits_{\Delta t \to 0}\dfrac{\lrsb{\vec n^{\sss B} \, \Delta \phi / 2 + \epsilon \, \vec k^{\sss B} \, \Delta \rho /2}^\wedge}{\Delta t}}\label{eq:SE3_time_derivative_zeta1b} 
\end{eqnarray}

Similar expressions based on \nm{\vec \tau^{\sss E} = \lrsb{\Delta \rho \, \vec k^{\sss E} \ \ \ \Delta \phi \, \vec n^{\sss E}}^T} can be found if left multiplying by the perturbation instead of right multiplying. The \nm{\vec {\dot M}\lrp{t}} and \nm{\vec {\dot \zeta}\lrp{t}} expressions (\ref{eq:SE3_homogeneous_dot}) and (\ref{eq::SE3_dual_quat_xi_dot}) are then directly obtained when defining the \emph{body twist} \nm{\vec \xi_{\sss EB}^{\sss B}} as the time derivative of the transform vector \nm{\vec \tau^{\sss B}} when viewed in local or body frame \nm{\FB}, and the \emph{spatial twist} \nm{\vec \xi_{\sss EB}^{\sss E}} as the time derivative of the transform vector \nm{\vec \tau^{\sss E}} when viewed in global or spatial frame \nm{\FE}:
\begin{eqnarray}
\nm{\vec \xi_{\sss EB}^{\sss B}\lrp{t}} & = & \nm{\Delta \vec{\dot \tau}^{\sss B}\lrp{t} = \lim\limits_{\Deltat \to 0} \frac{\Delta \vec \tau^{\sss B}}{\Deltat} = \lim\limits_{\Deltat \to 0} \frac{\lrsb{\vec k^{\sss B} \, \Delta \rho \ \ \ \vec n^{\sss B} \, \Delta \phi}^T}{\Deltat}}\label{eq:SE3_time_derivative_xiEBB} \\
\nm{\vec \xi_{\sss EB}^{\sss E}\lrp{t}} & = & \nm{\Delta \vec{\dot \tau}^{\sss E}\lrp{t} = \lim\limits_{\Deltat \to 0} \frac{\Delta \vec \tau^{\sss E}}{\Deltat} = \lim\limits_{\Deltat \to 0} \frac{\lrsb{\vec k^{\sss E} \, \Delta \rho \ \ \ \vec n^{\sss E} \, \Delta \phi}^T}{\Deltat}}\label{eq:SE3_time_derivative_xiEBE}  
\end{eqnarray}

The twist \nm{\vec \xi = \lrsb{\vec \nu \ \ \vec \omega}^T} represents the motion velocity and is composed by the angular velocity \nm{\vec \omega} defined in section \ref{sec:RigidBody_rotation_calculus_derivatives} and the linear velocity \nm{\vec \nu}. The twist physical meaning is revealed by obtaining its expressions when viewed in both the local and spatial frames. The body twist \nm{\xiEBBskew \in \mathfrak{se}\lrp{3}} corresponding to the rigid body motion \nm{\MEB\lrp{t} \in \mathbb{SE}\lrp{3}} responds to (\ref{eq:SE3_homogeneous_twist_body}):
\neweq{\xiEBBskew = \begin{bmatrix} \nm{\wEBBskew} & \nm{\nuEBB} \\ \nm{\vec 0_3^T} & 0 \end{bmatrix} = \MEBinv \; \MEBdot = \begin{bmatrix} \nm{\REBtrans} & \nm{- \REBtrans \, \TEBE} \\ \nm{\vec 0_3^T} & 1 \end{bmatrix} \begin{bmatrix} \nm{\REBdot} & \nm{\vec {\dot T}_{\sss EB}^{\sss E}} \\ \nm{\vec 0_3^T} & 0 \end{bmatrix} = \begin{bmatrix} \nm{\REBtrans \; \REBdot} & \nm{\REBtrans \; \vec {\dot T}_{\sss EB}^{\sss E}} \\ \nm{\vec 0_3^T} & 0 \end{bmatrix}}{eq:SE3_time_derivative_twist_body}

Its physical interpretation is that the angular component \nm{\wEBB} is indeed the (\ref{eq:SO3_dcm_omega_body}) angular velocity \nm{\vec \omega_{\sss EB}} as viewed from the body frame, and the linear component \nm{\nuEBB} is the linear velocity of the body frame origin \nm{\vec {\dot T}_{\sss EB}} also viewed from the body frame (\ref{eq:SO3_dcm_transform}, \ref{eq:SO3_dcm_inverse}) \cite{Murray1994}. The global twist \nm{\xiEBEskew \in \mathfrak{se}\lrp{3}} is determined by means of (\ref{eq:SE3_homogeneous_twist_space}):
\begin{eqnarray}
\nm{\xiEBEskew} & = & \nm{\begin{bmatrix} \nm{\wEBEskew} & \nm{\nuEBE} \\ \nm{\vec 0_3^T} & 0 \end{bmatrix} = \MEBdot \; \MEBinv = \begin{bmatrix} \nm{\REBdot} & \nm{\vec {\dot T}_{\sss EB}^{\sss E}} \\ \nm{\vec 0_3^T} & 0 \end{bmatrix} \begin{bmatrix} \nm{\REBtrans} & \nm{- \REBtrans \, \TEBE} \\ \nm{\vec 0_3^T} & 1 \end{bmatrix}} \nonumber \\
& = & \nm{\begin{bmatrix} \nm{\REBdot \; \REBtrans} & \nm{\vec {\dot T}_{\sss EB}^{\sss E} - \REBdot \; \REBtrans \; \vec T_{\sss EB}^{\sss E}} \\ \nm{\vec 0_3^T} & 0 \end{bmatrix}} \label{eq:SE3_time_derivative_twist_space}
\end{eqnarray}

The \nm{\xiEBEskew} physical interpretation is not intuitive, however. While the angular component \nm{\wEBE} is the (\ref{eq:SO3_dcm_omega_space}) angular velocity \nm{\wEB} as viewed from the spatial frame, its linear component \nm{\nuEBE} is not the velocity of the body frame origin \nm{\vec {\dot T}_{\sss EB}} viewed in the \nm{\FE} frame (\nm{\vec {\dot T}_{\sss EB}^{\sss E}}), but the velocity, viewed in the spatial frame, of a possibly imaginary point of the rigid body which at time \emph{t} is traveling through the origin of the spatial frame \cite{Murray1994}. Chapter \ref{cha:Composition} may facilitate the understanding of this concept.

Note that the transformation or motion of the twist (relationship between \nm{\xiEBE} and \nm{\xiEBB}), and hence that of the linear velocities \nm{\nuEBE} and \nm{\nuEBB}, is not given by the motion action \nm{\vec g_{\mathcal M*}} (\ref{eq:Motion_maps}) but by the adjoint map \nm{\vec{Ad}_{\mathcal M}} described in section \ref{sec:RigidBody_motion_adjoint}. Unlike in the case of rotations, these two maps do not coincide.

%%%%%%%%%%%%%%%%%%%%%%%%%%%%%%%%%%%%%%%%%%%%%%%%%%%%%%%%%%%%%%%%%%%%%%%%
%%%%%%%%%%%%%%%%%%%%%%%%%%%%%%%%%%%%%%%%%%%%%%%%%%%%%%%%%%%%%%%%%%%%%%%%
%%%%%%%%%%%%%%%%%%%%%%%%%%%%%%%%%%%%%%%%%%%%%%%%%%%%%%%%%%%%%%%%%%%%%%%%
% SECTION      RIGID BODY MOTION POINT VELOCITY
%%%%%%%%%%%%%%%%%%%%%%%%%%%%%%%%%%%%%%%%%%%%%%%%%%%%%%%%%%%%%%%%%%%%%%%%
%%%%%%%%%%%%%%%%%%%%%%%%%%%%%%%%%%%%%%%%%%%%%%%%%%%%%%%%%%%%%%%%%%%%%%%%
%%%%%%%%%%%%%%%%%%%%%%%%%%%%%%%%%%%%%%%%%%%%%%%%%%%%%%%%%%%%%%%%%%%%%%%%

\section{Rigid Body Motion Point Velocity}\label{sec:RigidBody_motion_velocity}

There exists a direct relationship between the velocity of a point belonging to a rigid body and the elements of its tangent space, this is, the twist in \nm{\mathfrak{se}\lrp{3}}. This relationship is independent of the \nm{\mathbb{SE}\lrp{3}} representation, although the homogeneous matrix is employed in the expressions below. If \nm{\pBbar = \lrsb{\pB \ \ 1}^T} are the fixed coordinates of a point belonging to the \nm{\FB} rigid body, the point spatial coordinates \nm{\pEbar = \lrsb{\pE \ \ 1}^T} can be obtained by means of (\ref{eq:SE3_homogeneous_transform}):
\neweq{\pEbar\lrp{t} = \vec g_{\mathcal M_{EB}(t)}\lrp{\pBbar} = \MEB\lrp{t} \; \pBbar} {eq:SE3_velocity1} 

The velocity of a point is the time derivative of its spatial or global coordinates. As \nm{\vec {\bar p}} is fixed to \nm{\FB}, its time derivative is zero \nm{\lrp{\pBbardot = \vec 0}}, so its velocity viewed in the spatial frame responds to:
\neweq{\vpEbar\lrp{t} = \pEbardot\lrp{t} = \MEBdot\lrp{t} \; \pBbar}{eq:SE3_velocity2}

Although \nm{\MEBdot} maps the point body coordinates to its spatial velocity per (\ref{eq:SE3_velocity2}), its high dimension makes it inefficient. By making use of the spatial and body twists (\nm{\vec \xi_{\sss EB}^{{\sss E}\wedge}, \, \vec \xi_{\sss EB}^{{\sss B}\wedge}}) introduced in (\ref{eq:SE3_homogeneous_dot}), the velocity of a point \nm{\pBbar} viewed in \nm{\FE} can be obtained as follows:
\begin{eqnarray}
\nm{\vpEbar\lrp{t}} & = & \nm{\xiEBEskew\lrp{t} \; \MEB\lrp{t} \; \pBbar = \xiEBEskew\lrp{t} \; \pEbar\lrp{t}} \label{eq:SE3_velocity_v_e} \\
\nm{\vpEbar\lrp{t}} & = & \nm{\MEB\lrp{t} \; \xiEBBskew\lrp{t} \; \pBbar} \label{eq:SE3_velocity_v_e_bis} 
\end{eqnarray}

The velocity of \nm{\pBbar} viewed in \nm{\FB} can then be obtained by means of the vector action map:
\neweq{\vpBbar\lrp{t} = \vec g_{\mathcal M_{EB(t)}*}^{-1} \big(\vpEbar\lrp{t}\big) = \MEBinv\lrp{t} \; \vpEbar\lrp{t} = \xiEBBskew\lrp{t} \; \pBbar} {eq:SE3_velocity_v_b} 

Returning to Cartesian coordinates and introducing the angular and linear components of the twist results in:
\begin{eqnarray}
\nm{\vec v_{\sss p}^{\sss E}\lrp{t}} & = & \nm{\wEBEskew\lrp{t} \; \pE\lrp{t} + \nuEBE\lrp{t}}\label{eq:SE3_velocity_e} \\
\nm{\vec v_{\sss p}^{\sss B}\lrp{t}} & = & \nm{\wEBBskew\lrp{t} \; \pB + \nuEBB\lrp{t}}\label{eq:SE3_velocity_b}
\end{eqnarray}

The point velocity is hence the result of the sum of the linear velocity and the cross product between the angular velocity and the point coordinates. 

%%%%%%%%%%%%%%%%%%%%%%%%%%%%%%%%%%%%%%%%%%%%%%%%%%%%%%%%%%%%%%%%%%%%%%%%
%%%%%%%%%%%%%%%%%%%%%%%%%%%%%%%%%%%%%%%%%%%%%%%%%%%%%%%%%%%%%%%%%%%%%%%%
%%%%%%%%%%%%%%%%%%%%%%%%%%%%%%%%%%%%%%%%%%%%%%%%%%%%%%%%%%%%%%%%%%%%%%%%
% SECTION      RIGID BODY MOTION ADJOINT
%%%%%%%%%%%%%%%%%%%%%%%%%%%%%%%%%%%%%%%%%%%%%%%%%%%%%%%%%%%%%%%%%%%%%%%%
%%%%%%%%%%%%%%%%%%%%%%%%%%%%%%%%%%%%%%%%%%%%%%%%%%%%%%%%%%%%%%%%%%%%%%%%
%%%%%%%%%%%%%%%%%%%%%%%%%%%%%%%%%%%%%%%%%%%%%%%%%%%%%%%%%%%%%%%%%%%%%%%%

\section{Rigid Body Motion Adjoint}\label{sec:RigidBody_motion_adjoint}

The \emph{adjoint map} of a Lie group is defined in section \ref{subsec:algebra_lie_adjoint} as an action of the Lie group on its own Lie algebra that converts between the local tangent space and that at the identity. In the case of rigid body motion, both the transform vector and the twist belong to the tangent space, so \nm{\lrb{\vec{Ad}\lrp{}: \mathbb{SE}\lrp{3} \times \mathfrak{se}\lrp{3} \rightarrow \mathfrak{se}\lrp{3} \ | \ \vec{Ad}_{\mathcal M}\lrp{\vec \tau^{\wedge}} = \mathcal M \circ \vec \tau^{\wedge} \circ \mathcal{M}^{-1}, \ \vec{Ad}_{\mathcal M}\lrp{\vec \xi^{\wedge}} = \mathcal M \circ \vec \xi^{\wedge} \circ \mathcal{M}^{-1}}}. This is equivalent to \nm{\vec \zeta \otimes \vec \xi^{\wedge} \otimes \vec \zeta^{\ast}} for unit dual quaternions or \nm{\vec M \, \vec \xi^\wedge \, \vec M^{-1}} for homogeneous matrices, which represents the similarity transformation\footnote{Two square matrices \nm{\vec A} and \nm{\vec B} are called similar if \nm{\vec B = {\vec P}^{-1} \; \vec A \; \vec P} for some invertible square matrix \nm{\vec P}.} between the spatial and body twists \nm{\vec \xi_{\sss EB}^{{\sss E}\wedge}} and \nm{\vec \xi_{\sss EB}^{{\sss B}\wedge}}:
\neweq{\xiEBEskew = \MEB \; \xiEBBskew \; \MEBinv \rightarrow \left\{\begin{aligned} \nm{\wEBEskew} & = \nm{\REB \; \wEBBskew \; \REBtrans = \vec{Ad}_{\mathcal R_{EB}}\lrp{\wEBBskew}} \\ \nm{\nuEBE} & = \nm{\REB \; \nuEBB - \wEBEskew \; \TEBE = \REB \; \nuEBB + \TEBEskew \; \wEBE} \end{aligned} \right.}{eq:SE3_velocity_similarity}

The application of the vee operator results in the adjoint matrix:
\neweq{\xiEBE = \begin{bmatrix} \nm{\nuEBE} \\ \nm{\wEBE} \end{bmatrix} = \begin{bmatrix} \nm{\REB} & \nm{\TEBEskew \; \REB} \\ \nm{\vec 0_{3x3}^T} & \nm{\REB} \end{bmatrix} \, \begin{bmatrix} \nm{\nuEBB} \\ \nm{\wEBB} \end{bmatrix} = \begin{bmatrix} \nm{\REB} & \nm{\TEBEskew \; \REB} \\ \nm{\vec 0_{3x3}^T} & \nm{\REB} \end{bmatrix} \, \xiEBB = \vec{Ad}_{\mathcal M_{EB}} \, \xiEBB}{eq:SE3_velocity_similarity2}

As stated above, note that the adjoint map (\ref{eq:SE3_velocity_similarity2}) is different than the vector action \nm{\vec g_{\mathcal M*}} (\ref{eq:Motion_maps}), unlike the case of rotational motion described in section \ref{sec:RigidBody_rotation_adjoint}, in which they coincide.

A similar process leads to the inverse adjoint matrix (\nm{\vec{Ad}_{\mathcal M}^{-1} \, \vec \xi = \vec{Ad}_{\mathcal M^{-1}} \, \vec \xi}):
\neweq{\xiEBB = \vec{Ad}_{\mathcal M_{EB}}^{-1} \; \xiEBE = \begin{bmatrix} \nm{\REB^T} & \nm{- \REB^T \, \TEBEskew} \\ \nm{\vec 0_{3x3}^T} & \nm{\REB^T} \end{bmatrix} \; \xiEBE}{eq:SE3_dcm_velocity7} 

%%%%%%%%%%%%%%%%%%%%%%%%%%%%%%%%%%%%%%%%%%%%%%%%%%%%%%%%%%%%%%%%%%%%%%%%
%%%%%%%%%%%%%%%%%%%%%%%%%%%%%%%%%%%%%%%%%%%%%%%%%%%%%%%%%%%%%%%%%%%%%%%%
%%%%%%%%%%%%%%%%%%%%%%%%%%%%%%%%%%%%%%%%%%%%%%%%%%%%%%%%%%%%%%%%%%%%%%%%
% SECTION      RIGID BODY MOTION UNCERTAINTY AND COVARIANCE
%%%%%%%%%%%%%%%%%%%%%%%%%%%%%%%%%%%%%%%%%%%%%%%%%%%%%%%%%%%%%%%%%%%%%%%%
%%%%%%%%%%%%%%%%%%%%%%%%%%%%%%%%%%%%%%%%%%%%%%%%%%%%%%%%%%%%%%%%%%%%%%%%
%%%%%%%%%%%%%%%%%%%%%%%%%%%%%%%%%%%%%%%%%%%%%%%%%%%%%%%%%%%%%%%%%%%%%%%%

\section{Rigid Body Motion Uncertainty and Covariance}\label{sec:RigidBody_motion_covariance}

Following the analysis of uncertainty on Lie groups presented in section \ref{subsec:algebra_lie_covariance}, the definitions of local and global autocovariances for \nm{\mathbb{SE}\lrp{3}} elements around a nominal or expected rotation \nm{E\lrsb{\mathcal M} = \vec \mu_{\mathcal M} \in \mathbb{SE}\lrp{3}} are the following:
\begin{eqnarray}
\nm{\vec C_{\mathcal M \mathcal M}^{\sss B}} & = & \nm{E\lrsb{\Delta \vec \tau^{\sss B} \, \Delta \vec \tau^{{\sss B}T}} = E\lrsb{\lrp{\mathcal M \ominus \vec \mu_{\mathcal M}}\lrp{\mathcal M \ominus \vec \mu_{\mathcal M}}^T} \ \ \in \mathbb{R}^{6x6}}\label{eq:SE3_covariance_right_def} \\
\nm{\vec C_{\mathcal M \mathcal M}^{\sss E}} & = & \nm{E\lrsb{\Delta \vec \tau^{\sss E} \, \Delta \vec \tau^{{\sss E}T}} = E\lrsb{\lrp{\mathcal M \boxminus \vec \mu_{\mathcal M}}\lrp{\mathcal M \boxminus \vec \mu_{\mathcal M}}^T} \ \ \in \mathbb{R}^{6x6}}\label{eq:SE3_covariance_left_def} 
\end{eqnarray}

Note that although the notation refers to the covariance of the rigid body motion manifold \nm{\mathcal M \in \mathbb{SE}\lrp{3}}, the definition in fact refers to the covariance of the transform vectors \nm{\Delta \vec \tau^{\sss B}} or \nm{\Delta \vec \tau^{\sss E}} located in the tangent space, with its dimension (6) matching the number of degrees of freedom of the \nm{\mathbb{SE}\lrp{3}} manifold. The relationship between the local and global autocovariances responds to:
\neweq{\vec C_{\mathcal M \mathcal M}^{\sss E} = \vec{Ad}_{\mathcal M_{EB}} \ \vec C_{\mathcal M \mathcal M}^{\sss B} \ \vec{Ad}_{\mathcal M_{EB}}^T} {eq:SE3_covariance_left_relationship}

Given a function \nm{\lrb{f: \mathcal{M} \rightarrow \mathcal {N} \ | \ \mathcal {N} = f\lrp{\mathcal {M}} \in \mathbb{SE}\lrp{3}, \, \forall \mathcal {M} \in \mathbb{SE}\lrp{3}}} between two rigid body motions, the covariances are propagated as follows:
\begin{eqnarray}
\nm{\vec C_{\mathcal N \mathcal N}^{\sss B}} & = & \nm{\vec J_{\ds{\oplus \; \mathcal M}}^{\ds{\ominus \; f\lrp{\mathcal M}}} \ \vec C_{\mathcal M \mathcal M}^{\sss B} \ \vec J_{\ds{\oplus \; \mathcal M}}^{{\ds{\ominus \; f\lrp{\mathcal M}}},T} \ \ \ \ \ \ \ \in \mathbb{R}^{6x6}} \label{eq:SE3_covariance_right_propagation} \\
\nm{\vec C_{\mathcal N \mathcal N}^{\sss E}} & = & \nm{\vec J_{\ds{\boxplus \; \mathcal M}}^{\ds{\boxminus \; f\lrp{\mathcal M}}} \ \vec C_{\mathcal M \mathcal M}^{\sss E} \ \vec J_{\ds{\boxplus \; \mathcal M}}^{{\ds{\boxminus \; f\lrp{\mathcal M}}},T} \ \ \ \ \ \ \ \in \mathbb{R}^{6x6}} \label{eq:SE3_covariance_left_propagation}
\end{eqnarray}

%%%%%%%%%%%%%%%%%%%%%%%%%%%%%%%%%%%%%%%%%%%%%%%%%%%%%%%%%%%%%%%%%%%%%%%%
%%%%%%%%%%%%%%%%%%%%%%%%%%%%%%%%%%%%%%%%%%%%%%%%%%%%%%%%%%%%%%%%%%%%%%%%
%%%%%%%%%%%%%%%%%%%%%%%%%%%%%%%%%%%%%%%%%%%%%%%%%%%%%%%%%%%%%%%%%%%%%%%%
% SECTION      RIGID BODY MOTION JACOBIANS
%%%%%%%%%%%%%%%%%%%%%%%%%%%%%%%%%%%%%%%%%%%%%%%%%%%%%%%%%%%%%%%%%%%%%%%%
%%%%%%%%%%%%%%%%%%%%%%%%%%%%%%%%%%%%%%%%%%%%%%%%%%%%%%%%%%%%%%%%%%%%%%%%
%%%%%%%%%%%%%%%%%%%%%%%%%%%%%%%%%%%%%%%%%%%%%%%%%%%%%%%%%%%%%%%%%%%%%%%%

\section{Rigid Body Motion Jacobians}\label{sec:RigidBody_motion_calculus_jacobians}

Lie group Jacobians are introduced in section \ref{sec:algebra_lie_jacobians} based on the right and left Lie group derivatives of section \ref{subsec:algebra_lie_derivatives}, and in this section are customized for the \nm{\mathbb{SE}\lrp{3}} case, with table \ref{tab:RigidBody_motion_jacobians} representing the particularization of table \ref{tab:algebra_lie_jacobians} to the case of rigid body motions\footnote{To save space within table \ref{tab:RigidBody_motion_jacobians}, \nm{< ; >} within a matrix implies a different row. For example, \nm{\lrsb{a \ b; \ c \ d}} is equivalent to \nm{\begin{bmatrix} a & b \\ c & d \end{bmatrix}}.}. The various Jacobians listed in table \ref{tab:RigidBody_motion_jacobians} have been obtained by means of the chain rule, the expressions already introduced in this chapter, and those of section \ref{sec:algebra_lie}. Note that although in many cases the results internally include the rotation matrix, all Jacobians are generic and do not depend on the specific \nm{\mathbb{SE}\lrp{3}} parameterization.

In addition to the adjoint matrix, two other Jacobians are of particular importance as they appear repeatedly in table \ref{tab:RigidBody_motion_jacobians}. These are the right and left Jacobians of the capitalized exponential function, also known as simply the \emph{right Jacobian} \nm{\vec J_R\lrp{\vec \tau}} and the \emph{left Jacobian} \nm{\vec J_L\lrp{\vec \tau}}, and they evaluate the variation of the \nm{\mathfrak{se}\lrp{3}} tangent space provided by the output of the \nm{Exp\lrp{\vec \tau}} map (locally for \nm{\vec J_R} and globally for \nm{\vec J_L}) while moving along the \nm{\mathbb{SE}\lrp{3}} manifold with respect to the (Euclidean) variations within the original tangent space provided by \nm{\vec \tau}. Their closed forms as well as those of their inverses are included in table \ref{tab:RigidBody_motion_jacobians}, and have been obtained from \cite{Barfoot2014}; they are based on the \nm{\vec Q\lrp{\vec \tau}} matrix:
\begin{eqnarray}
\nm{\vec Q\lrp{\vec \tau}} & = & \nm{\vec Q\lrp{\vec s, \, \vec r} = \vec Q\lrp{\vec k \, \rho, \, \vec n \, \phi} = \dfrac{\widehat{\vec s}}{2} + \dfrac{\phi - \sin \phi}{\phi^3}\lrp{\widehat{\vec r} \, \widehat{\vec s} + \widehat{\vec s} \, \widehat{\vec r} + \widehat{\vec r} \, \widehat{\vec s} \, \widehat{\vec r}}} \nonumber \\ 
& \nm{-} & \nm{\dfrac{1 - \phi^2 /2 - \cos \phi}{\phi^4}\lrp{\widehat{\vec r}^2 \, \widehat{\vec s} + \widehat{\vec s} \, \widehat{\vec r}^2 - 3 \, \widehat{\vec r} \, \widehat{\vec s} \, \widehat{\vec r}}} \nonumber \\
& \nm{-} & \nm{\dfrac{1}{2} \lrsb{\dfrac{1 - \phi^2 /2 - \cos \phi}{\phi^4} - 3 \dfrac{\phi - \sin \phi - \phi^3 /6}{\phi^5}} \lrp{\widehat{\vec r} \, \widehat{\vec s} \, \widehat{\vec r}^2 + \widehat{\vec r}^2 \, \widehat{\vec s} \, \widehat{\vec r}} \ \ \ \in \mathbb{R}^{3x3}} \label{eq:SE3_jacobian_left_Q} 
\end{eqnarray}

It is also worth noting the special importance of the \nm{\vec J_{\ds{+ \; \vec \tau}}^{\ds{- \; g_{Exp\lrp{\vec \tau}}(\vec p)}}} Jacobian present at the bottom of table \ref{tab:RigidBody_motion_jacobians}, which represents the derivative of a transformed point with respect to perturbations in the Euclidean tangent space (not on the curved manifold) that generates the motion, as it enables tangent space optimization by calculus methods designed exclusively for Euclidean spaces. 
\renewcommand{\arraystretch}{1.5} % increase row height
\begin{center}
\begin{tabular}{lcccll}
	\hline
	Jacobian & & Table \ref{tab:algebra_lie_jacobians} & & \multicolumn{1}{c}{Expression} & Size \\
	\hline
	\nm{\vec J_{\ds{\oplus \; \mathcal M}}^{\ds{\ominus \; \mathcal M}^{-1}}}       				& = & \nm{- \vec{Ad}_{\mathcal M}} 			& = & \nm{- \lrsb{\vec R \ \ \Tskew \; \vec R; \ \ \vec{0}_{3x3} \ \ \vec R}} & \nm{\in \mathbb R^{6x6}} \\ 
	\nm{\vec J_{\ds{\boxplus \; \mathcal M}}^{\ds{\boxminus \; \mathcal M}^{-1}}}         		& = & \nm{- \vec{Ad}_{\mathcal M}^{-1}}		& = & \nm{- \lrsb{\vec R^T \ \ - \vec R^T \, \Tskew; \ \ \vec{0}_{3x3} \ \ \vec R^T}} & \nm{\in \mathbb R^{6x6}} \\ 

	\nm{\vec J_{\ds{\oplus \; \mathcal M}}^{\ds{\ominus \; \mathcal M \circ \mathcal N}}}		& = & \nm{\vec{Ad}_{\mathcal N}^{-1}}	   	& = & \nm{\lrsb{\vec R_{\mathcal N}^T \ \ - \vec R_{\mathcal N}^T \, \widehat{\vec T}_{\mathcal N}; \ \ \vec{0}_{3x3} \ \ \vec R_{\mathcal N}^T}} & \nm{\in \mathbb R^{6x6}} \\ 
	\nm{\vec J_{\ds{\boxplus \; \mathcal M}}^{\ds{\boxminus \; \mathcal M \circ \mathcal N}}}	& = & \nm{\vec I}	 						& = & \nm{\vec{I}_{6x6}}		 	& \nm{\in \mathbb R^{6x6}} \\ 
	\nm{\vec J_{\ds{\oplus \; \mathcal N}}^{\ds{\ominus \; \mathcal M \circ \mathcal N}}}		& = & \nm{\vec I}							& = & \nm{\vec{I}_{6x6}}           	& \nm{\in \mathbb R^{6x6}} \\ 
	\nm{\vec J_{\ds{\boxplus \; \mathcal N}}^{\ds{\boxminus \; \mathcal M \circ \mathcal N}}}	& = & \nm{\vec{Ad}_{\mathcal M}}			& = & \nm{\lrsb{\vec R_{\mathcal M} \ \ \widehat{\vec T}_{\mathcal M} \; \vec R_{\mathcal M}; \ \ \vec{0}_{3x3} \ \ \vec R_{\mathcal M}}} & \nm{\in \mathbb R^{6x6}} \\ 

	\nm{\vec J_{\ds{+ \; \vec \tau}}^{\ds{\ominus \; Exp\lrp{\vec \tau}}}}						& = & \nm{\vec J_R\lrp{\vec \tau}} 				& = & \nm{\vec J_L\lrp{- \vec \tau} = \vec J_L\lrp{- \vec s, \; - \vec r}} & \nm{\in \mathbb R^{6x6}} \\ 
	\nm{\vec J_R^{-1}\lrp{\vec \tau}}																&   & 										& = &  \nm{\vec J_L^{-1}\lrp{- \vec \tau} = \vec J_L^{-1}\lrp{- \vec s, \; - \vec r}} & \nm{\in \mathbb R^{6x6}} \\ 
	\nm{\vec J_{\ds{+ \; \vec \tau}}^{\ds{\boxminus \; Exp\lrp{\vec \tau}}}} 						& = & \nm{\vec J_L\lrp{\vec \tau}} 				& = & \nm{\Big[\vec J_L\lrp{\vec r} \ \ \vec Q\lrp{\vec \tau}; \ \ \vec{0}_{3x3} \ \ \vec J_L\lrp{\vec r}\Big]} & \nm{\in \mathbb R^{6x6}} \\ 
	\nm{\vec J_L^{-1}\lrp{\vec \tau}}																&   & 										& = & \nm{\Big[\vec J_L^{-1}\lrp{\vec r} \ \ - \vec J_L^{-1}\lrp{\vec r} \; \vec Q\lrp{\vec \tau} \, \vec J_L^{-1}\lrp{\vec r}; \ \ \vec{0}_{3x3} \ \ \vec J_L^{-1}\lrp{\vec r} \Big]}  & \nm{\in \mathbb R^{6x6}} \\ 
	
	\nm{\vec J_{\ds{\oplus \; \mathcal M}}^{\ds{- \; Log\lrp{\mathcal M}}}}        				& = & \nm{\vec J_R^{-1}\big(Log\lrp{\mathcal M}\big)} & & & \nm{\in \mathbb R^{6x6}} \\  
	\nm{\vec J_{\ds{\boxplus \; \mathcal M}}^{\ds{- \; Log\lrp{\mathcal M}}}}     				& = & \nm{\vec J_L^{-1}\big(Log\lrp{\mathcal M}\big)} & & & \nm{\in \mathbb R^{6x6}} \\  
	\nm{\vec J_{\ds{\oplus \; \mathcal M}}^{\ds{\ominus \; \mathcal M \oplus \vec \tau}}}			& = & \nm{\vec{Ad}_{Exp\lrp{\vec \tau}}^{-1}}	& = & \nm{\lrsb{\vec R\lrp{\vec r}^T \ \ - \vec R\lrp{\vec r}^T \Tskew\lrp{\vec s, \vec r}; \ \ \vec{0}_{3x3} \ \ \vec R\lrp{\vec r}^T}} & \nm{\in \mathbb R^{6x6}} \\ 
	\nm{\vec J_{\ds{\boxplus \; \mathcal M}}^{\ds{\boxminus \; \vec \tau \boxplus \mathcal M}}}		& = & \nm{\vec{Ad}_{Exp\lrp{\vec \tau}}}		& = & \nm{\lrsb{\vec R\lrp{\vec r} \ \ \Tskew\lrp{\vec s, \vec r} \vec R\lrp{\vec r}; \ \ \vec{0}_{3x3} \ \ \vec R\lrp{\vec r}}}  & \nm{\in \mathbb R^{6x6}} \\ 
  
	\nm{\vec J_{\ds{+ \; \vec \tau}}^{\ds{\ominus \; \mathcal M \oplus \vec \tau}}}         			& = & \nm{\vec J_R\lrp{\vec \tau}}           	   	& & & \nm{\in \mathbb R^{6x6}} \\ 	
	\nm{\vec J_{\ds{+ \; \vec \tau}}^{\ds{\boxminus \; \vec \tau \boxplus \mathcal M}}}     			& = & \nm{\vec J_L\lrp{\vec \tau}}           	   	& & & \nm{\in \mathbb R^{6x6}} \\ 	
	
	\nm{\vec J_{\ds{\oplus \; \mathcal M}}^{\ds{- \; \mathcal N \ominus \mathcal M}}} 	      		& = & \nm{- \vec J_L^{-1}\lrp{\mathcal N \ominus \mathcal M}}	& & & \nm{\in \mathbb R^{6x6}} \\ 	
	\nm{\vec J_{\ds{\boxplus \; \mathcal M}}^{\ds{- \; \mathcal N \boxminus \mathcal M}}} 	   		& = & \nm{- \vec J_R^{-1}\lrp{\mathcal N \boxminus \mathcal M}}	& & & \nm{\in \mathbb R^{6x6}} \\ 	
	\nm{\vec J_{\ds{\oplus \; \mathcal N}}^{\ds{- \; \mathcal N \ominus \mathcal R}}}    	    	& = & \nm{\vec J_R^{-1}\lrp{\mathcal N \ominus \mathcal M}}		& & & \nm{\in \mathbb R^{6x6}} \\
	\nm{\vec J_{\ds{\boxplus \; \mathcal N}}^{\ds{- \; \mathcal N \boxminus \mathcal R}}}  	    	& = & \nm{\vec J_L^{-1}\lrp{\mathcal N \boxminus \mathcal M}}	& & & \nm{\in \mathbb R^{6x6}} \\  	
	
	\nm{\vec J_{\ds{\oplus \; \mathcal M}}^{\ds{- \; g_{\mathcal M}(\vec p)}}}          	 		&   & 		& = & \nm{\lrsb{\vec R \ \ - \vec R \; \pskew}}    										& \nm{\in \mathbb R^{3x6}} \\ 
	\nm{\vec J_{\ds{\boxplus \; \mathcal M}}^{\ds{- \; g_{\mathcal M}(\vec p)}}}      				&   & 		& = & \nm{\lrsb{\vec I_{3x3} \ \ - \lrp{\vec R \; \vec p}^\wedge - \widehat{\vec T}}}	& \nm{\in \mathbb R^{3x6}} \\ 
	\nm{\vec J_{\ds{+ \; \vec p}}^{\ds{- \; g_{\mathcal M}(\vec p)}}}               				&   & 		& = & \nm{\vec R}                          												& \nm{\in \mathbb R^{3x3}} \\  
	\hline
\end{tabular}
\end{center}
	
\begin{center}
\begin{tabular}{lcccll}
	\hline
	Jacobian & & Table \ref{tab:algebra_lie_jacobians} & & \multicolumn{1}{c}{Expression} & Size \\
	\hline

	\nm{\vec J_{\ds{\oplus \; \mathcal M}}^{\ds{- \; g_{\mathcal M}^{-1}(\vec p)}}}          		&   & 		& = & \nm{\lrsb{- \vec I_{3x3} \ \ \big(\vec R^T\lrp{\vec p - \vec T}\big)^\wedge}} 	& \nm{\in \mathbb R^{3x6}} \\ 
	\nm{\vec J_{\ds{\boxplus \; \mathcal M}}^{\ds{- \; g_{\mathcal M}^{-1}(\vec p)}}}  				&   & 		& = & \nm{\lrsb{- \vec R^T \ \ \vec R^T \; \pskew}} 									& \nm{\in \mathbb R^{3x6}} \\ 
	\nm{\vec J_{\ds{+ \; \vec p}}^{\ds{- \; g_{\mathcal M}^{-1}(\vec p)}}}          	   			&   & 		& = & \nm{\vec R^T}																		& \nm{\in \mathbb R^{3x3}} \\  	

	\nm{\vec J_{\ds{\oplus \; \mathcal M}}^{\ds{- \; \vec{Ad}_{\mathcal M}(\vec \xi)}}} 			& = & \multicolumn{3}{l}{\nm{\lrsb{- \lrp{\vec R \; \vec \omega}^\wedge \vec R \ \ - \vec R \; \widehat{\nu} - \Tskew \; \vec R \; \omegaskew; \ \ \vec{0}_{3x3} \ \ - \vec R \; \omegaskew}}} & \nm{\in \mathbb R^{6x6}} \\ 
	\nm{\vec J_{\ds{\boxplus \; \mathcal M}}^{\ds{- \; \vec{Ad}_{\mathcal M}(\vec \xi)}}}			& = & \multicolumn{3}{l}{\nm{\lrsb{- \lrp{\vec R \; \vec \omega}^\wedge \ \ - \lrp{\vec R \; \vec \nu}^\wedge - \Tskew \lrp{\vec R \; \vec \omega}^\wedge + \lrp{\vec R \; \vec \omega}^\wedge \Tskew; \ \ \vec{0}_{3x3} \ \ - \lrp{\vec R \; \vec \omega}^\wedge}}} & \nm{\in \mathbb R^{6x6}} \\ 
	\nm{\vec J_{\ds{+ \; \vec \xi}}^{\ds{- \; \vec{Ad}_{\mathcal M}(\vec \xi)}}}					& = & \nm{\vec{Ad}_{\mathcal M}}		& = & \nm{\lrsb{\vec R \ \ \Tskew \; \vec R; \ \ \vec{0}_{3x3} \ \ \vec R}} & \nm{\in \mathbb R^{6x6}} \\ 

	\nm{\vec J_{\ds{\oplus \; \mathcal M}}^{\ds{- \; \vec{Ad}_{\mathcal M}^{-1}(\vec \xi)}}} 		& = & \multicolumn{3}{l}{\nm{\lrsb{\vec R^T \omegaskew \; \vec R \ \ \lrp{\vec R^T \vec \nu}^\wedge - \lrp{\vec R^T \Tskew \; \vec \omega}^\wedge; \ \ \vec{0}_{3x3} \ \ \lrp{\vec R^T \vec \omega}^\wedge}}} & \nm{\in \mathbb R^{6x6}} \\ 
	\nm{\vec J_{\ds{\boxplus \; \mathcal M}}^{\ds{- \; \vec{Ad}_{\mathcal M}^{-1}(\vec \xi)}}}		& = & \multicolumn{3}{l}{\nm{\lrsb{\vec R^T \omegaskew \ \ \vec R^T \widehat{\vec \nu} - \vec R^T \Tskew \; \omegaskew; \ \ \vec{0}_{3x3} \ \ \vec R^T \omegaskew}}} & \nm{\in \mathbb R^{6x6}} \\ 
	\nm{\vec J_{\ds{+ \; \vec \xi}}^{\ds{- \; \vec{Ad}_{\mathcal M}^{-1}(\vec \xi)}}}				& = & \nm{\vec{Ad}_{\mathcal M}^{-1}}	& = & \nm{\lrsb{\vec R^T \ \ - \vec R^T \, \Tskew; \ \ \vec{0}_{3x3} \ \ \vec R^T}}	& \nm{\in \mathbb R^{6x6}} \\ 	
	
	\nm{\vec J_{\ds{+ \; \vec \tau}}^{\ds{- \; g_{Exp\lrp{\vec \tau}}(\vec p)}}}					&   & 									& = & \nm{\Big[\vec J_L\lrp{\vec r} \ \ \vec Q\lrp{\vec \tau} - \lrsb{\vec R \, \vec p}^\wedge \, \vec J_L\lrp{\vec r} - \widehat{\vec T} \, \vec J_L\lrp{\vec r}\Big]} & \nm{\in \mathbb R^{3x6}} \\ 
	\nm{\vec J_{\ds{+ \; \vec \tau}}^{\ds{- \; g_{Exp\lrp{\vec \tau}}^{-1}(\vec p)}}}				&   & 									& = & \nm{\Big[- \vec J_L\lrp{- \vec r} \ \ - \vec Q\lrp{- \vec \tau} - \lrsb{\vec R^T \; \lrp{\vec T - \vec p}}^\wedge \, \vec J_L\lrp{- \vec r}\Big]} & \nm{\in \mathbb R^{3x6}} \\ 
	\hline
\end{tabular}
\end{center}
\captionof{table}{Rigid body motion Jacobians} \label{tab:RigidBody_motion_jacobians}
\renewcommand{\arraystretch}{1.0} % reset row height

%%%%%%%%%%%%%%%%%%%%%%%%%%%%%%%%%%%%%%%%%%%%%%%%%%%%%%%%%%%%%%%%%%%%%%%%
%%%%%%%%%%%%%%%%%%%%%%%%%%%%%%%%%%%%%%%%%%%%%%%%%%%%%%%%%%%%%%%%%%%%%%%%
%%%%%%%%%%%%%%%%%%%%%%%%%%%%%%%%%%%%%%%%%%%%%%%%%%%%%%%%%%%%%%%%%%%%%%%%
% SECTION      RIGID BODY MOTION DISCRETE INTEGRATION
%%%%%%%%%%%%%%%%%%%%%%%%%%%%%%%%%%%%%%%%%%%%%%%%%%%%%%%%%%%%%%%%%%%%%%%%
%%%%%%%%%%%%%%%%%%%%%%%%%%%%%%%%%%%%%%%%%%%%%%%%%%%%%%%%%%%%%%%%%%%%%%%%
%%%%%%%%%%%%%%%%%%%%%%%%%%%%%%%%%%%%%%%%%%%%%%%%%%%%%%%%%%%%%%%%%%%%%%%%

\section{Rigid Body Motion Discrete Integration}\label{sec:RigidBody_motion_integration}

The discrete integration with time of an element of a Lie group based on its Lie algebra is discussed in detail in section \ref{sec:algebra_integration}, which includes expressions for the Euler, Heun and Runge-Kutta methods. In the case of rigid body motion, the state vector includes the motion element \nm{\mathcal M \in \mathbb{SE}\lrp{3}} and its twist \nm{\vec \xi \in \mathbb{R}^6} contained in the tangent space, viewed either in the local (\nm{\xiEBB}) or global (\nm{\xiEBE}) frames. The Euler method expressions equivalent to (\ref{eq:algebra_integration_comp_X_euler}) and (\ref{eq:algebra_integration_comp_X_euler_left}) are shown below. Expressions for other integration schemes can easily be derived from those in section \ref{sec:algebra_integration}:
\begin{eqnarray}
\nm{\mathcal M_{k+1}} & \nm{\approx} & \nm{\mathcal M_k \oplus \lrsb{\Delta t \ \vec \xi_{{\sss EB}k}^{\sss B}} = \mathcal M_k \circ Exp\lrp{\Delta t \ \vec \xi_{{\sss EB}k}^{\sss B}}} \label{eq:SE3_integration_comp_X_euler} \\
\nm{\mathcal M_{k+1}} & \nm{\approx} & \nm{\lrsb{\Delta t \ \vec \xi_{{\sss EB}k}^{\sss E}} \boxplus \mathcal M_k = Exp\lrp{\Delta t \ \vec \xi_{{\sss EB}k}^{\sss E}} \circ \mathcal M_k} \label{eq:SE3_integration_comp_X_euler_left}
\end{eqnarray}

%%%%%%%%%%%%%%%%%%%%%%%%%%%%%%%%%%%%%%%%%%%%%%%%%%%%%%%%%%%%%%%%%%%%%%%%
%%%%%%%%%%%%%%%%%%%%%%%%%%%%%%%%%%%%%%%%%%%%%%%%%%%%%%%%%%%%%%%%%%%%%%%%
%%%%%%%%%%%%%%%%%%%%%%%%%%%%%%%%%%%%%%%%%%%%%%%%%%%%%%%%%%%%%%%%%%%%%%%%
% SECTION      RIGID BODY MOTION GAUSS-NEWTON OPTIMIZATION
%%%%%%%%%%%%%%%%%%%%%%%%%%%%%%%%%%%%%%%%%%%%%%%%%%%%%%%%%%%%%%%%%%%%%%%%
%%%%%%%%%%%%%%%%%%%%%%%%%%%%%%%%%%%%%%%%%%%%%%%%%%%%%%%%%%%%%%%%%%%%%%%%
%%%%%%%%%%%%%%%%%%%%%%%%%%%%%%%%%%%%%%%%%%%%%%%%%%%%%%%%%%%%%%%%%%%%%%%%

\section{Rigid Body Motion Gauss-Newton Optimization}\label{sec:RigidBody_motion_gauss_newton}

The minimization by means of the Gauss-Newton iterative method of the Euclidean norm of a nonlinear function whose input is a Lie group element is presented in section \ref{sec:algebra_gradient_descent}. In the case of rigid body motion, the resulting expressions for perturbations \nm{\Delta \vec \tau_{\sss EB}^{\sss E} \in \mathfrak{se}\lrp{3}} to an input motion \nm{\mathcal M \in \mathbb{SE}\lrp{3}} viewed in the global frame \nm{\FE} are shown in (\ref{eq:SE3_gauss_newton_iterative_left}) and (\ref{eq:SE3_gauss_newton_solution_left}), which are equivalent to the generic (\ref{eq:algebra_gradient_descent_iterative_lie_left}) and (\ref{eq:algebra_gradient_descent_solution_lie_left}). Refer to section \ref{sec:algebra_gradient_descent} for the meaning of the function Jacobian \nm{\vec J} and to section \ref{sec:RigidBody_motion_calculus_jacobians} for that of the left Jacobian \nm{\vec J_L}.
\begin{eqnarray}
\nm{\mathcal M_{k+1}} & \nm{\longleftarrow} & \nm{\Delta \vec \tau_{{\sss EB}k}^{\sss E} \boxplus \mathcal M_k = \Delta \vec \tau_{{\sss EB}k}^{\sss E} \circ Exp\lrp{\vec \tau_{{\sss EB}k}}} \label{eq:SE3_gauss_newton_iterative_left} \\
\nm{\Delta \vec \tau_{{\sss EB}k}^{\sss E}} & = & \nm{- \lrsb{\vec J_{Lk}^{-T} \, \vec J_k^T \, \vec J_k \, \vec J_{Lk}^{-1}}^{-1} \, \vec J_{Lk}^{-T} \, \vec J_k^T \, \vec{\mathcal E}_k} \label{eq:SE3_gauss_newton_solution_left} 
\end{eqnarray}    

If the perturbation is viewed in the local frame \nm{\FB}, (\ref{eq:algebra_gradient_descent_iterative_lie_right}) and (\ref{eq:algebra_gradient_descent_solution_lie_right}) are customized as follows, making use of the right Jacobian \nm{\vec J_R} defined in section \ref{sec:RigidBody_motion_calculus_jacobians}:
\begin{eqnarray}
\nm{\mathcal M_{k+1}} & \nm{\longleftarrow} & \nm{\mathcal M_k \oplus \Delta \vec \tau_{{\sss EB}k}^{\sss B} = Exp\lrp{\vec \tau_{{\sss EB}k}} \circ \Delta \vec \tau_{{\sss EB}k}^{\sss B}} \label{eq:SE3_gauss_newton_iterative_right} \\
\nm{\Delta \vec \tau_{{\sss EB}k}^{\sss B}} & = & \nm{- \lrsb{\vec J_{Rk}^{-T} \, \vec J_k^T \, \vec J_k \, \vec J_{Rk}^{-1}}^{-1} \, \vec J_{Rk}^{-T} \, \vec J_k^T \, \vec{\mathcal E}_k} \label{eq:SE3_gauss_newton_solution_right} 
\end{eqnarray}    

%%%%%%%%%%%%%%%%%%%%%%%%%%%%%%%%%%%%%%%%%%%%%%%%%%%%%%%%%%%%%%%%%%%%%%%%
%%%%%%%%%%%%%%%%%%%%%%%%%%%%%%%%%%%%%%%%%%%%%%%%%%%%%%%%%%%%%%%%%%%%%%%%
%%%%%%%%%%%%%%%%%%%%%%%%%%%%%%%%%%%%%%%%%%%%%%%%%%%%%%%%%%%%%%%%%%%%%%%%
% SECTION      RIGID BODY MOTION STATE ESTIMATION
%%%%%%%%%%%%%%%%%%%%%%%%%%%%%%%%%%%%%%%%%%%%%%%%%%%%%%%%%%%%%%%%%%%%%%%%
%%%%%%%%%%%%%%%%%%%%%%%%%%%%%%%%%%%%%%%%%%%%%%%%%%%%%%%%%%%%%%%%%%%%%%%%
%%%%%%%%%%%%%%%%%%%%%%%%%%%%%%%%%%%%%%%%%%%%%%%%%%%%%%%%%%%%%%%%%%%%%%%%

\section{Rigid Body Motion State Estimation}\label{sec:RigidBody_motion_SS}

The adaptation of the \hypertt{EKF} state estimation introduced in section \ref{sec:SS} to the case in which Lie group elements and their velocities are present is discussed in detail in section \ref{sec:algebra_SS}. For rigid body motion with local perturbations, it is necessary to replace \nm{\mathcal X \in \mathcal G} by \nm{\mathcal M \in \mathbb{SE}\lrp{3}}, \nm{\Delta \vec \tau^{\mathcal X} \in T_{\mathcal X}\mathcal G} by \nm{\Delta \vec \tau^{\sss B} \in \ \mathfrak{se}\lrp{3}}, \nm{\vec v^{\mathcal X} \in \mathbb{R}^m} by \nm{\vec \xi^{\sss B} \in \mathbb{R}^6}, \nm{\vec C_{{\mathcal {XX}}}^{\mathcal X} \in \mathbb{R}^{mxm}} by \nm{\vec C_{{\mathcal {MM}}}^{\sss B} \in \mathbb{R}^{6x6}}, and \nm{\vec J_{\ds{\oplus \; \mathcal X}}^{\ds{\ominus \; \mathcal X \oplus \vec \tau}}} by \nm{\vec J_{\ds{\oplus \; \mathcal M}}^{\ds{\ominus \; \mathcal M \oplus \vec \tau}}}. The particularizations for global perturbations are similar.

%%%%%%%%%%%%%%%%%%%%%%%%%%%%%%%%%%%%%%%%%%%%%%%%%%%%%%%%%%%%%%%%%%%%%%%%
%%%%%%%%%%%%%%%%%%%%%%%%%%%%%%%%%%%%%%%%%%%%%%%%%%%%%%%%%%%%%%%%%%%%%%%%
%%%%%%%%%%%%%%%%%%%%%%%%%%%%%%%%%%%%%%%%%%%%%%%%%%%%%%%%%%%%%%%%%%%%%%%%
% SECTION     APPLICATIONS OF THE VARIOUS MOTION REPRESENTATIONS
%%%%%%%%%%%%%%%%%%%%%%%%%%%%%%%%%%%%%%%%%%%%%%%%%%%%%%%%%%%%%%%%%%%%%%%%
%%%%%%%%%%%%%%%%%%%%%%%%%%%%%%%%%%%%%%%%%%%%%%%%%%%%%%%%%%%%%%%%%%%%%%%%
%%%%%%%%%%%%%%%%%%%%%%%%%%%%%%%%%%%%%%%%%%%%%%%%%%%%%%%%%%%%%%%%%%%%%%%%

\section{Applications of the Various Motion Representations}\label{sec:RigidBody_motion_applications}

This chapter discusses six different parameterizations of the rigid body motion or special Euclidean group \nm{\mathbb{SE}\lrp{3}}: the affine representation, the homogeneous matrix, the transform vector, the unit dual quaternion, the half transform vector, and the screw. Although in theory all of them can be employed for each of the purposes described in this chapter, and the required expressions derived, each parameterization has its own advantages and disadvantages, being suited for certain purposes but not recommended for others. 
\begin{itemize}
\item The affine representation \nm{\lrp{\mathcal R, \, \vec T}} based on either the rotation matrix or the unit quaternion is the most natural parameterization for rigid body motion. It can be employed to track the motion over its manifold, although other options are preferred. Many difficulties arise from its complex nature as a composition between an \nm{\mathbb{SO}\lrp{3}} rotation and a \nm{\mathbb{R}^3} vector, such as the complex inverse and concatenation, the different nature of the point and vector actions, the lack of simple plus and minus operators to deal with perturbations, and the need to continuously keep track of both components when moving over the manifold.

\item The homogeneous matrix \nm{\vec M} is a generalization of the rotation matrix for the case of rigid body motion that not only linearizes the transformation of coordinates at the expense of bigger size, but also adopts matrix algebra for the inversion and concatenation of transformations. A second advantage is that the transformations of vectors and points share the same map when employing homogeneous coordinates. Additionally, it provides a clear connection with the tangent space, together with the exponential and logarithmic maps, and plus and minus operators, which are not complex. Its main inconvenients are the huge size (16), the expense of maintaining the internal rotation matrix orthogonal if allowed to deviate from the manifold, and the need to work with homogeneous coordinates for both points and vectors. Its high cost precludes its use to track the motion over its manifold, although most implementations continuously compute its components (the rotation matrix and the translation vector) if the adjoint matrix or the Jacobian blocks are required. 

\item The unit dual quaternion \nm{\vec \zeta} is the preferred parameterization to track the motion over its manifold, even if it is necessary to obtain the rotation matrix and the translation vector for the adjoint and Jacobian blocks. It possesses a significant size advantage (8) with respect to the homogeneous matrix, although it is not cheap to recover its structure if allowed to deviate from the manifold. Unit dual quaternions are the least natural of the rigid body motion representations, being necessary to convert to a different \nm{\mathbb{SE}\lrp{3}} representation for visualization. While the inverse and concatenation are simple and linear, the motion actions for points and vectors are bilinear and require slightly different expressions when inverting them, which presents a disadvantage with respect to the the homogeneous matrix. A significant advantage is given by its direct relationship with the screw and associated \hypertt{ScLERP} capabilities. Unit dual quaternion expressions are more complex than those of the homogeneous matrix, and present a slightly less obvious connection with the tangent space. 

\item The main advantage of the transform vector \nm{\vec \tau} is that it belongs to the \nm{\mathfrak{se}\lrp{3}} tangent space while simultaneously being an \nm{\mathbb{SE}\lrp{3}} representation. It is hence indicated for those uses related with incremental motion changes by means of the exponential map together with the plus and minus operators (periodically adding the perturbations to the unit dual quaternion tracking the motion), such as discrete integration, optimization, and state estimation. The norms of its angular and linear components \nm{\lrp{\phi, \rho}} are the most adequate metrics for evaluating the distance (or estimation error) between two rigid bodies. Although it benefits from its straightforward inverse when used as a perturbation, its geometric appeal, and its small dimension (6), its usage for other applications is discouraged by its complex nonlinear kinematics, coordinate transformation, and composition, which are not shown in this chapter. 

\item The half transform vector \nm{\vec \Psi} is so similar (half) to the transform vector that its usage is not recommended in order to avoid confusion. Its only real application as the tangent space of the unit dual quaternion is in practice solved by dividing the transform vector by two when necessary.

\item The screw \nm{\vec S}, in addition to simultaneously belonging to the \nm{\mathfrak{se}\lrp{3}} tangent space and the \nm{\mathbb{SE}\lrp{3}} manifold, has the advantage that it clearly separates the influence of the motion direction from that of its magnitude, and as such it enables the definition of powers and \hypertt{ScLERP}, which can not be obtained with any of the other representations. The dimension is not big (8) and the inversion is straightforward, but all other possible expressions, including motion and concatenation, are very complex and not shown in this chapter.
\end{itemize}

%% file: files_arxiv/ch06_multiple.tex
\chapter{Relative Motion of Multiple Rigid Bodies} \label{cha:Composition}

This chapter develops expressions for the positions, velocities, and accelerations (both linear and angular) that refer to different reference systems or rigid bodies, which are in continuous motion (translation and rotation) among themselves. It relies on the analysis of the motion of rigid bodies performed in chapters \ref{cha:Rotate} and \ref{cha:Motion} to determine the relationships among these vectors when viewed in different reference frames.
\input{scripts/a8_multiple/multiple_frames}

The below expressions are based on the three reference systems shown in figure \ref{fig:SO3_ref_systems}: an inertial reference system \nm{F_0 \{\vec 0_0, \, \vec i_1^0, \, \vec i_2^0, \, \vec i_3^0\}}\footnote{An \emph{inertial frame} is that in which bodies whose net force acting upon them is zero do not experience any acceleration and remain either at rest or moving at constant velocity in a straight line. Any reference frame can be considered inertial for the analysis of the motion of a given object if the accelerations (linear and angular) of that frame with respect to an accepted inertial system can be discarded when compared with the accelerations characteristic of the movement being studied.}, and two non-inertial systems \nm{F_1 \{\vec 0_1, \, \vec i_1^1, \, \vec i_2^1, \, \vec i_3^1\}} and \nm{F_2 \{\vec 0_2, \, \vec i_1^2, \, \vec i_2^2, \, \vec i_3^2\}}, where \nm{\lrp{\Tzeroone, \ \vzeroone, \ \azeroone}} are the position, linear velocity, and linear acceleration of \nm{\vec O_1} with respect to \nm{\vec O_0}, \nm{\lrp{\Tzerotwo, \ \vzerotwo, \ \azerotwo}} those of \nm{\vec O_2} with respect to \nm{\vec O_0}, and \nm{\lrp{\Tonetwo, \ \vonetwo, \ \aonetwo}} those of \nm{\vec O_2} with respect to \nm{\vec O_1}. Similarly, \nm{\lrp{\wzeroone, \ \alphazeroone}} are the angular velocity and angular acceleration of \nm{F_1} with respect to \nm{F_0}, \nm{\lrp{\wzerotwo, \ \alphazerotwo}} those of \nm{F_2} with respect to \nm{F_0}, and \nm{\lrp{\wonetwo, \ \alphaonetwo}} those of \nm{F_2} with respect to \nm{F_1}. Consider also that \nm{\Rzeroone}, \nm{\Rzerotwo}, and \nm{\Ronetwo} are the rotation matrices (section \ref{sec:RigidBody_rotation_dcm}) among the three different rigid bodies.
\begin{itemize}

\item \textbf{Composition of Position}. The relationship between the linear position vectors \nm{\Tzerotwo}, \nm{\Tonetwo}, and \nm{\Tzeroone} can be established by vector arithmetics when expressed in the same reference frame, or by coordinate transformation (\ref{eq:SO3_dcm_transform}) when not so:
\neweq{\Tzerotwozero = \Tonetwozero + \Tzeroonezero = \Rzeroone \; \Tonetwoone + \Tzeroonezero}{eq:MOT_Comp_Pos1}

\item \textbf{Composition of Linear Velocity}. The derivation with time of (\ref{eq:MOT_Comp_Pos1}) results in:
\neweq{\Tzerotwozerodot = \Rzeroonedot \; \Tonetwoone +\Rzeroone \; \Tonetwoonedot + \Tzeroonezerodot}{eq:MOT_Comp_LinearVel1}

Replacing the rotation matrix time derivative by (\ref{eq:SO3_dcm_dot}) results in:
\neweq{\Tzerotwozerodot =\Rzeroone \; \wzerooneoneskew \; \Tonetwoone +\Rzeroone \; \Tonetwoonedot + \Tzeroonezerodot}{eq:MOT_Comp_LinearVel2}

Reordering, replacing the position time derivatives with their respective velocities, and employing (\ref{eq:SO3_dcm_dot}), results in the relationship between the linear velocity vectors \nm{\vzerotwo}, \nm{\vonetwo}, and \nm{\vzeroone} expressed in the inertial frame \nm{F_0}:
\begin{eqnarray}
\nm{\vzerotwozero} & = & \nm{\Rzeroone \; \vonetwoone + \vzeroonezero +\Rzeroone \; \wzerooneoneskew \; \Tonetwoone} \label {eq:MOT_Comp_LinearVel3} \\
\nm{\vzerotwozero} & = & \nm{\vonetwozero + \vzeroonezero + \wzeroonezeroskew \; \Tonetwozero} \label{eq:MOT_Comp_LinearVel4} 
\end{eqnarray}

\item \textbf{Composition of Linear Acceleration}. The derivation with time of (\ref{eq:MOT_Comp_LinearVel3}) results in:
\neweq{\vzerotwozerodot = \Rzeroonedot \; \vonetwoone +\Rzeroone \; \vonetwoonedot + \vzeroonezerodot + \Rzeroonedot \; \wzerooneoneskew \; \Tonetwoone +\Rzeroone \; \wzerooneoneskewdot \; \Tonetwoone +\Rzeroone \; \wzerooneoneskew \; \Tonetwoonedot} {eq:MOT_Comp_LinearAcc1}

Replacing the rotation matrix time derivative by (\ref{eq:SO3_dcm_dot}) results in:
\neweq{\vzerotwozerodot =\Rzeroone \; \wzerooneoneskew \; \vonetwoone +\Rzeroone \; \vonetwoonedot + \vzeroonezerodot +\Rzeroone \; \wzerooneoneskew \; \wzerooneoneskew \; \Tonetwoone + \Rzeroone \; \wzerooneoneskewdot \; \Tonetwoone + \Rzeroone \; \wzerooneoneskew \; \Tonetwoonedot} {eq:MOT_Comp_LinearAcc2}

Reordering, replacing the position, linear velocity, and angular velocity time derivatives with their respective velocities, linear accelerations and angular accelerations, and employing (\ref{eq:SO3_dcm_dot}), results in the relationship between the linear acceleration vectors \nm{\azerotwo}, \nm{\aonetwo}, and \nm{\azeroone} expressed in the inertial frame \nm{F_0}:
\begin{eqnarray}
\nm{\azerotwozero} & = & \nm{\Rzeroone \; \aonetwoone + \lrp{ \azeroonezero + \Rzeroone \; \alphazerooneoneskew \; \Tonetwoone + \Rzeroone \; \wzerooneoneskew \; \wzerooneoneskew \; \Tonetwoone } +  2 \; \Rzeroone \; \wzerooneoneskew \; \vonetwoone} \label{eq:MOT_Comp_LinearAcc3} \\
\nm{\azerotwozero} & = & \nm{\aonetwozero + \lrp{ \azeroonezero + \alphazeroonezeroskew \; \Tonetwozero + \wzeroonezeroskew \; \wzeroonezeroskew \; \Tonetwozero } +  2 \; \wzeroonezeroskew \; \vonetwozero} \label{eq:MOT_Comp_LinearAcc4}
\end{eqnarray}

The term on the left hand side is called absolute acceleration, while the three right hand side terms are usually named relative, transport, and Coriolis accelerations, respectively.

\item \textbf{Composition of Angular Velocity}. The relationship among the different frames angular velocities is given by the rotation matrix composition rule (\ref{eq:SO3_dcm_concatenation}), which can be derivated with respect to time:
\neweq{\Rzerotwo = \Rzeroone \, \Ronetwo \ \rightarrow \ \Rzerotwodot = \Rzeroonedot \, \Ronetwo + \Rzeroone \, \Ronetwodot} {eq:MOT_Comp_AngularVel1}

Replacing the rotation matrix time derivatives by (\ref{eq:SO3_dcm_dot}) and employing the rotational motion adjoint (\ref{eq:SO3_dcm_velocity5}) results in:
\neweq{\Rzerotwo \; \wzerotwotwoskew = \wzeroonezeroskew \; \Rzeroone \; \Ronetwo + \Rzeroone \; \Ronetwo \; \wonetwotwoskew = \wzeroonezeroskew \; \Rzerotwo + \Rzerotwo \; \wonetwotwoskew = \Rzerotwo \; \wzeroonetwoskew + \Rzerotwo \; \wonetwotwoskew} {eq:MOT_Comp_AngularVel2}

The relationship among the angular velocity vectors \nm{\wzerotwo}, \nm{\wonetwo}, and \nm{\wzeroone} is hence the following:
\neweq{\wzerotwozero = \wonetwozero + \wzeroonezero = \Rzeroone \; \wonetwoone + \wzeroonezero}{eq:MOT_Comp_AngularVel3}

\item \textbf{Composition of Angular Acceleration}. The derivation with time of (\ref{eq:MOT_Comp_AngularVel3}) results in:
\neweq{\wzerotwozerodot = \Rzeroonedot \; \wonetwoone + \Rzeroone \; \wonetwoonedot + \wzeroonezerodot}{eq:MOT_Comp_AngularAcc1}

Replacing the rotation matrix time derivatives by (\ref{eq:SO3_dcm_dot}) results in:
\neweq{\wzerotwozerodot = \Rzeroone \; \wzerooneoneskew \; \wonetwoone + \Rzeroone \; \wonetwoonedot + \wzeroonezerodot}{eq:MOT_Comp_AngularAcc2}

Reordering, replacing the angular velocity time derivatives with their respective angular accelerations, and employing (\ref{eq:SO3_dcm_dot}), results in the relationship between the angular acceleration vectors \nm{\alphazerotwo}, \nm{\alphaonetwo}, and \nm{\alphazeroone} expressed in the inertial frame \nm{F_0}:
\begin{eqnarray}
\nm{\alphazerotwozero} & = & \nm{\Rzeroone \; \alphaonetwoone + \alphazeroonezero + \Rzeroone \; \wzerooneoneskew \; \wonetwoone} \label {eq:MOT_Comp_AngularAcc3} \\
\nm{\alphazerotwozero} & = & \nm{\alphaonetwozero + \alphazeroonezero + \wzeroonezeroskew \; \wonetwozero} \label {eq:MOT_Comp_AngularAcc4} 
\end{eqnarray}

\end{itemize}

The final expressions of the previous compositions (\ref{eq:MOT_Comp_Pos1}, \ref{eq:MOT_Comp_LinearVel4}, \ref{eq:MOT_Comp_LinearAcc4}, \ref{eq:MOT_Comp_AngularVel3}, and \ref{eq:MOT_Comp_AngularAcc4}) are all expressed in the inertial frame \nm{F_0}, but they are also valid in any other frame as long as all its components are converted into that frame\footnote{Note that it is not the same to compute a time derivative (velocity, acceleration, or angular acceleration) in the inertial frame and then convert it into a different frame, than to directly compute the derivative in a non-inertial frame.}:
\begin{eqnarray}
\nm{\Tzerotwo} & = & \nm{\Tonetwo + \Tzeroone} \label{eq:MOT_Comp_PosFinal} \\
\nm{\vzerotwo} & = & \nm{\vonetwo + \vzeroone + \wzerooneskew \; \Tonetwo} \label{eq:MOT_Comp_LinearVelFinal} \\ 
\nm{\azerotwo} & = & \nm{\aonetwo + \lrp{ \azeroone + \alphazerooneskew \; \Tonetwo + \wzerooneskew \; \wzerooneskew \; \Tonetwo } +  2 \; \wzerooneskew \; \vonetwo} \label{eq:MOT_Comp_LinearAccFinal} \\
\nm{\wzerotwo} & = & \nm{\wonetwo + \wzeroone}\label{eq:MOT_Comp_AngularVelFinal} \\
\nm{\alphazerotwo} & = & \nm{\alphaonetwo + \alphazeroone + \wzerooneskew \; \wonetwo} \label {eq:MOT_Comp_AngularAccFinal}
\end{eqnarray}

%% file: scripts/a8_multiple/multiple_frames.tex
\begin{figure}[h]
\centering
\begin{tikzpicture}[auto, node distance=2cm,>=latex']
	\filldraw [black] (+0.0, +0.0) circle [radius=2pt] node [above left=1pt] {\nm{\vec O_0}};
    \draw [->] (+0.0,+0.0) -- (-1.2,-0.5) node [left=3pt]  {\nm{\vec i_1^0}};
	\draw [->] (+0.0,+0.0) -- (+1.4,-0.5) node [right=3pt] {\nm{\vec i_2^0}};
	\draw [->] (+0.0,+0.0) -- (+0.0,+1.5) node [left=3pt]  {\nm{\vec i_3^0}};
	
	\filldraw [black, xshift=3.5cm, yshift=0.5cm] (+0.0,+0.0) circle [radius=2pt] node [below right=1pt] {\nm{\vec O_1}};
    \draw [xshift=3.5cm, yshift=0.5cm] [->] (+0.0,+0.0) -- (-0.6,-1.0) node [right=3pt] {\nm{\vec i_1^1}};
	\draw [xshift=3.5cm, yshift=0.5cm] [->] (+0.0,+0.0) -- (+1.4,+0.2) node [right=3pt] {\nm{\vec i_2^1}};
	\draw [xshift=3.5cm, yshift=0.5cm] [->] (+0.0,+0.0) -- (-0.4,+1.3) node [right=3pt] {\nm{\vec i_3^1}};
	
	\filldraw [black, xshift=1.5cm, yshift=2.0cm] (+0.0,+0.0) circle [radius=2pt] node [above left=0.5pt] {\nm{\vec O_2}};
    \draw [xshift=1.5cm, yshift=2.0cm] [->] (+0.0,+0.0) -- (+1.0,-0.5) node [above=3pt] {\nm{\vec i_1^2}};
	\draw [xshift=1.5cm, yshift=2.0cm] [->] (+0.0,+0.0) -- (+0.5,+1.0) node [right=1pt] {\nm{\vec i_2^2}};
	\draw [xshift=1.5cm, yshift=2.0cm] [->] (+0.0,+0.0) -- (-1.3,+0.0) node [above=1pt] {\nm{\vec i_3^2}};
	
	\draw [dashed, thick] [->] (+0.0,+0.0) -- (+3.5,+0.5) node [pos=0.7, above=1pt] {\nm{\Tzeroone}};
	\draw [dashed, thick] [->] (+0.0,+0.0) -- (+1.5,+2.0) node [pos=0.8, left=1pt]  {\nm{\Tzerotwo}};
	\draw [dashed, thick] [->] (+3.5,+0.5) -- (+1.5,+2.0) node [pos=0.6, left=2pt]  {\nm{\Tonetwo}};
\end{tikzpicture}
\caption{Reference system for combination of movements}
\label{fig:SO3_ref_systems}
\end{figure}
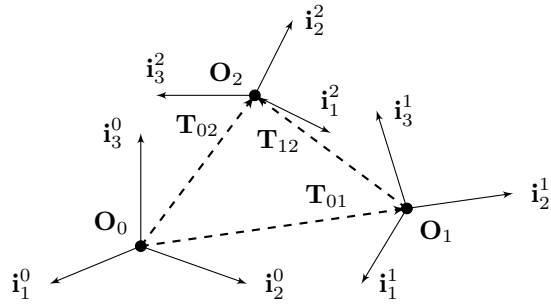